\documentclass{article}
\usepackage{iclr2025_conference, times}

\newcounter{notecounter}
\usepackage[utf8]{inputenc} % allow utf-8 input
\usepackage[T1]{fontenc}    % use 8-bit T1 fonts
\usepackage{hyperref}
\hypersetup{
     colorlinks=true,
     linkcolor=blue,
     filecolor=blue,
     citecolor=blue,
     urlcolor=blue,
     }
\usepackage{framed}
\usepackage{multirow}
\usepackage{url}
\usepackage{amsfonts}       % blackboard math symbols
\usepackage{nicefrac}       % compact symbols for 1/2, etc.
\usepackage{microtype}      % microtypography
\usepackage{xcolor}         % colors
\usepackage{multicol}
\usepackage{pythonhighlight}
\usepackage{graphicx}
\usepackage{subfigure}
\usepackage{tikz}
\usetikzlibrary{arrows.meta, positioning}
\usepackage{caption}
\usepackage{amsthm}
\usepackage{bbm}
\usepackage{comment}
\usepackage{listings}
\usepackage{bm}
\usepackage{booktabs}
\usepackage{algorithm,algorithmic}
\usepackage{subfigure}
\usepackage{mathtools}
\usepackage[export]{adjustbox}
\usepackage{stmaryrd}
\usepackage{booktabs} % for professional tables
\usepackage{xifthen}
\usepackage{xargs}
\usepackage{mathtools}
\usepackage{amsmath,amssymb}
\usepackage{hyperref}
\usepackage{cleveref}
\usepackage{titlesec}
\usepackage{wrapfig}
\titleformat*{\subparagraph}{\itshape}
\newtheorem{thm}{Theorem}[section]

\newtheorem{lemma}[thm]{Lemma}
\newtheorem{remark}[thm]{Remark}
\newtheorem{example}[thm]{Example}

\numberwithin{equation}{section}

\usepackage{amsmath}
\usepackage{amssymb}
\usepackage{mathtools}
\usepackage{amsthm}

\usepackage{xspace}

\def\rset{\mathbb{R}}
\def\nset{\mathbb{N}}
\def\rmd{\mathrm{d}}
\def\rme{\mathrm{e}}
\def\normpdf{\mathcal{N}}
\def\eqsp{\,}
\def\Id{\mathbf{I}}
\def\zero{\mathbf{0}}

\def\pE{\mathbb{E}}

\def\cov{\Sigma}
\def\var{v}
\def\indic{\mathbbm{1}}
\def\eqdef{\vcentcolon=}
\def\iid{i.i.d.}
\def\simiid{\overset{\mathrm{i.i.d.}}{\sim}}
\def\wrt{w.r.t.}
\def\diag{\mbox{diag}}
\def\normconst{\mathcal{Z}}
\def\law{\mathsf{Law}}

\def\trace{\operatorname{tr}}

\def\rhs{r.h.s.}

\newcommand{\intset}[2]{\llbracket #1, #2 \rrbracket}
\newcommand{\kldivergence}[2]{\mathsf{KL}(#1 \parallel #2)}

\def\algoname{{\sc{Midpoint Guidance Posterior Sampling}}}

\def\algo{{\sc{MGPS}}}
\def\pgdm{{\sc{PGDM}}}
\def\dps{{\sc{DPS}}}
\def\diffpir{{\sc{DiffPir}}}
\def\traineddiff{{\sc{TrainedDiff}}}
\def\ddnm{{\sc{DDNM}}}
\def\resample{{\sc{ReSample}}}
\def\psld{{\sc{PSLD}}}

\def\reddiff{{\sc{RedDiff}}}

\def\ffhq{{\texttt{FFHQ}}}
\def\imagenet{{\texttt{ImageNet}}}

\def\gauss{\mathrm{N}}
\def\bfA{\mathbf{A}}
\def\stdobs{\sigma_\obs}
\def\obs{{\boldsymbol{y}}}
\def\target{\pi}
\def\param{\theta}
\def\prior{q}
\def\dimobs{{d_\obs}}
\def\dimx{{d}}
\def\datadistr{q}
\newcommandx{\acp}[2][2=]{\ifthenelse{\equal{#2}{}}{\alpha_{#1}}{\alpha_{#1:#2}}}
\newcommandx{\hpredx}[3][2=0,3=\param]{\smash{\boldsymbol{m}^{#3} _{#2|#1}}}
\newcommandx{\predx}[2][2=0]{\smash{\boldsymbol{m} _{#2|#1}}}
\newcommandx{\prednoise}[2][2=\param]{\smash{\boldsymbol{\epsilon}^{#2} _{#1}}}
\newcommandx{\score}[2][2=\param]{\boldsymbol{s}^{#2} _{#1}}
\newcommand{\post}[2]{\ifthenelse{\equal{#2}{}}{\pi_{#1}}{\pi_{#1}(#2)}}
\newcommandx{\fw}[4][4=]{\ifthenelse{\equal{#3}{}}{q^{#4} _{#1}}{q^{#4} _{#1}(#3|#2)}}
\newcommandx{\fwmarg}[3][3=]{\ifthenelse{\equal{#2}{}}{q^{#3} _{#1}}{q^{#3} _{#1}(#2)}}
\newcommandx{\hmrg}[2][2=\ell]{\smash{\hat\pi^{#2} _{#1}}}
\newcommandx{\bw}[4][4=]{\ifthenelse{\equal{#3}{}}{q^{#4} _{#1}}{q^{#4} _{#1}(#3|#2)}}
\newcommandx{\bwp}[4][4=\param]{\ifthenelse{\equal{#3}{}}{p^{#4} _{#1}}{p^{#4} _{#1}(#3|#2)}}
\newcommandx{\hpibw}[4][4=\param]{\ifthenelse{\equal{#3}{}}{\smash{\hat\pi^{#4} _{#1}}}{\smash{\hat\pi^{#4} _{#1}(#3|#2)}}}
\newcommandx{\hpibwD}[4][4=\param]{\ifthenelse{\equal{#3}{}}{\smash{\hat\pi^{#4} _{#1}}}{\smash{\hat\pi^{#4} _{#1}(#3)}}}
\newcommandx{\vpibw}[4][4=\vparam]{\ifthenelse{\equal{#3}{}}{\lambda^{#4} _{#1}}{\lambda^{#4} _{#1}(#3|#2)}}
\newcommandx{\vpibwD}[4][4=\vparam]{\ifthenelse{\equal{#3}{}}{\lambda^{#4} _{#1}}{\lambda^{#4} _{#1}(#3)}}
\newcommand{\hpotn}[2]{\ifthenelse{\equal{#2}{}}{\hat{p}^\param _{#1}(\obs|\cdot)}{\hat{p}^\param _{#1}(\obs|#2)}}
\newcommand{\tpotn}[2]{\ifthenelse{\equal{#2}{}}{\tilde{p} _{#1}(\obs|\cdot)}{\tilde{p} _{#1}(\obs|#2)}}
\newcommand{\potn}[2]{\ifthenelse{\equal{#2}{}}{p_{#1}(\obs|\cdot)}{p_{#1}(\obs|#2)}}
\newcommandx{\pibw}[4][4=]{\ifthenelse{\equal{#3}{}}{\smash{\target}^{#4} _{#1}}{\smash{\target}^{#4} _{#1}(#3|#2)}}
\newcommandx{\fwtrans}[4][4=]{\ifthenelse{\equal{#2}{}}{q_{#1}}{q _{#1}(#3|#2)}}

\newcommandx{\vmu}[1][1=]{\hat{\boldsymbol{\mu}}^{#1}}
\newcommandx{\vstd}[1][1=]{\hat\sigma^{#1}}
\newcommandx{\vlstd}[1][1=]{\hat{\boldsymbol{\rho}}^{#1}}
\def\vX{\hat{\mathbf{X}}}
\def\vparam{{\boldsymbol{\varphi}}}
\def\by{\boldsymbol{y}}
\def\bx{\boldsymbol{x}}
\def\bX{\mathbf{X}}
\def\bY{\mathbf{Y}}
\def\bz{\boldsymbol{z}}
\def\bZ{\mathbf{Z}}
\def\beps{\boldsymbol{\epsilon}}
\def\meanBridge{\boldsymbol{m}}
\def\encoder{\mathcal{E}}
\def\decoder{\mathcal{D}}
\def\tmidfn{\ell}
\newcommand{\tmid}[1]{{\tmidfn_{#1}}}
\def\ngrad{M}
\newcommand{\revision}[1]{#1}

\newcommand{\mean}[1]{m_{#1}}
\newcommand\blfootnote[1]{%
  \begingroup
  \renewcommand\thefootnote{}\footnote{#1}%
  \addtocounter{footnote}{-1}%
  \endgroup
}
\iclrfinalcopy
\title{Variational Diffusion Posterior Sampling with Midpoint Guidance}

\author{%
Badr Moufad${}^{*, 1}$ \, Yazid Janati${}^{*, 1}$ \, Lisa Bedin${}^{*, 1}$\\
\bf{Alain Durmus${}^{1}$ \, Randal Douc${}^{2}$ \, Eric Moulines${}^{1}$ \, Jimmy Olsson${}^{3}$} \\
${}^{1}$Ecole polytechnique\, ${}^{2}$Télécom SudParis \, ${}^{3}$KTH University
}

\begin{document}
  \maketitle

  \begin{abstract}

  %   Diffusion models have recently shown significant potential in solving Bayesian inverse problems by using these models as priors. However, sampling from the resulting denoising posterior distributions, denoted as $\pi$, remains challenging. To tackle this, prominent approaches decompose the problem into sampling from a surrogate model that corresponds to a diffusion model targeting $\pi$. Sampling from this model involves sampling from a sequence of intractable transitions. Existing methods decompose the associated score into two parts: the prior score and an intractable guidance term. While the former is replaced by the pre-trained score of the considered diffusion model, the guidance term must be estimated.
  % In this work, we propose a novel approach that leverages a decomposition of the transitions. Unlike previous methods, our decomposition allows for a trade-off between the complexity of the intractable guidance term and that of the prior transitions. We validate the proposed approach through extensive experiments on both linear and non-linear inverse problems, including challenging cases with latent diffusion models as priors.
Diffusion models have recently shown considerable potential in solving Bayesian inverse problems when used as priors. However, sampling from the resulting
denoising posterior distributions remains a challenge as it involves intractable terms. To tackle this issue, state-of-the-art approaches formulate the problem as that of sampling from a surrogate diffusion model targeting the posterior and decompose its scores into two terms: the prior score and an intractable guidance term. While the former is replaced by the pre-trained score of the considered diffusion model, the guidance term has to be estimated. In this paper, we propose a novel approach that utilises a decomposition of the transitions which, in contrast to previous methods, allows a trade-off between the complexity of the intractable guidance term and that of the prior transitions. We validate the proposed approach through extensive experiments on linear and nonlinear inverse problems, including challenging cases with latent diffusion models as priors. \revision{We then demonstrate its applicability to various modalities and its promising impact on public health by tackling cardiovascular disease diagnosis through the reconstruction of incomplete electrocardiograms.} The code is publicly available at \url{https://github.com/yazidjanati/mgps}.
%, and demonstrate itseffectiveness in reconstructing electrocardiogram (ECG) from partial measurements for accurate cardiac diagnosis.

% We illustrate the proposed approach through extensive experiments on both linear and non-linear inverse problems with challenging nonlinearities, including inverse problems with latent diffusion models as prior.
  \end{abstract}

  \blfootnote{${}^{*}$ authors contributed equally.}
  \blfootnote{Correspondence: $\{\texttt{firstname.secondname@polytechnique.edu}\} $}

  %%%%%%%% sections %%%%%%%%
  \section{Introduction}
  % Inverse problems are now ubiquitous in many fields, such as ... \alain{cite + example} \badr{This first sentence is very similar to the first sentence in DAPS}. These problems aim to reconstruct signals from incomplete or noisy measurements. These problems are often ill-posed, i.e.,  multiple solutions fit the observed data due to the noise and complexity of the measurement process.
 Inverse problems aim to reconstruct signals from incomplete and noisy observations and are prevalent across various fields. In signal and image processing, common examples include signal deconvolution, image restoration, and tomographic image reconstruction \cite{stuart2010inverse,idier2013bayesian}. Other applications extend to protein backbone motif scaffolding \cite{watson2023novo} and urban mobility modeling \cite{jiang2023motiondiffuser}. Due to the presence of noise and the inherent complexity of the measurement process, these problems are typically ill-posed, meaning they have an infinite number of possible solutions. While many of these solutions may fit the observed data, only a few align with the true underlying signal.
  %In such cases, the challenge is not merely to find any solution that matches the observations, but to find one that adheres to the inherent structure or properties of the original signal. For example, when part of a signal is missing, the reconstruction should not only be consistent with the observed data but should also reflect the natural symmetries or patterns of the whole.
  Bayesian inverse problems provide a principled framework for addressing the challenges of signal reconstruction by incorporating prior knowledge, enabling more plausible and meaningful solutions. The \emph{a priori} knowledge about the signal $\bx$ to be reconstructed is captured in the prior distribution $\prior(\bx)$, while the information about the observation $\obs$ is encoded in the likelihood function $p(\obs|\bx)$. Given these components, solving the inverse problem boils down to sampling from the posterior distribution $\pi(\bx) \propto p(\obs|\bx) \prior(\bx)$, which integrates both prior knowledge and observational data.

  The choice of the prior distribution is crucial for achieving accurate reconstructions. If the goal is to reconstruct high-resolution data, the prior must be able to model data with similar fine detail and complexity. Recently, denoising diffusion models (DDM) \cite{sohl2015deep, song2019generative,ho2020denoising,song2021score} have emerged as a powerful approach in this context. These models can generate highly realistic and detailed reconstructions and are becoming increasingly popular as priors.
  Using a DDM, a sample $\bX _0$ being approximately distributed according to a given data distribution $\prior$ of interest can be generated by iteratively denoising an initial sample $\bX _n$ from a standard Gaussian distribution.
  %They generate a sample $\bX _0$ by iteratively denoising an initial sample $\bX _n$ from a standard Gaussian distribution.
The denoising process $(\bX _k)_{k = n}^0$ consists of the iterative refinement of $\bX _k$ for $k \in \intset{1}{n}$ based on a parametric approximation of the score $\nabla \log q_k$, where $q_k(\bx _k) = \int \prior(\bx _0) \fwtrans{k|0}{\bx _0}{\bx _k} , \rmd \bx _0$, with $\fwtrans{k|0}{\bx _0}{\cdot}$ being a Gaussian noising transition.

Denoising Diffusion Models (DDMs) form a highly powerful class of prior distributions, but their use introduces significant challenges in posterior sampling. Specifically, since a DDM prior does not admit an explicit and tractable density, conventional Markov chain Monte Carlo (MCMC) methods, such as the Metropolis--Hastings algorithm and its variants \cite{besag:1994,Neal2011}, cannot be applied in general. Furthermore, gradient-based MCMC methods often prove inefficient, as they tend to get trapped in local modes of the posterior.
%It would be possible to run MCMC on the extended space to circumvent the requirement of evaluating the marginal density, their memory footprint and shortcomings in high dimensions make them computationally prohibitive in this context.

The problem of sampling from the posterior in Bayesian inverse problems with DDM priors has recently been addressed in several papers \cite{kadkhodaie2020solving, song2021score, kawar2021snips, kawar2022denoising, lugmayr2022repaint, ho2022video, chung2023diffusion, song2022pseudoinverse}. These methods typically modify the denoising process to account for the observation $\obs$. One principled approach for accurate posterior sampling involves skewing the denoising process at each stage $k$ using the score $\nabla \log \post{k}{}$, where $\post{k}{}$ is defined analogously to $\fwmarg{k}{}$, with $\prior$ replaced by the posterior $\post{}{}$. This guides the denoising process in a manner that is consistent with the observation. Notably, the posterior score $\nabla \log \post{k}{}$ decomposes into two components: the prior score and an additional intractable term commonly referred to as \emph{guidance}. Previous works leverage \emph{pre-trained} DDMs for the prior score, while providing various \emph{training-free} approximations for the guidance term \cite{ho2022video, chung2023diffusion, song2022pseudoinverse, finzi2023user}. This framework enables solving a wide range of inverse problems for a given prior. However, despite the notable success of these methods, the approximations often introduce significant errors, leading to inaccurate reconstructions when the posterior distribution is highly multimodal, the measurement process is strongly nonlinear, or the data is heavily contaminated by noise.

  % \begin{wrapfigure}{r}{5.0cm}
  %   \centering
  %   \vspace{-0.25cm}
  %   \includegraphics[width=.3\textwidth]{figures/illustration_tmid.pdf}
  %   \vspace{-0.1cm}
  %   \captionsetup{font=small}
  %   \caption{Illustration of intermediate step decomposition.}
   %   \label{fig:illus-tmid-decompistion}
  %   \vspace{-0.3cm}
  % \end{wrapfigure}

\paragraph{Our contribution.} We begin with the observation that the posterior denoising step, which transforms a sample $\bX _{k+1}$ into a sample $\bX _k$, does not necessarily require the conditional scores $\nabla \log \post{k+1}{}$ or $\nabla \log \post{k}{}$.
Instead, we demonstrate that this step can be decomposed into two intermediate phases: first, denoising $\bX _{k+1}$ into an intermediate \emph{midpoint} state $\bX _\tmid{k}$, where $\tmid{k} < k$, and then noising back to obtain $\bX _k$, unconditionally on the observation $\obs$.
This decomposition introduces an additional degree of freedom, as it only requires the estimation of the guidance term at the intermediate step $\tmid{k}$ rather than step $k+1$.
Building on this insight, we introduce \algoname\, (\algo), a novel diffusion posterior sampling scheme that explicitly leverages this approach.
Our algorithm develops a principled approximation of the denoising transition by utilizing a Gaussian variational approximation, combined with the guidance approximation proposed by \citet{chung2023diffusion} at the intermediate steps $\tmid{k}$.
The strong empirical performance of our method is demonstrated through an extensive set of experiments.
In particular, we validate our approach on a Gaussian mixture toy example, as well as various linear and nonlinear image reconstruction tasks--inpainting, super-resolution, phase retrieval, deblurring, JPEG dequantization, high-dynamic range--using both pixel-space and latent diffusion models (LDM).
Finally, we demonstrate the versatility of our approach by applying it to cardiovascular diagnosis using reconstructed electrocardiograms, showing that MGPS achieves significant improvements over competing methods.
%We propose using posterior sampling to complete ECGs to 12 leads, addressing incomplete data and enhancing anomaly detection. This approach is unprecedented in completing ECGs for better diagnosis.
%their generalizability and societal impact by considering imputation tasks for 12-lead electrocardiograms (ECGs).
%}
%Our results demonstrate that \emph{training-free} posterior sampling performs on par with diffusion models specifically trained for ECG imputation~\cite{alcarazImputation}.
%Moreover, \algo\ demonstrates significantly improved performance compared to competitors for cardiovascular diagnosis from reconstructed ECGs

  % \input{sections/background}
  \section{Posterior sampling with DDM prior}
  \paragraph{Problem setup.}
In this paper, we focus on the approximate sampling from a density of the form
\begin{equation}
 \label{eq:posterior_def}
 \post{}{\bx} \eqdef \potn{}{\bx} \prior(\bx) / \normconst
 \eqsp,
\end{equation}
where $\potn{}{}: \rset^{\dimx} \to \rset$ is a \emph{non-negative} and \emph{differentiable} likelihood that \emph{can be evaluated pointwise}, $\prior$ is a prior distribution, and $\smash{\normconst \eqdef \int \potn{}{\bx} \prior(\bx) \, \rmd \bx}$ is the normalizing constant.
We are interested in the case where a DDM $\bwp{0}{}{}$ for the prior has been pre-trained for the prior based on \iid\ observations from $\prior$.
As a by-product, we also have parametric approximations $(\score{k})_{k = 1} ^n$ of the \emph{scores} $(\nabla \log \fwmarg{k}{})_{k = 1}^n$, where the marginals $(\fwmarg{k}{})_{k=1}^n$ are defined as $\fwmarg{k}{\bx _k} \eqdef \int \datadistr(\bx _0) \, \fwtrans{k|0}{\bx _0}{\bx _k} \, \rmd \bx _0$ with $\fwtrans{k|0}{\bx _0}{\bx _k} \eqdef \normpdf(\bx _k; \sqrt{\acp{k}} \bx _0, \var_k \Id_\dimx)$ and $\var _k \eqdef 1 - \acp{k}$.
% \jimmy{Not.}
% and $(\acp{k})_{k = 0}^n$n $\var _k \eqdef 1 - \acp{k}$
Typically, the sequence $(\acp{k})_{k = 0}^n$ is a decreasing sequence with $\acp{0} = 1$ and $\acp{n}$ approximately equals zero for $n$ large enough.

These score approximations enable the generation of new samples from $\datadistr$ according to one of the DDM sampling schemes \cite{song2019generative,ho2020denoising,song2021ddim,song2021score,karras2022elucidating}.
All these approaches boil down to simulating a Markov chain $(\bX _k)_{k = n}^0$ backwards in time starting from a standard Gaussian $\bwp{n}{}{}[] \eqdef \gauss(\zero_\dimx, \Id_{\dimx})$ and following the Gaussian transitions
$\bwp{k|k+1}{\bx _{k+1}}{\bx _{k}} = \normpdf(\bx _k; \hpredx{k+1}[k](\bx _{k+1}), \var_{k|k+1} \Id_{\dimx})$,
where the mean functions $\hpredx{k+1}[k]$ can be derived from the learned DDM, and $\var_{k|k+1}>0$ are fixed and pre-defined variances.
% where $k$ goes from $n - 1$ to zero,
% for some mean functions $\hpredx{k+1}[k]$, which refer to the learned values, and some fixed variances $\var_{k|k+1}>0$.
% The initial distribution of the chain is $\bwp{n}{}{}[] \eqdef \normpdf(\zero, \Id_{\dimx})$.
The backward transitions are designed in such a way that the marginal law of $\bX _k$ approximates $\fwmarg{k}{}$; see \Cref{sec:ddpm-background} for more details on the form of the backward transitions.
% For more details on the form of the backward transitions, see \Cref{sec:ddpm-background}.

In our problem setup, we assume we only have access to the approximated scores of the prior and no observation from neither posterior $\post{}{}$ nor the prior $\prior$.
This setup encompasses of solving an inverse problem without training a conditional model from scratch based on a paired signal/observation dataset; a requirement typically imposed by conditional DDM frameworks \cite{song2021score,batzolis2021conditional,tashiro2021csdi, saharia2022image}.
On the other hand, we require access to $\potn{}{}$.
This setup includes many applications in Bayesian inverse problems, \emph{e.g.}, image restoration,
%such as Super Resolution, JPEG dequantization, phase retrieval, \dots,
motif scaffolding in protein design \cite{trippe2023diffusion,wu2023practical}, and trajectory control in traffic-simulation frameworks \cite{jiang2023motiondiffuser}.

\paragraph{Conditional score.}
Since we have access to a DDM model for $\datadistr$, a natural approach to sampling from $\post{}{}$ is to define a DDM approximation based on the pre-trained score approximations for $\prior_k$.
% defining a DDM approximation for $\post{}{}$ using the pre-trained score approximations for $\prior_k$ provides a natural approach to sampling from $\post{}{}$.
% ; see \Cref{sec:vdps-algo} for a more detailed discussion.
Following the DDM approach, the basic idea here is to sample sequentially from the marginals $\post{k}{\bx _k} \eqdef \int \post{}{\bx _0} \fwtrans{k|0}{\bx _0}{\bx _k} \, \rmd \bx _0$ by relying on approximations of the conditional scores $(\nabla \log \post{k}{})_{k = 1}^n$.
% The latter can be expressed in terms of the unconditional scores by using the identity
% \begin{equation}
%   \label{eq:posterior-forward}
%   \post{k}{\bx _k} \propto \int \potn{}{\bx _0} \datadistr(\bx _0) \fwtrans{k|0}{\bx _0}{\bx _k} \, \rmd \bx _0 \propto \potn{k}{\bx _k} \fwmarg{k}{\bx _k} \eqsp,
% \end{equation}
% with
% \begin{equation}
% \label{eq:def_gk}
%     \potn{k}{\bx _k} \eqdef \int \potn{}{\bx _0} \bw{0|k}{\bx _k}{\bx _0} \, \rmd \bx _0 \eqsp,
% \end{equation}
% where we have defined the backward transition kernel as $\bw{0|k}{\bx _k}{\bx _0} \eqdef \datadistr(\bx _0) \fw{k|0}{\bx _0}{\bx _k} \big/ \fwmarg{k}{\bx _k}$. It then follows that
% \begin{equation*}
%   \nabla \log \post{k}{\bx _k} = \nabla \log \fwmarg{k}{\bx _k} + \nabla \log \potn{k}{\bx _k} \eqsp.
% \end{equation*}
The latter can be expressed in terms of the unconditional scores by using %the identity
\begin{equation}
  \label{eq:posterior-forward}
  \post{k}{\bx _k} \propto \int \potn{}{\bx _0} \datadistr(\bx _0) \fwtrans{k|0}{\bx _0}{\bx _k} \, \rmd \bx _0
  \eqsp,
  % \propto \potn{k}{\bx _k} \fwmarg{k}{\bx _k} \eqsp,
\end{equation}
where after defining the backward transition kernel $\bw{0|k}{\bx _k}{\bx _0} \eqdef \datadistr(\bx _0) \fw{k|0}{\bx _0}{\bx _k} \big/ \fwmarg{k}{\bx _k}$ yields
\begin{equation}
\label{eq:def_gk}
  \post{k}{\bx _k} \propto \potn{k}{\bx _k} \fwmarg{k}{\bx _k}\eqsp,
  \quad \text{where} \quad
  \potn{k}{\bx _k} \eqdef \int \potn{}{\bx _0} \bw{0|k}{\bx _k}{\bx _0} \, \rmd \bx _0
  \eqsp.
\end{equation}
% where we have defined the backward transition kernel as $\bw{0|k}{\bx _k}{\bx _0} \eqdef \datadistr(\bx _0) \fw{k|0}{\bx _0}{\bx _k} \big/ \fwmarg{k}{\bx _k}$.
It then follows that
% $\nabla \log \post{k}{\bx _k} = \nabla \log \fwmarg{k}{\bx _k} + \nabla \log \potn{k}{\bx _k}$.
\begin{equation*}
  \nabla \log \post{k}{\bx _k} = \nabla \log \fwmarg{k}{\bx _k} + \nabla \log \potn{k}{\bx _k} \eqsp.
\end{equation*}
While we have a parametric approximation of the first term on the \rhs, the bottleneck is the second term, which is intractable due to the integration of $\potn{}{}$ against the conditional distribution $\smash{\bw{0|k}{\bx _k}{\cdot}}$.

\paragraph{Existing approaches.} To circumvent this computational bottleneck, previous works have proposed approximations that involve replacing $\bw{0|k}{\bx _k}{\cdot}$ in \eqref{eq:def_gk} with either a Dirac delta mass \cite{chung2023diffusion} or a Gaussian approximation \cite{ho2022video,song2022pseudoinverse}
\begin{equation}
  \label{eq:guidance-gaussian-approximation}
  \smash{\bw{0|k}{\bx _k}{\bx _0} \approx \normpdf(\bx _0; \hpredx{k}(\bx _k), \var _{0|k} \Id_\dimx)}
  \eqsp,
\end{equation}
where $\hpredx{k}$ is a parametric approximation of $\smash{\predx{k}(\bx _k) \eqdef \int \bx _0 \, \bw{0|k}{\bx _k}{\bx _0} \, \rmd \bx _0}$ and $\var _{0|k}$ is a tuning parameter. Using Tweedie's formula \cite{robbins1956empirical}, it holds that $\predx{k}(\bx _k) = ( - \bx _k + \sqrt{\acp{k}} \, \nabla \log \fwmarg{k}{\bx _k} ) / \var_k$, which suggests the approximation $\hpredx{k}(\bx _k) \eqdef (- \bx _k + \sqrt{\acp{k}} \score{k}(\bx _k)) / \var_k \eqsp.$
As for the variance parameter $\var_{0|k}$, \citet{ho2022video} suggest setting $\var_{0|k} = \var_k / \acp{k}$, while \citet{song2022pseudoinverse} use $\var_{0|k} = \var_k$.
When $\potn{}{}$ is the likelihood of a linear model, \emph{i.e.}, $\potn{}{\bx} \eqdef \normpdf(\obs; \bfA \bx, \stdobs^2 \Id_\dimobs)$, for $\bfA \in \rset^{\dimobs \times \dimx}$ and $\stdobs >0$, it can be exactly integrated against the Gaussian approximation \eqref{eq:guidance-gaussian-approximation}, yielding $\potn{k}{\bx _k} \approx \normpdf(\obs; \bfA \hpredx{k}(\bx _k), \stdobs^2 \Id_\dimobs + \var_{0|k}\bfA \bfA^\intercal)$.
\citet{chung2023diffusion}, on the other hand, use the pointwise approximation, yielding $\potn{k}{\bx _k} \approx \potn{}{\hpredx{k}(\bx _k)}$.
We denote by $\tpotn{k}{}$ the resulting approximation stemming from any of these methods. Approximate samples from the posterior distribution are then drawn by simulating a backward Markov chain $(\bX _k)_{k = n}^0$,  where $\bX _n \sim \bwp{n}{}{}[]$ and then, given $\bX _{k+1}$, $\bX _k$ is obtained via the update
\begin{equation}
  \label{eq:guidance-update}
  \mathbf{\bX}_k \eqdef \tilde{\bX}_k + w_{k+1}(\bX _{k+1}) \nabla \log \tpotn{k+1}{\bX _{k+1}} \eqsp, \quad \text{where} \quad \tilde{\bX}_k \sim \bwp{k|k+1}{\bX _{k+1}}{\cdot} \eqsp,
\end{equation}
and $w_{k+1}$ is a method-dependent weighting function. For instance, \citet{chung2023diffusion} use $w_{k+1}(\bX _{k+1}) = \zeta \stdobs^2 / \|\obs - \bfA \hpredx{k+1}(\bX _{k+1}) \|$, with $\zeta \in [0, 1]$.
We note that the update \eqref{eq:guidance-update} in general involves a vector-Jacobian product, \emph{e.g.}, considering the approximation suggested by \citet{chung2023diffusion}, $\nabla_{\bx _k} \log \tilde{p} _k(\obs|\bx_k) = \nabla_{\bx _k} \hpredx{k}(\bx _k)^\intercal \nabla_{\bx _0} \left. \log \potn{}{\bx _0} \right|_{\bx _0 = \hpredx{k}(\bx _k)}$, which incurs an additional computational cost compared to an unconditional diffusion step.

  \section{The \algo\ algorithm}
  \label{sec:vdps-algo}
  %\note{YJ}{Présenter la méthode sans critiquer les autres.}

% In this section we present our method, \algo, which addresses a key limitation of existing popular DPS-like sampling methods. As we highlight in the next section and also in \Cref{sec:study_appro\bx _DPS}, the approximations $\tilde{g}_k$ of the likelihoods $g_k$ degrade as $k$ increases. To mitigate this issue, our main idea is to introduce a novel Markovian update that replaces \eqref{eq:guidance-update}. This update relies only on estimating $\potn{k}{}$ for small $k$, thereby reducing the impact of larger $k$ on the estimation error.
% To circumvent this issue, our core idea is to move forward to an intermediate timestep $\tmid{k}$, closer to $0$ at which we compute the likelihood approximation, thus allowing us to mitigate the impact of $k$ on this approximation.
In this section we propose a novel scheme for the approximate inference of $\post{}{}$. We start by presenting a midpoint decomposition of the backward transition that allows us to trade adherence to the prior backward dynamics for improved guidance approximation.
% \paragraph{Midpoint decomposition.}
% First, define for all $(j, k) \in \intset{0}{n}^2$ such that $j<k$ the DDPM forward process $\fwtrans{k|j}{\bx _j}{\bx _k} \eqdef \normpdf(\bx _{k}; (\acp{k}/\acp{j})^{1/2} \bx _j, (1 - \acp{k}/\acp{j}) \Id_\dimx)$.
% Furthermore, consider the joint distribution
% \begin{equation}
%     \label{eq:posterior_forward}
%   \post{0:n}{\bx _{0:n}} \eqdef \post{}{\bx _0} \prod_{k = 0}^{n-1} \fwtrans{k+1|k}{\bx _k}{\bx _{k+1}}
% \end{equation}
% on $(\rset^\dimx)^{n+1}$ obtained by initializing the forward process with the posterior \eqref{eq:posterior_def} of interest.

\revision{We preface our description of the \algo\ algorithm with some additional notations.} First, consider the DDPM forward process with transitions given by   
$\fwtrans{k|j}{\bx _j}{\bx _k} \eqdef \normpdf(\bx _{k}; \sqrt{\acp{k}/\acp{j}} \bx _j, (1 - \acp{k}/\acp{j}) \Id_\dimx)$ for all $(j, k) \in \intset{0}{n}^2$ such that $j<k$. We also define the joint law  
\begin{equation}
\label{eq:posterior_forward}
    \post{0:n}{\bx _{0:n}} \eqdef \post{}{\bx _0} \prod_{k = 0}^{n-1} \fwtrans{k+1|k}{\bx _k}{\bx _{k+1}} \eqsp,
\end{equation}
of the forward process when initialized with the posterior  \eqref{eq:posterior_def} of interest. Note that with these definitions, $\post{k}{}$ (given by \eqref{eq:def_gk}) is the marginal of $\post{0:n}{\bx _{0:n}}$ w.r.t. $\bx_k$.  
%Note that the identity $\smash{\fwtrans{k|0}{\bx _0}{\bx _k} = \int \prod_{j = 0}^{k-1} \fwtrans{j+1|j}{\bx _j}{\bx _{j+1}} \, \rmd \bx _{1:k-1}}$ implies that the $k$-th marginal of this distribution is $\post{k}{}$ as defined in \eqref{eq:posterior_forward}.
% We initiate our approach by replacing the initial problem of sampling from \eqref{eq:posterior_def} with that of sampling from t
The time-reversed decomposition of \eqref{eq:posterior_forward} writes
\begin{equation}
    \label{eq:pi-backward}
    \post{0:n}{\bx _{0:n}} \eqdef \post{n}{\bx _n} \prod_{k = 0}^{n-1} \pibw{k|k+1}{\bx _{k+1}}{\bx _{k}} \eqsp,
\end{equation}
\revision{where, more generally, for all $(i, j) \in \intset{0}{n}^2$ such that $i < j$,
\begin{equation}
    \label{eq:pi-backward-transition}
    \pibw{i|j}{\bx _{j}}{\bx _{i}} \eqdef \post{i}{\bx _i}\fwtrans{j|i}{\bx _i}{\bx _{j}} \big/ \post{j}{\bx _{j}} \propto \potn{i}{\bx_i} \fwtrans{i|j}{\bx_j}{\bx_i} \eqsp,
  \end{equation}
  where the marginals are given by \eqref{eq:posterior-forward}, is the conditional density of $\bX_i$ given $\bX_j = \bx_j$ and $\by$.}
%   Moreover, note that $\pibw{i|j}{\bx _{j}}{\bx _{i}} \propto \potn{i}{\bx _i} \fwtrans{i|j}{\bx _j}{\bx _i}$.
\begin{wrapfigure}{r}{5.5cm}
    \centering
    \resizebox{0.4\textwidth}{!}{
    \begin{tikzpicture}[scale=0.4, node distance=3cm, every node/.style={circle, draw, thick, fill=blue!5, minimum size=1.5cm, font=\Large},
        >={Stealth[round]},
        thick]

    % Nodes
    \node (x0) at (0, 0) {$\mathbf{\bX}_0$};
    \node (ylk) [right of=x0] {$\mathbf{\bX}_{\ell_k}$};
    \node (xk) [right of=ylk] {$\mathbf{\bX}_k$};
    \node (xk1) [right=.5cm of xk] {$\mathbf{\bX}_{k+1}$};
    \node (y) [above=1.cm of x0] {$\boldsymbol{y}$};

    \draw[thick, black] (x0) -- (ylk) node[midway, minimum size=1.45cm, fill=white!50, draw=white!80, font=\small, pos=0.5, sloped] {$\cdots$};
    \draw[thick, black] (ylk) -- (xk) node[midway, minimum size=1.45cm, fill=white!50, draw=white!80, font=\small, pos=0.5, sloped] {$\cdots$};
    % \draw[->, thick, black] (ylk) -- (xk);
    % \draw[->, thick, green] (xk) -- (xk1);

    % Dashed edges
    \draw[->,  thick, orange] (x0) -- (y);
    \draw[->,  thick, red] (ylk) -- (y);
    \draw[->,  thick, blue] (xk) -- (y);

    % Red dashed edges
    \draw[->,  thick, orange] (xk1) to[bend right=40] (x0);
    \draw[->,  thick, blue] (xk1) to[bend right=40] (xk);
    \draw[->,  thick, red] (xk1) to[bend right=40] (ylk);
    % \draw[->, dashed, thick, red] (ylk) to[bend right=40] (xk);
    % \draw[->, dashed, thick, green] (x0) to[bend right=40] (xk);
    \end{tikzpicture}
    }
    \captionsetup{font=small}
    \caption{For each color, the different solid arrows indicate different conditional densities that need to be approximated for a given choice of the midpoint $\tmid{k}$. The longer the arrow, the more difficult it is to approximate the corresponding conditional density. By placing the $\tmid{k}$ midway between zero and $k$, the shortest arrows are obtained.}
    \label{fig:illustration}
    \vspace{-.6cm}
        % Schematic overview of different conditionals involved in a  decomposition of $\pibw{k|k+1}{\bX_{k+1}}{\cdot}$. Each arrow indicates a conditional distribution.}
\end{wrapfigure}
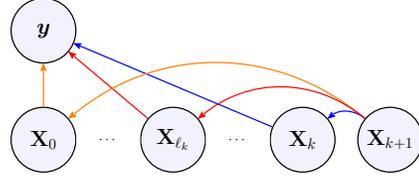
Using the  decomposition \eqref{eq:pi-backward}, an exact draw $\bX _0$ from \eqref{eq:posterior_def} is obtained by first drawing $\bX _n$ from $\post{n}{}$ and then simulating recursively $\bX_k$ from the transitions $\smash{\pibw{k|k+1}{\bX_{k+1}}{\cdot}}$ for all $k$. \revision{However, according to \eqref{eq:pi-backward-transition}, such transitions involve the intractable likelihood $\potn{k}{\bx_k}$ which is hard to approximate accurately when $k$ is large.}

\paragraph{Midpoint decomposition.} \revision{The transition \eqref{eq:pi-backward-transition} has two components: the conditional density $\potn{i}{\bx_i}$ and the prior transition density $\fwtrans{i|j}{\bx_j}{\bx_i}$. The approximation $\potn{i}{\bx_i} \approx \potn{}{\hpredx{i}(\bx_i)}$ of \citet{chung2023diffusion} is accurate when $i$ is small, whereas the Gaussian approximation of $\fwtrans{i|j}{\bx_j}{\bx_i}$ proposed by \citet{ho2020denoising} is accurate when the difference $| i -j |$ is small.
Thus, in order to combine the best of both worlds, we introduce a decomposition that transfers the problem of sampling from $\pibw{k|k+1}{}{}$ to that of sampling from $\pibw{\tmid{k}|k+1}{}{}$, \emph{i.e.}, drawing $\bX_\tmid{k}$ given $\bX_{k+1}$ and $\obs$, where $\tmid{k} < k$. The parameter $\tmid{k}$ balances the approximation errors of the two conditional densities in \eqref{eq:pi-backward-transition}; see \Cref{fig:illustration}.}
In order to make this construction, define, for each $\ell \in \intset{0}{k}$, the bridge kernel $\fwtrans{k|\ell, k+1}{\bx _\ell, \bx _{k+1}}{\bx _k} \propto \fwtrans{k|\ell}{\bx _\ell}{\bx _k}\fwtrans{k+1|k}{\bx _k}{\bx _{k+1}}$, which is Gaussian; see \Cref{sec:ddpm-background}.
By convention, $\fwtrans{k|k, k+1}{\bx _{k}, \bx _{k+1}}{\cdot} = \delta_{\bx _k}$.
\begin{lemma}
    \label{lem:pi_bwker}
    For all $k \in \intset{1}{n-1}$ and $\tmid{k} \in \intset{0}{k}$ it holds that
    \begin{equation}
        \pibw{k|k+1}{\bx _{k+1}}{\bx _k} = \int \fwtrans{k|\tmid{k}, k+1}{\bx _\tmid{k}, \bx _{k+1}}{\bx _k} \pibw{\tmid{k}|k+1}{\bx _{k+1}}{\bx _\tmid{k}} \, \rmd \bx _\tmid{k} \eqsp.
        \label{eq:backward-intermediate-decomp}
    \end{equation}
\end{lemma}
The proof is found in \Cref{apdx:backward-intermediate-decomp}. Lead by the decomposition provided by \Cref{lem:pi_bwker}, we define
\begin{equation}
    \label{eq:pi-transition-approximation}
    \hpibw{k|k+1}{\bx _{k+1}}{\bx _k}[\tmidfn] \eqdef \int \fwtrans{k|\tmid{k}, k+1}{\bx _\tmid{k}, \bx _{k+1}}{\bx _k} \hpibw{\tmid{k}|k+1}{\bx _{k+1}}{\bx _\tmid{k}}[\param] \, \rmd \bx _{\tmid{k}} \eqsp,
    %\eqsp,
\end{equation}
where $\hpibw{\tmid{k}|k+1}{\bx _{k+1}}{\bx _\tmid{k}}[\param] \propto \hpotn{\tmid{k}}{\bx _\tmid{k}} \bwp{\tmid{k}|k+1}{\bx _{k+1}}{\bx _\tmid{k}}$ with $\hpotn{\tmid{k}}{\bx _\tmid{k}} \eqdef \potn{}{\hpredx{\tmid{k}}(\bx _\tmid{k})}$ and $\bwp{\tmid{k}|k+1}{\bx _{k+1}}{\bx_\tmid{k}} \eqdef \fwtrans{\tmid{k}|0, k+1}{\hpredx{k+1}(\bx _{k+1}), \bx _{k+1}}{\bx_\tmid{k}}$. Finally, for any sequence $\ell \eqdef (\tmid{k})_{k = 1} ^n$ satisfying $\tmid{k} \leq k$,  we define our surrogate model for $\post{0:n}{}$ and the posterior \eqref{eq:posterior_def} as 
\begin{equation}
    \label{eq:pi-surrogate}
    \hpibw{0:n}{}{}[\tmidfn](\bx _{0:n}) \eqdef \hpibw{n}{}{}[](\bx _n) \prod_{k = 0}^{n-1} \hpibw{k|k+1}{\bx _{k+1}}{\bx _k}[\tmidfn]  \eqsp,   \quad \hpibw{0}{}{}[\tmidfn](\bx _0) \eqdef \int \hpibw{0:n}{}{}[\tmidfn](\bx _{0:n}) \, \rmd \bx _{1:n} \eqsp,
\end{equation}
where $\hpibw{n}{}{}[](\bx _n) = \normpdf(\bx _n; \zero_\dimx, \Id_\dimx)$ and $\hpibw{0|1}{\bx _1}{\cdot}[\tmidfn] \eqdef \delta_{\hpredx{1}(\bx _1)}$ \cite[similarly to ][]{ho2020denoising,song2021ddim}.
\revision{Since the bridge kernel $\fwtrans{k|\tmid{k}, k+1}{}{}$  is Gaussian, it can be easily sampled. Hence, we only need to focus on the approximate sampling from $\smash{\pibw{\tmid{k}|k+1}{}{}}$. Moreover, since the decomposition \eqref{eq:backward-intermediate-decomp} is valid for any $\tmid{k}$, it can be chosen to better balance 
%in a way that better balances 
the approximation errors, as discussed above.} 

\paragraph{Choice of the sequence $(\tmid{k})_{k = 1} ^{n-1}$.}
The accuracy of the surrogate model \eqref{eq:pi-surrogate} ultimately depends on the design of the sequence $(\tmid{k})_{k = 1} ^{n-1}$.
In fact, for a given $k$, note that a decrease in $\tmid{k}$ ensures that the approximation 
$\hpotn{\tmid{k}}{\bx _\tmid{k}}$
%$\potn{}{\hpredx{\tmid{k}}(\cdot)}$
%\circ \hpredx{\tmid{k}}$ 
of $\potn{\tmid{k}}{\bx _\tmid{k}}$ becomes more accurate. For instance, setting $\tmid{k} = 0$
\begin{wrapfigure}{r}{7.0cm}
    \centering
    \vspace{-0.0cm}
    \includegraphics[width=.5\textwidth]{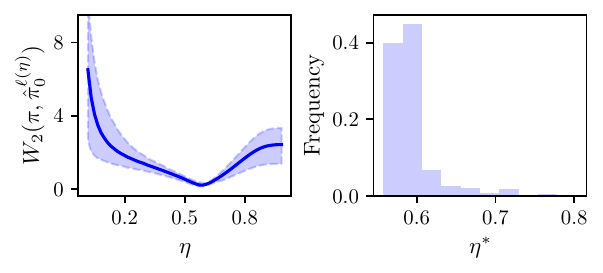}
    \vspace{-0.7cm}
    \captionsetup{font=small}
    \caption{Left: average $W_2$ with $10\%$--$90\%$ quantile range.
    Right: distribution of the minimizing $\eta^*$.}
    \label{fig:tmid-gaussian-prior}
    \vspace{-0.2cm}
\end{wrapfigure}
eliminates any error of the approximate likelihood (since $\potn{0}{} = \potn{}{}$); however, decreasing $\tmid{k}$ can cause the distribution $\smash{\pibw{0|k+1}{\bx_{k+1}}{\cdot}}$ to become strongly multimodal, especially in the early stages of the diffusion process, making the DDPM Gaussian approximation less accurate.
Conversely, setting $\tmid{k}$ closer to $k$, similar to the DPS approach of \citet{chung2023diffusion}, improves the accuracy of the approximation of $\smash{\fwtrans{\tmid{k}|k+1}{}{}}$, but impairs the approximation $\hpotn{\tmid{k}}{\bx _\tmid{k}}$ of $\smash{\potn{\tmid{k}}{\bx _\tmid{k}}}$. 
The choice of $\tmid{k}$ is therefore subject to a particular trade-off, which we illustrate using the following Gaussian toy example. When considering the parameterization $\tmidfn^\eta _{k} = \lfloor \eta k \rfloor$ with $\eta \in [0,1]$, we show that the Wasserstein-2 distance between $\post{}{}$ and $\hpibw{0}{}{}[\tmidfn(\eta)]$ reaches its minimum %\wrt\ $\eta$ 
for $\eta$ around $0.5$,
%$$\eta \approx 0.5$,
which confirms that the additional flexibility obtained by introducing the midpoint can provide a surrogate model \eqref{eq:pi-surrogate} with reduced approximation error.

\begin{example}
\label{example:gaussian}
Consider a linear inverse problem $\bY = \bfA \bX + \sigma_\obs \bZ$ with Gaussian noise and prior $\prior = \normpdf(\boldsymbol{m}, \mathbf{\cov})$, where $\boldsymbol{m} \in \rset^\dimx$ and 
$\mathbf{\cov} \in \rset^{d \times d}$ is positive definite. 
%$\mathbf{\cov} \in \mathcal{S}^{++}_\dimx$.
% \jimmy{Maybe skip boldface for $\bfA$ and $\Id_\dimx$ for consequence?}
% \badr{You are right, but we better Boldface \Sigma instead}
In this setting, we have access to the exact denoiser $\predx{k}{}$ in a closed form, which allows us to take $\hpredx{k} = \predx{k}$.
Moreover, since the transition \eqref{eq:pi-transition-approximation}
%\eqref{eq:backward-surrogate-model} 
admits an explicit expression in this case, we can quantify the approximation error of \eqref{eq:pi-surrogate} {\wrt} $\post{}{}$ for any sequence $(\tmid{k})_{k = 1}^{n-1}$.
Note that in this case the true backward transitions $\fwtrans{\tmid{k}|k+1}{}{}$ are Gaussian and have the same mean as the Gaussian DDPM transitions, but differ in their covariance.
All necessary derivations and computational details can be found in \Cref{apdx:gaussian-example}.
We consider sequences $\tmidfn(\eta) \eqdef (\tmidfn^\eta _k)_{k = 1} ^n$ with $\tmidfn^{\eta}_k = \lfloor \eta k \rfloor$, $\eta \in [0, 1]$, and compute the Wasserstein-2 distance $W_2(\post{}{}, \hpibw{0}{}{}[\tmidfn(\eta)])$ as a function of $\eta$ on the basis of $500$ samples $(\bfA, \boldsymbol{m}, \mathbf{\cov})$ for $\dimx = 100$.
The left panel of \Cref{fig:tmid-gaussian-prior} displays the average error $W_2(\post{}{}, \hpibw{0}{}{}[\tmidfn(\eta)])$ as a function of $\eta$, while the right panel displays the associated distribution of $\eta^* \eqdef argmin _{\eta \in [0, 1]} W_2(\post{}{}, \hpibw{0}{}{}[\tmidfn(\eta)])$.
\Cref{fig:tmid-gaussian-prior} illustrates that the smallest approximation error for the surrogate model \eqref{eq:pi-surrogate} is reached at intermediate values close to $\eta=0.5$ rather than $\eta=1$, which corresponds to a DPS-like approximation.
\end{example}

% As shown by \Cref{example:gaussian},  It now remains to clarify how to sample from $\hpibw{0:n}{}{}[\tmidfn]$.
% I'm in favor of moving algo 1 (pseudo code) to appendix.
% Doing so will free up plenty of space.
% }
% \paragraph{Sampling from $\hpibw{0:n}{}{}[\tmidfn]$ with a variational approximation approach.}
\paragraph{Variational approximation.}
To sample from $\hpibw{0:n}{}{}[\tmidfn]$ in \eqref{eq:pi-surrogate}, 
we focus on simulating approximately and recursively (over $k$) the Markov chain with transition densities \eqref{eq:pi-transition-approximation}.  
%we focus on generating approximate samples from the sequence $(\hpibw{k}{}{}[\tmidfn])_{k = 0}^n$ of marginal distributions of $ \hpibw{0:n}{}{}[\tmidfn] $ in \eqref{eq:pi-transition-approximation}, by backward recursion over $k$. 
Ideally, for $k=n$, we simply sample from $\hpibw{n}{}{}$; then, recursively, assuming that we have access to an approximate sample $\bX _{k+1}$ from $\hpibw{k+1}{}{}[\tmidfn]$, the next state $\bX _k$ is drawn from $\hpibw{k|k+1}{\bX _{k+1}}{\cdot}[\tmidfn]$.
However, as $\hpibw{k|k+1}{\bX _{k+1}}{\cdot}[\tmidfn]$ is intractable, 
%due to its definition \eqref{eq:pi-transition-approximation}, 
we propose using the Gaussian approximation that we specify next.

In the following, let $k \in \intset{1}{n}$ and $\tmid{k} \in \intset{0}{k}$ be fixed. For $\smash{\vparam = (\vmu_{\tmid{k}}, \vlstd_\tmid{k}) \in (\rset^\dimx)^2}$, consider the Gaussian variational distribution
%We consider the Gaussian variational distribution $\smash{\vpibw{k|k+1}{}{}}$ with $\smash{\vparam = (\vmu_{\tmid{k}}, \vlstd_\tmid{k}) \in (\rset^\dimx)^2}$ where for any $\bx _k,\bx _{k+1}\in\rset^d$,
\begin{align}
    \label{eq:variational-bridge}
    \vpibw{k|k+1}{\bx _{k+1}}{\bx _k} & \eqdef \int \fwtrans{k|\tmid{k}, k+1}{\bx _\tmid{k}, \bx _{k+1}}{\bx _k} \vpibwD{\tmid{k}|k+1}{}{\bx _\tmid{k}} \, \rmd \bx _\tmid{k} \eqsp, \\
    \vpibw{\tmid{k}|k+1}{}{}(\bx _\tmid{k}) & \eqdef \normpdf(\bx _\tmid{k}; \vmu_\tmid{k}, \diag(\rme^{2 \vlstd _\tmid{k}})) \eqsp. \label{eq:def:lambda}
\end{align}
In definition \eqref{eq:def:lambda} the exponential function is applied element-wise to the vector $\smash{\vlstd_\tmid{k}}$ and $\smash{\diag(\rme^{\vlstd_\tmid{k}})}$ is the diagonal matrix with diagonal entries $\smash{\rme^{\vlstd_\tmid{k}}}$. Based on the family $\smash{\{ \vpibw{k|k+1}{}{}\, :\, \vparam \in (\rset^\dimx)^2\}}$ and an approximate sample $\bX _{k+1}$ from $\hpibw{k+1}{}{}[\tmidfn]$, we then seek the best fitting parameter $\vparam$ that  minimizes the upper bound on the backward KL divergence
%\begin{equation*}
%    \int
%\kldivergence{\vpibw{k|k+1}{\bx _{k+1}}{\cdot}}{\hpibw{k|k+1}{\bx _{k+1}}{\cdot}[\tmidfn] }  \hpibwD{k+1}{}{\bx _{k+1}}[\tmidfn] \, \rmd \bx _{k+1} \eqsp.
%\end{equation*}
obtained using the data-processing inequality
\begin{equation*}
    \kldivergence{\vpibw{k|k+1}{\bX _{k+1}}{\cdot}}{\hpibw{k|k+1}{\bX _{k+1}}{\cdot}[\tmidfn]} \leq \kldivergence{\vpibw{\tmid{k}|k+1}{}{}}{\hpibw{\tmid{k}|k+1}{\bX _{k+1}}{\cdot}[\param]} =\vcentcolon \mathcal{L}_k (\vparam; \bX _{k+1})
    \eqsp.
\end{equation*}
Here the gradient of the upper bound
%\wrt\ $\vparam$
is given by
\begin{equation*}
    % \label{eq:objective-gradient}
    \nabla_\vparam \mathcal{L}_k (\vparam; \bX _{k+1})  = - \pE \big[ \nabla_\vparam \log \hpotn{\tmid{k}}{\vmu_{\tmid{k}} + \diag(\rme^{\vlstd_\tmid{k}}) \bZ}\big] + \nabla_\vparam \kldivergence{\vpibw{\tmid{k}|k+1}{}{}}{\bwp{\tmid{k}|k+1}{\bX _{k+1}}{\cdot}}
    \eqsp,
\end{equation*}
where $\bZ \sim \gauss(\zero_\dimx, \Id_\dimx)$ is independent of $\bX _{k+1}$.
The first term is dealt with using the  reparameterization trick \cite{kingma:welling:2013} and can be approximated using a Monte Carlo estimator based on a single sample $\bZ_{k+1}$; we denote this estimate by $\smash{\nabla_\vparam \widetilde{\mathcal{L}}_k(\vparam, \bZ_{k+1}; \bX _{k+1})}$.
The second term is the gradient of the KL divergence between two Gaussian distributions and can thus be computed in a closed form.
Similarly to the previous approaches of \citet{ho2022video,chung2023diffusion,song2022pseudoinverse}, our gradient estimator involves a vector-Jacobian product of the denoising network.

\begin{algorithm}[t]
    \caption{\algoname}
    \begin{algorithmic}[1]
       \STATE {\bfseries Input:} $(\tmid{k})_{k = 1} ^n$ with $\tmid{n} = n$ and $\tmid{1} = 1$; number $\ngrad$ of gradient steps.
       %gradient steps $G$, Langevin steps $K$.
       \STATE $\bX _n \sim \normpdf(\zero_\dimx, \Id_\dimx), \, \vX_n \leftarrow \bX _n$
       \FOR{$k = n-1$ {\bfseries to} $1$}
            \STATE $\vmu_{\tmid{k}} \leftarrow \frac{\sqrt{\acp{\tmid{k}}} (1 - \acp{k+1} / \acp{\tmid{k}})}{1 - \acp{k+1}} \hpredx{\tmid{k+1}}(\vX_{\tmid{k+1}}) + \frac{\sqrt{\acp{k+1} / \acp{\tmid{k}}}(1 - \acp{\tmid{k}})}{1 - \acp{k+1}} \bX _{k+1}$ \label{line:mu_init}
            \STATE $ \vlstd_\tmid{k} \leftarrow \frac{1}{2} \log\frac{(1 - \acp{k+1} / \acp{\tmid{k}}) (1 - \acp{\tmid{k}})}{1 - \acp{k+1}}$
            \FOR{$j = 1$ {\bfseries to} $\ngrad$}
            \STATE $\bZ \sim \normpdf(\zero_\dimx, \Id_\dimx)$
            \STATE $(\vmu_\tmid{k}, \vlstd_\tmid{k}) \leftarrow \mathsf{OptimizerStep}(\nabla_\vparam \widetilde{\mathcal{L}}_k(\cdot, \bZ; \bX _{k+1});\, \vmu_\tmid{k}, \vlstd_\tmid{k})$
            \ENDFOR
            \STATE $\bZ_\tmid{k}, \bZ_k \simiid \normpdf(\zero_\dimx, \Id_\dimx)$
            \STATE $\vX_\tmid{k} \leftarrow \vmu_\tmid{k} + \diag(\rme^{\vlstd_\tmid{k}}) \bZ_\tmid{k}$ \label{line:xhat_def}
            \STATE $\bX _{k} \sim \fwtrans{k|\tmid{k}, k+1}{\vX_\tmid{k}, \bX_{k+1}}{\cdot}$ (See \eqref{eq:bridge_kernel})
       \ENDFOR
       \STATE $\bX _0 \leftarrow \hpredx{1}(\bX _1)$
    \end{algorithmic}
    \label{algo:algo}
\end{algorithm}

We now provide a summary of the \algo\ algorithm, whose pseudocode is given in \Cref{algo:algo}. Given $(\tmid{k})_{k = 1}^{n-1}$, \algo\ proceeds by simulating a Markov chain $(\bX _k)_{k = n}^0$ starting from $\bX _n \sim \normpdf(\zero_\dimx, \Id_\dimx)$.
Recursively, given the state $\bX _{k+1}$, the state $\bX _k$ is obtained by
\begin{enumerate}
    \item minimizing $\mathcal{L}_k(\cdot; \bX _{k+1})$ by performing $\ngrad$ stochastic gradient steps using the gradient estimator $\nabla_\vparam \widetilde{\mathcal{L}}_k(\cdot; \bX _{k+1})$, yielding a parameter $\vparam^*_k(\bX _{k+1})$,
    \item sampling $\vX_\tmid{k} \sim \vpibw{\tmid{k}|k+1}{}{}[\vparam^*_k]$, where we drop the dependence on $\bX _{k+1}$ in $\vparam^*_k$, and then $\bX _k \sim \fwtrans{k|\tmid{k}, k+1}{\vX_\tmid{k}, \bX _{k+1}}{\cdot}$.
\end{enumerate}
\begin{remark}
    While conditionally on $\vparam^*_k$ we have that $\smash{\vX_\tmid{k} \sim \vpibwD{\tmid{k}|k+1}{\bX _{k+1}}{}[\vparam^*_k]}$, the actual law of $\vX_\tmid{k}$ given $\bX _{k+1}$ is not a Gaussian distribution as we need to marginalize over that of $\vparam^* _k$ due to the randomness in the stochastic gradients.
\end{remark}

A well-chosen initialization for the Gaussian variational approximation parameters is crucial for achieving accurate fitting with few optimization steps, ensuring the overall runtime of \algo\ remains competitive with existing methods. Given $\hat{\bX}_\tmid{k+1}$ sampled from $\smash{\vpibw{\tmid{k+1}|k+2}{}{}[\vparam^* _{k+1}]}$ at the previous step, we choose the initial parameter $\vparam_k$ so that
\begin{equation}
    \label{eq:init_strategy}
    \vpibw{\tmid{k}|k+1}{}{}[\vparam_k](\bx _k)  = \smash{\fwtrans{\tmid{k}|0, k+1}{\hpredx{\tmid{k+1}}(\vX_\tmid{k+1}), \bX _{k+1}}{\bx _k}}
    \eqsp.
\end{equation}
This initialization is motivated by noting that $\fwtrans{\tmid{k}|0, k+1}{\boldsymbol{m}^\pi _{0|k+1}(\bX _{k+1}), \bX _{k+1}}{\cdot}$, the DDPM approximation of $\pibw{\tmid{k}|k+1}{\bX _{k+1}}{\cdot}$, where $\boldsymbol{m}^\pi _{0|k+1}$ is a supposed denoiser for the posterior, is a reasonable candidate for the initialization. As it is intractable, we replace it with  $\hpredx{\tmid{k+1}}(\vX_\tmid{k+1})$, our best current guess. Finally, in \Cref{sec:warm_start} we devise a warm-start approach that improves the initialization during the early steps of the algorithm. This has proved advantageous in the most challenging problems.

\paragraph{Related works.} We now discuss existing works that have similarities with our algorithm, \algo.
%We postpone the discussion of other existing approaches to solve the problem at hand to \Cref{sec:litt_review}.
In essence, our method tries to reduce the approximation error incurred by DPS-like approximations.
This has also been the focus of many works in the literature, which we review below.

When $\potn{}{}$ is a likelihood associated to a linear inverse problem with Gaussian noise, \citet{finzi2023user,stevens2023removing,boys2023tweedie} leverage the fact that the Gaussian projection of $\smash{\fwtrans{0|k}{\bx _k}{\cdot}}$ minimizing the forward KL divergence
%is $\smash{ \normpdf(\predx{k}(\bx _k), \Sigma_{0|k}(\bx _k))}$ where $\smash{\Sigma_{0|k}(\bx _k)}$ is the second moment computed using $\fwtrans{0|k}{\bx _k}{\cdot}$ and
can be approximated using the pre-trained denoisers; see \citet[Theorem 1]{meng2021estimating}.
In the considered Gaussian setting, the likelihood $\potn{}{}$ can be exactly integrated against the estimated Gaussian projection.
However, this involves a matrix inversion that may be prohibitively expensive.
\citet{boys2023tweedie} circumvent the latter by using a diagonal approximation which involves the computation of $\dimx$ vector-Jacobian products.
For general likelihoods $\potn{}{}$, \citet{song2023loss} use the Gaussian approximations of \citet{ho2022video,song2022pseudoinverse} to estimate $\potn{k}{}$ for non-linear likelihoods $\potn{}{}$ using a vanilla Monte Carlo estimate.
\citet{zhang2023towards} also consider the surrogate transitions \eqref{eq:pi-transition-approximation} with $\tmid{k} = k$, and, given an approximate sample $\bX _{k+1}$ from $\hat{\pi}^{\ell}_{k+1}$, estimate the next sample $\bX _k$ maximizing $\bx _{k}\mapsto \hpibw{{k}|k+1}{\bX _{k+1}}{\bx _{k}}[\param]$ using gradient ascent.
They also consider improved estimates of the likelihood $\potn{k}{}$ running a few diffusion steps.
However, it is shown \cite[last subtable in Table 2]{zhang2023towards} that this brings minor improvements at the expense of a sharp increase in computational cost, which is mainly due to backpropagation over the denoisers. \revision{See also \Cref{apdx:dps-link} for a discussion on how our method relates to the \dps\ \cite{chung2023diffusion} in the case where $\tmid{k} = k$.}
%at the cost of large increase in the computational cost stemming essentially from the backpropagation over the denoisers.

% In contrast to these methods that aim to improve to reduce the approximations errors by improving the covariance approximation, \citet{janati2024divide} leverage the fact that transitions $\smash{\fwtrans{\ell|k}{\bx _k}{\cdot}}$ with $\ell > 0$ are provably easier to approximate than $\smash{\fwtrans{0|k}{\bx _k}{\cdot}}$ \cite[Proposition (4.2)]{janati2024divide}.
% They introduce intermediate and user-defined likelihoods at a sparse number of timesteps $(\tau_k)_{k = 0} ^L \subset \intset{0}{n}$ with $\tau_0 = 0$, and propose a method that only requires the approximation of the integrals $\int \potn{\tau_k}{\bx _{\tau_k}} \fwtrans{\tau_k | j}{\bx _j}{\bx _{\tau_k}} \, \rmd \bx _{\tau_k}$ for $j \in \intset{\tau_k + 1}{\tau_{k+1}}$.
% These problems can be solved efficiently as long as the length of each subinterval is small.
% Our work share the same principles of this method which aim to reduce the approximation error by shrinking the interval over which the approximation is performed.

Finally, a second line of work considers the distribution path $(\hat{\pi}_k)_{k = n} ^0$, where $\hat\pi _k(\bx _k) \propto \potn{}{\predx{k}(\bx _k)} \fwmarg{k}{\bx _k}$ for $k \in \intset{0}{n-1}$ and $\hat\pi_n = \normpdf(\zero_\dimx, \Id_\dimx)$.
This path bridges the Gaussian distribution and the posterior of interest.
Furthermore, if one is able to accurately sample from $\hat\pi_{k+1}$, then these samples can be used to initialize a sampler targeting the next distribution $\hat\pi_k$. As these distributions are expected to be close, the sampling from $\hat\pi_k$ can also be expected to be accurate. Repeating this process yields approximate samples from $\post{}{}$ \emph{regardless} of the approximation error in the likelihoods.
This approach is pursued by \citet{rozet2023score},  who combines the update \eqref{eq:guidance-update} with a Langevin dynamics targeting $\hat\pi_k$.
% \badr{I think we are good here, ne need to open the discussion about SMC methods what do you think ?}
\citet{wu2023practical} use instead sequential Monte Carlo \cite{gordon-salmond-smith} to recursively build empirical approximations of each $\hat\pi_k$ by evolving a sample of $N$ particles.
  \section{Experiments}
  
\begin{figure}
    \center
    \includegraphics[width=1\textwidth]{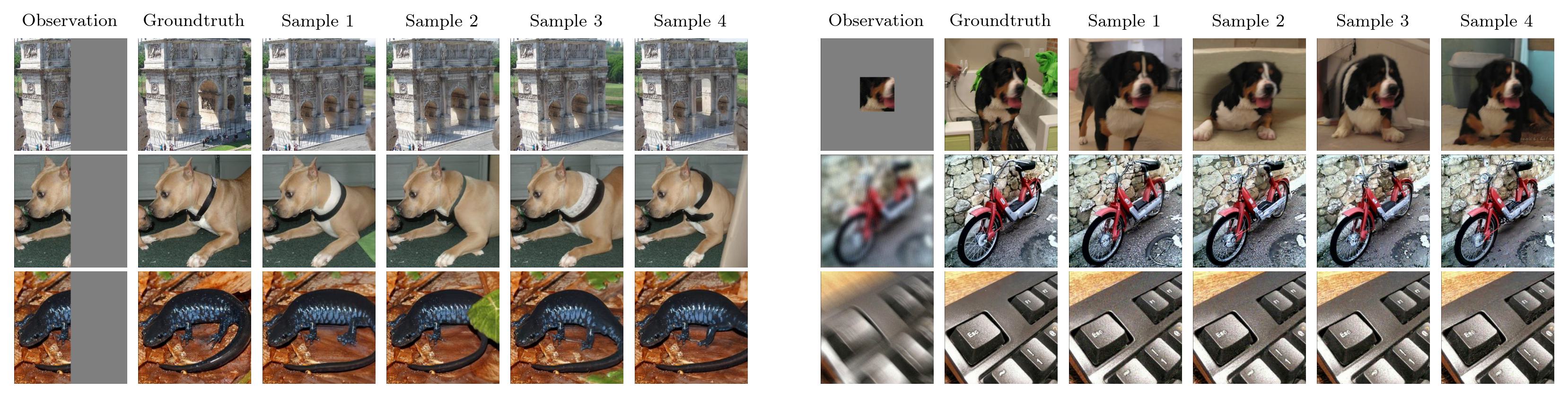}
    \caption{\algo\ sample images for half mask (left), expand task, Gaussian blur and motion blur (right) on the \imagenet\ dataset.}
    \label{fig:ddm_samples}
\end{figure}

We now evaluate our algorithm on three different problems and compare it with several competitors.
We begin in benchmarking our method on toy Gaussian-mixture targets and image experiments with both pixel space and latent diffusion models. We review the latter in \Cref{subsec:ldms}.
Finally, we perform inpainting experiments on ECG data.
% \begin{wraptable}{r}{7.2cm}
%     \centering
%     \vspace{-0.2cm}
%     \caption{XXX.}
%     \vspace{-0.2cm}
%     \begin{tabular}{lcc}
%         \toprule
%         &$\dimx=20, \dimobs=1$ & $\dimx=200, \dimobs=1$ \\
%         \midrule
%         \algo     & $2.91 \pm 0.74$ & $4.04 \pm 1.00$ \\
%         \dps     & $XX \pm XX$ & $XX \pm XX$ \\
%         \pgdm     & $XX \pm XX$ & $XX \pm XX$ \\
%         \diffpir & $XX \pm XX$ & $XX \pm XX$ \\
%         \ddnm  & $XX \pm XX$ & $XX \pm XX$ \\
%         \reddiff  & $XX \pm XX$ & $XX \pm XX$ \\
%         \bottomrule
%         \end{tabular}
%     \captionsetup{font=small}
%     \label{table:sw-gm}
%     \vspace{-0.3cm}
% \end{wraptable}
For experiments based on pixel-space diffusion models, \algo\ is benchmarked against state-of-the art methods in the literature: \dps\ \cite{chung2023diffusion}, \pgdm\ \cite{song2022pseudoinverse}, \ddnm\ \cite{wang2023zeroshot}, \diffpir\ \cite{zhu2023denoising}, and \reddiff\ \cite{mardani2024a}.
Regarding experiments using latent diffusion models, we compare against \psld\ \cite{rout2024solving} and \resample\ \cite{song2024solving}. Full details on the hyperparameters can be found in \Cref{sec:implem_details}. \revision{Our code for reproducing all the experiments is publicly available.\footnote{Code available at \url{https://github.com/yazidjanati/mgps}}}
% We discuss all the methods we consider in \Cref{sec:litt_review}. In all experiments, we use $\tmid{k} = \lfloor 0.5 k \rfloor$ and Adam \cite{kingma2014adam} with a learning rate of $3 \cdot 10^{-2}$ as optimizer in \Cref{algo:algo}.
% However, we vary the number of gradient steps depending on the difficulty of the dataset; see \Cref{sec:implem_details} for more details.
% Finally, for the imaging and ECG data experiments we use as initialization the warm start strategy described in \Cref{sec:warm_start}.

\begin{figure}[t]
    \begin{minipage}[t]{0.50\textwidth}
        \centering
        \includegraphics[width=.97\textwidth]{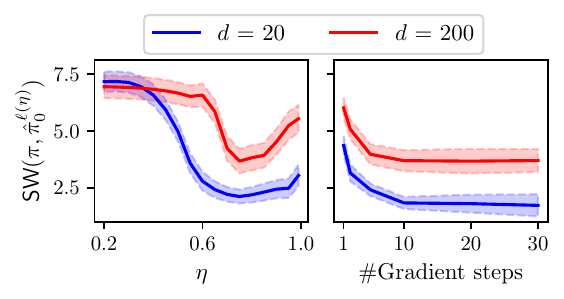}
        \vspace{-.3cm}
        \captionsetup{font=small}
        \captionof{figure}{Left: SW as a function of $\eta$ with $\tmid{k} = \lfloor \eta k \rfloor$. Right: SW as a function of the number of gradient steps, for a specific choice of $(\tmid{k})_k$. }
        \label{fig:ablation-gm-gradient-steps-eta}
    \end{minipage}
\raisebox{95pt}[0pt][0pt]{
    \begin{minipage}[t]{0.5\textwidth}
        \centering
        \captionsetup{font=small}
        \captionof{table}{95 \% confidence interval for the SW on the GM experiment.}
    \resizebox{.9\textwidth}{!}{
        \begin{tabular}{lcc}
        \toprule
        &$\dimx=20, \dimobs=1$ & $\dimx=200, \dimobs=1$ \\
        \midrule
        \algo     & ${\bf{2.11}} \pm 0.30$ & ${\bf{3.66}} \pm 0.53$ \\
        \dps     & $8.93 \pm 0.49$ & $9.15 \pm 0.44$ \\
        \pgdm     & $\underline{2.44} \pm 0.36$ & $\underline{5.23} \pm 0.38$ \\
        \ddnm & $4.24 \pm 0.37$ & $7.10 \pm 0.50$ \\
        \diffpir  & $4.14 \pm 0.42$ & $8.43 \pm 0.92$ \\
        \reddiff  & $6.70 \pm 0.45$ & $8.35 \pm 0.39$ \\
        \bottomrule
        \end{tabular}
        \label{table:sw-gm}
    }
    \end{minipage}
}
\end{figure}

\paragraph*{Gaussian mixture.}
% \begin{wrapfigure}{r}{7.0cm}
%     \centering
%     \vspace{-0.2cm}
%     \includegraphics[width=.5\textwidth]{figures/ablation--.pdf}
%     \vspace{-0.7cm}
%     \captionsetup{font=small}
%     \caption{Right:XXX. Left:XXX}
%     \label{fig:ablation-gm-gradient-steps-eta}
%     \vspace{-0.3cm}
% \end{wrapfigure}
We first evaluate the accuracy of our method and the impact of the hyperparameters on a toy linear inverse problem with Gaussian mixture (GM) as $\prior$ and a Gaussian likelihood $\potn{}{\bx} \eqdef \normpdf(\obs; \bfA \bx, \stdobs^2 \Id_\dimx)$, where $\bfA \in \rset^{\dimobs\times \dimx}$ and $\stdobs = 0.05$.  We repeat the experiment of \citet[App B.3]{cardoso2023monte},
where the prior is a GM with 25 well-separated strata.
%where the prior is a 25 component GM with well seperated modes.
% See Appendix XXX for more details about the component means, covariance and weights.
For this specific example, both the posterior and the denoiser $\predx{k}$ are available in a closed form. In particular, the posterior can be shown to be a GM. The full details are provided in \Cref{sec:gm-appendix}. We consider the settings $(20, 1)$ and $(200, 1)$ for the dimensions $(\dimx, \dimobs)$. For each method, we reconstruct $1000$ samples and then compute the sliced Wasserstein (SW) distance to exact samples from the posterior distribution. We repeat this procedure $100$ times with randomly drawn matrices $\bfA$, and consider the resulting averaged SW distance.
In every replication, the observation is generated as $\bY = \bfA \bX + \stdobs \bZ$ where $\bZ \sim \normpdf(\zero_\dimobs, \Id_\dimobs)$ and $\bX \sim \prior$. The results are reported in \Cref{table:sw-gm}.
It can be observed that \algo\ with $\eta=0.75$ outperforms all baselines in both settings. Although we have tuned the parameters of \dps\, it still exhibits considerable instability and often diverges. To account for these instabilities we set the SW to $10$ when it diverges. %This explains the significantly high SW.

\emph{Ablations.} \quad
Next, we perform an ablation study on the total number of gradient steps and the choice of the sequence $(\tmid{k})_{k = 1}^{n-1}$. The right-hand plot in \Cref{fig:ablation-gm-gradient-steps-eta} shows the average SW distance as a function of the number of gradient steps per denoising step when using $\tmid{k} = \lfloor 3k/4 \rfloor$.
For $\dimx = 20$, we observe a monotonic decrease of the SW distance. For $\dimx = 200$, the SW distance decreases and then stabilizes after 10 gradient steps, indicating that in this case, \algo\ performs well without requiring many additional gradient steps.
We also report the average SW as a function of $\eta \in [0,1]$ in the left-hand plot of \Cref{fig:ablation-gm-gradient-steps-eta}, following the same sequence choices as in \Cref{example:gaussian}, \emph{i.e.}, $\tmidfn^\eta_k = \lfloor \eta k \rfloor$. In both settings, the best SW is achieved at $\eta = 0.75$. In dimension $d = 200$, the SW at $\eta = 0.75$ is nearly twice as good as at $\eta = 1$, which corresponds to a DPS-like approximation. This demonstrates that the introduced trade-off leads to non-negligeable performance gains.

% \begin{wrapfigure}{r}{7.0cm}
%     \centering
%     \vspace{-0.2cm}
%     \includegraphics[width=.5\textwidth]{figures/runtime-gpu-ffhq-ldm.pdf}
%     \vspace{-0.7cm}
%     \captionsetup{font=small}
%     \caption{The runtime and GPU memory usage on \ffhq\ dataset with latent diffusion models.}
%     \label{fig:runtime-ldm}
%     \vspace{-0.3cm}
% \end{wrapfigure}

% For the gradient steps, we plot in \Cref{fig:ablation-gm-gradient-steps} the average SW distance as a function of the number of gradient steps per denoising step. It shows that the SW distance decreases monotonically as we increase the number of gradient steps. As for the impact of $(\tmid{k})_{k = 1}^{n-1}$, we consider the same choice of sequences as in \Cref{example:gaussian} i.e., $\tmidfn^\eta _k = \lfloor \eta k \rfloor$ and plot the average SW as a function of $\eta \in [0,1]$. The results are displayed in Figure XXX. We observe a trade-off similar to \Cref{example:gaussian}.
% We randomly create a set of $100$ problem instances $(\obs, \bfA, \prior)$ with prior having 25 components and then perform posterior sampling with different choices of $\tmidfn$ before averaging the results across the problem instance.

\paragraph*{Images.}
We evaluate our algorithm on a wide range of linear and nonlinear inverse problems with noise level $\stdobs = 0.05$.
As linear problems we consider: image inpainting with a box mask of shape $150 \times 150$ covering the center of the image and half mask covering its right-hand side; image Super Resolution (SR) with factors $\times 4$ and $\times 16$; image deblurring with Gaussian blur; motion blur with $61 \times 61$ kernel size.
We use the same experimental setup as \citet[Section 4]{chung2023diffusion} for the last two tasks.
Regarding the nonlinear inverse problems on which we benchmark our algorithm, we consider: JPEG dequantization with quality factor $2\%$; phase retrieval with oversampling factor $2$; non-uniform deblurring emulated via the forward model of \citet{tran2021non-uniform-deblurring}; high dynamic range (HDR) following \citet[Section 5.2]{mardani2024a}.
Since the JPEG operator is not differentiable, we instead use the differentiable JPEG framework from \citet{shin2017jpeg}. We note that only \dps\ and \reddiff\ apply to nonlinear problems.

%\emph{Datasets and models.}\quad
%We test our algorithm on the $256 \times 256$ versions of the \ffhq \ \cite{karras2019ffhq} and \imagenet \ \cite{deng2009imagenet} datasets.
%We chose randomly $50$ images from each to be used in the evaluation of the algorithms.
%We leverage pre-trained DDMs that are publicly available.
%\textcolor{red}{TODO}
%For \ffhq, we use the pixel-space DDM of \citet{choi2021ilvr} and the LDM of \citet{rombach2022high} with VQ4 autoencoder.
%We use the model by \citet{dhariwal2021diffusion} for \imagenet.
\emph{Datasets and evaluation.}\quad
We test our algorithm on the $256 \times 256$ versions of the \ffhq \ \cite{karras2019ffhq} and \imagenet \ \cite{deng2009imagenet} datasets using publicly available pre-trained DDMs. For \ffhq, we use the pixel-space DDM of \citet{choi2021ilvr} and the LDM of \citet{rombach2022high} with the VQ4 autoencoder. For \imagenet, we use the model by \citet{dhariwal2021diffusion}. \revision{
Due to computational constraints, we first evaluate \algo\ and competitors on all tasks using 50 random images per dataset. We then focus on the 5 most challenging tasks, evaluating \algo\ and the top 2 competitors on 1k images in \Cref{table:extended-metrics-1k}.
%\emph{Evaluation.}\
For all tasks, we report the LPIPS \cite{zhang2018lpips} between the reference image and the reconstruction, averaged over 50 images. For the five tasks tested on 1k images, we also report the FID in \Cref{table:extended-metrics-1k}. Although pixel-wise metrics like PSNR and SSIM are less informative for multimodal posterior distributions, they are available in \Cref{table:extended-metrics-1k,table:extended-metrics-ffhq,table:extended-metrics-imagenet,table:extended-metrics-ffhq-ldm}.}
%We use LPIPS \cite{zhang2018lpips} as the metric to evaluate the quality of the reconstructions. Given that the posterior distribution is expected to be highly multimodal for most of the tasks, pixel-wise comparisons with the ground truth, such as PSNR and SSIM, provide limited insight into reconstruction quality. \revision{We nonetheless provide them in the extended tables \Cref{table:extended-metrics-ffhq}, \ref{table:extended-metrics-imagenet} and \ref{table:extended-metrics-ffhq-ldm} in the appendix.}
%For each inverse problem, we compute the LPIPS between the ground-truth image and a reconstruction and then average over the dataset.
For all the considered competitors, we use the hyperparameters proposed in their official implementations if available, otherwise we manually tune them to achieve the best reconstructions (implementation details in \Cref{sec:implem_details}).
For phase retrieval, due to the task’s inherent instability, we follow the approach of \citet{chung2023diffusion} by selecting the best reconstruction out of four; further discussion on this can be found in \citet[Appendix C.6]{chung2023diffusion}.
% To perform the experiments, we use two GPUs NVIDIA L40S having each 48 GB of memory.

\emph{Results.} \quad We report the results in \Cref{table:combined-lpips} for \ffhq\ and \imagenet\ with DDM prior, and \Cref{table:lpips-ffhq-ldm} for \ffhq\ with LDM prior.
These reveal that \algo\ consistently outperforms the competing algorithms across both linear and nonlinear problems. Notably, on nonlinear tasks, \algo\ achieves up to a twofold reduction in LPIPS compared to the other methods with DDM prior. It effectively manages the additional nonlinearities introduced by using LDMs and surpasses the state-of-the-art, as demonstrated in \Cref{table:lpips-ffhq-ldm}. Reconstructions obtained with \algo\ are displayed in \Cref{fig:ddm_samples} and \Cref{fig:ldm_samples}. Further examples and comparisons with other competitors are provided in \Cref{subsec:sample-images}. It is seen that \algo\ provides high quality reconstructions even for the most challenging examples. For the DDM prior, the results are obtained using the warm-start approach, see \Cref{algo:algo2}, and setting $\tmid{k} = \lfloor k / 2 \rfloor$ on all the tasks based on the \ffhq\  dataset. As for the \imagenet\ dataset, we use the same configuration on all the tasks, except for Gaussian deblur and motion deblur. For these tasks we found that using $\tmid{k} = \lfloor k/2 \rfloor \indic_{k \geq \lfloor n/2 \rfloor} + k \indic_{k < \lfloor n/2 \rfloor}$ improves  performance. We use a similar strategy for the LDM prior. In \Cref{sec:runtime-memory}, we measure the runtime and GPU memory requirements for each algorithm. The memory requirement of our algorithm is the same as that of \dps\ and \pgdm, \revision{but the runtime of \algo\ with $n=300$ is slightly larger.} With fewer diffusion steps, the runtime is halved compared to \dps\ and \pgdm\ and comparable to that of \diffpir, \ddnm\ and \reddiff. Still, \algo\ remains competitive and consistently outperforms the baselines, particularly on nonlinear tasks. See \Cref{table:extended-metrics-ffhq} and \ref{table:extended-metrics-imagenet} for detailed results using $n \in \{50, 100\}$ diffusion steps.
\begin{table}[t]
    \centering
    \caption{Mean LPIPS for various linear and nonlinear imaging tasks on the \ffhq\ and \imagenet\  $256 \times 256$ datasets with $\stdobs = 0.05$.}
    \resizebox{\textwidth}{!}{
    \begin{tabular}{lcccccc|cccccc}
        \toprule
        & \multicolumn{6}{c}{\bf{\ffhq}} & \multicolumn{6}{c}{\bf{\imagenet}} \\
        \cmidrule(lr){2-7} \cmidrule(lr){8-13}
        \textbf{Task} & \algo\ & \dps & \pgdm & \ddnm & \diffpir & \reddiff & \algo\ & \dps & \pgdm & \ddnm & \diffpir & \reddiff \\
        \midrule
        SR ($\times 4$)        & \bf{0.09} & \bf{0.09} & 0.33 & 0.14 & \underline{0.13} & 0.36 & \bf{0.30} & 0.41 & 0.78 & \underline{0.34} & 0.36 & 0.56 \\
        SR ($\times 16$)       & \underline{0.26} & \bf{0.24} & 0.44 & 0.30 & 0.28 & 0.51 & \underline{0.53} & \bf{0.50} & 0.60 & 0.70 & 0.63 & 0.83 \\
        Box inpainting         & \bf{0.10} & 0.19 & 0.17 & \underline{0.12} & 0.18 & 0.19 & \bf{0.22} & 0.34 & 0.29 & \underline{0.28} & \underline{0.28} & 0.36 \\
        Half mask              & \bf{0.20} & 0.24 & 0.26 & \underline{0.22} & 0.23 & 0.28 & \bf{0.29} & 0.44 & 0.38 & 0.38 & \underline{0.35} & 0.44 \\
        Gaussian Deblur        & \underline{0.15} & 0.16 & 0.87 & 0.19 & \bf{0.12} & 0.23 & \underline{0.32} & 0.35 & 1.00 & 0.45 & \bf{0.29} & 0.54 \\
        Motion Deblur          & \bf{0.13} & 0.16 & $-$ & $-$ & $-$ & 0.21 & \bf{0.22} & 0.39 & $-$ & $-$ & $-$ & 0.40 \\
        \cmidrule(lr){1-13}
        JPEG (QF = 2)          & \bf{0.16} & 0.39 & 1.10 & $-$ & $-$ & \underline{0.32} & \bf{0.42} & 0.63 & 1.31 & $-$ & $-$ & \underline{0.51} \\
        Phase retrieval        & \bf{0.11} & 0.46 & $-$ & $-$ & $-$ & \underline{0.25} & \bf{0.47} & 0.62 & $-$ & $-$ & $-$ & \underline{0.60} \\
        Nonlinear deblur       & \bf{0.23} & \underline{0.52} & $-$ & $-$ & $-$ & 0.66 & \bf{0.44} & 0.88 & $-$ & $-$ & $-$ & \underline{0.67} \\
        High dynamic range     & \bf{0.07} & 0.49 & $-$ & $-$ & $-$ & \underline{0.20} & \bf{0.10} & 0.85 & $-$ & $-$ & $-$ & \underline{0.21} \\
        \bottomrule
    \end{tabular}
    }
    \label{table:combined-lpips}
\end{table}

% \emph{Runtime and Memory.}\quad \yazid{put a single phrase for this}
% For each dataset, we compute an average across the different tasks of the runtime and GPU memory consumption of the considered algorithm.
% We note that the time and memory cost of \algo\ is competitive with the other algorithms while providing good reconstruction as mentioned in \emph{Evaluation}; refer to \Cref{sec:runtime-memory} for the description and the results of this experiment.
\begin{minipage}[t]{0.60\textwidth}
    \centering
    \includegraphics[width=1.\textwidth]{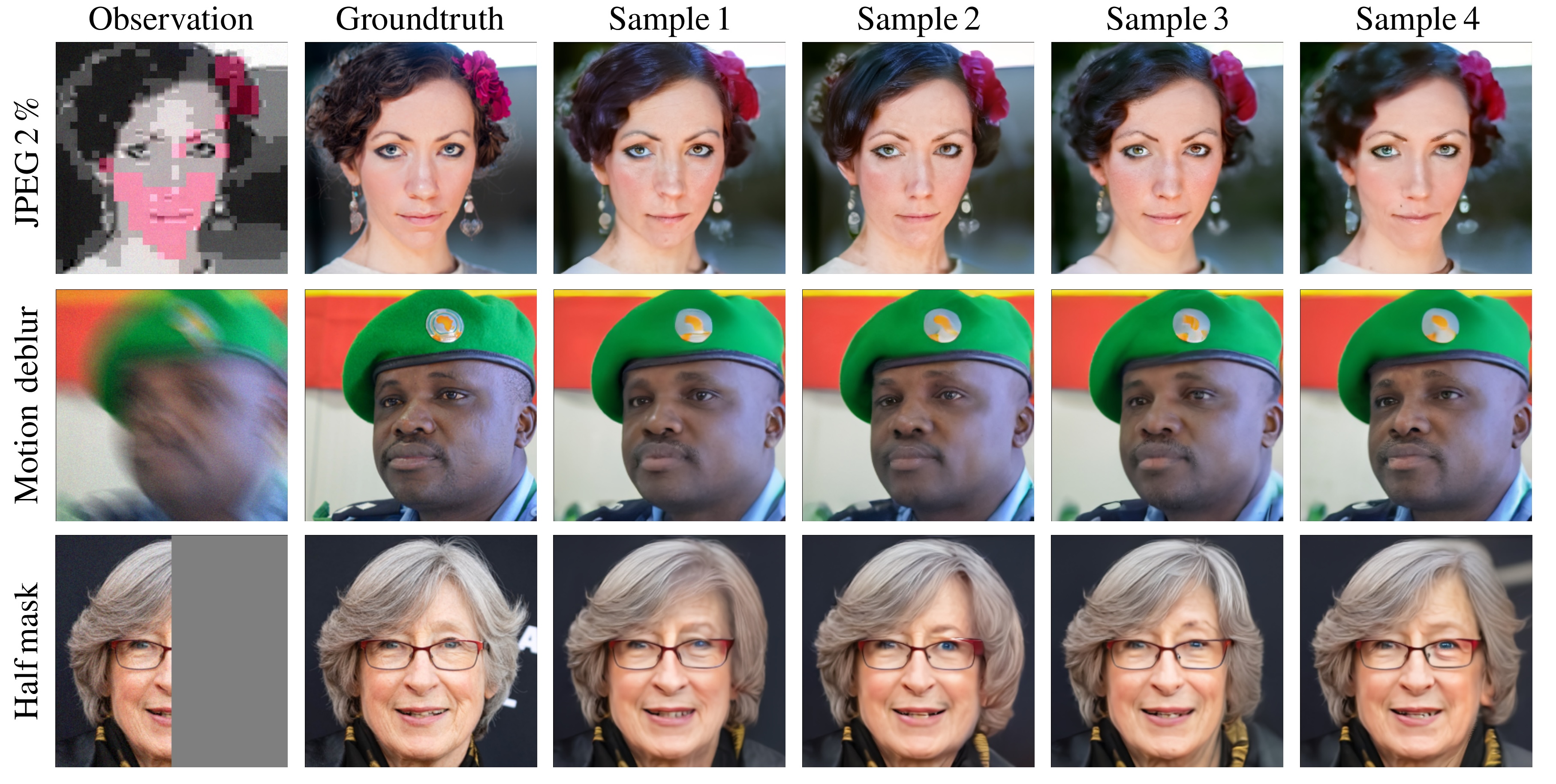}
    \centering
    \captionsetup{font=small}
    \captionof{figure}{\algo\ samples with LDM on \ffhq\ dataset.}
    \label{fig:ldm_samples}
\end{minipage}
% \hspace{0.02\textwidth}
\raisebox{55pt}[0pt][0pt]{
\begin{minipage}{0.38\textwidth}
    \centering
    \vspace{0pt}
    \captionsetup{font=small}
    \captionof{table}{Mean LPIPS with LDM on \ffhq.}
    \resizebox{1.\textwidth}{!}{
        \begin{tabular}{lccccc}
            \toprule
            Task & \algo\ & \resample & \psld \\
            \midrule
            SR ($\times 4$) & \bf{0.11} & \underline{0.20} &  0.22 \\
            SR ($\times 16$) & \bf{0.30} & 0.36  &  \underline{0.35} \\
            Box inpainting & \bf{0.16} & \underline{0.22} &  0.26 \\
            Half mask & \bf{0.25} & \underline{0.30} & 0.31 \\
            Gaussian Deblur & \underline{0.16} & \bf{0.15} & 0.35  \\
            Motion Deblur & \bf{0.18} & \underline{0.19} & 0.41  \\
            \cmidrule(lr){1-4}
            JPEG (QF = 2) & \bf{0.20} & 0.26 & $-$  \\
            Phase retrieval & \bf{0.34} & 0.41 &  $-$  \\
            Nonlinear deblur & \bf{0.26} & 0.30 & $-$  \\
            High dynamic range & \bf{0.15} & \bf{0.15} & $-$ \\
            \bottomrule
        \end{tabular}
        \label{table:lpips-ffhq-ldm}
        }
\end{minipage}
}
\paragraph{ECG.} We now explore posterior sampling algorithms beyond image tasks to show their versatility and public health impact. Cardiovascular diseases cause one-third of global deaths, and better detection can improve management. Wearables like smartwatches can enhance diagnosis by capturing brief symptom episodes, particularly for conditions like paroxysmal atrial fibrillation (AF), which may go undetected during routine medical visits. However, they offer only a partial ECG view (Lead I instead of 12 leads) and a recent study showed the Apple Watch detected AF in only 34 of 90 episodes \cite{AppleWatchAF}.
To address this limitation, we propose using posterior sampling algorithms to reconstruct incomplete electrocardiograms (ECG). An ECG is an electrical recording of the heart's activity in which the signals generated by the heartbeats are recorded in order to diagnose various cardiac conditions such as cardiac arrhythmias and conduction abnormalities.
Unlike static images, ECGs constitute complex time-series data consisting of 12 electrical signals acquired using 10 electrodes, 4 of which are attached to the limbs and record the 'limb leads', while the others record the 'precordial leads' around the heart.
%Unlike static images, ECGs are complex time-series data consisting of 12 electrical signals acquired using 10 electrodes, 4 of them placed on the limbs capture the ‘limb leads’, and the other around the hearts capture the ‘precordial leads’.
We study two conditional generation problems in ECGs. The first is a forecasting or missing-block (MB) reconstruction problem, where one half of the ECG is reconstructed from the other (see \cref{fig:MB-ecg}). This task evaluates the algorithm's ability to capture temporal information to predict a coherent signal. The second problem is an inpainting or missing-leads (ML) reconstruction, where the entire ECG is reconstructed from the lead I (see \cref{fig:leadI_NSR}). The question is whether we can capture the subtle information contained in lead I and reconstruct a coherent ECG with the same diagnosis as the real ECG. This task is challenging because lead I, being acquired with limb electrodes far from the heart, may contain very subtle features related to specific cardiac conditions.
%Following the architecture proposed in \cite{alcarazImputation,alcaraz2023diffusion},
We train a state-space diffusion model \cite{goel2022sashimi} to generate ECGs using the 20k training ECGs from the \textsc{ptb-xl} dataset~\cite{ptbxl}, and benchmark the posterior sampling algorithm on the 2k test ECGs (see \cref{sec:ecg-appendix}). We report the Mean Absolute Error (MAE) and the Root Mean Squared Error (RMSE) between ground-truth and reconstructions in \Cref{table:rmse-ecg}. We demonstrate that a diffusion model trained to generate ECGs can serve as a prior to solve imputation tasks without additional training or fine-tuning, yielding superior results to a diffusion model trained conditionally on the MB task~\cite{alcarazImputation}. The rationale for this result is discussed in \Cref{sec:ecg-discussion}.
We report in \Cref{table:bacc-ecg} the balanced accuracy of diagnosing Left Bundle Branch Block (LBBB), Right Bundle Branch Block (RBBB), Atrial Fibrillation (AF), and Sinus Bradycardia (SB) using the downstream classifier proposed in~\cite{Strodthoff2020ecgbenchmarking} (see \cref{sec:ecg-metrics}), applied to both ground-truth and to reconstructed samples from lead I.
See \cref{sec:samples-ecg} for sample figures. \algo\ outperforms all other posterior sampling algorithms with just 50 diffusion steps.

\begin{table}[ht]
    \centering
    \caption{\small MAE and RMSE for missing block task on the \textsc{ptb-xl} dataset.}
    \resizebox{\textwidth}{!}{
    \begin{tabular}{lcccccccc}
        \toprule
        Metric & \algo\ & \dps & \pgdm & \ddnm & \diffpir & \reddiff & \traineddiff \\
        \midrule
       % RBM & $\mathbf{0.130\pm4\mathrm{e}\!{-3}}$ & TODO & TODO & TODO & TODO & TODO & $\mathbf{0.131\pm3\mathrm{e}\!{-3}}$\\
        MAE & $\underline{0.111\pm2\mathrm{e}\!{-3}}$ &$0.117\pm4\mathrm{e}\!{-3}$ & $0.118\pm2\mathrm{e}\!{-3}$ & $\mathbf{0.103\pm2\mathrm{e}\!{-3}}$ & $0.115\pm2\mathrm{e}\!{-3}$ & $0.171\pm3\mathrm{e}\!{-3}$ & $0.116\pm2\mathrm{e}\!{-3}$\\
        RMSE & $\mathbf{0.225\pm4\mathrm{e}\!{-3}}$ &$0.232\pm4\mathrm{e}\!{-3}$ & $0.233\pm4\mathrm{e}\!{-3}$ & $\mathbf{0.224\pm4\mathrm{e}\!{-3}}$ & $0.233\pm4\mathrm{e}\!{-3}$ & $0.287\pm5\mathrm{e}\!{-3}$ & $0.266\pm3\mathrm{e}\!{-3}$\\
%        ML & $0.195\pm5\mathrm{e}\!{-3}$ & $1.90 \pm0.5$ & $0.191\pm5\mathrm{e}\!{-3}$ & $0.193\pm5\mathrm{e}\!{-3}$ & $0.205\pm5\mathrm{e}\!{-3}$ & $0.198\pm4\mathrm{e}\!{-3}$ & -\\
        %MB + ML & TODO & TODO & TODO & TODO & TODO & TODO & -\\
        \bottomrule
    \end{tabular}
    \label{table:rmse-ecg}
    }
\end{table}
% \begin{table}[ht]
%     \centering
%     \caption{Balanced accuracy score for downstream diagnosis from 12-lead ECGs reconstructed from lead I with the classification model from \cite{Strodthoff2020ecgbenchmarking}.} % trained on the \textsc{ptb-xl} dataset.}
%     \resizebox{0.8\textwidth}{!}{
%     \begin{tabular}{lccccccc|l}
%         \toprule
%         Diagnosis & \algo${}_{50}$ & \algo${}_{300}$  & \dps & \pgdm & \ddnm & \diffpir & \reddiff & Ground-truth \\
%         \midrule
%         RBBB & \textbf{0.782} & \textbf{0.853} & 0.500 & 0.645 & 0.709 & 0.531 & \underline{0.723} & 0.965 \\
%         LBBB & \textbf{0.908} &\textbf{0.954} & 0.856 & \underline{0.895} & 0.857 & 0.851 & 0.876 & 0.999 \\
%         AF & \textbf{0.887} & \textbf{0.926} & 0.583 & 0.628 & 0.800 & \underline{0.840} & 0.831 & 0.949 \\
%         SB & \textbf{0.752} & \textbf{0.746} & 0.513 & 0.634 & 0.703 & 0.588 & \underline{0.713}& 0.797 \\
%         \bottomrule
%     \end{tabular}
%     \label{table:bacc-ecg_old}
%     }
% \end{table}
% \begin{figure}[!ht]
% \centering
%         \includegraphics[width=.5\textwidth]{figures/ECG/IMPUTATION_lead_seed0_GOOD.pdf}
%         \captionof{figure}{\small ECG reconstruction from lead I. Ground-truth in blue, %5th-95th
%         $5\%$--$95\%$
%         quantile range in green, random sample in orange}
%         \label{fig:leadI_NSR_old}
% \end{figure}

\begin{minipage}[t]{0.60\textwidth}
    \centering
    \includegraphics[width=1.\textwidth]{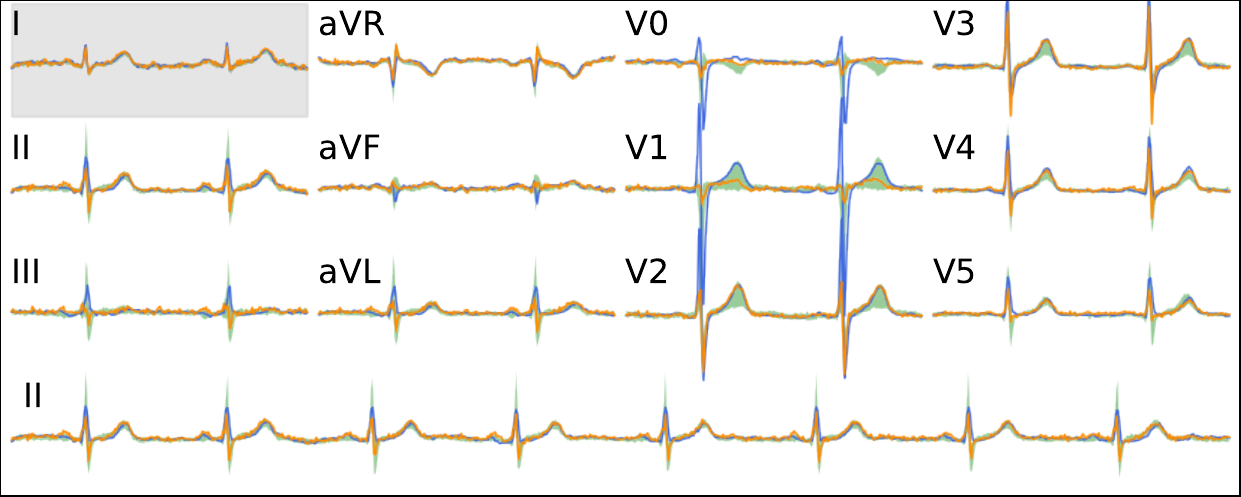}
    \centering
    \captionsetup{font=small}
    \captionof{figure}{\small 10s ECG reconstruction from lead I. Ground-truth in blue,
        $10\%$--$90\%$
        quantile range in green, random sample in orange.}
    \label{fig:leadI_NSR}
\end{minipage}
% \hspace{0.02\textwidth}
\raisebox{35pt}[0pt][0pt]{
\begin{minipage}{0.38\textwidth}
    \centering
    \vspace{0pt}
    \captionsetup{font=small}
    \captionof{table}{Balanced acc. downstream diagnosis from ECG reconstructed from lead I.}
    \resizebox{1.\textwidth}{!}{
        \begin{tabular}{lcccc}
            \toprule
            Method & RBBB & LBBB & AF & SB \\
            \midrule
            \algo${}_{50}$ & \underline{0.81} & \underline{0.92} & \bf{0.94} & \underline{0.66}\\
            \algo${}_{300}$ & \bf{0.90} & \bf{0.93} & \underline{0.92} & \bf{0.66}\\
            \dps & 0.54 & 0.84 & 0.79 & 0.50\\
            \pgdm & 0.65 & 0.87 & 0.88 & 0.55\\
            \ddnm & 0.71 & 0.83 & 0.86 & 0.59\\
            \diffpir & 0.57 & 0.80 & 0.77 & 0.53\\
            \reddiff & 0.73 & 0.86 & 0.88 & 0.60\\
            \midrule
            Ground-truth & 0.99 & 0.98 & 0.94 & 0.70\\
            \bottomrule
        \end{tabular}
        \label{table:bacc-ecg}
        }
\end{minipage}
}
%%% Local Variables:
%%% mode: LaTeX
%%% TeX-master: "../main_iclr"
%%% End:

  \label{sec:experiments}
  \section{Conclusion}
  We have introduced \algo, a novel posterior sampling algorithm designed to solve general inverse problems using both diffusion models and latent diffusion models as a prior. Our approach is based on trading off the approximation error in the prior backward dynamics for a more accurate approximation of the guidance term.
  This strategy has proven to be effective in a variety of numerical experiments across various tasks.
  %This strategy has demonstrated clear benefits in toy examples and has been further validated through extensive experiments across various tasks.
  The results show that \algo\ consistently performs competitively and often even superior to state-of-the-art methods.
  %The results show that \algo\ consistently delivers competitive, and often superior, performance compared to state-of-the-art methods.

\paragraph{Limitations and future directions.} Although the proposed method is promising, it has certain limitations that invite further exploration. First, a detailed analysis of how the algorithm’s performance depends on the choice of the intermediate time steps $(\tmid{k})_{k = 1} ^{n-1}$ remains a challenging but critical subject for future research. Although our image and ECG experiments suggest that setting $\tmid{k} = \lfloor k/2 \rfloor$ leads to significant performance gains on most tasks, we have observed that using an adaptive sequence $\tmid{k} = \lfloor \eta_k k \rfloor$, where $\eta_k$ increases as $k$ decreases, further enhances results on certain tasks, such as Gaussian and motion deblurring, as well as when using latent diffusion models. A second direction for improvement lies in refining the approximation of $\potn{k}{}$ which we believe could lead to overall algorithmic improvements. Specifically, devising an approximation $\hpotn{k}{}$ such that $\nabla \log \hpotn{k}{}$ does not require a vector-Jacobian product could significantly reduce the algorithm’s runtime. Lastly, we see potential in using the two-stage approximation introduced in our warm-start strategy (see \Cref{algo:algo2}) at every diffusion step. Although this technique is promising, it currently leads to instability as $k$ decreases. Understanding and resolving this instability will be a key focus of our future work.
  %%%%%%%%%%%%%%%%%%%%%%%%%%%%%%%
\paragraph{Acknowledgements. } The work of Y.J. and B.M. has been supported by Technology Innovation Institute (TII), project Fed2Learn. The work of Eric Moulines has been partly funded by the European Union (ERC-2022-SYG-OCEAN-101071601). Views and opinions expressed are however those of the author(s) only and do not necessarily reflect those of the European Union or the European Research Council Executive Agency. Neither the European Union nor the granting authority can be held responsible for them. We would like to thank IHU-LIRYC for the computing power made available to us.
  % \newpage
  % \clearpage
  \bibliography{bibliography}
  \bibliographystyle{iclr2025_conference}

  \newpage
  \appendix
  \section{Background on denoising diffusion models}
\subsection{Denoising diffusion probabilistic models}
\label{sec:ddpm-background}

In this section, we provide further background on DDMs based on the DDPM framework \cite{ho2020denoising,dhariwal2021diffusion,song2021ddim}. We rely on definitions provided in the main text.

DDPMs define generative models for $\prior$ relying only on parametric approximations $(\hpredx{t})_{t = 1}^T$ of the denoisers $(\predx{t})_{t = 1} ^T$. These approximate denoisers are usually defined through the parameterization
\begin{equation}
    \label{eq:denoiser-parameterization}
\hpredx{t}(\bx _t) = (\bx _t - \sqrt{1 - \acp{t}} \prednoise{t}(\bx _t)) / \sqrt{\acp{t}}
\end{equation}
and trained by minimizing the denoising loss
\begin{equation}
    \label{eq:denoising_objective}
    \sum_{t = 1}^T  w_t \pE \left[ \| \beps_t - \prednoise{t}(\sqrt{\acp{t}} \bX _0 + \sqrt{1 - \acp{t}} \beps_t ) \|^2 \right]
    \eqsp,
\end{equation}
w.r.t. the neural network parameter $\param$,
where $(\beps_t)_{t = 1}^T$ are i.i.d. standard normal vectors, $\bX _0 \sim \prior$, and $(w_t)_{t = 1}^T$ are some nonnegative weights.
% $
% \beps_t(\bx _t) \eqdef \int \bz \, \mu_{0|t}(\bz | \bx _t) \rmd \bz
% $
% where $\mu_{0|t}(\bz | \bx _t) \propto \prior(\frac{\bx _t - \sqrt{1 - \acp{t}} \bz}{\sqrt{\acp{t}}})\normpdf(\bz; \zero_\dimx, \Id_\dimx)$. It is seen, using a change of variables, that the true denoiser satisfies $\predx{t}(\bx _t) = (\bx _t - \sqrt{1 - \acp{t}} \beps_{t}(\bx _t)) / \sqrt{\acp{t}}$ thus justifying the parameterization \eqref{eq:denoiser-parameterization}.
%$\beps_t \simiid \gauss(\zero_\dimx, \Id_\dimx)$ and $(w(t))_{t=0}^T$ are some weights.
Having trained the denoisers, the  generative model for $\prior$ is defined as follows. Let $( t_{k} )_{k = 0} ^n$ be an increasing sequence of time steps in $\intset{0}{T}$ with $t_0 = 0$. We assume that $t_n$ is large enough so that $\fwmarg{t_n}{\bx _{t_n}} = \int \prior(\bx _0) \fwtrans{t_n|0}{\bx _0}{\bx _{t_n}} \, \rmd \bx _0$  is approximately the density of a multivariate standard normal distribution. For convenience we assign the index $k$ to any quantity depending on $t_k$; \emph{e.g.}, we denote $\fwmarg{t_k}{}$ by $\fwmarg{k}{}$. Consider the backward decomposition
\begin{align*}
    \fwmarg{0:n}{\bx _{0:n}}  = \prior(\bx _0) \prod_{k = 0}^{n-1} \fwtrans{k+1|k}{\bx _k}{\bx _{k+1}}  = \fwmarg{n}{\bx _n} \prod_{k = 0}^{n-1} \fwtrans{k|k+1}{\bx _{k+1}}{\bx _k}
\end{align*}
of the forward process initialized at $\prior$, where $\fwtrans{k|k+1}{\bx _{k+1}}{\bx _k} \propto \fwmarg{k}{\bx _k} \fwtrans{k+1|k}{\bx _k}{\bx _{k+1}}$. Next, for $(j, \ell, k) \in \intset{0}{n}^3$ such that $j < \ell < k$, define
\begin{align}
    \label{eq:mean_bridge}
    \meanBridge_{\ell|j, k}(\bx _j, \bx _k) & \eqdef \frac{\sqrt{\acp{\ell}{} / \acp{j}}(1 - \acp{k} / \acp{\ell})}{1 - \acp{k} / \acp{j}} \bx _j + \frac{\sqrt{\acp{k} / \acp{\ell}} (1 - \acp{\ell} / \acp{j})}{1 - \acp{k} / \acp{j}} \bx _k \eqsp, \\
    \var_{\ell|j, k} & \eqdef \frac{(1 - \acp{\ell} / \acp{j})(1 - \acp{k} / \acp{\ell})}{1 - \acp{k} / \acp{j}} \eqsp. \label{eq:std_bridge}
\end{align}
Then the bridge kernel is 
\begin{align}
    \label{eq:bridge_kernel}
    \fwtrans{\ell|j, k}{\bx _j, \bx _k}{\bx _\ell} & = \fwtrans{\ell|j}{\bx _j}{\bx _\ell} \fwtrans{k|\ell}{\bx _\ell}{\bx _k} \big/ \fwtrans{k|j}{\bx _j}{\bx _k} \nonumber \\
    & = \normpdf(\bx_\ell; \meanBridge_{\ell|j, k}(\bx _j, \bx _k), \var_{\ell|j,k} \Id_\dimx) \eqsp.
\end{align}
Using the bridge kernel, $\fwmarg{0:n}{}$ is approximated using the variational approximation
%DDPM posits the following variational approximation of $\fwmarg{0:n}{}$
\begin{equation*}
    \bwp{0:n}{}{}(\bx _{0:n}) = \bwp{n}{}{}(\bx _n) \prod_{k = 0}^{n-1} \bwp{k|k+1}{\bx _{k+1}}{\bx _k} \eqsp,
\end{equation*}
where $\bwp{0|1}{\bx _1}{\bx _0} = \normpdf(\bx _0; \hpredx{1}(\bx _1), \var_{0|1}\Id_\dimx)$, $\var_{0|1}$ being a tunable parameter, and
\begin{equation}
    \label{eq:ddpm-transition}
\bwp{k|k+1}{\bx _{k+1}}{\bx _k} \eqdef \fwtrans{k|0, k+1}{\hpredx{k+1}(\bx _{k+1}), \bx _{k+1}}{\bx _k} \eqsp \quad k \in \intset{1}{n-1} \eqsp.
\end{equation}
When $n = T$, the denoising objective \eqref{eq:denoising_objective}
corresponds to the KL divergence $\kldivergence{\fwmarg{0:n}{}}{\bwp{0:n}{}{}}$ for a specific choice of weights $(w_t)_t$. In practice, a DDPM is trained using the objective \eqref{eq:denoising_objective} with large $T$ but at inference $n$ is usually much smaller.
% Finally, in order to get some intuition about the parameterization \eqref{eq:ddpm-transition}, note that the Gaussian projection of $\fwtrans{k|k+1}{\bx _{k+1}}{\cdot}$ minimizing $\kldivergence{\fwtrans{k|k+1}{\bx _{k+1}}{\cdot}}{}$ Applying the tower property, it is seen that
% $$
% \int \bx _k \, \fwtrans{k|k+1}{\bx _{k+1}}{\bx _k} \rmd \bx _k = \meanBridge_{k|0, k+1}(\predx{k+1}(\bx _{k+1}), \bx _{k+1})
% $$
% and hence, \eqref{eq:ddpm-transition} is understood as a suboptimal Gaussian projection
\subsection{Latent diffusion models}
\label{subsec:ldms}
Latent diffusion models (LDM) \cite{rombach2022high} define a DDM in a latent space. Let $\encoder: \rset^\dimx \to \rset^p$ be an encoder function and $\decoder: \rset^p \to \rset^\dimx$ a decoder function. We assume that these functions satisfy $\decoder_\sharp \encoder_\sharp \prior \approx \prior$, where, for instance, $\encoder_\sharp q$ denotes the law of the random variable $\encoder(\bX)$ where $\bX \sim \prior$. A LDM approximating $\prior$ is given by $\decoder_\sharp \bwp{0}{}{}$, where $\bwp{0}{}{}$ is a diffusion model trained on samples from $\encoder_\sharp \prior$. Finally, when solving inverse problems with LDMs, we assume that the target distribution is instead
$$
    \post{}{\bx} \propto \potn{}{\bx} \decoder_\sharp \encoder_\sharp \prior(\bx) \eqsp.
$$
Let $\bX \sim \post{}{}$, $D \sim \decoder_\sharp \encoder_\sharp \prior$, $E \sim \encoder_\sharp \prior$, and $\bZ \sim \overline{\pi}$, where $\overline{\pi}(\bz) \propto \potn{}{\decoder_\sharp(\bz)} \encoder_\sharp \prior(\bz)$. For any bounded function $f$ on $\rset^\dimx$, we have, following the definition of $\post{}{}$, that
\begin{align*}
    \pE[f(\bX)]  = \frac{\pE[\potn{}{D} f(D)]}{\pE[\potn{}{D}]} = \frac{\pE[\potn{}{\decoder(E)} f(\decoder(E))]}{\pE[\potn{}{\decoder(E)}]} = \pE[f(\decoder(\bZ))] \eqsp.
\end{align*}
Hence $\law(\bX) = \law(\decoder(\bZ))$. As a result, to sample approximately from $\post{}{}$, we first sample approximately from $\overline{\pi}$ using any diffusion posterior sampling algorithm with pre-trained DDM for $\encoder_\sharp \prior$, then decode the obtained samples using $\decoder$.
\subsection{Midpoint decomposition}
\label{apdx:backward-intermediate-decomp}
Before we proceed with the midpoint decomposition, we first recall that under the joint distribution obtained by initializing the posterior $\post{}{}$ with the forward process (see \eqref{eq:posterior_forward}), it holds that for all $i < j$,
\begin{equation}
\label{eq:reversal}
        \pibw{i}{}{}(\bx _{i}) \fwtrans{j|i}{\bx _{i}}{\bx _{j}} = \pibw{j}{}{}(\bx _{j}) \pibw{i|j}{\bx _{j}}{\bx _{i}}
        %\eqsp, \quad \forall i < j
        \eqsp,
\end{equation}
where $\pibw{i|j}{\bx _j}{\bx _i} \eqdef \post{i}{\bx _i} \fwtrans{j|i}{\bx _i}{\bx _j} / \post{j}{\bx _j}$ and integrates to one.
\begin{proof}[Proof of \Cref{lem:pi_bwker}]
Let $(\ell, k) \in \intset{0}{n}$ be such that $\ell < k$.
Applying repeatedly the definition of the bridge kernel \eqref{eq:bridge_kernel} and the identity \eqref{eq:reversal},
\begin{align*}
    % forward
    \fwtrans{k|\ell, k+1}{\bx _\ell, \bx _{k+1}}{\bx _k} & =
    \frac{\fwtrans{k|\ell}{\bx _\ell}{\bx _k} \fwtrans{k+1|k}{\bx _k}{\bx _{k+1}}}{\fwtrans{k+1|\ell}{\bx _\ell}{\bx _{k+1}}} \\
    & = \frac{\pibw{\ell}{}{}(\bx _\ell) \fwtrans{k|\ell}{\bx _\ell}{\bx _k} \fwtrans{k+1|k}{\bx _k}{\bx _{k+1}}}{\pibw{\ell}{}{}(\bx _\ell) \fwtrans{k+1|\ell}{\bx _\ell}{\bx _{k+1}}} \\
    %\\
    & = \frac{\pibw{\ell|k}{\bx _{k}}{\bx _{\ell}} \pibw{k|k+1}{\bx _{k+1}}{\bx _{k}}  \pibw{k+1}{}{}(\bx _{k+1})}{\pibw{\ell|k+1}{\bx _{k+1}}{\bx _{\ell}} \pibw{k+1}{}{}(\bx _{k+1})} \\
    & = \frac{\pibw{\ell|k}{\bx _{k}}{\bx _{\ell}} \pibw{k|k+1}{\bx _{k+1}}{\bx _{k}}}{\pibw{\ell|k+1}{\bx _{k+1}}{\bx _{\ell}}}
    \enspace.
\end{align*}
It then follows that
\begin{align*}
    \pibw{k|k+1}{\bx _{k+1}}{\bx _{k}}
    & = \int \pibw{\ell|k}{\bx _k}{\bx _\ell} \pibw{k|k+1}{\bx _{k+1}}{\bx _k} \, \rmd \bx _\ell  \\
    & = \int \fwtrans{k|\ell, k+1}{\bx _\ell, \bx _{k+1}}{\bx _k} \pibw{\ell|k+1}{\bx _{k+1}}{\bx _{\ell}}  \, \rmd \bx _\ell
    \eqsp.
\end{align*}
\end{proof}

  \section{The Gaussian case}
\label{apdx:gaussian-example}

\subsection{Derivation}

% --- local vars ---
\def\cov{\mathbf{\Sigma}}
\def\mean{\boldsymbol{m}}
\def\likelihood{\bfA}
\def\covBridge{\sigma^2}
\def\meanConditional{\hat{\mean}}
\def\potBias{\boldsymbol{b}}
\def\potLikelihood{\hat{\likelihood}}
\def\covPosterior{\mathbf{\Gamma}}
\def\matrixXk{\mathbf{M}}
\def\bfM{\mathbf{M}}
\def\hpimean{\hat{\boldsymbol{\mu}}}
\def\bfH{\mathbf{H}}
\def\bc{\boldsymbol{c}}
% ---
In this section we derive the recursions verified by the first and second moments of the marginal distribution $\hpibw{0}{}{}[\tmidfn]$ of the surrogate model \eqref{eq:pi-surrogate} in the simplified setting of \Cref{example:gaussian}. We recall that in this specific example we assume that $\prior = \gauss(\mean, \cov)$ where $(\mean, \cov) \in \rset^\dimx \times \mathcal{S}^{++} _\dimx$ and $\potn{}{}: \bx \mapsto \normpdf(\obs; \bfA \bx, \stdobs^2 \Id_\dimobs)$.

\paragraph{Denoiser and DDPM transitions.} Since we are dealing with a Gaussian prior, the denoiser $\predx{k}$ can be computed in closed form for any $k \in \intset{1}{n}$. Using \citet[Eqn.~2.116]{bishop2006pattern}, we have that
    \begin{align*}
        \bw{0|k}{\bx _k}{\bx _0}
            & \propto \prior(\bx _0) \fwtrans{k|0}{\bx _0}{\bx _k} \\
            & = \normpdf \big(\bx _0;
                \cov_{0|k} \big( (\sqrt{\acp{k}} / \var_k) \bx _k + \cov^{-1} \mean \big), \cov_{0|k} \big),
    \end{align*}
where $\cov_{0|k} \eqdef ((\acp{k} / \var_k) \Id + \cov^{-1})^{-1}$. Hence,
$$
    \predx{k}(\bx _k) = \cov_{0|k}\big( (\sqrt{\acp{k}} / \var_k) \bx _k + \cov^{-1} \mean \big),
$$
% Indeed, we have that $\fwmarg{k}{} = \gauss(\sqrt{\acp{k}} \mean, \cov_k)$ where $\cov_k = \var_k \Id + \acp{k} \cov$ and hence, by Tweedie's formula
% $$
%     \predx{k}(\bx _k) = \frac{\bx _k + \var_k \nabla_{\bx _k} \log \fwmarg{k}{\bx _k}}{\sqrt{\acp{k}}} = \frac{\bx _k - \var_k \cov^{-1} _k (\bx _k - \sqrt{\acp{k}} \mu)}{\sqrt{\acp{k}}} \eqsp.
% $$
and we assume in the remainder of this section that $\hpredx{k} = \predx{k}$. From the expression of $\fwtrans{0|k}{\bx _k}{\cdot}$ we can immediately derive the more general backward transitions $\fwtrans{\ell|k}{\bx _k}{\cdot}$ for $\ell \in \intset{1}{k-1}$ by noting that
$$
\fwtrans{\ell|k}{\bx _k}{\bx _\ell} = \int \fwtrans{\ell | 0, k}{\bx _0, \bx _k}{\bx _\ell} \fwtrans{0|k}{\bx _k}{\bx _0} \, \rmd \bx _0 \eqsp.
$$
From this, \eqref{eq:mean_bridge}, \eqref{eq:std_bridge}, and the law of total expectation and covariance, it follows that $\fwtrans{\ell|k}{\bx _k}{\cdot} = \gauss(\mean_{\ell|k}(\bx _k), \cov_{\ell|k}(\bx _k))$, where
\begin{align*}
    \mean_{\ell|k}(\bx _k) = \meanBridge_{\ell|0, k}(\predx{k}(\bx _k), \bx _k) \eqsp, \quad \cov_{\ell | k} = \frac{\acp{\ell}(1 - \acp{k} / \acp{\ell})^2}{(1 - \acp{k})^2} \cov_{0|k} + \var _{\ell|0, k} \Id_{\dimx} \eqsp.
\end{align*}
On the other hand, the DDPM transitions are
\begin{equation}
    \label{eq:ddpm-gaussian}
\bwp{\ell|k}{\bx _k}{\cdot} = \fwtrans{\ell|0, k}{\predx{k}(\bx _k), \bx _k}{\cdot} = \gauss(\meanBridge_{\ell|0, k}(\predx{k}(\bx _k), \bx _k), \var_{\ell|0, k} \Id_\dimx),
\end{equation}
which shows that in this case, the true transitions and approximate ones differ only by their covariance.

\paragraph{Moments recursion.} For $k \in \intset{0}{n}$ we let $\hpibw{k}{}{}[\tmidfn]$ denote the $\bx _k$ marginal of the surrogate model \eqref{eq:pi-surrogate}. We remind the reader that $\hpibw{n}{}{}[\tmidfn] = \gauss(\zero_\dimx, \Id_\dimx)$. The marginals satisfy the recursion
$$
    \hpibw{k}{}{}[\tmidfn](\bx _k) = \int \hpibw{k|k+1}{\bx _{k+1}}{\bx _k}[\tmidfn] \hpibw{k+1}{}{}[\tmidfn](\bx _{k+1}) \, \rmd \bx _{k+1}\eqsp, \quad k \in \intset{0}{n-1} \eqsp.
$$
Since $\hpotn{k}{\bx _k} = \normpdf(\obs; \bfA \hpredx{k}(\bx _k), \stdobs^2 \Id_\dimobs)$ and $\hpredx{k}$ is linear in $\bx _k$, it is easily seen that $\hpibw{\tmid{k}|k+1}{\bx _{k+1}}{\cdot}$ is the density of a Gaussian distribution. Consequently, by definition \eqref{eq:pi-transition-approximation} and the definition \eqref{eq:bridge_kernel} of the bridge kernel, this is also the case for $\hpibw{k|k+1}{\bx _{k+1}}{\cdot}[\tmidfn]$. Now assume that
\begin{align}
    \hpibw{k+1}{}{}[\tmidfn](\bx _{k+1}) & = \normpdf(\bx _{k+1}; \hpimean^\tmidfn _{k+1}, \hat\cov^\tmidfn _{k+1}) \eqsp,\\
    \hpibw{k|k+1}{\bx _{k+1}}{\bx _k}[\tmidfn] & = \normpdf(\bx _k; \bfM^\tmidfn _{k|k+1} \bx _{k+1} + \bc^\tmidfn _{k|k+1}, \hat{\cov}^\tmidfn_{k|k+1}) \eqsp, \label{eq:hbw-explicit}
\end{align}
where $\bfM^\tmidfn _{k|k+1} \in \rset^{\dimx \times \dimx}$, $\hat\cov^\tmidfn _{k|k+1} \in \mathcal{S}^{++} _\dimx$, and $\bc^\tmidfn _{k|k+1} \in \rset^{\dimx}$.
Using the definition \eqref{eq:bridge_kernel} of the bridge kernel we find that $\hpibw{k}{}{}[\tmidfn] = \gauss(\hpimean_k, \hat\cov_k)$, where
\begin{align*}
    \hpimean^\tmidfn _k & = \bfM^\tmidfn _{k|k+1} \hpimean^\tmidfn _{k+1} + \bc^\tmidfn _{k|k+1} \eqsp, \\
    \hat\cov^\tmidfn _k & = \bfM^\tmidfn _{k|k+1} \hat\cov^\tmidfn _{k+1} \bfM^\intercal _{k|k+1} + \hat\cov^\tmidfn _{k|k+1} \eqsp.
\end{align*}
Iterating these updates until reaching $k = 0$, starting from the initialization $\hpimean^\tmidfn _n = \zero_\dimx$ and $\hat\cov^\tmidfn _n = \Id_\dimx$, yields the desired moments of the surrogate posterior $\pibw{0}{}{}[\tmidfn]$. It now remains to show that the backward transition $\hpibw{k|k+1}{\bx _{k+1}}{\cdot}[\tmidfn]$ writes in the form \eqref{eq:hbw-explicit} and identify $\bfM^\tmidfn _{k|k+1}$ and $\bc^\tmidfn _{k|k+1}$.

First, we write the approximate likelihood in the form
$$
    \hpotn{k}{\bx _k} = \normpdf(\obs; \hat\bfA_k \bx _k + \potBias_k, \stdobs^2 \Id_\dimobs),
$$
where
$$
\potLikelihood_k = (\sqrt{\acp{k}} / \var_k)  \likelihood \cov_{0|k}, \quad \potBias_k = \likelihood \cov_{0|k} \cov^{-1} \mean 
\eqsp.
$$
We also denote by $\boldsymbol{m}^\param _{\ell|k}(\bx _k)$ the mean of the Gaussian distribution with density given by the DDPM transition $\bwp{\ell|k}{\bx _k}{\cdot}$ in \eqref{eq:ddpm-gaussian}. We have that
$$
    \boldsymbol{m}^\param _{\tmid{k}|k+1}(\bx _{k+1}) = \bfH_{\tmid{k}|k+1} \bx _{k+1} + \boldsymbol{h}_{\tmid{k}|k+1} \eqsp,
$$
where
\begin{align}
    \bfH_{\tmid{k}|k+1} & \eqdef \frac{\acp{\tmid{k}} (1 - \acp{k+1} / \acp{\tmid{k}})}{\var_\tmid{k} \var_{k+1}} \cov_{0|k} + \frac{\sqrt{\acp{k+1}}(1 - \acp{\tmid{k}})}{\var_{k+1}} \Id_\dimx \eqsp,\\
    \boldsymbol{h}_{\tmid{k}|k+1} & \eqdef   \frac{\sqrt{\acp{\tmid{k}}} (1 - \acp{k+1} / \acp{\tmid{k}})}{\var_{k+1}} \cov_{0|k} \cov^{-1} \mean 
    \eqsp.
\end{align}
Then, applying \cite[Eqn. 2.116]{bishop2006pattern}, we get
\begin{align*}
        \hpibw{\tmid{k}|k+1}{\bx _{k+1}}{\bx _{\tmid{k}}} = \normpdf(\bx _{\tmid{k}}; \widetilde\bfM^\param _{\tmid{k}|k+1} \bx _{k+1} + \tilde{\bc}^\param _{\tmid{k}|k+1}
            ,
            \covPosterior_{\tmid{k}|k+1}
            ) \eqsp,
    % \label{eq:g-conditional-backward}
\end{align*}
where
\begin{align*}
    \covPosterior_{\tmid{k}|k+1} & \eqdef \big(\var^{-1} _{\tmid{k}|0, k+1} \Id + \sigma_{\obs}^{-2} \potLikelihood_{\tmid{k}}^\top \potLikelihood_{\tmid{k}}\big)^{-1} \eqsp, \\
    \widetilde\bfM^\param _{\tmid{k}|k+1} & \eqdef \var^{-1} _{\tmid{k}|0, k+1}  \covPosterior_{\tmid{k}|k+1} \bfH _{\tmid{k}|k+1} \eqsp, \\
    \tilde{\bc}^\param _{\tmid{k}|k+1} & \eqdef \covPosterior_{\tmid{k}|k+1} \big[ \sigma_{\obs}^{-2} \potLikelihood_{\tmid{k}}^\top (\obs - \potBias_{\tmid{k}}) + \var^{-1} _{\tmid{k}|0, k+1} \boldsymbol{h}_{\tmid{k}|k+1} \big] \eqsp.
\end{align*}
Finally, following \eqref{eq:pi-transition-approximation} we find that
\begin{align*}
    \bfM^\tmidfn _{k|k+1} & = \frac{\sqrt{\acp{k} / \acp{\tmid{k}}} (1 - \acp{k+1} / \acp{k})}{1 - \acp{k+1} / \acp{\tmid{k}}} \widetilde\bfM^\param _{\tmid{k}|k+1} + \frac{\sqrt{\acp{k+1} / \acp{k}} (1  - \acp{k} / \acp{\tmid{k}})}{1 - \acp{k+1} / \acp{\tmid{k}}} \Id_\dimx \eqsp, \\
    \bc^\tmidfn _{k|k+1} & = \frac{\sqrt{\acp{k} / \acp{\tmid{k}}} (1 - \acp{k+1} / \acp{k})}{1 - \acp{k+1} / \acp{\tmid{k}}} \tilde{\bc}^\param _{\tmid{k}|k+1} \eqsp.
\end{align*}

\subsection{Experimental setup}

% ---
\def\G{\mathbf{G}}
% ---

In the experiment described in \Cref{example:gaussian}, we set $\dimx = 100$ and generate $500$ instances of inverse problems $(\likelihood, \mean, \cov)$.
For each instance, we compute the Wasserstein-2 distance between the resulting posterior distribution $\post{}{}$ and $\hpibw{0}{}{}[\tmidfn(\eta)]$.

\paragraph{Prior.}
The mean of the prior is sampled from a standard Gaussian distribution.
To generate the covariance $\cov \in \mathcal{S}^{++} _\dimx$, we first draw a matrix $\G \in \rset^{\dimx \times \dimx}$ with \iid\ entries sampled from $\gauss(0, 1)$.
Then, for a better conditioning of $\cov$, we normalize the columns of $\G$ and set $\cov = \bar{\lambda}^2 \Id + \G \G^\top$, where $\bar{\lambda}^2$ is the mean of the squared singular values of $\G$.

\paragraph{Likelihood.}
To sample an ill-posed problem we generate a rank-deficient matrix $\likelihood \in \rset^{\dimx_{\obs} \times \dimx}$ with $\dimx_{\obs} \leq \dimx$.
We sample uniformly $\dimx_{\obs}$ from the interval $\intset{\dimx / 10}{\dimx}$ and draw the entries of $\likelihood$ \iid\ from $\gauss(0, 1)$.
Regarding $\sigma_\obs$, we sample it uniformly from the interval $[0.1, 0.5]$.

Finally, the resulting posterior is also Gaussian \cite[Eqn.~2.116]{bishop2006pattern}
\begin{equation*}
    \post{}{}(\bx)  \propto \potn{}{\bx} p(\bx) \propto \normpdf(\bx; \mean_{\obs}, \cov_{\obs})
    \eqsp,
\end{equation*}
where $\cov_{\obs} = (\cov^{-1} + (1 / \stdobs^2) \likelihood^\top \likelihood)^{-1}$ and  $ \mean_{\obs} = \cov_{\obs} \big( (1 / \stdobs^2)\likelihood^\top y + \cov^{-1} \mean \big)$.
The Wasserstein-2 distance between the true posterior and $\hpibw{0}{}{}[\tmidfn(\eta)]$ is thus \cite{olkin1982wasserstein2}
\begin{equation*}
    W_2(\post{}{}, \hpibw{0}{}{}[\tmidfn])^2 =
        \| \mean_{\obs} - \hpimean^\tmidfn_{0} \|^2
        + \trace \big( \cov_{\obs} + \hat\cov^\tmidfn_0 - 2 (\cov^{\frac{1}{2}}_{\obs} \hat\cov^\tmidfn_0 \cov^{\frac{1}{2}}_{\obs})^{\frac{1}{2}}\big)
    \eqsp.
\end{equation*}

  \def\lossws{\mathcal{L}^{\mathrm{ws}} _{1|\tmid{k}}}
\def\estlossws{\widetilde{\mathcal{L}}^{\mathrm{ws}} _{1|\tmid{k}}}
\def\wsvmu{\tilde{\boldsymbol{\mu}}}
\def\wsvlstd{\tilde{\boldsymbol{\rho}}}
\def\wsvparam{\psi}

\section{More details on \algo}
\subsection{Details on the loss}
\revision{First, by the Data Processing inequality \cite[Example 2]{van2014renyi},
\begin{align*} 
& \kldivergence{\vpibw{k|k+1}{\bx _{k+1}}{\cdot}}{\hpibw{k|k+1}{\bx _{k+1}}{\cdot}[\tmidfn]} \\
&  \leq \int \log \frac{\fwtrans{k|\tmid{k}, k+1}{\bx _\tmid{k}, \bx _{k+1}}{\bx _k} \vpibwD{\tmid{k}|k+1}{}{\bx _\tmid{k}}}{\fwtrans{k|\tmid{k}, k+1}{\bx _\tmid{k}, \bx _{k+1}}{\bx _k} \hpibw{\tmid{k}|k+1}{\bx _{k+1}}{\bx _\tmid{k}}[\param]} \fwtrans{k|\tmid{k}, k+1}{\bx _\tmid{k}, \bx _{k+1}}{\bx _k} \vpibwD{\tmid{k}|k+1}{}{\bx _\tmid{k}} \, \rmd \bx_k \rmd \bx_\tmid{k}\\
&  = \int \log \frac{\vpibwD{\tmid{k}|k+1}{}{\bx _\tmid{k}}}{\hpibw{\tmid{k}|k+1}{\bx _{k+1}}{\bx _\tmid{k}}[\param]} \fwtrans{k|\tmid{k}, k+1}{\bx _\tmid{k}, \bx _{k+1}}{\bx _k} \vpibwD{\tmid{k}|k+1}{}{\bx _\tmid{k}} \, \rmd \bx_k \rmd \bx_\tmid{k} \\
& = \kldivergence{\vpibw{\tmid{k}|k+1}{}{}}{\hpibw{\tmid{k}|k+1}{\bx _{k+1}}{\cdot}[\param]}
\end{align*}
The gradient of 
$
\vparam \mapsto \kldivergence{\vpibw{\tmid{k}|k+1}{}{}}{\hpibw{\tmid{k}|k+1}{\bx _{k+1}}{\cdot}[\param]} 
$
for a given $\bx _{k+1}$ writes }
\begin{multline}
    \label{eq:loss-grad}
    \nabla_\vparam \mathcal{L}_k (\vparam; \bX _{k+1}) \\ = - \pE \big[ \nabla_\vparam \log \hpotn{\tmid{k}}{\vmu_{\tmid{k}} + \diag(\rme^{\vlstd_\tmid{k}}) \bZ}\big] + \nabla_\vparam \kldivergence{\vpibw{\tmid{k}|k+1}{}{}}{\bwp{\tmid{k}|k+1}{\bX _{k+1}}{\cdot}}
\end{multline}
and the second term can be expressed in a closed form since both distributions are Gaussians, \emph{i.e.},
\begin{multline*}
    \nabla_\vparam \kldivergence{\vpibw{\tmid{k}|k+1}{}{}}{\bwp{\tmid{k}|k+1}{\bx _{k+1}}{\cdot}}
    \\ = \nabla_\vparam \left[ - \sum_{j=1}^\dimx \vlstd_{\tmid{k}, j} + \frac{\| \vmu_\tmid{k} - \meanBridge_{\tmid{k}|0, k+1}(\hpredx{k+1}(\bx _{k+1}), \bx _{k+1})\|^2 + \sum_{j = 1} ^\dimx \frac{\rme^{2 \vlstd_{\tmid{k}, j}}}{\var_{\tmid{k}|0, k+1}}}{2 \var_{\tmid{k}|0, k+1}}\right] \eqsp.
\end{multline*}
\subsection{Warm start}
\label{sec:warm_start}
In this section we describe the warm-start approach discussed in the main paper. The complete algorithm with the warm-start procedure is given in \Cref{algo:algo2}. The original version presented in \Cref{algo:algo} relies on first sampling approximately $\bX _{\tmid{k}}$, given $\bX _{k+1}$, from the surrogate transition $\hpibw{\tmid{k}|k+1}{\bX _{k+1}}{\cdot}$ and then sampling $\bX _k$ from $\fwtrans{k|\tmid{k}, k+1}{\bX _\tmid{k}, \bX _{k+1}}{\cdot}$. In our warm-start approach,  which we apply only during the first iterations of the algorithm (see the hyperparameter settings in \Cref{table:hyperparams-algo}), we draw inspiration from the decomposition
\begin{equation}
    \label{eq:extended-decomp}
    \pibw{k|k+1}{\bx _{k+1}}{\bx _k} = \int \fwtrans{k|1, k+1}{\bx _1, \bx _{k+1}}{\bx _k} \pibw{1|\tmid{k}}{\bx _\tmid{k}}{\bx _1} \pibw{\tmid{k}|k+1}{\bx _{k+1}}{\bx _\tmid{k}} \, \rmd \bx _1 \rmd \bx _\tmid{k} \eqsp.
\end{equation}
It suggests introducing a second intermediary step that involves sampling approximately from the transition $\smash{\bX _1 \sim \pibw{1|\tmid{k}}{\bX _\tmid{k}}{\cdot}}$. $\bX _k$ is then sampled from the bridge $\fwtrans{k|1, k+1}{\bX _1, \bX _{k+1}}{\cdot}$.

In order to draw approximate samples from $\pibw{1|\tmid{k}}{\bX _{\tmid{k}}}{\cdot}$ we leverage again a Gaussian variational approximation $\smash{\vpibw{1|\tmid{k}}{}{}[\wsvparam] \eqdef \gauss(\wsvmu_1, \diag(\rme^{2 \wsvlstd_1}))}$, which we fit by minimizing a proxy of the KL divergence between $\vpibw{1|\tmid{k}}{}{}[\wsvparam]$ and $\pibw{1|\tmid{k}}{\bX _\tmid{k}}{\cdot}$ of which the gradient \wrt\ $\wsvparam \eqdef (\wsvmu_1, \wsvlstd_1)$ writes
\begin{multline}
    \label{eq:ws-proxy-grad}
    \nabla_\wsvparam \lossws(\wsvparam; \bX _\tmid{k}) _{|\wsvparam _0}  \eqdef - \nabla_\wsvparam \left[ \sum_{j = 1}^\dimx \wsvlstd_{1, j} - \frac{\| \bX _\tmid{k} - (\acp{\tmid{k}}/\acp{1})^{1/2} \wsvmu_1 \|^2 + (\acp{\tmid{k}} / \acp{1}) \sum_{j=1}^\dimx \rme^{\wsvlstd^2 _{1, j}}}{2 (1 - \acp{\tmid{k}} / \acp{1})} \right] _{|\wsvparam_0} \\
     - \pE \left[ \nabla_\wsvparam (\wsvmu_1 + \diag(\rme^{\wsvlstd_1}) \bZ)^\intercal _{| \wsvparam_0} (\nabla_x \log \potn{}{} + \score{1})(\wsvmu_1 + \diag(\rme^{\wsvlstd_1}) \bZ \, {}_{| \wsvparam_0})\right] \eqsp.
\end{multline}
This expression corresponds to the gradient \wrt\ $\wsvparam$ of $\kldivergence{\vpibw{1|\tmid{k}}{}{}[\wsvparam]}{\pibw{1|\tmid{k}}{\bX _\tmid{k}}{\cdot}}$ combined with the score approximation $\score{1}$ of $\nabla_x \log \fwmarg{1}{}$ and the mild likelihood approximation $\nabla_x \log \potn{1}{}$ of $\nabla_x \log \potn{}{}$. Indeed, we have that
\begin{multline*}
    \kldivergence{\vpibw{1|\tmid{k}}{}{}[\wsvparam]}{\pibw{1|\tmid{k}}{\bX _\tmid{k}}{\cdot}} = - \pE_{\vpibw{1|\tmid{k}}{}{}[\wsvparam]} \big[ \log \potn{1}{\bX _1} + \log \fwmarg{1}{\bX _1} \big] \\ + \int \log \frac{\vpibw{1|\tmid{k}}{}{}[\wsvparam](\bx _1)}{\fwtrans{\tmid{k}|1}{\bx _1}{\bX _\tmid{k}}} \vpibw{1|\tmid{k}}{}{}[\wsvparam](\bx _1) \, \rmd \bx _1 \eqsp.
\end{multline*}
The second term can be computed in a closed form and its gradient corresponds to the first term in \eqref{eq:ws-proxy-grad}. As for the first term, under standard differentiability assumptions and applying the raparameterization trick, we find that
\begin{align*}
    \nabla_\wsvparam & \pE_{\vpibw{1|\tmid{k}}{}{}} \big[ \log \potn{1}{\bX _1} + \log \fwmarg{1}{\bX _1} \big]_{|\wsvparam_0}\\
     & = \nabla_\wsvparam \pE \left[ \big(\log \potn{1}{} + \log \fwmarg{1}{}\big)(\wsvmu_1 + \diag(\rme^{\wsvlstd_1}) \bZ)  \right] _{| \wsvparam_0} \\
    & = \pE \left[ \nabla_\wsvparam (\wsvmu_1 + \diag(\rme^{\wsvlstd_1}) \bZ)^\intercal _{| \wsvparam_0} (\nabla_x \log \potn{1}{} + \nabla_x \log \fwmarg{1}{})(\wsvmu_1 + \diag(\rme^{\wsvlstd_1}) \bZ \, {}_{| \wsvparam_0})\right] \eqsp.
\end{align*}
Plugging the previous approximations yields the second term in \eqref{eq:ws-proxy-grad}. The warm-start procedure is summarized in \Cref{algo:warm_start}. We also use a single sample Monte Carlo estimate to perform the optimization. Finally, at step $k$ and given $\vX_\tmid{k}$, the initial parameter $\wsvparam_k$ of the variational approximation is chosen so that
$$
    \vpibw{1|\tmid{k}}{}{}[\wsvparam_k](\bx _1) = \fwtrans{1|0, \tmid{k}}{\hpredx{\tmid{k}}(\vX_\tmid{k}), \vX_\tmid{k}}{\bx _1} \eqsp.
$$
After having sampled $\tilde{\bX}_1 \sim \vpibw{1|\tmid{k}}{}{}[\wsvparam^{*} _k]$, we use it to first sample $\bX _k \sim \fwtrans{k|1, k+1}{\tilde{\bX}_1, \bX _{k+1}}{\cdot}$ and then initialize the next variational approximation $\vpibw{\tmid{k-1}|k}{}{}$. The initial parameter $\vparam_{k-1}$ is set so that
$$
\vpibw{\tmid{k-1}|k}{}{}[\vparam_{k-1}](\bx _\tmid{k-1}) = \fwtrans{\tmid{k-1}|0, k}{\tilde{\bX}_1, \bX _k}{\bx _\tmid{k-1}} \eqsp,
$$
see line \ref{ws:init} in \Cref{algo:algo2}.
\begin{algorithm}[t]
    \caption{$\mathsf{WarmStart}$}
    \begin{algorithmic}[1]
        \STATE {\bfseries Input:} step $k$, samples $(\vX_\tmid{k}, \bX _{k+1})$, gradient steps $\ngrad$
        \STATE $\wsvmu_1 \leftarrow \meanBridge_{1|0, \tmid{k}}(\hpredx{\tmid{k}}(\vX_\tmid{k}), \vX_\tmid{k}) \eqsp, \quad \wsvlstd_1 \leftarrow \frac{1}{2} \log \var_{1| 0, \tmid{k}}$
            \FOR{$j = 1$ {\bfseries to} $\ngrad$}
            \STATE $\bZ \sim \gauss(\zero_\dimx, \Id_\dimx)$
            \STATE $(\wsvmu_1, \wsvlstd_1) \leftarrow \mathsf{OptimizerStep}(\nabla_\wsvparam \estlossws(\cdot, \bZ; \bX _\tmid{k});\, \wsvmu_1, \wsvlstd_1)$
            \ENDFOR
        \STATE $\vX_1 \leftarrow \vmu_1 + \diag(\rme^{\vlstd_1}) \bZ_1$ where $\bZ_1 \sim \gauss(\zero_\dimx, \Id_\dimx)$
        \STATE $\bX _k \leftarrow \meanBridge_{k|1, k+1}(\vX_1, \bX _{k+1}) + \var^{1/2} _{k|1, k+1} \bZ_k$
        \STATE {\bfseries Output:} $\bX _k, \vX_1$
    \end{algorithmic}
    \label{algo:warm_start}
\end{algorithm}
\begin{algorithm}[t]
    \caption{\algo\ with warm start strategy}
    \begin{algorithmic}[1]
       \STATE {\bfseries Input:} $(\tmid{k})_{k = 1} ^n$ with $\tmid{n} = n$, $\tmid{1} = 1$, gradient steps $(\ngrad_k)_{k = 1} ^{n-1}$, warm start threshold $w$.
       %gradient steps $G$, Langevin steps $K$.
       \STATE $\bX _n \sim \normpdf(\zero_\dimx, \Id_\dimx), \, \vX_n \leftarrow \bX _n$
       \STATE $\vX_0 \leftarrow \hpredx{n}(\vX_n)$
       \FOR{$k = n-1$ {\bfseries to} $1$}
            % \STATE $\vX_0 \leftarrow \vX_1 \indic_{k + 1 \geq w} + \hpredx{\tmid{k+1}}(\vX_\tmid{k+1}) \indic_{k+1 < w}$
            \STATE $\vmu_{\tmid{k}} \leftarrow \meanBridge_{\tmid{k}|0, k+1}(\vX_0, \bX _{k+1}), \quad \vlstd_\tmid{k} \leftarrow \frac{1}{2} \log \var_{\tmid{k}|0, k+1}$ \label{ws:init}
            \FOR{$j = 1$ {\bfseries to} $\ngrad_k$}
            \STATE $\bZ \sim \gauss(\zero_\dimx, \Id_\dimx)$
            \STATE $(\vmu_\tmid{k}, \vlstd_\tmid{k}) \leftarrow \mathsf{OptimizerStep}(\nabla_\vparam \widetilde{\mathcal{L}}_\tmid{k}(\cdot, \bZ; \bX _{k+1});\, \vmu_\tmid{k}, \vlstd_\tmid{k})$
            \ENDFOR
            \STATE $\bZ_\tmid{k}, \bZ_k \simiid \gauss(\zero_\dimx, \Id_\dimx)$
            \STATE $\vX_\tmid{k} \leftarrow \vmu_\tmid{k} + \diag(\rme^{\vlstd_\tmid{k}}) \bZ_\tmid{k}$
            \vspace{.03cm}
            \IF{$k \geq w$}
            \STATE $(\bX _k, \vX_1) \leftarrow \mathsf{WarmStart}(k, \vX_\tmid{k}, \bX _{k+1}, \ngrad_k)$
            \STATE $\vX_0 \leftarrow \vX_1$
            \ELSE
            \STATE $\bX _{k} \leftarrow \meanBridge_{k|\tmid{k}, k+1}(\vX_\tmid{k}, \bX _{k+1}) + \var^{1/2}_{k|\tmid{k}, k+1} \bZ_k$
            \STATE $\vX_0 \leftarrow \hpredx{\tmid{k}}(\bX _\tmid{k})$
            \ENDIF
       \ENDFOR
       \STATE {\bfseries Output:} $\vX_0$
    \end{algorithmic}
    \label{algo:algo2}
\end{algorithm}

\subsection{Difference with \dps}
\label{apdx:dps-link}
\revision{In this section we detail the differences between \algo\ and the \dps\ algorithm proposed in \cite{chung2023diffusion}. While \dps\ cannot be seen as an instantition of our methodology, we highlight the main differences by deriving a specific case of \algo\ that closely resembles \dps.}

\revision{We assume that $\tmid{k} = k$ for all $k \in \intset{1}{n-1}$ and that $\potn{}{\bx} = \normpdf(\obs; \mathcal{A}(\bx), \stdobs^2 \Id_\dimobs)$ where $\mathcal{A}: \rset^\dimx \to \rset^\dimobs$.
We consider the same optimization procedure as in \Cref{algo:algo} but we: (i) optimize only the mean parameter $\vmu_{k}$ and fix the covariance of the variational approximation to $\var_{k|k+1} \Id_\dimx$, (ii) perform a single stochastic optimization step.}

\revision{
Following that, the initialization strategy \eqref{eq:init_strategy} boils down to setting
\begin{equation}
    \label{eq:dps-init}
    \vpibw{k|k+1}{\bx _{k+1}}{\bx _k}[\vparam_k] = \bwp{k|k+1}{\bx _{k+1}}{\bx _k}
\end{equation}
which means that the inital mean parameter is $\vmu[0] _k \eqdef \meanBridge_{k|0,k+1}(\hpredx{k+1}(\bx _{k+1}), \bx _{k+1})$. With this initialization $\vparam_k \mapsto \kldivergence{\vpibw{k|k+1}{\bx _{k+1}}{\cdot}[\vparam_k]}{\bwp{k|k+1}{\bx _{k+1}}{\cdot}}$ has its global optimum attained at $\vparam_k = \vmu[0] _k$. Thus, the gradient \eqref{eq:loss-grad} writes $\nabla_{\vparam_k} \mathcal{L}_k(\vparam_k; \bx _{k+1}) _{|\vmu[0] _k} = - \pE \big[ \nabla_{\vparam_k} \log \hpotn{k}{\vmu_{k} + \sqrt{\var_{k|k+1}} \bZ} _{| \vmu[0] _k} \big].$
Next, updating the parameters of the variational approximation using a single stochastic gradient descent step with the gradient estimate $- \nabla_{\vparam_k} \log \hpotn{k}{\vmu_{k} + \sqrt{\var_{k|k+1}} \bZ_k} _{| \vmu[0] _k}$, where $\bZ_k \sim \gauss(\zero_\dimx, \Id_\dimx)$ and step-size
$$
\gamma_k = \frac{\zeta \stdobs^2}{\|\obs - \mathcal{A}\big(\hpredx{k}(\vmu[0] _{k} + \sqrt{\var_{k|k+1}} \bZ_k)\big)\|}
$$
yields the variational approximation
$$
    \vpibw{k|k+1}{\bx _{k+1}}{\bx _k}[\vparam^* _k] = \normpdf(\bx _k; \vmu[0]_k - \zeta \nabla_{\vparam_k} \|\obs - \mathcal{A}\big(\hpredx{k}(\vmu _{k} + \sqrt{\var_{k|k+1}} \bZ_k)\big)\|_{|\vmu[0] _k}, \var_{k|k+1} \Id_\dimx) \eqsp.
$$
In order to simply the expression we have used the fact that
$$
    \frac{- \nabla_{\vparam_k} \log \hpotn{k}{\vmu_{k} + \sqrt{\var_{k|k+1}} \bZ_k} _{| \vmu[0] _k}}{\|\obs - \mathcal{A}\big(\hpredx{k}(\vmu[0] _{k} + \sqrt{\var_{k|k+1}} \bZ_k)\big)\|} = \frac{1}{\stdobs^2}\nabla_{\vparam_k} \|\obs - \mathcal{A}\big(\hpredx{k}(\vmu _{k} + \sqrt{\var_{k|k+1}} \bZ_k)\big)\| _{| \vmu[0] _k}
    \eqsp.
$$
Therefore, given $\bX _{k+1}$, a draw from the variational approximation in \algo\ is obtained following the update
$$
    \bX _k = \tilde{\bX}_k - \zeta \nabla_{\bx _k} \| \obs - \mathcal{A}(\hpredx{k}(\bx _k))\|_{|\bx _k=\tilde{\bX}^\prime _k} \eqsp, \quad (\tilde{\bX}_k, \tilde{\bX}^\prime _k) \simiid \bwp{k|k+1}{\bX _{k+1}}{\cdot} \eqsp.
$$
On the other hand, the \dps\ transition is given by
$$
    \vpibw{k|k+1}{\bx _{k+1}}{\bx _k}[\scriptsize \mathsf{DPS}] \eqdef \normpdf(\bx _k; \vmu[0] _k - \zeta \nabla_{\bx _{k+1}} \| \obs - \mathcal{A}(\hpredx{k+1}(\bx _{k+1})) \|, \var_{k|k+1} \Id_\dimx) \eqsp.
$$
and, hence a sample $\bX ^{\scriptsize \mathsf{DPS}} _k$ is drawn following
$$
\bX ^{\scriptsize \mathsf{DPS}} _k = \tilde{\bX}_k - \zeta \nabla_{\bx _{k+1}} \| \obs - \mathcal{A}(\hpredx{k+1}(\bx _{k+1}))\|_{|\bx _{k+1}=\bX _{k+1}} \eqsp, \quad \tilde{\bX}_k \sim \bwp{k|k+1}{\bX _{k+1}}{\cdot} \eqsp.
$$
% To further draw the parallel between this special case of \algo\ and \dps, note that given $\bX _{k+1}$, a sample $\bX _k \sim \vpibw{k|k+1}{\bX _{k+1}}{\cdot}[\vparam_k]$ writes
% $$
%     \bX _k = \tilde{\bX}_k - \zeta \nabla_{\bx _k} \| \obs - \mathcal{A}(\hpredx{k}(\bx _k))\|_{|\bx _k=\tilde{\bX}^\prime _k} \eqsp, \quad \mathrm{where} \quad (\tilde{\bX}_k, \tilde{\bX}^\prime _k) \simiid \vpibw{k|k+1}{\bX _{k+1}}{\cdot}
% $$
% whereas a sample $\bX ^{\scriptsize \mathsf{DPS}} _k$ from the DPS transition writes
% $$
% \bX ^{\scriptsize \mathsf{DPS}} _k = \tilde{\bX}_k - \zeta \nabla_{\bx _{k+1}} \| \obs - \mathcal{A}(\hpredx{k+1}(\bx _{k+1}))\|_{|\bx _{k+1}=\bX _{k+1}} \eqsp, \quad \mathrm{where} \quad \tilde{\bX}_k \sim \vpibw{k|k+1}{\bX _{k+1}}{\cdot} \eqsp.
% $$
The difference between \algo\ and \dps\ is in: (i) the diffusion step used for the denoiser ($k$ for \algo, $k+1$ for \dps), (ii) the sample where the gradient is evaluated.
}

  \section{Experiments}

In this section we provide the implementation details on our algorithm as well as the algorithms we benchmark against.
We use the %privilege using 
the hyperparameters recommended by the authors and tune them on each dataset if they are not provided.

\subsection{Implementation details and hyperparameters for \algo}
\label{sec:implem_details}

We implement \Cref{algo:algo} with $\tmid{k} = \lfloor k/2 \rfloor$ and use the Adam optimizer \cite{kingma2014adam} with a learning rate of 0.03 for optimization. The number of gradient steps is adjusted based on the complexity of the task: posterior sampling with the \imagenet\ DDM prior or \ffhq\ LDM prior is more challenging and therefore requires additional gradient steps. Detailed hyperparameters are provided in \Cref{table:hyperparams-algo}.

The warm-start strategy outlined in \Cref{sec:warm_start} improved reconstruction plausibility and eliminated potential artifacts. A similar effect was observed when performing multiple gradient steps ($M=20$) during the initial stages. For latent-space models, switching the intermediate step to $\tmid{k} = k$ for the second half of the diffusion process has been crucial and significantly enhanced reconstruction quality by mitigating the smoothing effect, which often removes important details. A similar strategy has been useful for the Gaussian deblurring and motion deblurring tasks on the \imagenet\ dataset. 

% \begin{table}[ht]
%     \centering
%     \caption{The hyperparameters used in \algo\ for the considered datasets.}
%     \vspace{-0.2cm}
%     \resizebox{\textwidth}{!}{
%     \begin{tabular}{lcccccc}
%         \toprule
%         & Warm start &  $\tmid{k}$  & Learning rate & \#Gradient steps & Skip stepsize & \#Skip gradient steps \\
%         \midrule
%         \ffhq & \color{green}\checkmark &  $\lfloor k/2 \rfloor$ & $0.03$  & 2 & 10 & 20
%         \\
%         \ffhq\ (LDM) & \color{green}\checkmark & $\lfloor k/2 \rfloor \indic_{k > \lfloor n/2 \rfloor} + k \indic_{k \leq \lfloor n/2 \rfloor}$ & $0.03$ & 5 & 10 & 10
%         \\
%         \imagenet & \color{green}\checkmark & $\lfloor k/2 \rfloor$  &  $0.03$ & 2 & 20 & 10
%         \\
%         \bottomrule
%     \end{tabular}
%     \label{table:hyperparams-algo}
%     }
% \end{table}

\begin{table}[ht]
    \centering
    \caption{The hyperparameters used in \algo\ for the considered datasets.}
    \vspace{-0.2cm}
    \resizebox{\textwidth}{!}{
    \begin{tabular}{lcccccc}
        \toprule
        & Warm start & Threshold ($w$) & Diffusion steps & $\tmid{k}$  & Learning rate & Gradient steps  \\
        % \midrule
        % Gaussian mixture & \color{green}\checkmark & $700$ &  $\lfloor k/2 \rfloor$ & $0.03$  & $\ngrad_k = \begin{cases}
        %     20 &  \text{ if } k \leq n - 5 \\
        %     20 &  \text{ if } k \mod 10 = 0   \\
        %     2  &  \text{ otherwise} \\
        % \end{cases}$
        \midrule
        {\bf{\ffhq}} & \color{blue}\Large\checkmark & $\lfloor 3n/4 \rfloor$ & $n \in \{50, 100, 300\}$ & $\lfloor k/2 \rfloor$ & $0.03$  & $\ngrad_k = \begin{cases}
            20 &  \text{ if } k \geq n - 5 \\
            20 &  \text{ if } k \mod 10 = 0   \\
            2  &  \text{ otherwise} \\
        \end{cases}$
        \\
        \midrule
        %  $\lfloor k/2 \rfloor \indic_{k > \lfloor n/2 \rfloor} + k \indic_{k \leq \lfloor n/2 \rfloor}$
        {\bf \ffhq}\ (LDM) & \color{blue}\Large\checkmark &  $\lfloor 3n/4 \rfloor$ &$n \in \{50, 100, 300\}$ & $\tmid{k} = \begin{cases}
            \lfloor k / 2 \rfloor & \text{ if }  k > \lfloor n/2 \rfloor \\
            k & \text{ otherwise }
        \end{cases}$ & $0.03$ & $\ngrad_k = \begin{cases}
            20 & \text{ if } k \geq n - 5 \\
            10 & \text{ if } k \mod 10 = 0   \\
            5  & \text{ otherwise} \\
            \end{cases}$
        \\
        \midrule
        {\bf \imagenet} & \color{blue}\Large\checkmark &  $\lfloor 3n/4 \rfloor$ &$n \in \{50, 100, 300\}$ & $\lfloor k/2 \rfloor$  &  $0.03$ & $\ngrad_k = \begin{cases}
            20 &  \text{ if } k \geq n - 5 \\
            10 &  \text{ if } k \mod 20 = 0   \\
            2  &  \text{ otherwise} \\
            \end{cases}$
        \\
        \midrule
        {\bf Gaussian Mixture} & \color{blue}\Large$\times$ & -- & $n = 300$ &$\lfloor k/2 \rfloor$ & $0.1$ & $\ngrad_k = \begin{cases}
            20 &  \text{ if } k \geq n - 5 \\
            20 &  \text{ if } k \mod 10 = 0   \\
            2  &  \text{ otherwise} \\
            \end{cases}$
        \\
        \midrule
        {\bf \textsc{ptb-xl}\ (ECG)} & \color{blue}\Large\checkmark &  $\lfloor 3n/4 \rfloor$ & $n \in \{50, 300\}$ &$\lfloor k/2 \rfloor$  &  $0.03$ & $\ngrad_k = \begin{cases}
            20 &  \text{ if } k \geq n - 5 \\
            5 &  \text{ if } k \mod 20 = 0   \\
            5  &  \text{ otherwise} \\
            \end{cases}$
        \\
        \bottomrule
    \end{tabular}
    \label{table:hyperparams-algo}
    }
\end{table}

\subsection{Implementation details of competitors}
\label{sec:competitors}
\paragraph*{DPS.}
We implemented \citet[Algorithm~1]{chung2023diffusion} and refer to \citet[App.~D]{chung2023diffusion} for the values of its hyperparameters.
After tuning, we adopt $\gamma = 0.2$ for JPEG $2\%$ and $\gamma = 0.07$ for High Dynamic Range tasks.

\paragraph{DiffPIR.}
We implemented \citet[Algorithm 1]{zhu2023denoising} to make it compatible with our existing code base.
We used the hyperparameters recommended in the official, released version\footnote{\url{https://github.com/yuanzhi-zhu/DiffPIR}}.
Unfortunately, we did not manage to make the algorithm converge for nonlinear problems.
While the authors give some guidelines to handle such problems \cite[Eqn. (13)]{zhu2023denoising}, examples are missing in the paper and the released code.
Similarly, we do not run it on the motion deblur task as the FFT-based solution provided in \cite{zhu2023denoising} is only valid for circular convolution, yet we opted for the experimental setup of \citet{chung2023diffusion} which uses convolution with reflect padding.

\paragraph*{DDNM.}
We adapted the implementation in the released code\footnote{\url{https://github.com/wyhuai/DDNM}} to our code base.
Since \ddnm\ utilizes the pseudo-inverse of the degradation operator, we noticed that it is unstable for operators whose SVD are prone to numerical errors, such as Gaussian Blur with wide convolution kernel.

\paragraph*{RedDiff.}
We used the implementation of \reddiff\ available in the released code\footnote{\url{https://github.com/NVlabs/RED-diff}}.
On nonlinear problems, for which the pseudo-inverse of the observation is not available, we initialized the variational optimization with a sample from the standard Gaussian distribution. 

\paragraph*{PGDM.}
We relied on the implementation provided in \reddiff's code as some of its authors are co-authors of \pgdm.
The implementation features a subtle difference with \citet[Algorithm 1]{song2022pseudoinverse}: in the last line of the algorithm, the guidance term $g$ is multiplied by $\sqrt{\alpha}_t$ but in the implementation it is multiplied by $\sqrt{\alpha_{t-1}\alpha_t}$.
This modification stabilizes the algorithm on most tasks. For JPEG $2\%$ we found that it worsens the performance. In this case we simply multiply by $\sqrt{\alpha}_t$, as in the original algorithm. 

\paragraph*{PSLD}
We implemented the \psld\ algorithm provided in \citet[Algorithm (2)]{rout2024solving} and used the hyperparameters provided by the authors in the publicly available implementation\footnote{\url{https://github.com/LituRout/PSLD}}.

\paragraph*{ReSample}
We used the original code provided by the authors\footnote{\url{https://github.com/soominkwon/resample}} and modified it to expose several hyperparameters that were not directly accessible in the released version.
Specifically, we exposed the tolerance $\varepsilon$ and the maximum number of iterations $N$ for solving the optimization problems related to hard data consistency, as well as the scaling factor for the variance of the stochastic resampling distribution $\gamma$.
Our experiments revealed that the algorithm is highly sensitive to $\varepsilon$.
We found that setting it equal to the noise level of the inverse problem gave the best reconstruction across tasks and noise levels.
We set the maximum number of gradient iterations to $N=200$ to make the algorithm less computationally prohibitive.
Finally, we tuned $\gamma$ for each task but found it has less impact on the quality of the reconstruction compared to $\varepsilon$.

\subsubsection{Runtime and memory}
\label{sec:runtime-memory}
To get the runtime and GPU memory consumption of an algorithm on a dataset, we average these two metrics over both samples of the dataset and the considered tasks in \Cref{sec:experiments}.
% To measure the runtime, we use the \texttt{Python} function \texttt{time.perf\_counter}.

\begin{figure}[htb]
    \centering
    \subfigure{
        \includegraphics[height=0.2\textwidth]{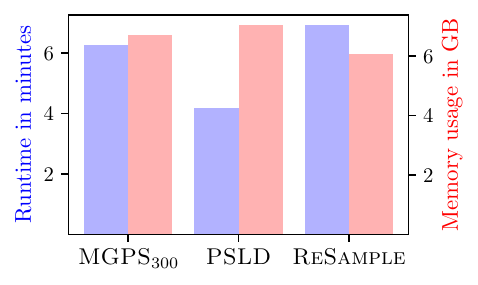}   
    }
    \subfigure{
        \includegraphics[height=.2\textwidth]{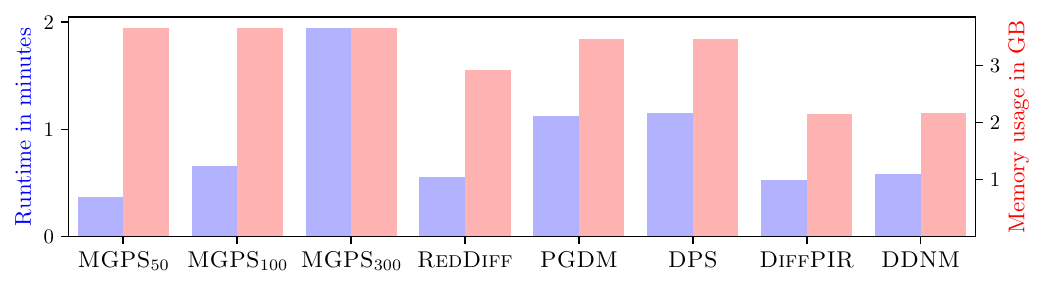}
        }
    \subfigure{
            \includegraphics[height=.2\textwidth]{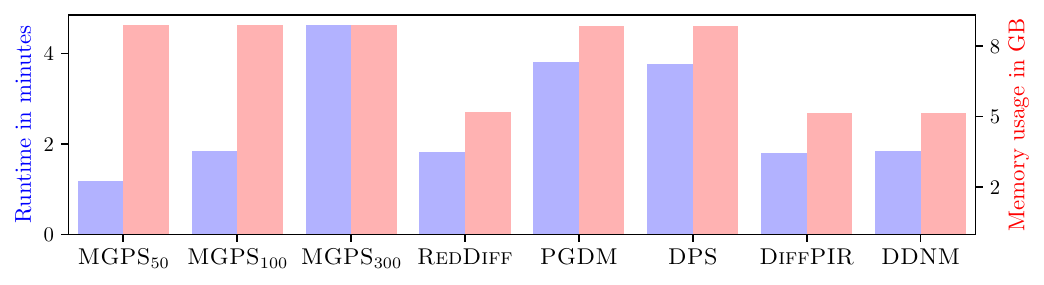}
        }
    \caption{Runtime and memory requirement of the considered algorithms on datasets: \ffhq\ latent space (1\textsuperscript{st} row), \ffhq\ pixel space (2\textsuperscript{nd} row), and \imagenet\ (3\textsuperscript{rd} row).
    The left axis displays the runtime in minutes, whereas the right axis the GPU memory requirement in Gigabytes (GB).}
    \label{fig:runtime}
\end{figure}

% \begin{figure}[!htb]
%     \centering
%     \begin{minipage}{.59\textwidth}
%         \centering
%         \includegraphics[width=0.36\textheight]{figures/runtime-gpu-ffhq.pdf}
%         \label{fig:prob1_6_2}
%     \end{minipage}%
%     \begin{minipage}{0.41\textwidth}
%         \centering
%         \includegraphics[width=0.25\textheight]{figures/runtime-gpu-ffhq-ldm.pdf}
%         \label{fig:prob1_6_1}
%     \end{minipage}
%     \caption{The runtime and GPU memory usage on \ffhq\ dataset. Left: The pixel space diffusion model. Right: The latent space diffusion model. \badr{runtime is just a placeholder}}
% \end{figure}

% \begin{figure}
%     \centering
%     \vspace{-0.2cm}
%     \includegraphics[width=.59\textwidth]{figures/runtime-gpu-imagenet.pdf}
%     \vspace{-0.1cm}
%     \captionsetup{font=small}
%     \caption{The runtime and GPU memory usage on \imagenet\ dataset.}
%     \label{fig:runtime-imagenet}
%     \vspace{-0.3cm}
% \end{figure}

% \begin{figure}
%     \centering
%     \vspace{-0.2cm}
%     \includegraphics[width=.4\textwidth]{figures/runtime-gpu-ffhq-ldm.pdf}
%     \vspace{-0.1cm}
%     \captionsetup{font=small}
%     \caption{The runtime and GPU memory usage on \ffhq\ dataset with LDM.}
%     \label{fig:runtime-ldm}
%     \vspace{-0.3cm}
% \end{figure}

% \begin{figure}
%     \centering
%     \vspace{-0.2cm}
%     \includegraphics[width=.5\textwidth]{figures/runtime-gpu-ffhq.pdf}
%     \vspace{-0.1cm}
%     \captionsetup{font=small}
%     \caption{The runtime and GPU memory usage on \ffhq\ dataset.}
%     \label{fig:runtime-ffhq}
%     \vspace{-0.3cm}
% \end{figure}

\subsection{Gaussian mixtures}
\label{sec:gm-appendix}

% --- locals
\def\mean{\boldsymbol{m}}
\def\cov{\sigma^2}
\def\covPostior{\bar{\mathbf{\Sigma}}}
\def\likelihood{\bfA}

In this section, we elaborate more on the Gaussian mixture experiment in \Cref{sec:experiments}, where we consider a linear inverse problem with a Gaussian mixture as prior.
Therefore, the likelihood is $\potn{}{}: \bx \mapsto \normpdf(\obs; \likelihood \bx, \stdobs^2 \Id_\dimobs)$.
Recall that a Gaussian mixture with $C \in \nset$ components, whose weights, means, and covariances are $w_i > 0$, $\mean_i \in \rset^{\dimx}$, and $\cov_i \Id_{\dimx}$, respectively, has the density
$$
\prior(\bx) = \sum_{i=1}^{C} w_i \normpdf(\bx; \mean_{i}, \cov_i \Id_{\dimx})
\eqsp.
$$
% In this section, we first derive the expressions of both the DDPM denoiser and the posterior.
% Then we describe the experimental setup.
\paragraph{Denoiser.}
The denoiser is obtained via Tweedie's formula \cite{robbins1956empirical},
% We leverage Tweedie's formula \cite{robbins1956empirical} to get the expression of the denoiser:
$$
\predx{k}(\bx _k) = \big( \bx _k + (1 - \acp{k}) \nabla \log \prior_k(\bx _k) \big) / \sqrt{\acp{k}},
$$
where the densities of the marginals $(\prior_k)_{k=1}^{n}$ are straightforward,
\begin{equation*}
    \begin{aligned}
        \prior_k(\bx _k)
            & = \int \prior(\bx _0) \fwtrans{k|0}{\bx _0}{\bx _k} \ \rmd \bx _0
            = \sum_{i=1}^{C} w_i \int \normpdf(\bx _0; \mean_{i}, \cov_i \Id_{\dimx}) \fwtrans{k|0}{\bx _0}{\bx _k} \ \rmd \bx _0 \\
            & = \sum_{i=1}^{C} w_i  \normpdf(\bx _k; \sqrt{\acp{k}} \mean_{i}, (\acp{k}\cov_i + \var_k) \Id_{\dimx}) \eqsp.  \\
    \end{aligned}
\end{equation*}

\paragraph{Posterior.}
Furthermore, following \citet[Eqn.~2.116]{bishop2006pattern}, the posterior can be shown to be a Gaussian mixture
\begin{equation*}
    \post{}{\bx} \propto g(\bx) \prior(\bx) \propto \sum_{i=1}^{C} \bar{w}_i \normpdf(\bx; \bar{\mean}_{i}, \covPostior_i)
    \eqsp,
\end{equation*}
with new weights, covariances, and means given by
\begin{equation*}
    \begin{aligned}
        \bar{w}_i       & = w_i \normpdf(\obs; \likelihood \mean_i, \stdobs^2 \Id_{\dimx_{\obs}} + \cov_i \likelihood \likelihood^\top)\eqsp,\\
        \covPostior_i   & = \big( (1 / \cov_i) \Id_{\dimx} + (1 / \stdobs^2)\likelihood^\top \likelihood \big)^{-1}\eqsp,\\
        \bar{\mean}_{i} & = \covPostior_i \big( (1 / \stdobs^2) A^\top \obs + (1 / \cov_i) \mean_i \big)\eqsp, \\
    \end{aligned}
\end{equation*}
where the weights are un-normalized.

\paragraph{Experimental setup.}
We consider two setups where $\dimx \in \{ 20, 200 \}$ and generate Gaussian mixtures of $C = 25$ components with means $(\mean_i)_{i=1}^{C} = \{ (8i, 8j, \ldots, 8i, 8j) \in \rset^{\dimx}: (i, j) \in \intset{-2}{2}^2 \}$, unit covariances, and weights $w_i$ drawn from a uniform distribution on $[0, 1]$ then normalized to sum to $1$.
The linear inverse problem $(\obs, \likelihood)$ is generated by first drawing the matrix $\likelihood \in \rset^{1 \times \dimx}$ with entries \iid\ from $\gauss(0, 1)$, then sampling $\bx ^*$ from the prior $\prior$ and finally computing $\obs = \likelihood \bx ^* + \stdobs \varepsilon$ with $\varepsilon \sim \gauss(0, 1)$ to get the observation.
The standard deviation of the inverse problem is fixed to $\stdobs = 0.05$ across all problem instances.

To assess the performance of each algorithm, we draw $2000$ samples and compare against $2000$ samples from the true posterior distribution using the Sliced Wasserstein (SW) distance by averaging over $10^4$ slices.
In \Cref{table:sw-gm} we report the average SW and the 95\% confidence interval over $100$ replicates.

\subsection{Extended image experiments}

\revision{To strengthen our conclusions and calculate the FID, we reran \algo\ and the two closest competitors on 1000 images for the following five tasks: half mask, JPEG, motion deblur, nonlinear deblur, and high dynamic range, using the three priors (\ffhq, \imagenet, \ffhq-LDM). The \Cref{table:extended-metrics-1k} shows that \algo\ significantly outperforms the other competitors, including for the FID, with no significant change in other metrics compared to what we calculated on a smaller dataset.}

We extend, in Tables \ref{table:extended-metrics-ffhq} and \ref{table:extended-metrics-imagenet}, the results in \Cref{table:combined-lpips} by including \algo\ with $n = 50$ and $n=100$. In the setting $n = 100$, the runtime of \algo\ is twice lower than that of \dps, \pgdm\ and comparable to that of \diffpir, \ddnm\ and \reddiff, see \Cref{fig:runtime}. It is also outperforming them on most of the tasks, especially the nonlinear ones, as is seen in the table below. In the setting $n = 50$, \algo\ has the lowest runtime among all the methods while maintaining competitive performance. 

Finally, in \Cref{subsec:sample-images}, we present sample reconstructions on the various tasks. For each algorithm, we generate five samples and select the four most visually appealing ones. Completely black or white images correspond to failure cases of \dps. 

\begin{table}[ht]
    \centering
    \vspace{0pt}
    \captionsetup{font=small}
    \captionof{table}{Mean and confidence intervals of LPIPS/PSNR/SSIM values and FID value for various linear and nonlinear imaging tasks on 1k images for the 3 priors. Best is in \textbf{bold}.}
    \resizebox{\textwidth}{!}{
        \begin{tabular}{l ccc c ccc c cc }
            \toprule
             & \multicolumn{3}{c}{\ffhq} && \multicolumn{3}{c}{\imagenet} && \multicolumn{2}{c}{\ffhq-LDM} \\
    %      \cmidrule(lr){2-4} \cmidrule(lr){6-8} \cmidrule(lr){10-11}
             Metric & \algo\ & \ddnm & \diffpir && \algo\ & \ddnm & \diffpir && \algo\ & \resample \\
             \midrule
         \multicolumn{11}{c}{Half mask}  \\
        \midrule
            FID & \textbf{27.0} & 38.6 & 45.2& & \textbf{40.0} & 50.0 & 57.0 & & \textbf{49.5} & 66.6\\
            LPIPS & $\mathbf{0.19\pm0.00}$ & $0.23\pm0.00$ & $0.25\pm0.00$& & $\mathbf{0.30\pm0.00}$ & $0.38\pm 0.01$ & $0.40\pm0.01$ & &$\mathbf{0.26\pm 0.00}$ & $0.30\pm0.00$ \\
            PNSR & $15.9\pm0.2$ & $\mathbf{16.3\pm0.2}$ & $16.1\pm0.3$& & $15.0\pm0.1$ & $\mathbf{16.0\pm0.1}$ & $15.8\pm0.1$ & &$\mathbf{15.6\pm0.1}$ & $15.7\pm0.1$ \\
            SSIM & $0.70\pm0.00$ & $\mathbf{0.74\pm0.00}$ & $0.72\pm0.01$& & $0.63\pm0.00$ & $\mathbf{0.68\pm0.01}$ & $0.67\pm0.01$ & &$\mathbf{0.69\pm0.00}$ & $0.67\pm0.00$
            \\
                        \midrule \\
            \\
            & \algo\ & \ddnm & \diffpir && \algo\ & \ddnm & \diffpir && \algo\ & \resample \\
        \midrule
         \multicolumn{11}{c}{Motion deblur}  \\
        \midrule
            FID & \textbf{29.7} & 36.7 & 77.0& & \textbf{35.3} & 55.0 & 87.3 & &\textbf{44.6} & 51.8 \\
            LPIPS & $\mathbf{0.12\pm0.00}$ & $0.17\pm0.00$ & $0.22\pm0.00$& & $\mathbf{0.20\pm0.01}$ & $0.40\pm0.01$ & $0.39\pm0.01$ & &$\mathbf{0.19\pm0.00}$ & $0.20\pm0.00$ \\
            PNSR & $26.7\pm0.1$ & $24.1\pm0.1$ & $\mathbf{27.4\pm0.1}$& & $\mathbf{24.4\pm0.1}$ & $21.4\pm0.1$ & $24.2\pm0.1$ & &$26.4\pm0.1$ & $\mathbf{26.7\pm0.1}$ \\
            SSIM & $\mathbf{0.77\pm0.00}$ & $0.70\pm0.01$ & $0.71\pm0.00$ & & $\mathbf{0.67\pm0.01}$ & $0.55\pm0.01$ & $0.61\pm0.00$ & &$\mathbf{0.76\pm0.00}$ & $0.72\pm0.00$
            \\
             \midrule
            \\
         \multicolumn{11}{c}{JPEG (QF = 2)}  \\
        \midrule
            FID & \textbf{31.6} & 87.6 & 109& & \textbf{61.4} & 128.8 & 92.8 & &\textbf{45.0} & 65.3 \\
            LPIPS & $\mathbf{0.15\pm0.00}$ & $0.37\pm0.00$ & $0.33\pm0.01$ & & $\mathbf{0.40\pm0.01}$ & $0.60\pm0.01$ & $0.61\pm0.00$ & & $\mathbf{0.21\pm0.00}$ & $0.26\pm0.01$ \\
            PNSR & $\mathbf{25.2\pm0.1}$ & $19.0\pm0.2$ & $24.5\pm0.1$& & $\mathbf{22.2\pm0.1}$ & $16.7\pm0.1$ & $22.2\pm0.1$ & &$24.6\pm0.1$ & $\mathbf{24.8\pm0.1}$ \\
            SSIM & $\mathbf{0.73\pm0.01}$ & $0.55\pm0.02$ & $0.70\pm0.00$& & $\mathbf{0.60\pm0.01}$ & $0.41\pm0.02$ & $0.60\pm0.01$ & &$\mathbf{0.71\pm0.00}$ & $0.66\pm0.01$
            \\
            \midrule
            \\
         \multicolumn{11}{c}{Nonlinear deblur}  \\
        \midrule
            FID & \textbf{50.8} & 164 & 88.4& & 113 & 272 & \textbf{112} & &$\mathbf{69.2}$ & 71.5 \\
            LPIPS & $\mathbf{0.23\pm0.01}$ & $0.51\pm 0.02$ & $0.68\pm0.01$& & $\mathbf{0.43\pm0.01}$ & $0.83\pm0.01$ & $0.66\pm0.01$ & &$\mathbf{0.26\pm0.01}$ & $0.32\pm0.01$ \\
            PNSR & $\mathbf{24.3\pm0.2}$ & $16.2\pm0.5$ & $21.9\pm0.1$& & $\mathbf{22.2\pm0.2}$ & $9.9\pm0.4$ & $20.7\pm0.2$ & & $23.9\pm0.1$ & $\mathbf{24.2\pm0.1}$ \\
            SSIM & $\mathbf{0.70\pm0.01}$ & $0.45\pm0.02$ & $0.42\pm0.01$& & $\mathbf{0.58\pm0.01}$ & $0.41\pm0.01$ & $0.241\pm0.01$ & &$\mathbf{0.69\pm0.01}$ & $0.67\pm0.01$
            \\
            \midrule
            \\
         \multicolumn{11}{c}{High dynamic range}  \\
        \midrule
            FID & \textbf{20.9} & 153 & 47.5& & \textbf{20.2} & 316 & 35.7 & &44.2 & \textbf{38.7} \\
            LPIPS & $\mathbf{0.08\pm0.01}$ & $0.40\pm0.04$ & $0.20\pm0.01$& & $\mathbf{0.11\pm0.01}$ & $0.83\pm0.02$ & $0.20\pm0.01$ & &$0.14\pm0.00$ & $\mathbf{0.12\pm0.00}$ \\
            PNSR & $\mathbf{27.0\pm0.1}$ & $18.7\pm0.2$ & $21.7\pm0.1$& & $\mathbf{26.3\pm0.2}$ & $9.9\pm0.2$ & $21.9\pm0.1$ & &$25.5\pm0.1$ & $\mathbf{26.0\pm0.1}$ \\
            SSIM & $\mathbf{0.83\pm0.01}$ & $0.55\pm0.04$ & $0.72\pm0.01$& & $\mathbf{0.83\pm0.01}$ & $0.23\pm0.02$ & $0.71\pm0.01$ & &$0.80\pm0.01$ & $\mathbf{0.83\pm0.01}$
            \\
            \bottomrule
        \end{tabular}
        \label{table:extended-metrics-1k}
        }
\end{table}

\begin{table}[ht]
    \centering
    \caption{Mean LPIPS/PSNR/SSIM values for various linear and nonlinear imaging tasks on the \textsc{FFHQ} $256 \times 256$ dataset. Best is in \textbf{bold} and second best is \underline{underlined}.}
    \resizebox{0.9\textwidth}{!}{
    \begin{tabular}{lccccccccc}
        \toprule
        Task & \algo${}_{50}$ &\algo${}_{100}$ & \algo${}_{300}$ & \dps & \pgdm & \ddnm & \diffpir & \reddiff \\
        % --- lpips
        \midrule
        \multicolumn{9}{c}{LPIPS \ $\downarrow$} \\
        \midrule
        SR ($\times 4$) & 0.13 & \underline{0.10} &\bf{0.09} & \bf{0.09} & 0.33 & 0.14 & 0.13 & 0.36 \\
        SR ($\times 16$) & 0.27 & \underline{0.26} & \underline{0.26} & \bf{0.24} & 0.44 & 0.30 & 0.28 & 0.51 \\
        Box inpainting & 0.16 & \underline{0.12} & \bf{0.10} & 0.19 & 0.17 & \underline{0.12} & 0.18 & 0.19 \\
        Half mask & 0.24 & \underline{0.22} & \bf{0.20} & 0.24 & 0.26 & \underline{0.22} &  0.23 & 0.28  \\
        Gaussian Deblur & 0.21 & 0.18 & \underline{0.15} & 0.16 & 0.87 & 0.19 & \bf{0.12} & 0.26 \\
        Motion Deblur & 0.19 & \underline{0.15} & \bf{0.13} & 0.16 & $-$ & $-$ & $-$ & 0.21 \\
        \\ % \cmidrule(lr){1-9}
        JPEG (QF = 2) & 0.20 & \underline{0.17} & \bf{0.16} & 0.39 & 1.10 & $-$ & $-$ & \underline{0.32}\\
        Phase retrieval & 0.20 & \underline{0.14} &\bf{0.11} & 0.46 & $-$ & $-$ & $-$ & \underline{0.25} \\
        Nonlinear deblur & \bf{0.23} & \bf{0.23} &\bf{0.23} & \underline{0.52} & $-$ & $-$ & $-$ & 0.66 \\
        High dynamic range & 0.13 & \underline{0.09} &\bf{0.07} & 0.49 & $-$ & $-$ & $-$ & \underline{0.20} \\
        % --- psnr
        \midrule
        \multicolumn{9}{c}{PSNR \ $\uparrow$}  \\
        \midrule
        SR ($\times 4$)    & 27.83 & 27.79 & 27.79 & \underline{28.24} & 23.34 & \textbf{29.52} & 27.17 & 27.25 \\
        SR ($\times 16$)   & 20.45 & 20.34 & 20.22 & 20.67 & 17.65 & \textbf{22.43} & 20.75 & \underline{21.91} \\
        Box inpainting     & 21.55 & 22.22 & \textbf{22.68} & 18.39 & 21.13 & \underline{22.35} & 21.96 & 21.79 \\
        Half mask          & 15.10 & 15.32 & 15.54 & 14.82 & 16.03 & \underline{16.16} & 15.17 & \textbf{16.21} \\
        Gaussian Deblur    & 25.09 & 25.19 & 25.89 & 24.20 & 13.36 & \underline{26.69} & 25.89 & \textbf{26.72} \\
        Motion Deblur      & 26.07 & \underline{26.64} & 26.48 & 24.24 &  $-$  &  $-$  &  $-$  & \textbf{27.58} \\
        \\ % \cmidrule(lr){1-9}
        JPEG (QF = 2)      & \underline{25.00} & \textbf{25.23} & 24.94 & 18.50 & 12.76 &$-$&$-$& 24.42 \\
        Phase retrieval    & 24.20 & \underline{26.60} & \textbf{27.25} & 14.87 &  $-$  &$-$&$-$& 24.85 \\
        Nonlinear deblur   & 24.16 & \underline{24.21} & \textbf{24.37} & 15.89 &  $-$  &$-$&$-$& 21.97 \\
        High dynamic range & 24.77 & \underline{26.07} & \textbf{27.74} & 16.83 &  $-$  &$-$&$-$& 21.25 \\
        % -- ssim
        \midrule
        \multicolumn{9}{c}{SSIM \ $\uparrow$} \\
        \midrule
        SR ($\times 4$)  & 0.81 & 0.80 & 0.79 & \underline{0.81} & 0.50 & \textbf{0.85} & 0.77 & 0.70 \\
        SR ($\times 16$) & 0.58 & 0.57 & 0.55 & 0.59 & 0.38 & \textbf{0.67} & 0.60 & \underline{0.62} \\
        Box inpainting   & 0.80 & 0.81 & \underline{0.82} & 0.76 & 0.70 & \textbf{0.83} & 0.80 & 0.70 \\
        Half mask        & 0.69 & 0.70 & \underline{0.71} & 0.66 & 0.56 & \textbf{0.73} & 0.67 & 0.65 \\
        Gaussian Deblur  & 0.71 & 0.73 & 0.75 & 0.69 & 0.14 & \textbf{0.77} & 0.73 & \underline{0.76} \\
        Motion Deblur    & \underline{0.77} & \textbf{0.78} & \underline{0.77} & 0.71 & $-$  & $-$  & $-$  & 0.71 \\
        \\ % \cmidrule(lr){1-9}
        JPEG (QF = 2)      & \textbf{0.75} & \underline{0.74} & 0.73 & 0.51 & 12.76 &$-$&$-$& 0.71 \\
        Phase retrieval    & 0.74 & \underline{0.78} & \textbf{0.79} & 0.43 &  $-$  &$-$&$-$& 0.61 \\
        Nonlinear deblur   & \textbf{0.70} & \textbf{0.70} & \underline{0.69} & 0.46 &  $-$  &$-$&$-$& 0.42 \\
        High dynamic range & 0.79 & \underline{0.81} & \textbf{0.86} & 0.48 &  $-$  &$-$&$-$& 0.71 \\
        \bottomrule
    \end{tabular}
    \label{table:extended-metrics-ffhq}
    }
\end{table}

\begin{table}[ht]
    \centering
    \caption{Mean LPIPS/PSNR/SSIM values for various linear and nonlinear imaging tasks on the \textsc{Imagenet} $256 \times 256$ dataset. Best is in \textbf{bold} and second best is \underline{underlined}.}
    \resizebox{0.9\textwidth}{!}{
    \begin{tabular}{lccccccccc}
        \toprule
        Task & \algo${}_{50}$ &\algo${}_{100}$ & \algo${}_{300}$ & \dps & \pgdm & \ddnm & \diffpir & \reddiff \\
        % --- lpips
        \midrule
        \multicolumn{9}{c}{LPIPS \ $\downarrow$} \\
        \midrule
        SR ($\times 4$) & 0.36 & \underline{0.33} &\bf{0.30}  & 0.41  & 0.78 & 0.34 & 0.36 & 0.56 \\
        SR ($\times 16$) & 0.58 & 0.55 &\underline{0.53}  &  \bf{0.50} & 0.60 & 0.70 & 0.63  & 0.83 \\
        Box inpainting & 0.31 & \underline{0.26} &\bf{0.22} &  0.34 & 0.29 & 0.28 & 0.28  & 0.36  \\
        Half mask & 0.39 & \underline{0.34} &\bf{0.29} & 0.44 & 0.38 & 0.38 & 0.35  & 0.44 \\
        Gaussian Deblur & 0.35 & \bf{0.29} & \underline{0.32} & 0.35 & 1.00 & 0.45 & \bf{0.29} & 0.52 \\
        Motion Deblur & 0.35 & \underline{0.25} & \bf{0.22} & 0.39 & $-$ & $-$ & $-$ & 0.40 \\
       % \cmidrule(lr){1-9}
        JPEG (QF = 2) & 0.50 & \underline{0.46} &\bf{0.42} & 0.63 & 1.31 & $-$ & $-$ & 0.51 \\
        Phase retrieval & 0.54 & \underline{0.52} &\bf{0.47}  & 0.62 & $-$ & $-$ & $-$ & 0.60 \\
        Nonlinear deblur & 0.49 & \underline{0.47} &\bf{0.44} & 0.88  & $-$ & $-$ & $-$ & 0.67 \\
        High dynamic range & 0.22 & \underline{0.15} &\bf{0.10} & 0.85  & $-$ & $-$ & $-$ & 0.21 \\
        % --- psnr
        \midrule
        \multicolumn{9}{c}{PSNR \ $\uparrow$}  \\
        \midrule
        SR ($\times 4$)   & 24.68 & 24.70 & \underline{24.77} & 23.52 & 15.67 & \textbf{25.55} & 24.26 & 24.24 \\
        SR ($\times 16$)  & 18.56 & 18.42 & 18.04 & 18.22 & 15.80 & \textbf{20.43} & 19.37 & \underline{19.95} \\
        Box inpainting    & 17.52 & 17.95 & 18.25 & 14.34 & 17.35 & \textbf{20.08} & \underline{19.77} & 18.90 \\
        Half mask         & 14.98 & 15.14 & 15.60 & 14.65 & 14.36 & \textbf{17.06} & 15.79 & \underline{16.96} \\
        Gaussian Deblur   & 22.56 & 21.96 & 20.10 & 21.20 & 9.93  & \textbf{23.29} & 22.10 & \underline{23.27} \\
        Motion Deblur     & 23.91 & \textbf{25.03} & \underline{24.50} & 21.59 &  $-$  &  $-$  &  $-$  & 24.43 \\
        \\ % \cmidrule(lr){1-9}
        JPEG (QF = 2)      & 21.96 & \underline{22.17} & \textbf{22.44} & 16.11 & 5.29 &$-$&$-$& 22.15 \\
        Phase retrieval    & 16.36 & \underline{16.94} & \textbf{18.10} & 14.40 &  $-$ &$-$&$-$& 15.78 \\
        Nonlinear deblur   & 21.89 & \underline{22.23} & \textbf{22.36} & 8.49  &  $-$ &$-$&$-$& 20.76 \\
        High dynamic range & 23.93 & \underline{25.64} & \textbf{27.04} & 9.32  &  $-$ &$-$&$-$& 22.88 \\
        % -- ssim
        \midrule
        \multicolumn{9}{c}{SSIM \ $\uparrow$} \\
        \midrule
        SR ($\times 4$)   & 0.65 & \underline{0.66} & \underline{0.66} & 0.60 & 0.23 & \textbf{0.70} & 0.63 & 0.61 \\
        SR ($\times 16$)  & 0.41 & 0.38 & 0.35 & 0.40 & 0.24 & \textbf{0.50} & 0.46 & \underline{0.47} \\
        Box inpainting    & 0.71 & 0.74 & \underline{0.76} & 0.70 & 0.62 & \textbf{0.78} & 0.74 & 0.67 \\
        Half mask         & 0.61 & 0.63 & \underline{0.65} & 0.55 & 0.53 & \textbf{0.69} & 0.63 & 0.62 \\
        Gaussian Deblur   & 0.56 & 0.53 & 0.44 & 0.51 & 0.07 & \textbf{0.60} & 0.51 & \underline{0.57} \\
        Motion Deblur     & 0.64 & \textbf{0.69} & \underline{0.65} & 0.55 &  $-$ &  $-$ &  $-$ & 0.60 \\
        \\ % \cmidrule(lr){1-9}
        JPEG (QF = 2)      & 0.59 & \textbf{0.60} & \textbf{0.60} & 0.39 & 0.01 &$-$&$-$& 0.59 \\
        Phase retrieval    & 0.37 & \underline{0.40} & \textbf{0.43} & 0.29 &  $-$ &$-$&$-$& 0.26 \\
        Nonlinear deblur   & \underline{0.57} & \textbf{0.58} & \textbf{0.58} & 0.24 &  $-$ &$-$&$-$& 0.41 \\
        High dynamic range & 0.75 & \underline{0.81} & \textbf{0.84} & 0.25 &  $-$ &$-$&$-$& 0.73 \\
        % -- ssim
        \bottomrule
    \end{tabular}
    \label{table:extended-metrics-imagenet}
    }
\end{table}
\begin{table}[ht]
    \centering
    \vspace{0pt}
    \captionsetup{font=small}
    \captionof{table}{Mean LPIPS/PSNR/SSIM values for various linear and nonlinear imaging tasks on \ffhq \ $256 \times 256$ dataset with LDM prior. Best is in \textbf{bold} and second best is \underline{underlined}.}
    \resizebox{.9\textwidth}{!}{
        \begin{tabular}{l ccc c ccc c ccc }
            \toprule
             & \algo\ & \resample & \psld && \algo\ & \resample & \psld && \algo\ & \resample & \psld \\
            \cmidrule(lr){2-4} \cmidrule(lr){6-8} \cmidrule(lr){10-12}
            Task & \multicolumn{3}{c}{LPIPS \ $\downarrow$} && \multicolumn{3}{c}{PSNR \ $\uparrow$} && \multicolumn{3}{c}{SSIM \ $\uparrow$} \\
            \midrule
            SR ($\times 4$) & \bf{0.11} & \underline{0.20} &  0.22 && \textbf{28.46} & \underline{26.08} & 25.53 && \textbf{0.83} & 0.69 & \underline{0.70}  \\
            SR ($\times 16$) & \bf{0.30} & 0.36  &  \underline{0.35} && 20.64 & \underline{21.09} & \textbf{21.42} && \underline{0.57} & 0.56 & \textbf{0.63}  \\
            Box inpainting & \bf{0.16} & \underline{0.22} & 0.26 && \textbf{22.94} & 18.80 & \underline{20.39} && \textbf{0.79} & \underline{0.75} & 0.66  \\
            Half mask & \bf{0.25} & \underline{0.30} & 0.31 && \textbf{15.11} & 14.59 & \underline{14.75} && \textbf{0.69} & \underline{0.67} & 0.61  \\
            Gaussian Deblur & \underline{0.16} & \bf{0.15} & 0.35  && \textbf{27.57} & \underline{27.44} & 19.95 && \textbf{0.79} & \underline{0.75} & 0.47  \\
            Motion Deblur & \bf{0.18} & \underline{0.19} & 0.41  && \underline{26.49} & \textbf{26.85} & 18.14 && \textbf{0.77} & \underline{0.72} & 0.39  \\
            \\ % \cmidrule(lr){1-4}
            JPEG (QF = 2) & \bf{0.20} & 0.26 & $-$  && \textbf{24.75} & 24.33 &$-$ && \textbf{0.72} & 0.67 &$-$ \\
            Phase retrieval & \bf{0.34} & 0.41 &  $-$  && \textbf{22.21} & 19.05 &$-$ && \textbf{0.62} & 0.47 &$-$ \\
            Nonlinear deblur & \bf{0.26} & 0.30 & $-$  && 23.79 & \textbf{24.44} &$-$ && \textbf{0.70} & 0.68 &$-$ \\
            High dynamic range & \bf{0.15} & \bf{0.15} & $-$ && 25.17 & \textbf{25.42} & $-$ && 0.79 & \textbf{0.81} & $-$ \\
            \bottomrule
        \end{tabular}
        \label{table:extended-metrics-ffhq-ldm}
        }
\end{table}
\label{sec:img-appendix}
\subsection{ECG experiments}
\label{sec:ecg-appendix}
\subsubsection{Implementation details}
\paragraph{\textsc{ptb-xl} dataset}
We use 12-lead ECGs at a sampling frequency of 100 Hz from the \textsc{ptb-xl} dataset \cite{ptbxl}. For both training and generation, we do not use the augmented limb leads aVL, aVR and aVF as they can be obtained with the following relations: aVL=(I-III)/2, aVF=(II+III)/2 and aVR=-(I+II)/2. This leads to inputs of shape $T \times 9$ instead of $T \times 12$ (where $T$ is the length of the signal). For evaluation metrics, we reconstruct the augmented limb leads.
For the MB task, following \cite{alcarazImputation}, we use 2.56-second ECG random crops, hence the inputs are of shape $256\times 9$. For the ML task, we use the full 10-second ECGs and pad them into a tensor of shape $1024\times 9$. \Cref{table:shape-ecg} summarizes the input shapes for models and algorithms for each task. \Cref{table:ptbxl-splits} summarizes the distribution of train/val/test splits and the number of cardiac conditions (RBBB, LBBB, AF, SB) per split. Note that for posterior sampling, we only use the test set, and for both training and sampling we never use the cardiac condition.

\begin{table}[ht]
    \centering
    \caption{\small ECG size and input-shape per task.}
    \resizebox{0.6\textwidth}{!}{
    \begin{tabular}{lcccc}
        \toprule
       Task & ECG size (seconds) & Total leads & Used leads & Input shape\\
        \midrule
        MB & 2,56 & 12 & I, II, III, V1--6 & $256 \times 9$\\
        ML & 10 & 12 & I, II, III, V1--6 & $1024 \times 9$\\
        \bottomrule
    \end{tabular}
    \label{table:shape-ecg}
    }
\end{table}

\begin{table}[ht]
    \centering
    \caption{\small \textsc{ptb-xl} dataset description.}
    \resizebox{0.45\textwidth}{!}{
    \begin{tabular}{lccccc}
        \toprule
        Split & All & RBBB & LBBB & AF & SB \\
        \midrule
        Train & 17,403 & 432 & 428 & 1211 & 503\\
        Val & 2,183 & 55 & 54 & 151 & 64\\
        Test & 2,203 & 54 & 54 & 152 & 64\\
        \bottomrule
    \end{tabular}
    \label{table:ptbxl-splits}
    }
\end{table}

\paragraph{Diffusion model}
Following \cite{alcaraz2023diffusion,alcarazImputation} we used Structured State Space Diffusion (SSSD) models \cite{Gu2022SSSD,goel2022sashimi}.
SSSD generates a sequence $u(t)$ with the following differential equation denpending on the input seuqnece $\bx(t)$ and a hidden state $h(t)$:
\begin{align}
    h'(t) &= Ah(t) + Bx(t)\\
    u(t) &= Ch(t) + Dx(t)\eqsp,
\end{align}
where $A,B,C,D$ are transition matrices. We use the SSSD$^{SA}$ \cite{alcarazImputation} (also denoted as Sashimi \cite{goel2022sashimi}) architecture publicly available\footnote{\url{https://github.com/AI4HealthUOL/SSSD}}\footnote{\url{https://github.com/albertfgu/diffwave-sashimi/tree/master}}. We parametrize the matrix $A=\Lambda -pp^*$ where $p$ is a vector and $\Lambda$ is a diagonal matrix. This parametrization with  facilitates the control of the spectrum of $A$ to enforce stability (negative real part of eigenvalues).
We use a multi-scale architecture, with a stack of residual S4 blocks. The top tier processes the raw audio signal at its original sampling rate, while lower tiers process downsampled versions of the input signal. The outputs of the lower tiers are upsampled with up-pooling and down-pooling operations and combined with the input to the upper tier to provide a stronger conditioning signal. Hyper-parameters are described in \cref{table:params-ecg-SSSD}. The selected model is the model of the last epoch.
Training time is 30 hours for 2.5 seconds ECGs, 42 hours for 10 seconds ECGs.
\begin{table}[ht]
    \centering
    \caption{\small ECG diffusion generative model hyper-parameters.}
    \resizebox{0.4\textwidth}{!}{
    \begin{tabular}{lc}
        \toprule
        Hyper-parameter & Value \\
        \midrule
        Residual layers & 4 \\
Pooling factor & [1, 2, 2] \\
Feature expansion & 2 \\
Diffusion embedding dim. 1 & 128\\
Diffusion embedding dim. 2 & 512\\
Diffusion embedding dim. 3 & 512\\
Diffusion steps & 1000 \\
Optimizer & Adam \\
Number of iterations & 150k \\
Loss function & MSE \\
Learning rate & 0.002 \\
Batch size & 128 \\
Number of parameters & 16 millions \\
        \bottomrule
    \end{tabular}
    \label{table:params-ecg-SSSD}
    }
\end{table}
\paragraph{Algorithm parameters} We use the same parameters as described in \cref{table:hyperparams-algo} for \algo\ and the implementation described in \cref{sec:competitors} for competitors. Since we had to run the experiments over 2k samples, we set the number of diffusion steps such that the runtime approximates 30 seconds for generating 10 samples. This leads to the parameters shown in \cref{table:params-ecg}. For the ML experiment, we also tested \algo\ with 300 diffusion steps, which resulted in a runtime of 5 minutes and 30 seconds per batch of 10 samples. The results improved, but we already outperform competitors with just 50 diffusion steps and a 30-second runtime.
\begin{table}[ht]
    \centering
    \caption{\small Number of diffusion steps used in posterior sampling algorithms for ECG tasks.}
    \resizebox{0.6\textwidth}{!}{
    \begin{tabular}{ccccccc}
        \toprule
        \algo${}_{50}$ & \algo${}_{300}$  & \dps & \pgdm & \ddnm & \diffpir & \reddiff \\
        \midrule
         50 & 300 & 400 & 200 & 200 & 500 & 400 \\
        \bottomrule
    \end{tabular}
    \label{table:params-ecg}
    }
\end{table}
\paragraph{Evaluation metrics}
\label{sec:ecg-metrics}
For the MB task, for each observation, we generate 10 samples - instead of 100 as in in \cite{alcarazImputation}. We then compute the Mean Absolute Error (MAE) and Root Mean Squared Error (RMSE) between each generated sample and the ground-truth. The final score is the average of these errors over the 10 generated samples. We report the confidence intervals over the test set in \cref{table:rmse-ecg} and \cref{table:rmse-ecg-RMB}.
% For MB task we use the same evaluation metric as \cite{alcarazImputation} defined as follow:
% \begin{equation}
%     \frac{1}{N}\sum_{n=1}^N(\frac{2}{LT}\sum_{t=1}^{T}\sum_{l=1}^{L}\|(\hat{\bx}^n[t, l]-\bx[t, l])m[t, l]\|^p)^{\frac{1}{p}}\eqsp,
% \end{equation}
% where $p=1$ for Mean Absolute Error (MAE) and $p=2$ of Root Mean Squared Error (RMSE), and $N=10$ is the number of generated samples per observation, $T=256$ is the ECG length, $L=12$ is the number of leads, $(\hat{\bx}^n)_n$ are the generated samples, $\bx$ is the ground-truth $m$ is the mask of the observation.

For the ML task, we use a classifier trained to detect four cardiac conditions: Right Bundle Branch Block (RBBB), Left Bundle Branch Block (LBBB), Atrial Fibrillation (AF), and Sinus Bradycardia (SB) on the \textsc{ptb-xl} dataset. We follow the \textsc{XResNet1d50} described in \cite{Strodthoff2020ecgbenchmarking} with hyper-parameters reported in \cref{table:resnet-ecg}.
We apply the classifier to the ground-truth ECG of the test set and to the samples generated from lead I. As in the MB task, for each observation, 10 samples are generated. The output of the classifier is averaged over these 10 samples.
For each cardiac condition, we compute balanced accuracy to account for class imbalance (see \cref{table:ptbxl-splits}). The classification threshold is selected using \textsc{ptb-xl} validation-set.
\begin{table}[ht]
    \centering
    \caption{\small \textsc{XResNet1d50} downstream classifier hyper-parameters.}
    \resizebox{0.3\textwidth}{!}{
    \begin{tabular}{lc}
        \toprule
        Hyper-parameter & Value \\
        \midrule
Blocks & 4 \\
Layers per block & [3, 4, 6, 3] \\
Expansion & 4 \\
Stride & 1 \\
Optimizer & Adam \\
Learning rate & $0.001$\\
Batch size & $0.001$\\
Epochs & 100\\
        \bottomrule
    \end{tabular}
    \label{table:resnet-ecg}
    }
\end{table}
\subsubsection{Additional Results}
\paragraph{Discussion}
\label{sec:ecg-discussion}
In \cref{table:rmse-ecg}, we demonstrated that most posterior sampling algorithms outperform the trained diffusion model for the missing block reconstruction task. This result is particularly interesting as it suggests that training a diffusion model (which takes several days) is not necessary for this task.
However, when \cite{alcaraz2023diffusion} trained the model on the reconstruction of random missing blocks (RMB), where 50\% of each lead is independently removed, the model outperformed all posterior sampling algorithms on the MB task. We report in \cref{table:rmse-ecg-RMB} the results of the top two algorithms, as well as the model trained on the MB task and the RMB task. The significant improvement between RMB and MB can be seen as an enhancement due to data augmentation.
\begin{table}[ht]
    \centering
    \caption{\small MAE and RMSE for missing block task on the \textsc{ptb-xl} dataset.}
    \resizebox{0.7\textwidth}{!}{
    \begin{tabular}{lcccc}
        \toprule
        Metric & \algo\ & \ddnm & \traineddiff-MB & \traineddiff-RMB \\
        \midrule
        MAE & $0.111\pm2\mathrm{e}\!{-3}$ & $0.103\pm2\mathrm{e}\!{-3}$ & $0.116\pm2\mathrm{e}\!{-3}$ & $\mathbf{0.0879\pm2\mathrm{e}\!{-3}}$\\
        RMSE & $0.225\pm4\mathrm{e}\!{-3}$ & $0.224\pm4\mathrm{e}\!{-3}$ & $0.266\pm3\mathrm{e}\!{-3}$ & $\mathbf{0.217\pm6\mathrm{e}\!{-3}}$\\
        \bottomrule
    \end{tabular}
    \label{table:rmse-ecg-RMB}
    }
\end{table}
\paragraph{Generated samples}
\label{sec:samples-ecg}
\begin{figure}[h!]
    \centering
    \includegraphics[width=0.47\textwidth]{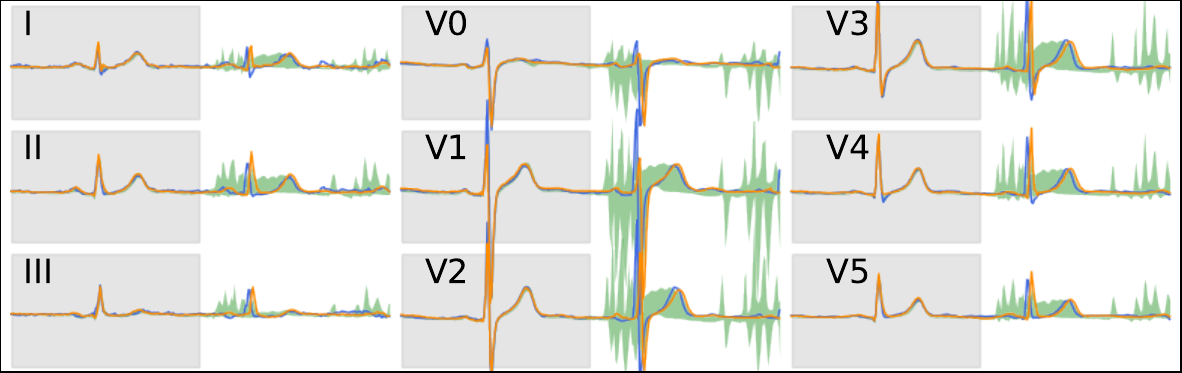}
        \caption{\small Missing block imputation with \algo\ on 2.56s 12-lead ECG. Ground-truth in blue,
        $10\%$--$90\%$
        quantile range in green, random sample in orange.}
    \label{fig:MB-ecg}
\end{figure}
\begin{figure}[h!]
    \centering
    \begin{subfigure}  %[b]{0.45\textwidth}
        \centering
        \includegraphics[width=0.47\textwidth]{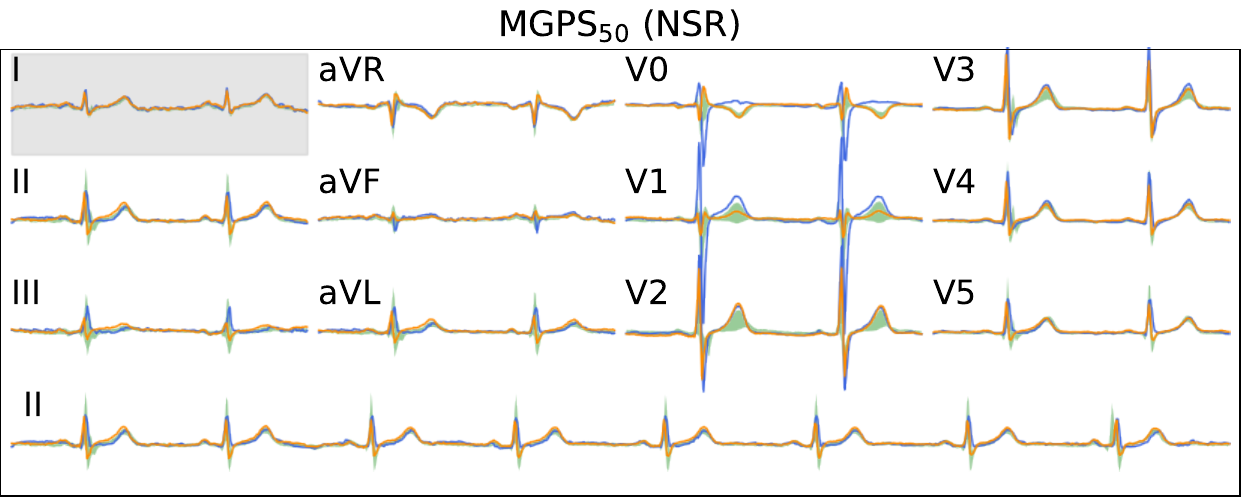}
        %\caption{Description de l'image 1}
       % \label{fig:image1}
    \end{subfigure}
    \hfill
    \begin{subfigure} %[b]{0.45\textwidth}
        \centering
        \includegraphics[width=0.47\textwidth]{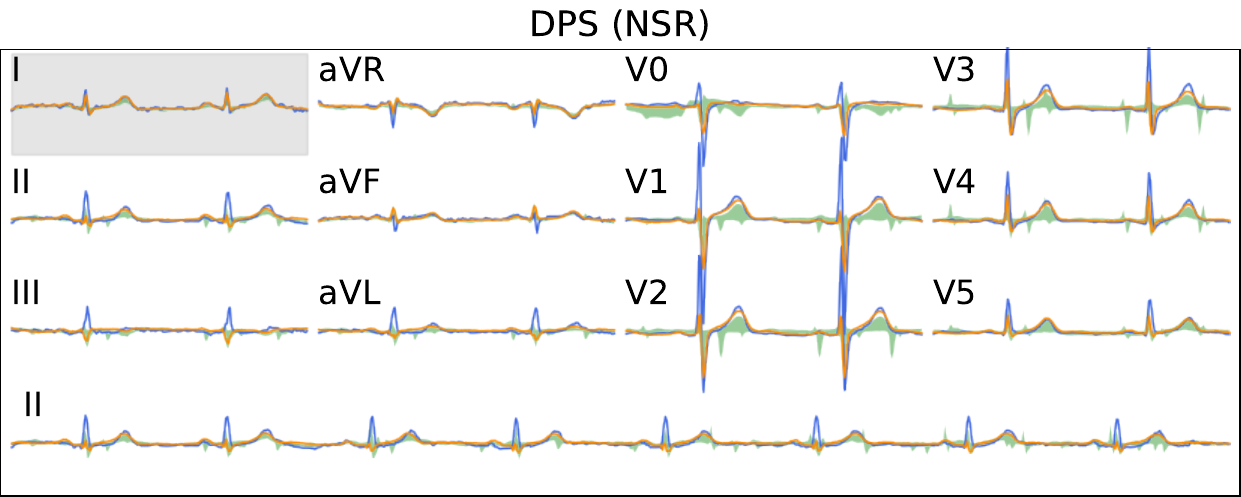}
        %\caption{Description de l'image 2}
       % \label{fig:image2}
    \end{subfigure}
   % \vspace{1cm} % Espace entre les deux rangées d'images
    \begin{subfigure}  %[b]{0.45\textwidth}
        \centering
        \includegraphics[width=0.47\textwidth]{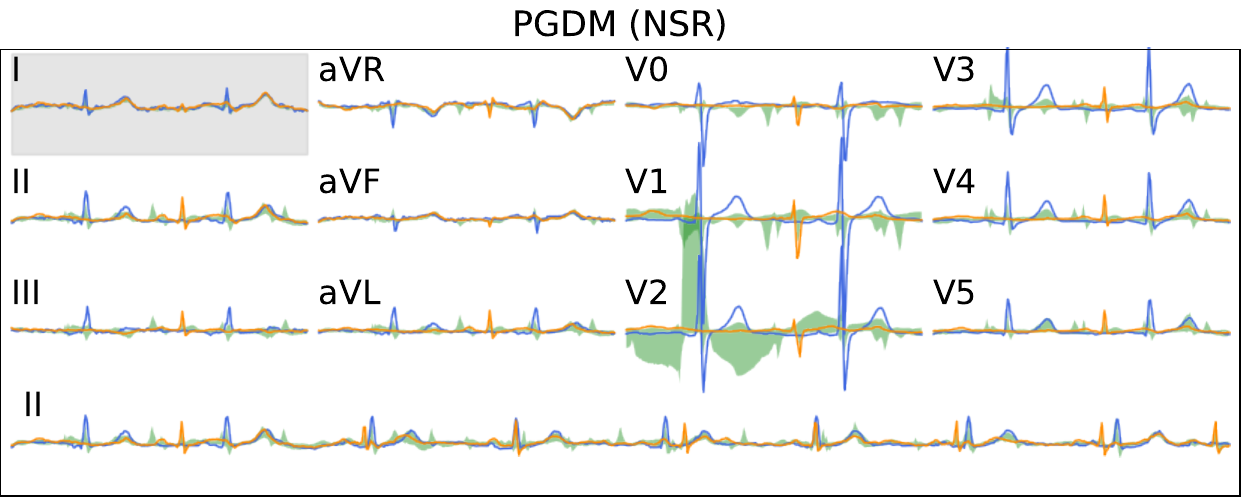}
        %\caption{Description de l'image 3}
       % \label{fig:image3}
    \end{subfigure}
    \hfill
    \begin{subfigure}  %[b]{0.45\textwidth}
        \centering
        \includegraphics[width=0.47\textwidth]{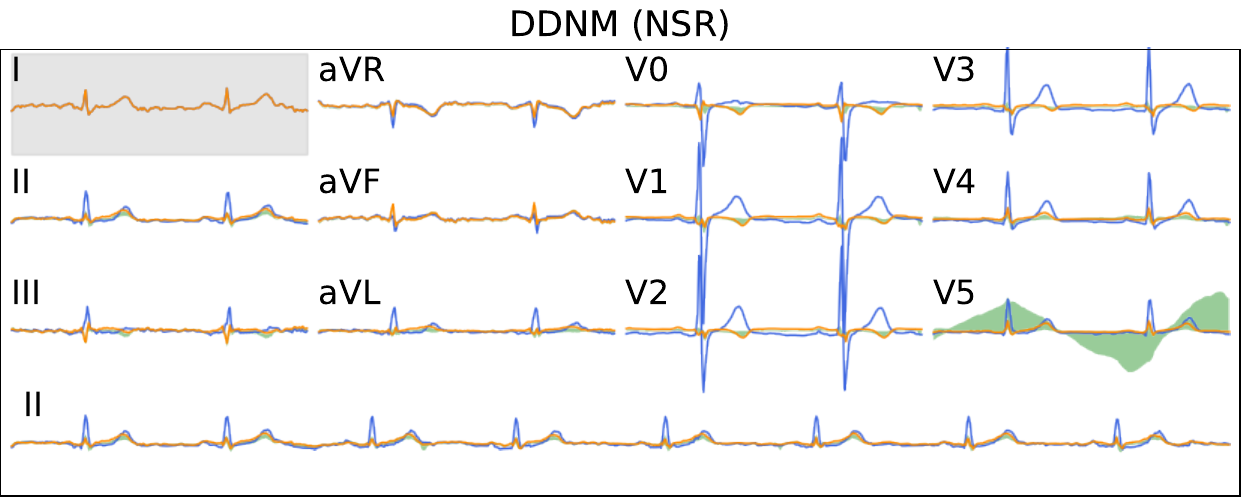}
       % \caption{Description de l'image 4}
       % \label{fig:image4}
    \end{subfigure}
        \begin{subfigure}  %[b]{0.45\textwidth}
        \centering
        \includegraphics[width=0.47\textwidth]{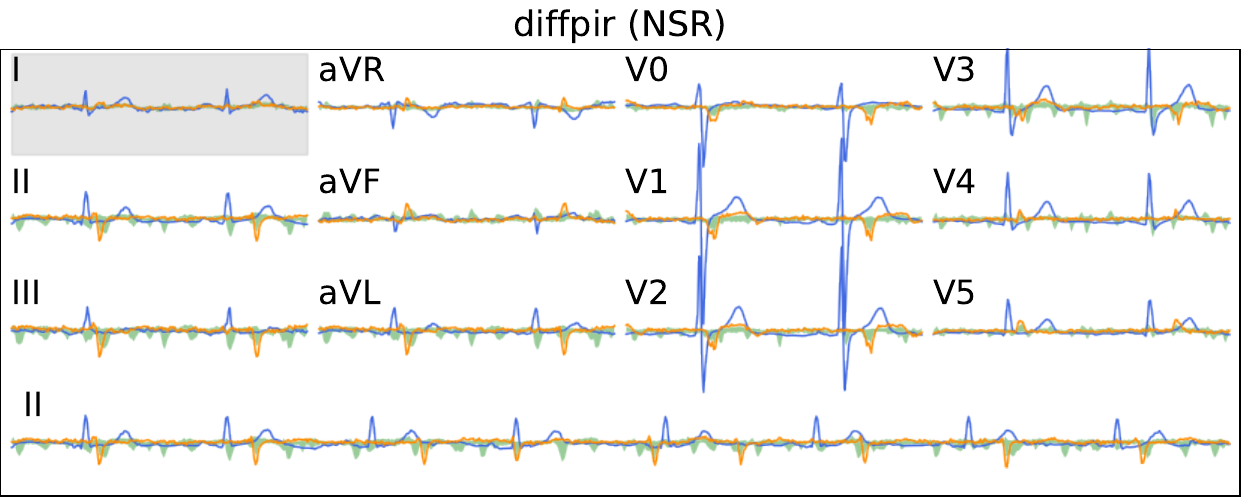}
        %\caption{Description de l'image 3}
       % \label{fig:image3}
    \end{subfigure}
    \hfill
    \begin{subfigure}  %[b]{0.45\textwidth}
        \centering
        \includegraphics[width=0.47\textwidth]{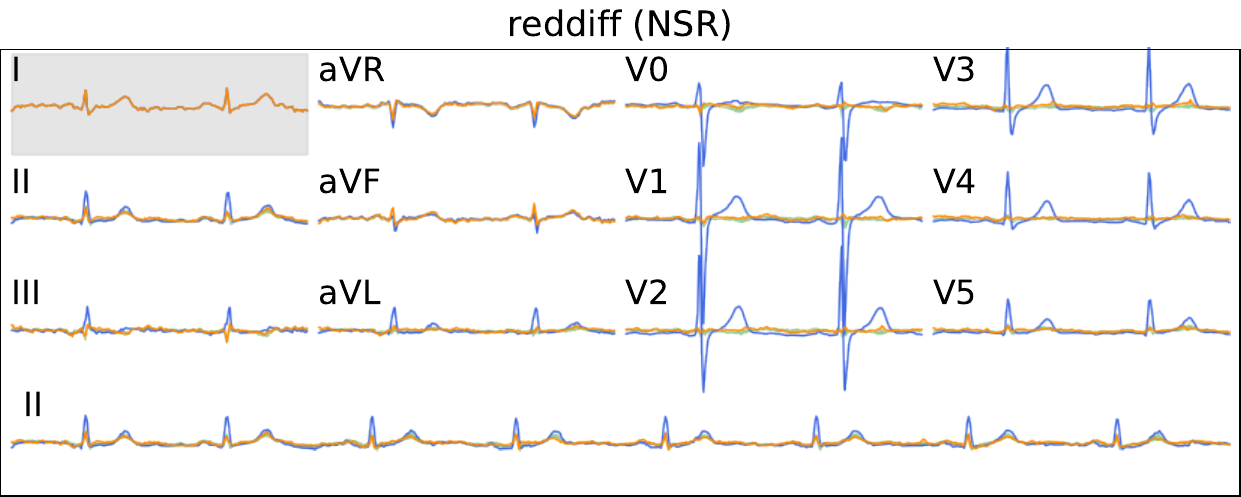}
       % \caption{Description de l'image 4}
       % \label{fig:image4}
    \end{subfigure}
    \caption{\small Missing lead reconstruction from lead I on 10s 12-lead Normal Sinus Rythm (NSR) ECGs. Ground-truth in blue,
        $10\%$--$90\%$
        quantile range in green, random sample in orange.}
    \label{fig:ML-NSR-ecg'}
\end{figure}

\begin{figure}[h!]
    \centering
    \begin{subfigure}  %[b]{0.45\textwidth}
        \centering
        \includegraphics[width=0.47\textwidth]{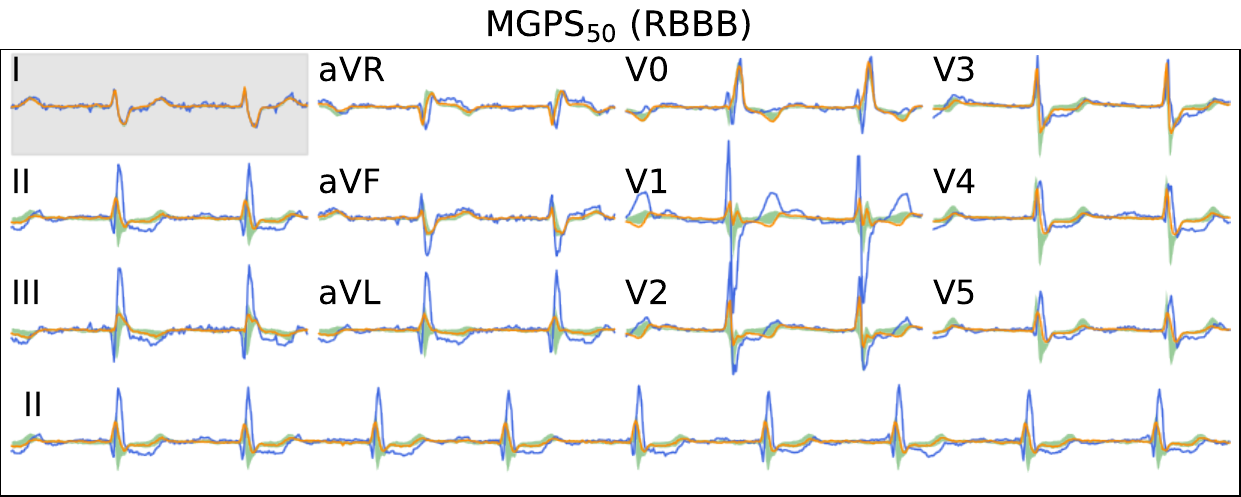}
        %\caption{Description de l'image 1}
       % \label{fig:image1}
    \end{subfigure}
    \hfill
    \begin{subfigure} %[b]{0.45\textwidth}
        \centering
        \includegraphics[width=0.47\textwidth]{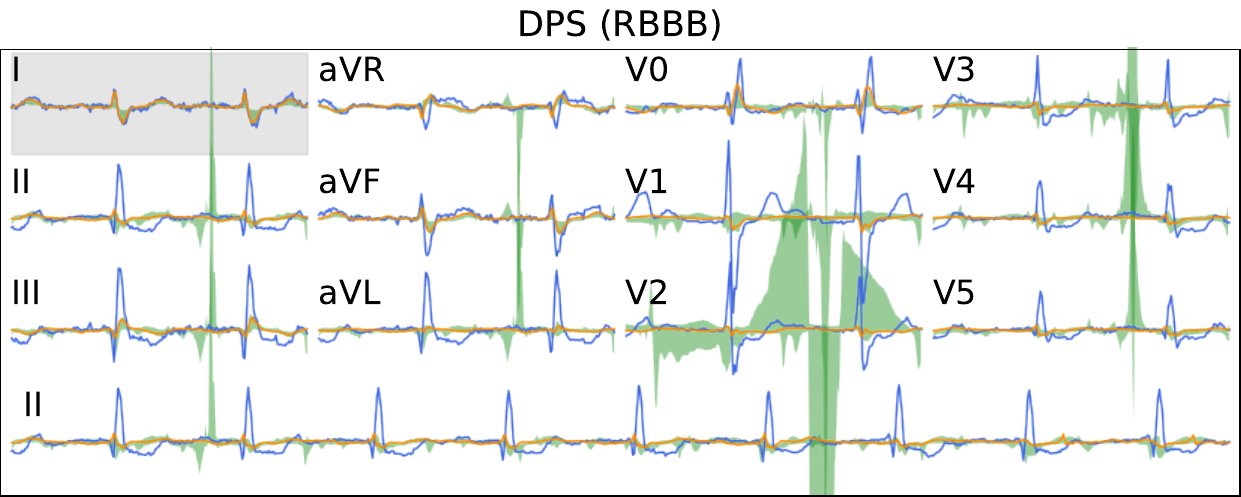}
        %\caption{Description de l'image 2}
       % \label{fig:image2}
    \end{subfigure}
   % \vspace{1cm} % Espace entre les deux rangées d'images
    \begin{subfigure}  %[b]{0.45\textwidth}
        \centering
        \includegraphics[width=0.47\textwidth]{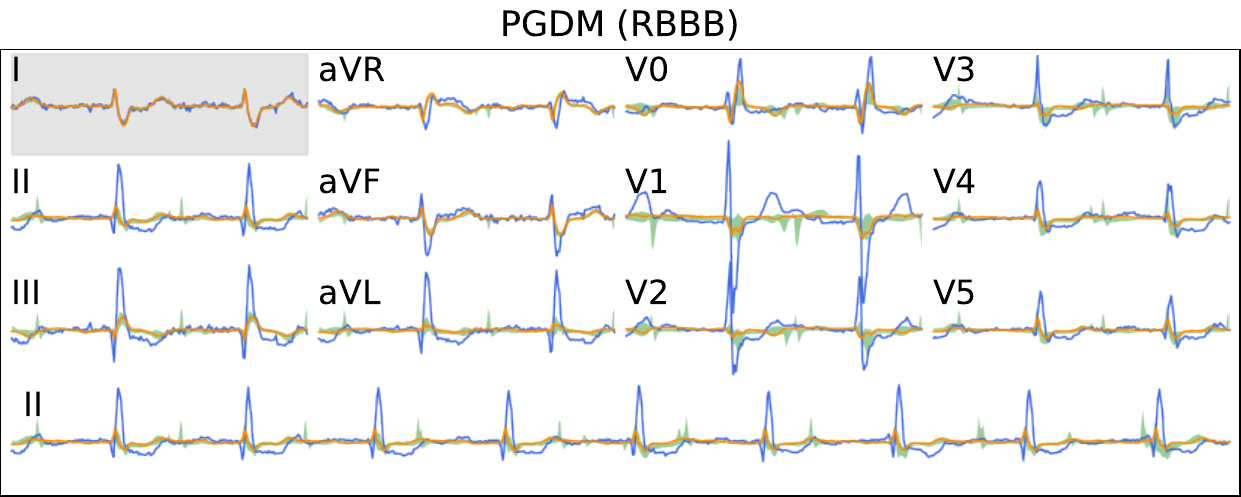}
        %\caption{Description de l'image 3}
       % \label{fig:image3}
    \end{subfigure}
    \hfill
    \begin{subfigure}  %[b]{0.45\textwidth}
        \centering
        \includegraphics[width=0.47\textwidth]{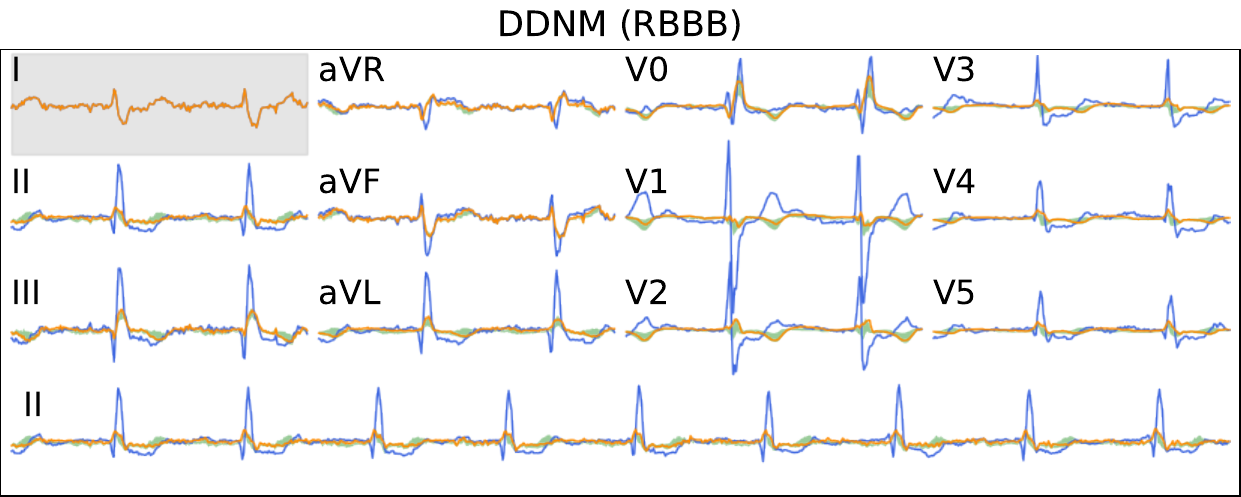}
       % \caption{Description de l'image 4}
       % \label{fig:image4}
    \end{subfigure}
        \begin{subfigure}  %[b]{0.45\textwidth}
        \centering
        \includegraphics[width=0.47\textwidth]{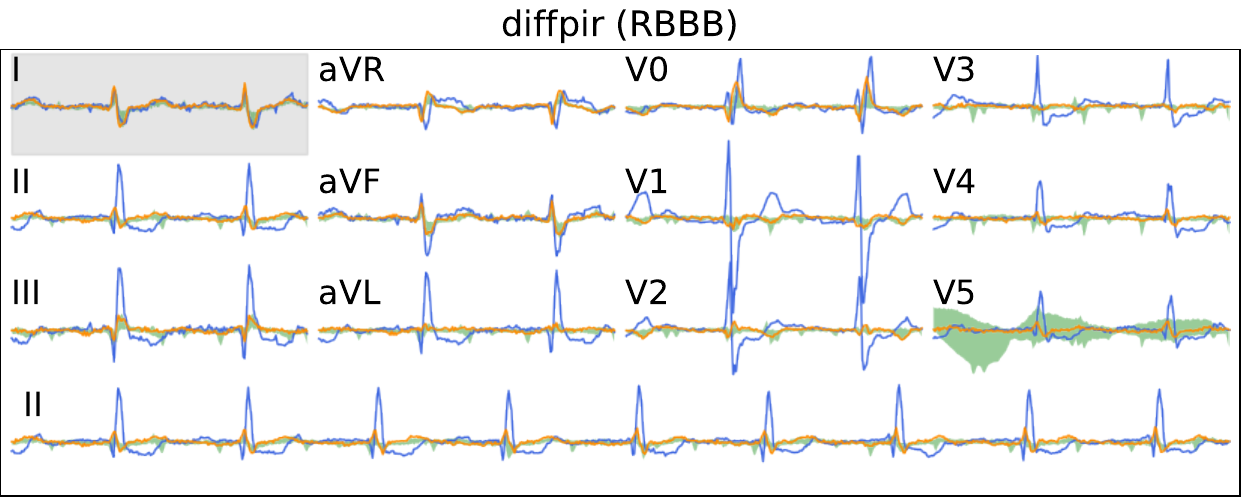}
        %\caption{Description de l'image 3}
       % \label{fig:image3}
    \end{subfigure}
    \hfill
    \begin{subfigure}  %[b]{0.45\textwidth}
        \centering
        \includegraphics[width=0.47\textwidth]{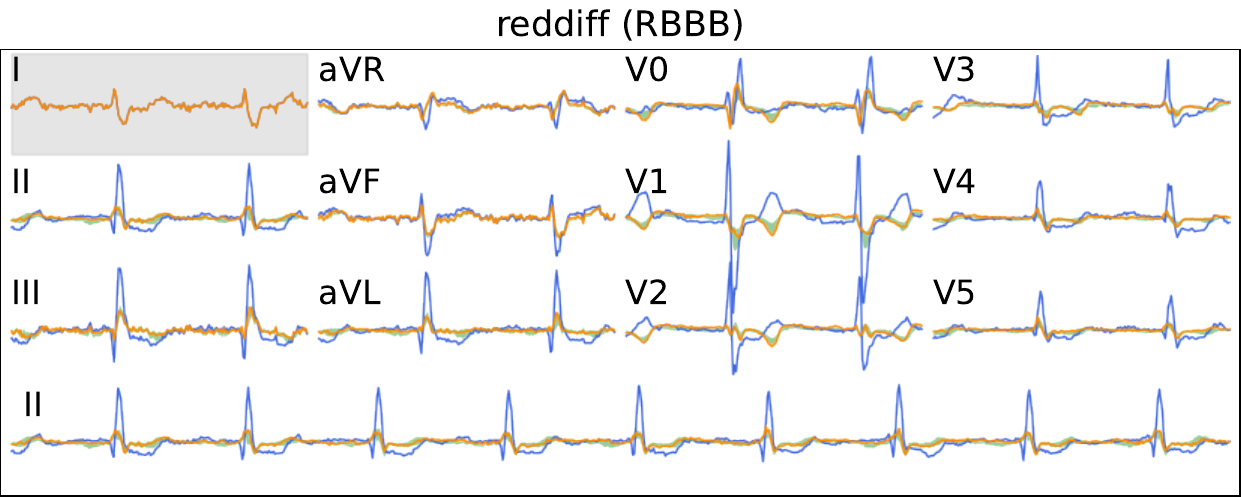}
       % \caption{Description de l'image 4}
       % \label{fig:image4}
    \end{subfigure}
    \caption{\small Missing lead reconstruction from lead I on 10s 12-lead Right Bundle Branch Block (RBBB) ECGs. Ground-truth in blue,
        $10\%$--$90\%$
        quantile range in green, random sample in orange.}
    \label{fig:ML-RBBB-ecg'}
\end{figure}

\begin{figure}[h!]
    \centering
    \begin{subfigure}  %[b]{0.45\textwidth}
        \centering
        \includegraphics[width=0.47\textwidth]{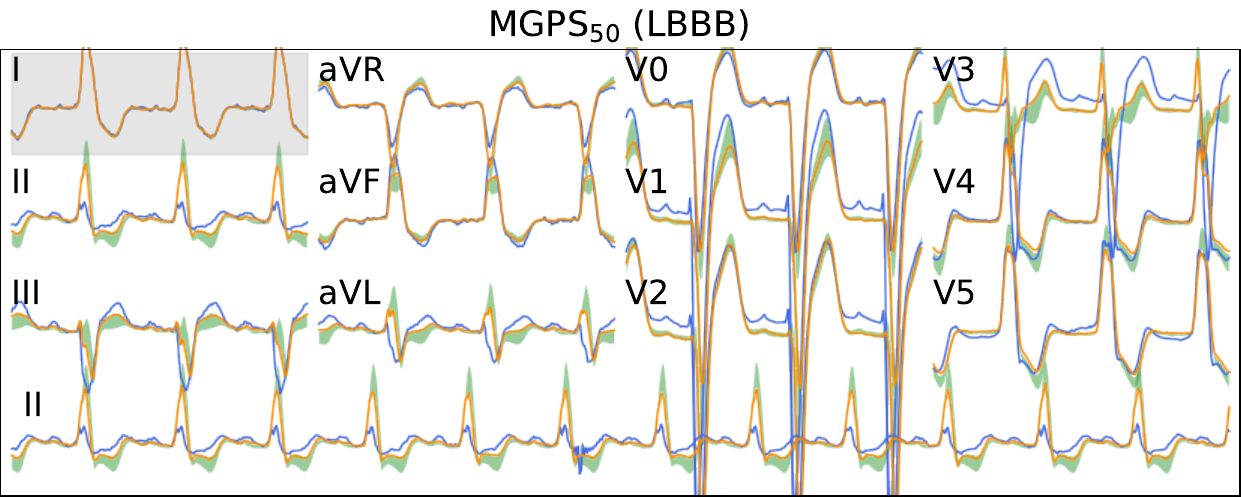}
        %\caption{Description de l'image 1}
       % \label{fig:image1}
    \end{subfigure}
    \hfill
    \begin{subfigure} %[b]{0.45\textwidth}
        \centering
        \includegraphics[width=0.47\textwidth]{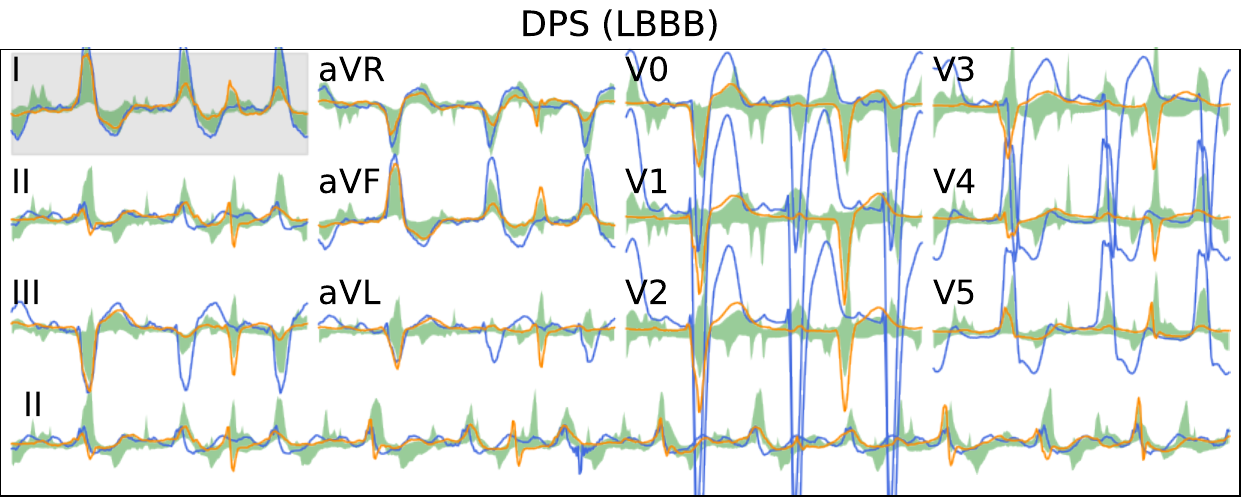}
        %\caption{Description de l'image 2}
       % \label{fig:image2}
    \end{subfigure}
   % \vspace{1cm} % Espace entre les deux rangées d'images
    \begin{subfigure}  %[b]{0.45\textwidth}
        \centering
        \includegraphics[width=0.47\textwidth]{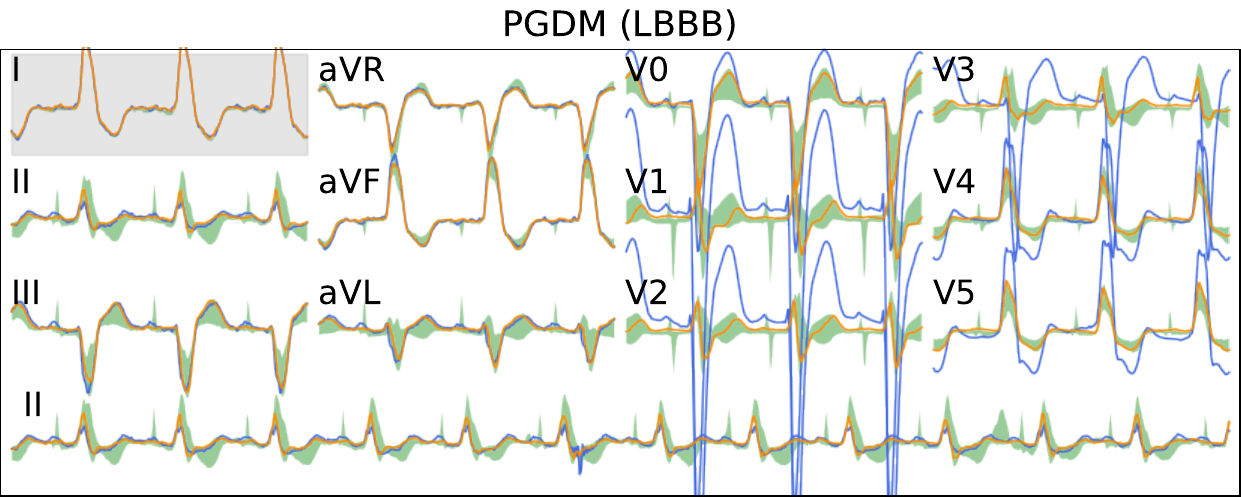}
        %\caption{Description de l'image 3}
       % \label{fig:image3}
    \end{subfigure}
    \hfill
    \begin{subfigure}  %[b]{0.45\textwidth}
        \centering
        \includegraphics[width=0.47\textwidth]{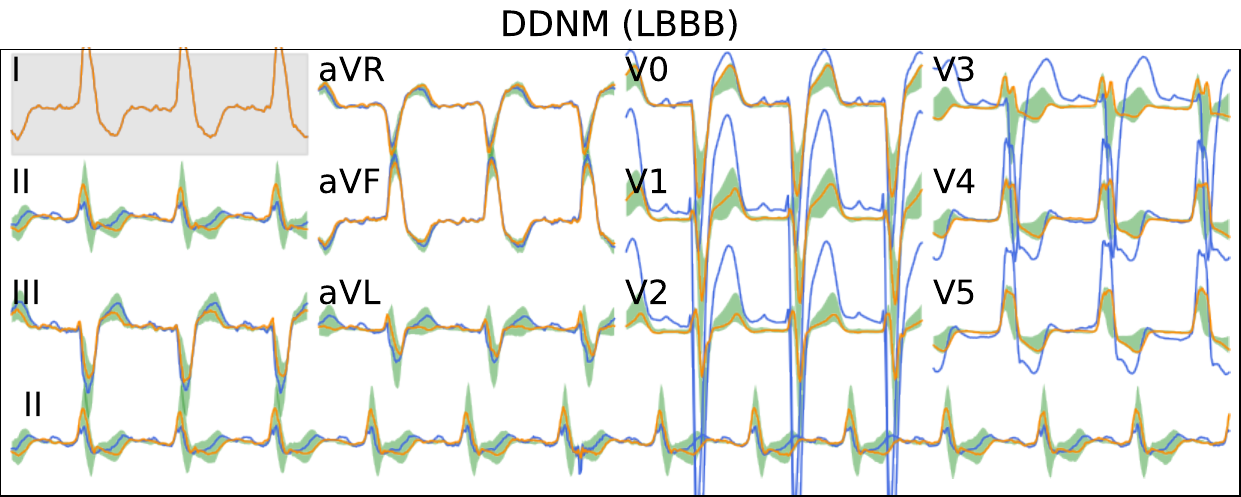}
       % \caption{Description de l'image 4}
       % \label{fig:image4}
    \end{subfigure}
        \begin{subfigure}  %[b]{0.45\textwidth}
        \centering
        \includegraphics[width=0.47\textwidth]{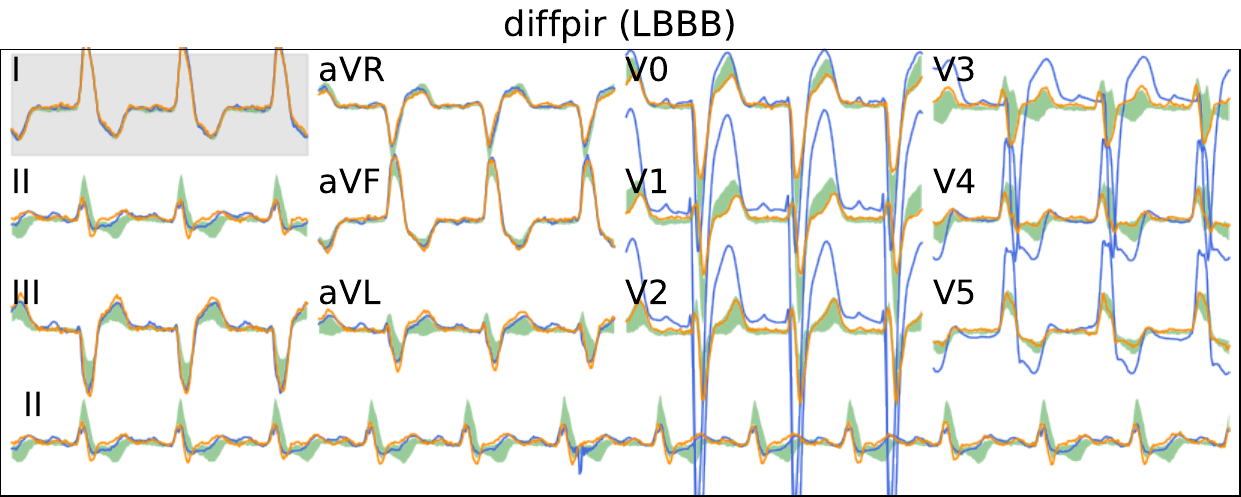}
        %\caption{Description de l'image 3}
       % \label{fig:image3}
    \end{subfigure}
    \hfill
    \begin{subfigure}  %[b]{0.45\textwidth}
        \centering
        \includegraphics[width=0.47\textwidth]{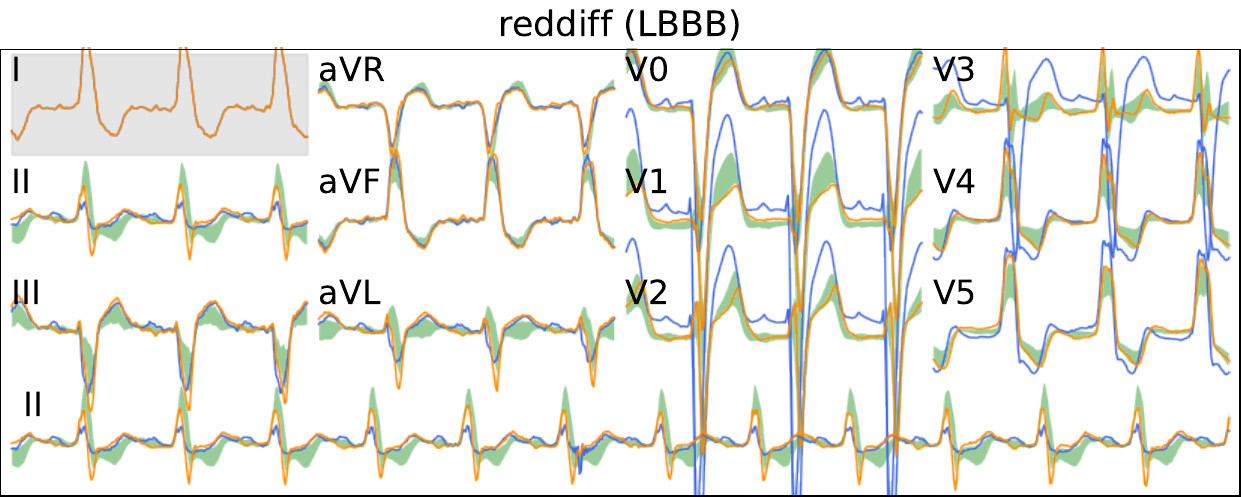}
       % \caption{Description de l'image 4}
       % \label{fig:image4}
    \end{subfigure}
    \caption{\small Missing lead reconstruction from lead I on 10s 12-lead Left Bundle Branch Block (LBBB) ECGs. Ground-truth in blue,
        $10\%$--$90\%$
        quantile range in green, random sample in orange.}
    \label{fig:ML-LBBB-ecg'}
\end{figure}

\begin{figure}[h!]
    \centering
    \begin{subfigure}  %[b]{0.45\textwidth}
        \centering
        \includegraphics[width=0.47\textwidth]{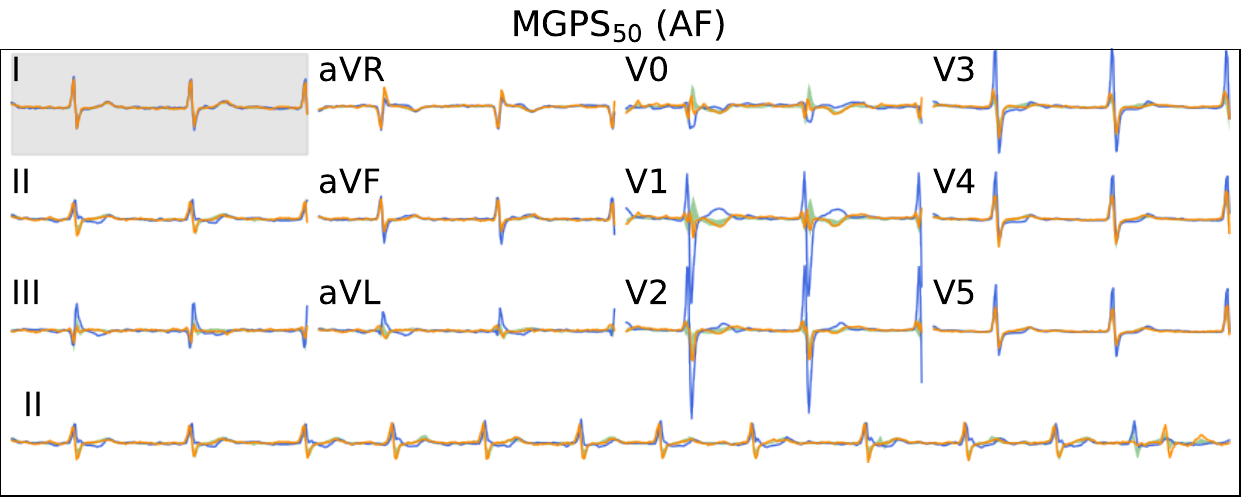}
        %\caption{Description de l'image 1}
       % \label{fig:image1}
    \end{subfigure}
    \hfill
    \begin{subfigure} %[b]{0.45\textwidth}
        \centering
        \includegraphics[width=0.47\textwidth]{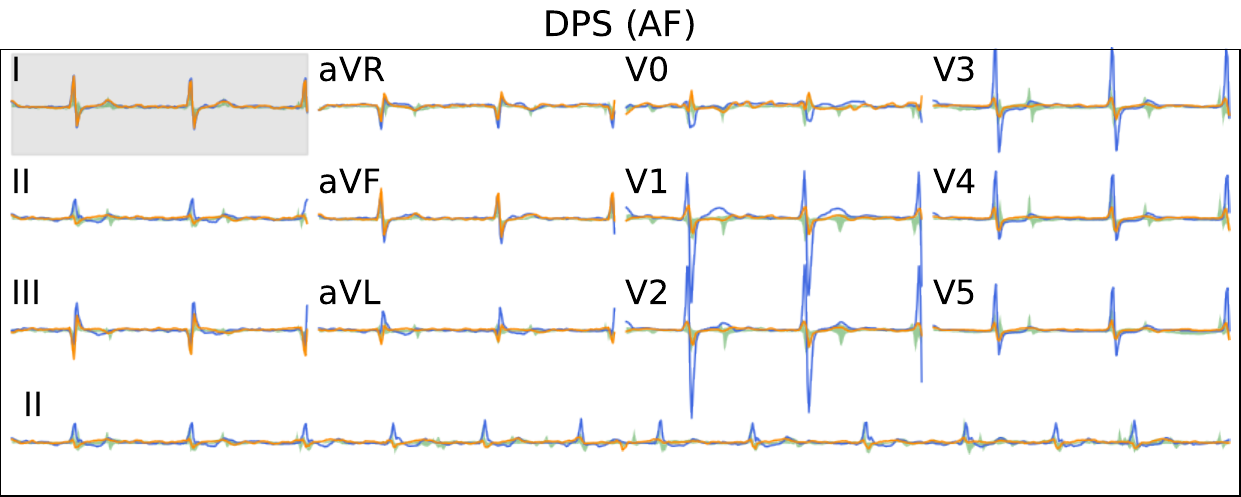}
        %\caption{Description de l'image 2}
       % \label{fig:image2}
    \end{subfigure}
   % \vspace{1cm} % Espace entre les deux rangées d'images
    \begin{subfigure}  %[b]{0.45\textwidth}
        \centering
        \includegraphics[width=0.47\textwidth]{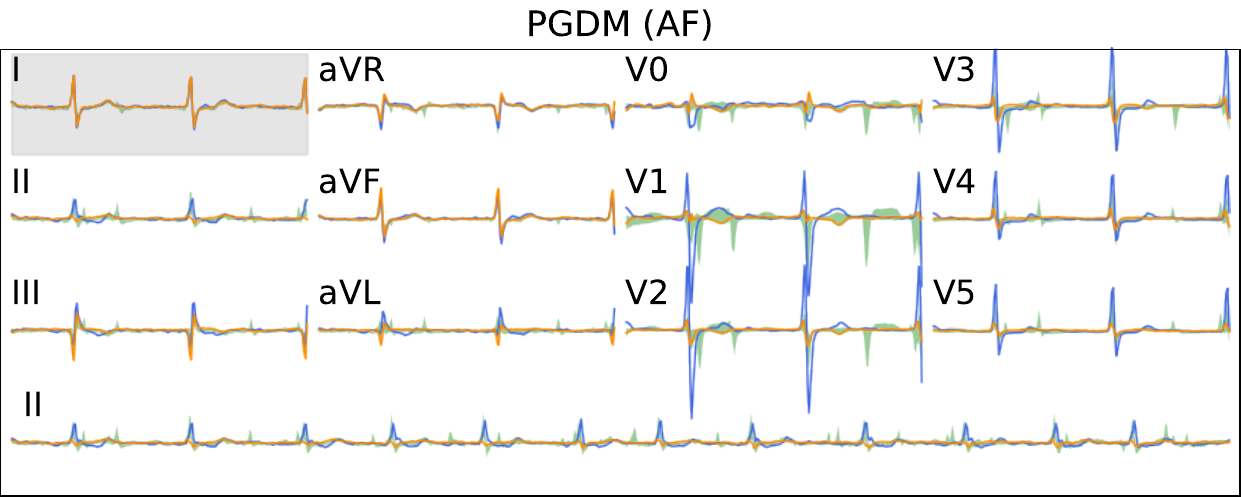}
        %\caption{Description de l'image 3}
       % \label{fig:image3}
    \end{subfigure}
    \hfill
    \begin{subfigure}  %[b]{0.45\textwidth}
        \centering
        \includegraphics[width=0.47\textwidth]{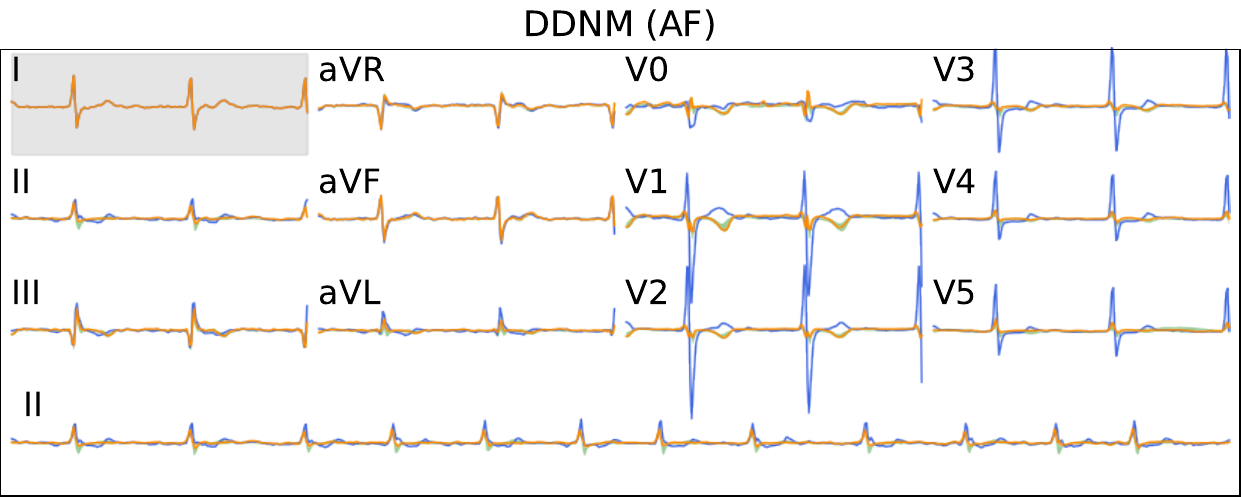}
       % \caption{Description de l'image 4}
       % \label{fig:image4}
    \end{subfigure}
        \begin{subfigure}  %[b]{0.45\textwidth}
        \centering
        \includegraphics[width=0.47\textwidth]{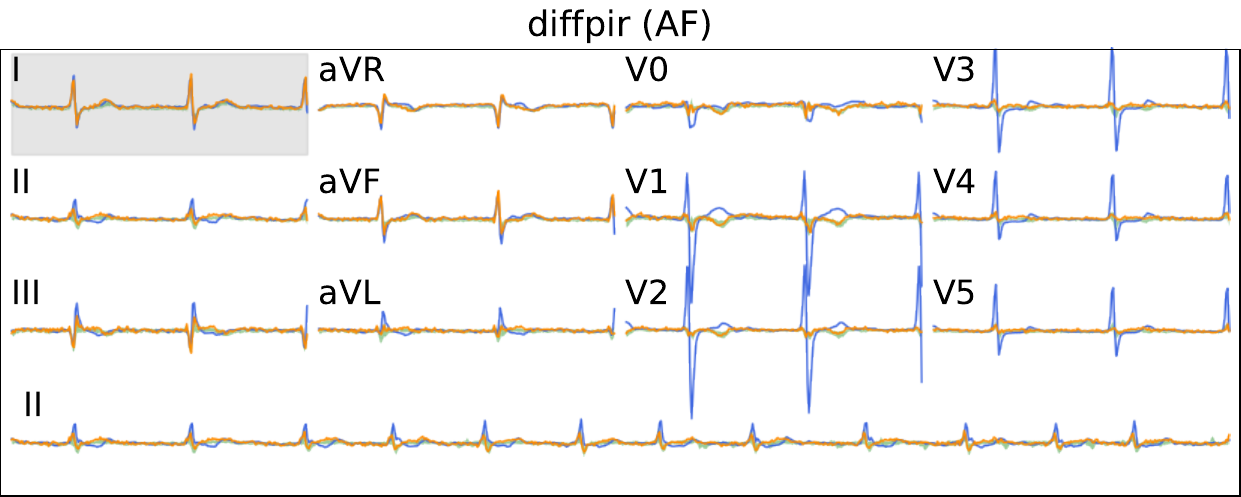}
        %\caption{Description de l'image 3}
       % \label{fig:image3}
    \end{subfigure}
    \hfill
    \begin{subfigure}  %[b]{0.45\textwidth}
        \centering
        \includegraphics[width=0.47\textwidth]{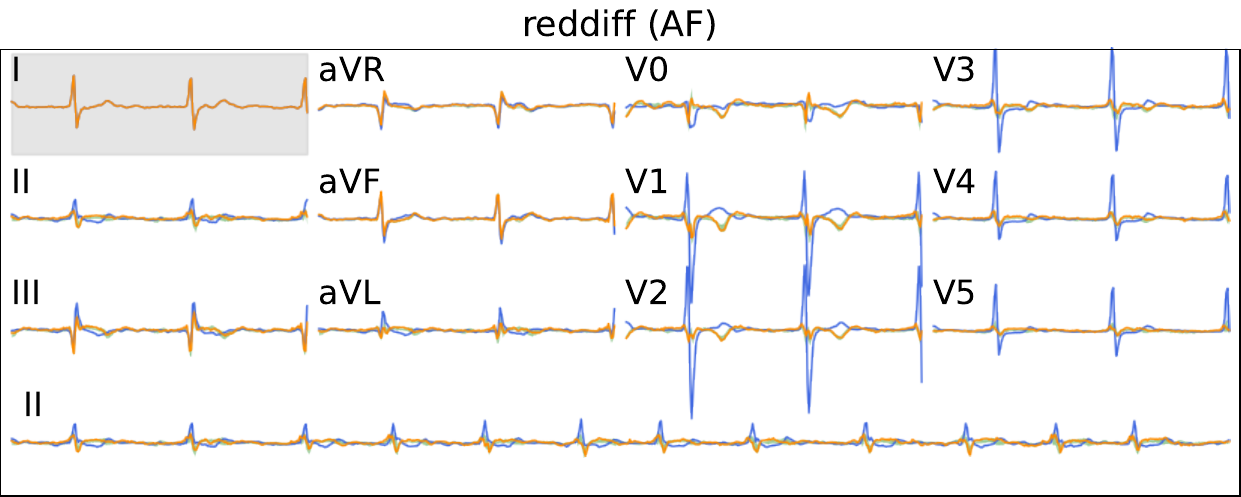}
       % \caption{Description de l'image 4}
       % \label{fig:image4}
    \end{subfigure}
    \caption{\small Missing lead reconstruction from lead I on 10s 12-lead Atrial Fibrillation (AF) ECGs. Ground-truth in blue,
        $10\%$--$90\%$
        quantile range in green, random sample in orange.}
    \label{fig:ML-AF-ecg'}
\end{figure}

\begin{figure}[h!]
    \centering
    \begin{subfigure}  %[b]{0.45\textwidth}
        \centering
        \includegraphics[width=0.47\textwidth]{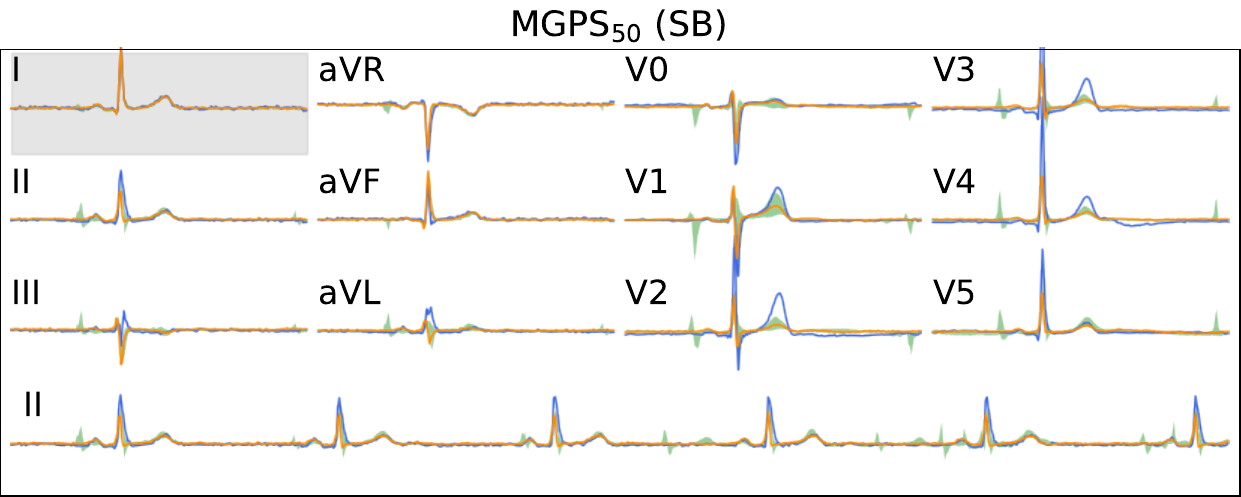}
        %\caption{Description de l'image 1}
       % \label{fig:image1}
    \end{subfigure}
    \hfill
    \begin{subfigure} %[b]{0.45\textwidth}
        \centering
        \includegraphics[width=0.47\textwidth]{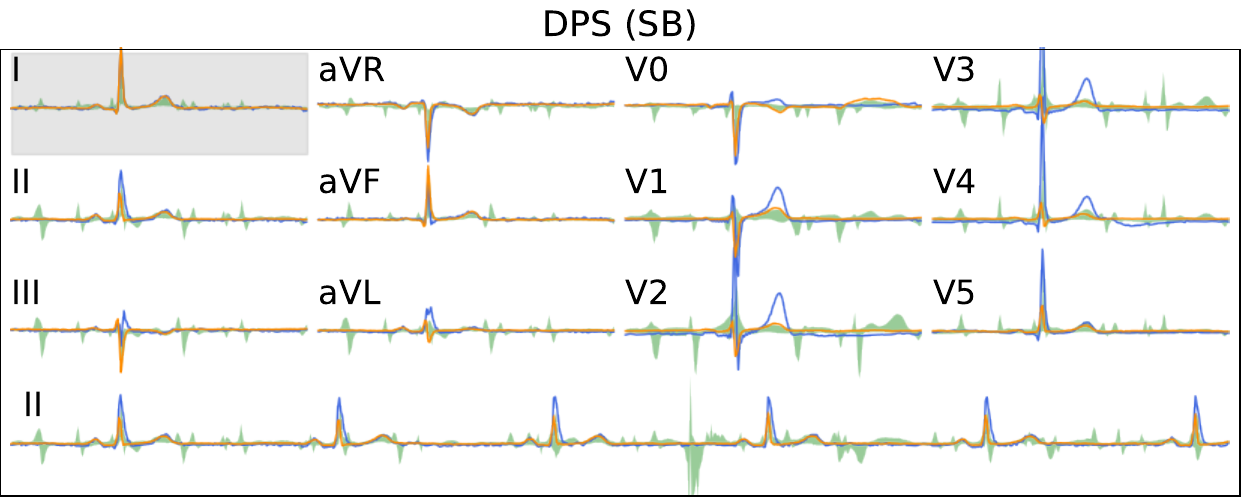}
        %\caption{Description de l'image 2}
       % \label{fig:image2}
    \end{subfigure}
   % \vspace{1cm} % Espace entre les deux rangées d'images
    \begin{subfigure}  %[b]{0.45\textwidth}
        \centering
        \includegraphics[width=0.47\textwidth]{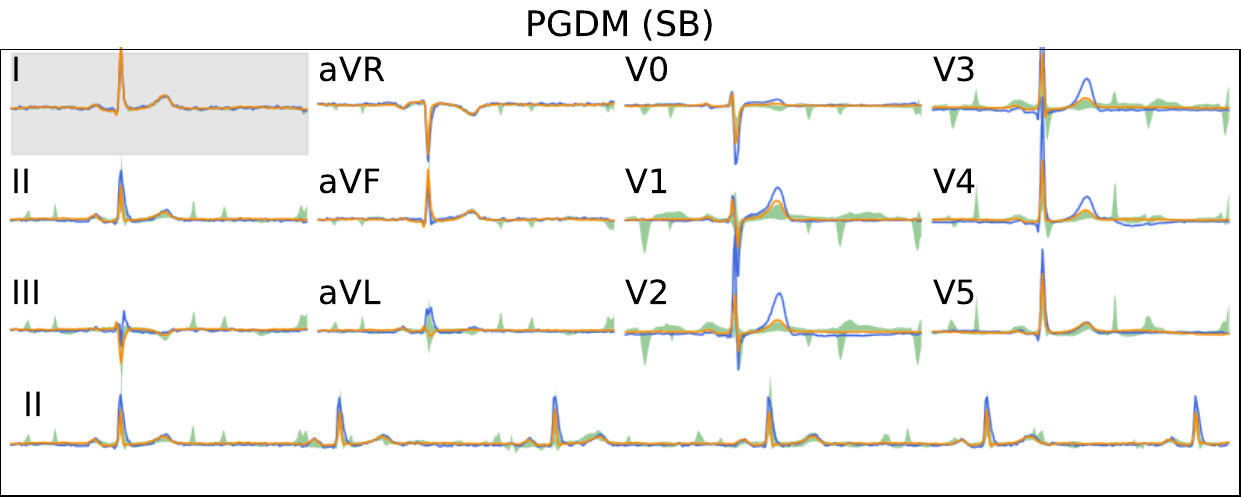}
        %\caption{Description de l'image 3}
       % \label{fig:image3}
    \end{subfigure}
    \hfill
    \begin{subfigure}  %[b]{0.45\textwidth}
        \centering
        \includegraphics[width=0.47\textwidth]{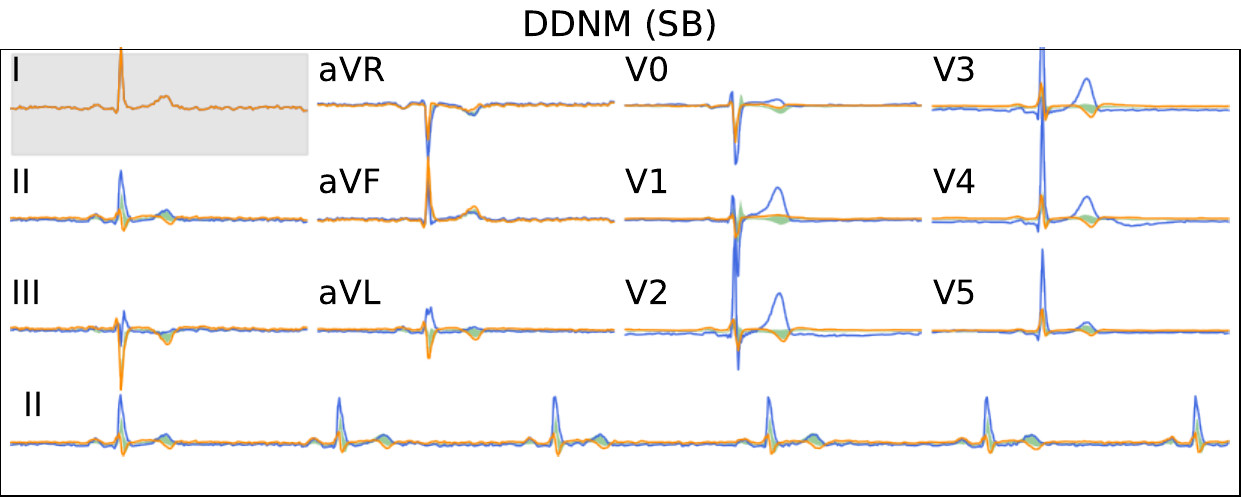}
       % \caption{Description de l'image 4}
       % \label{fig:image4}
    \end{subfigure}
        \begin{subfigure}  %[b]{0.45\textwidth}
        \centering
        \includegraphics[width=0.47\textwidth]{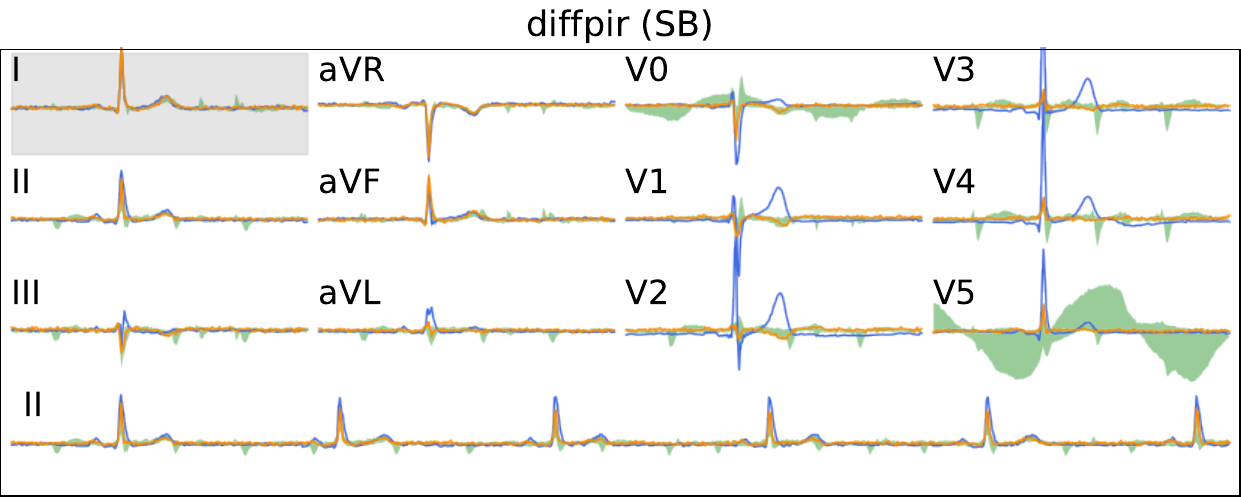}
        %\caption{Description de l'image 3}
       % \label{fig:image3}
    \end{subfigure}
    \hfill
    \begin{subfigure}  %[b]{0.45\textwidth}
        \centering
        \includegraphics[width=0.47\textwidth]{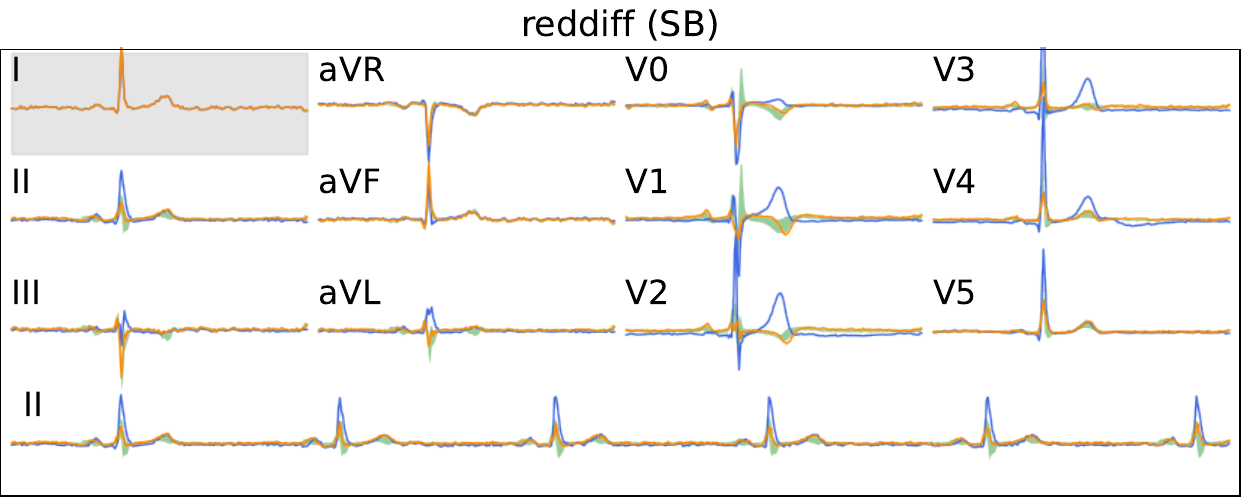}
       % \caption{Description de l'image 4}
       % \label{fig:image4}
    \end{subfigure}
    \caption{\small Missing lead reconstruction from lead I on 10s 12-lead Sinus Bradycardia (SB) ECGs. Ground-truth in blue,
        $10\%$--$90\%$
        quantile range in green, random sample in orange.}
    \label{fig:ML-SB-ecg'}
\end{figure}

\subsection{Sample images}
\label{subsec:sample-images}
\begin{figure}[htb]
    \centering
    \subfigure{
    \includegraphics[width=0.5\textwidth]{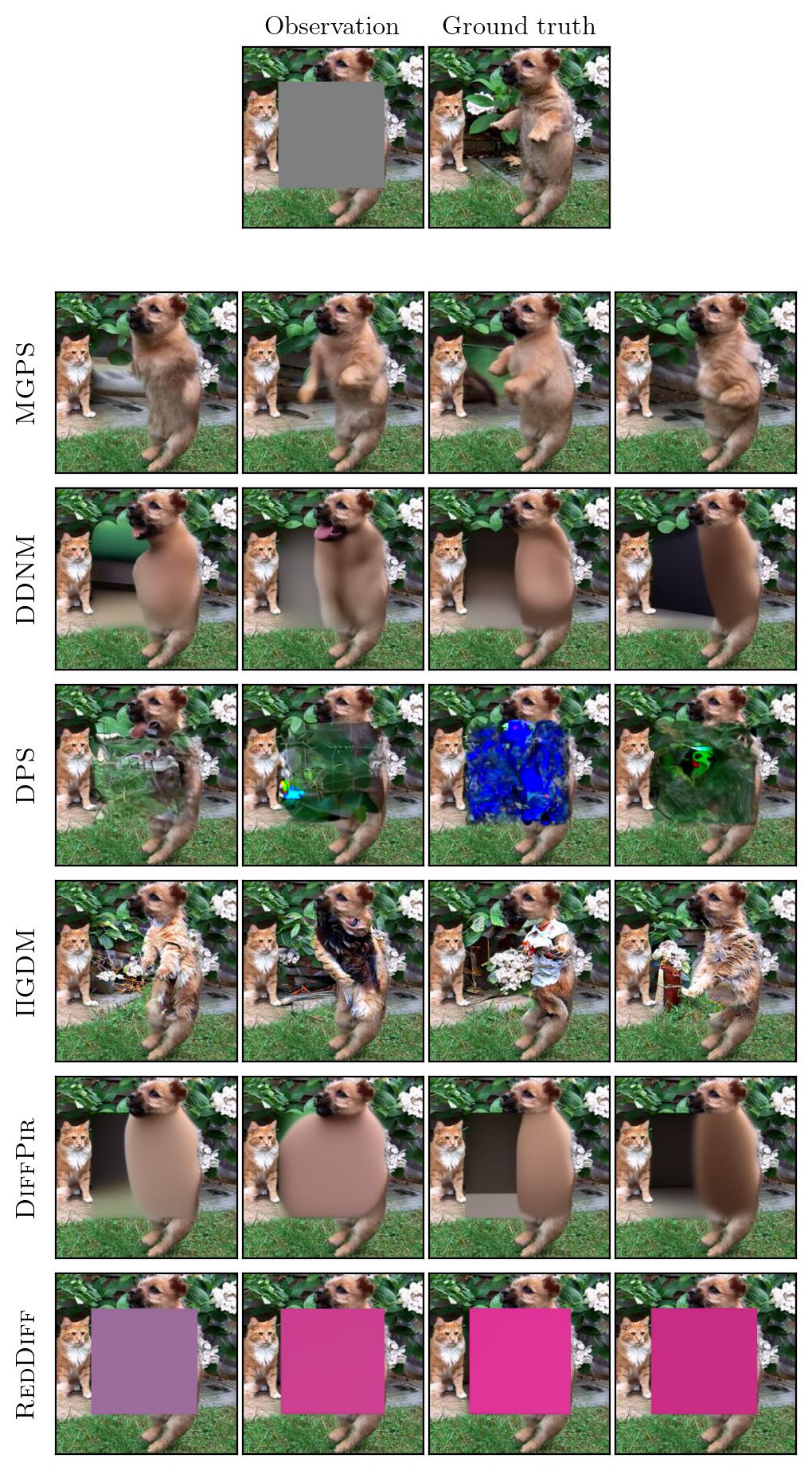}
    \includegraphics[width=0.5\textwidth]{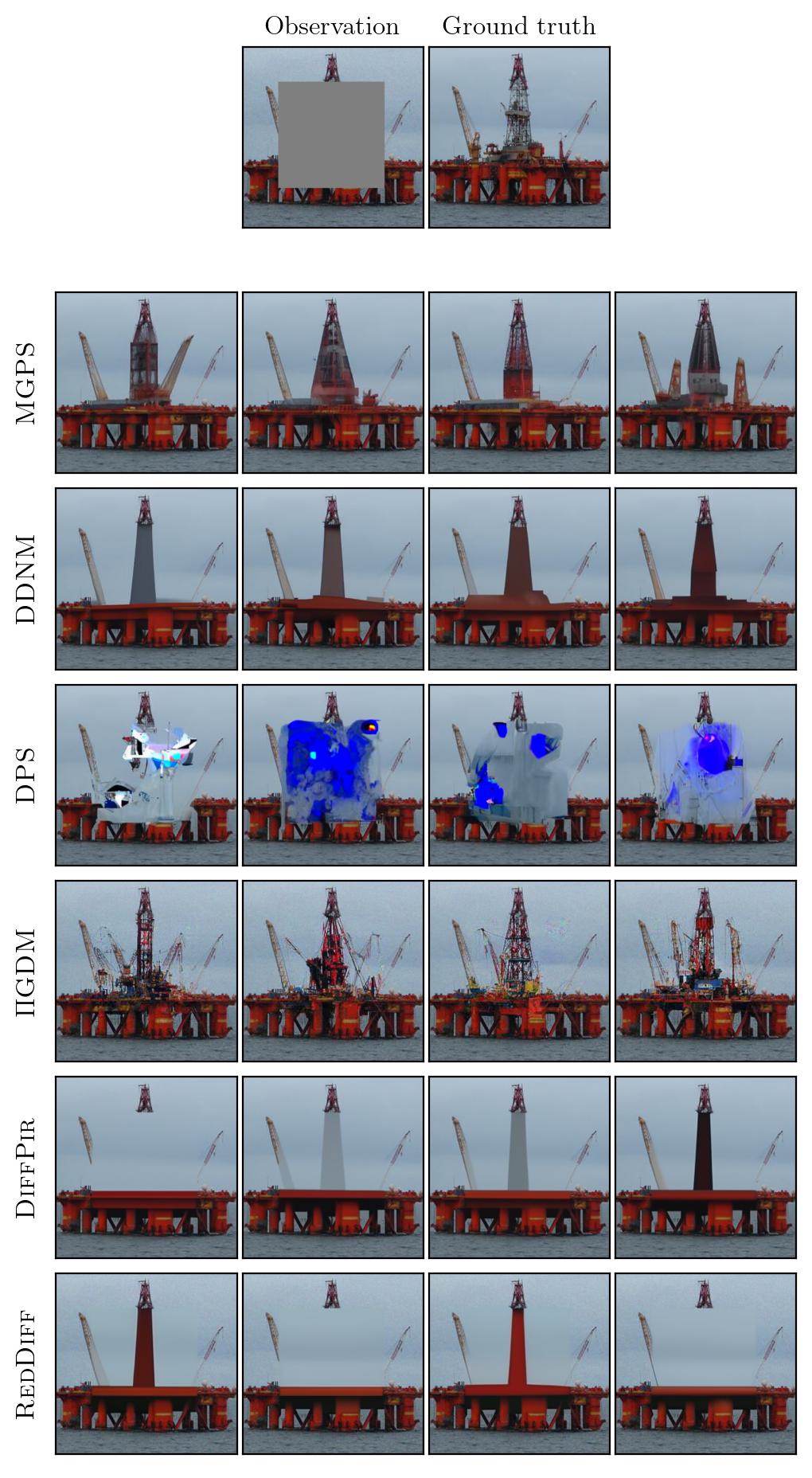}
    }
    \caption{Sample reconstructions with box mask on \imagenet\ dataset.}
    \label{fig:imagnet-inpainting-1}
\end{figure}
\begin{figure}[htb]
    \centering
    \subfigure{
        \includegraphics[width=0.42\textwidth]{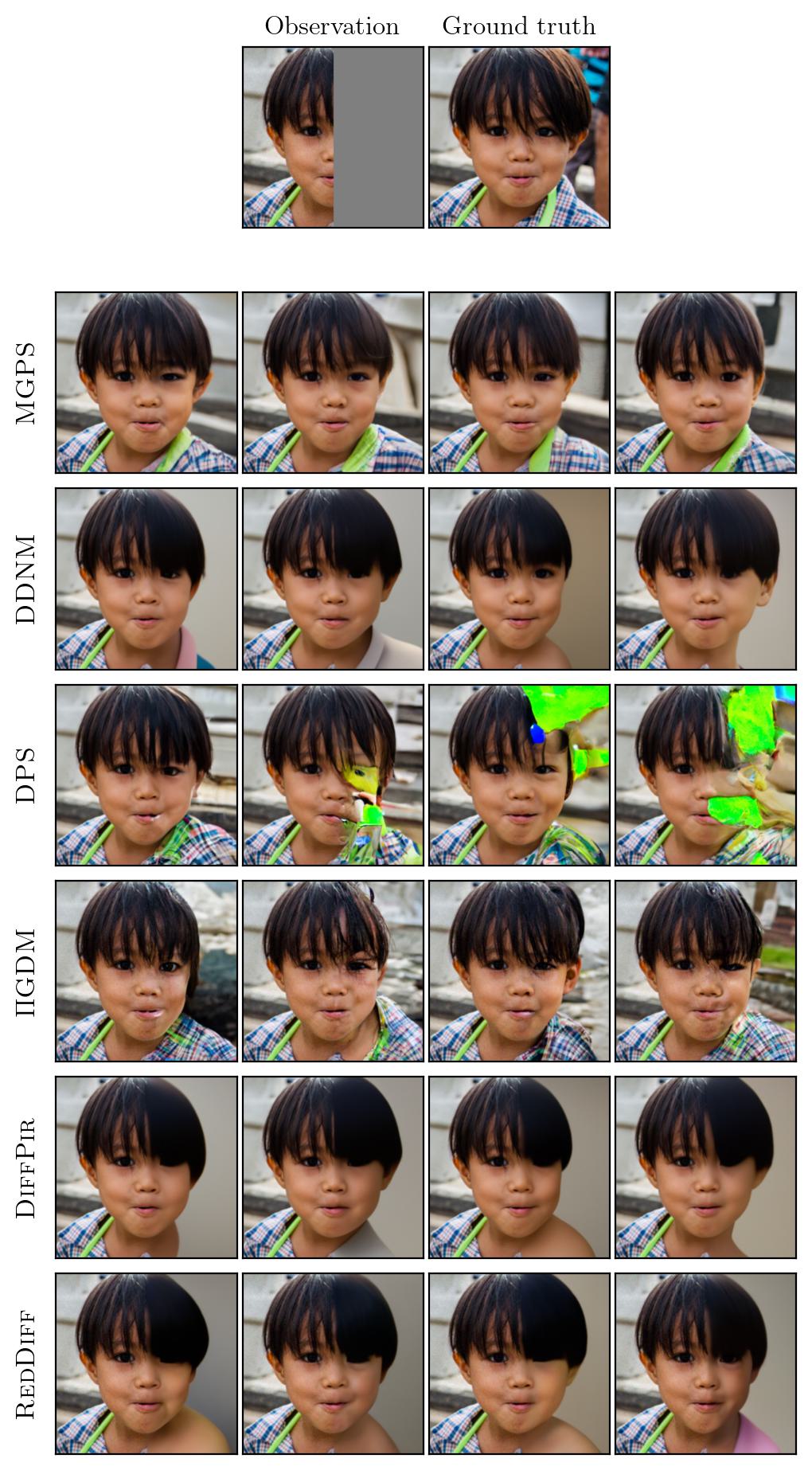}
        \includegraphics[width=0.42\textwidth]{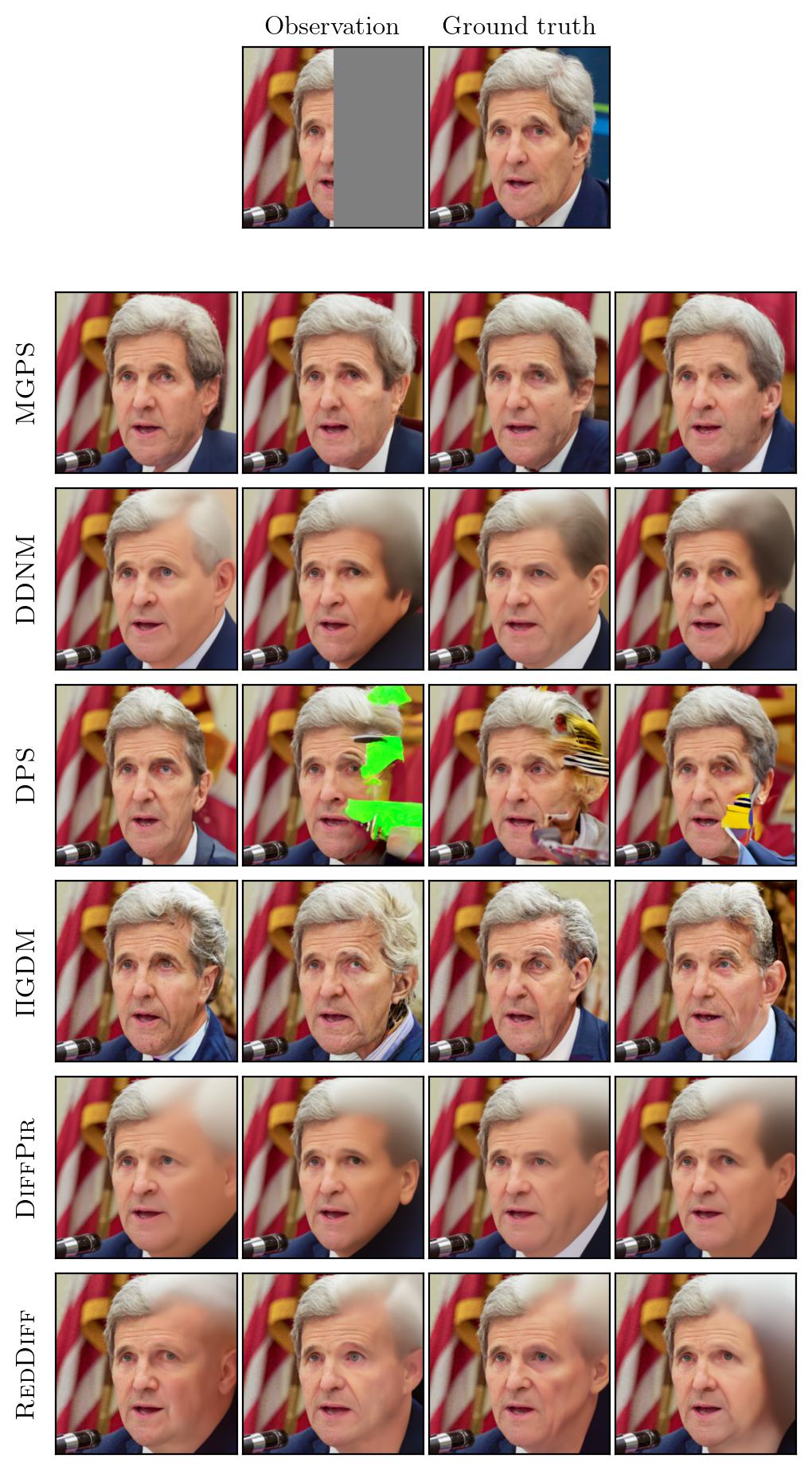}
    }
    \subfigure{
        \includegraphics[width=0.42\textwidth]{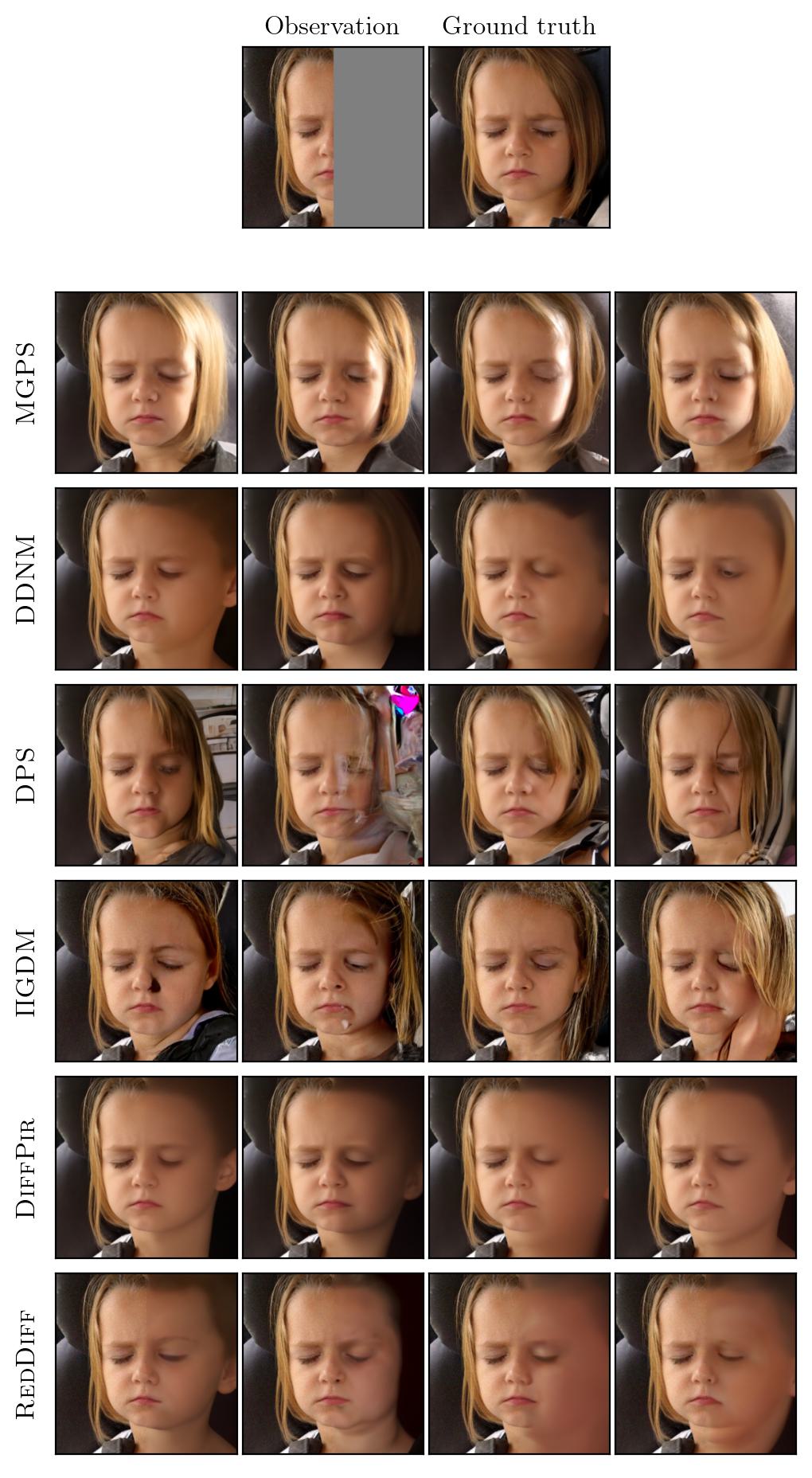}
        \includegraphics[width=0.42\textwidth]{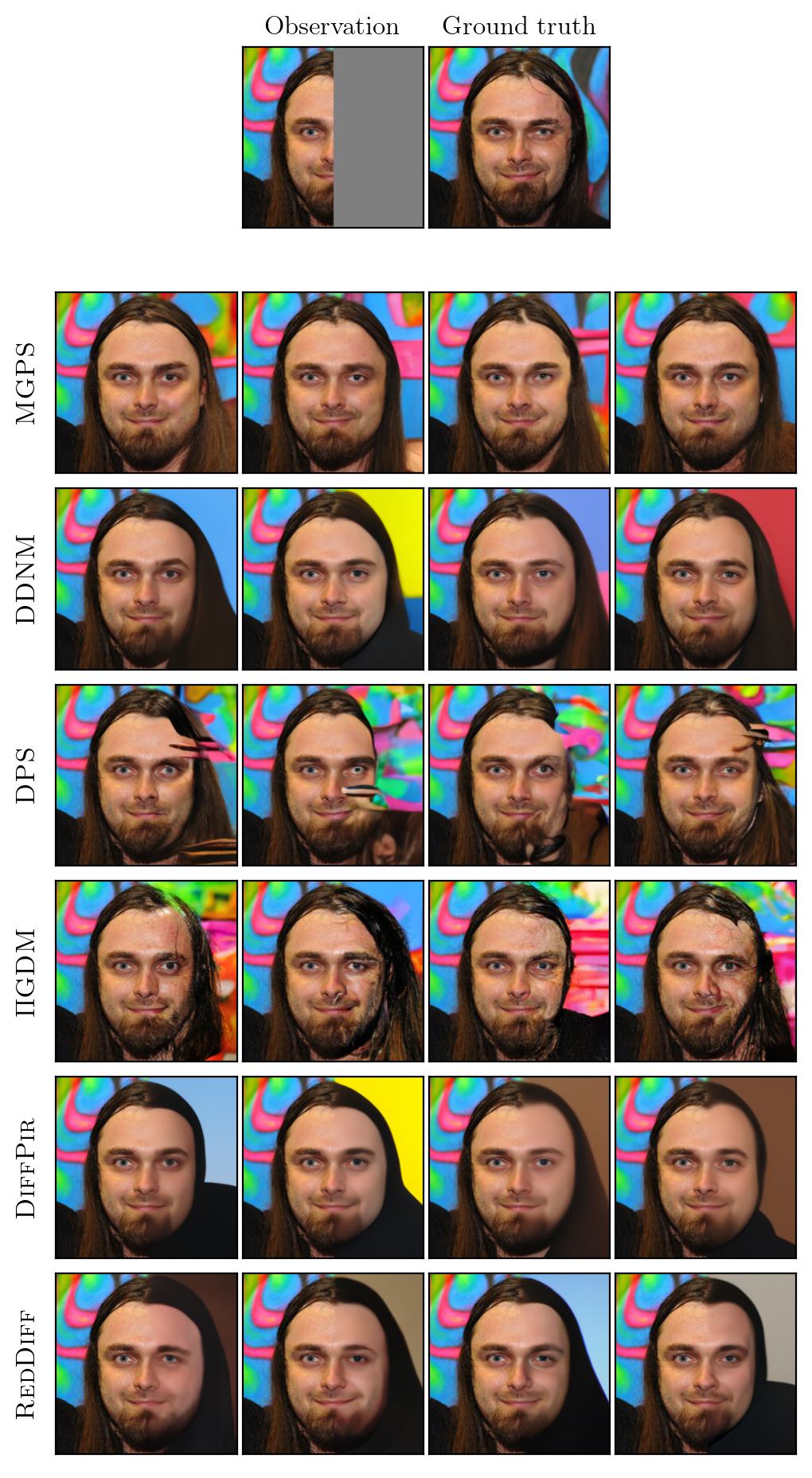}
    }
    % \hspace{0.05\textwidth} % Horizontal space between figures
    % \subfigure[Outpainting with half mask]{
    %     \includegraphics[width=0.45\textwidth]{figures/ldm_ims/outpainting_half/79.jpeg}
    %     \includegraphics[width=0.45\textwidth]{figures/ffhq_ldm_ims/outpainting_half/93.jpeg}
    % }
    % \subfigure[SR $4 \times$]{
    %     \includegraphics[width=0.45\textwidth]{figures/ffhq_ldm_ims/sr4/10.jpeg}
    %     \includegraphics[width=0.45\textwidth]{figures/ffhq_ldm_ims/sr4/45.jpeg}
    % }
    \caption{Sample reconstructions with half mask on \ffhq\ dataset.}
    \label{fig:ffhq-inpainting-1}
\end{figure}
\begin{figure}[htb]
    \centering
    \subfigure{
        \includegraphics[width=0.42\textwidth]{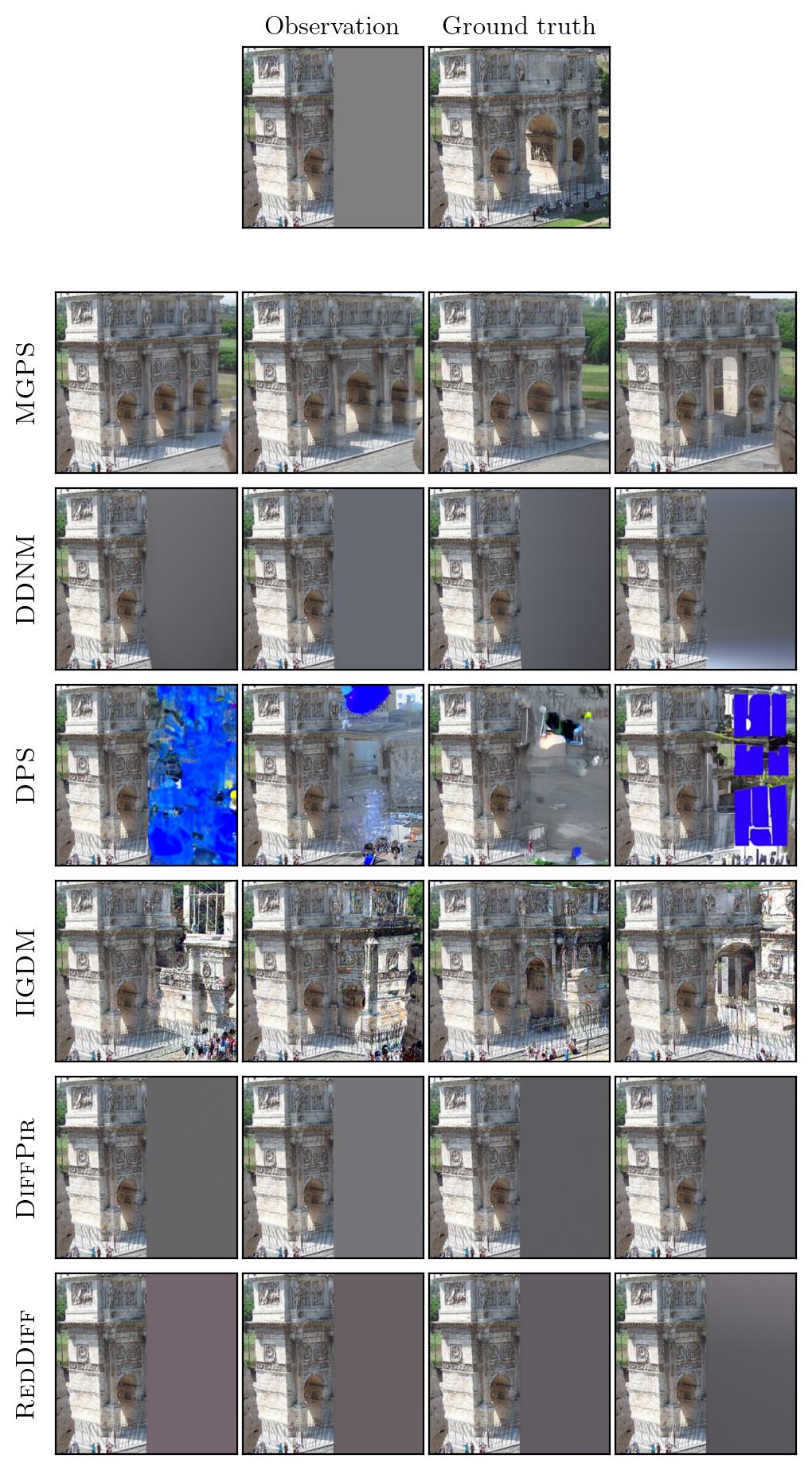}
        \includegraphics[width=0.42\textwidth]{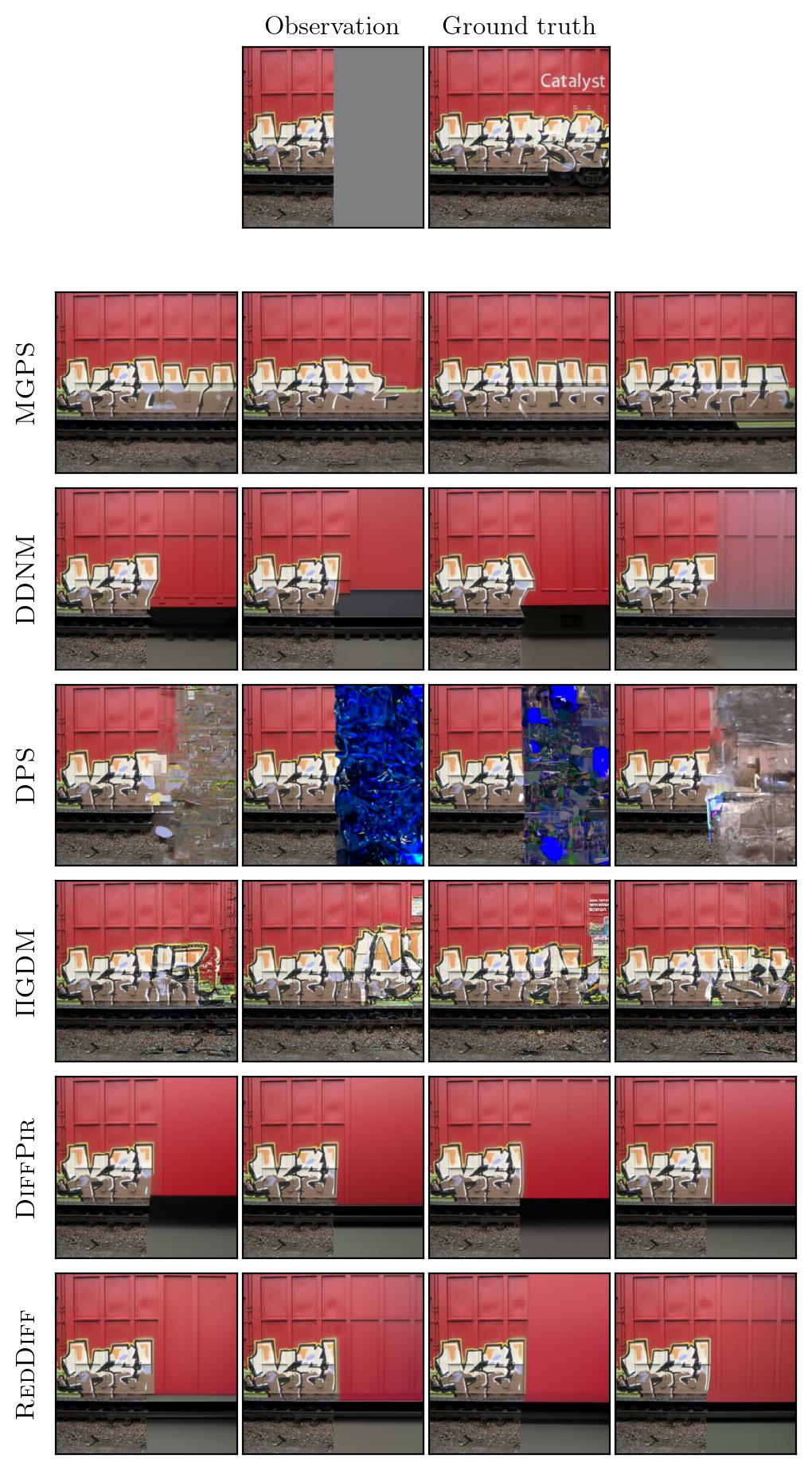}

    }
    \subfigure{
        \includegraphics[width=0.42\textwidth]{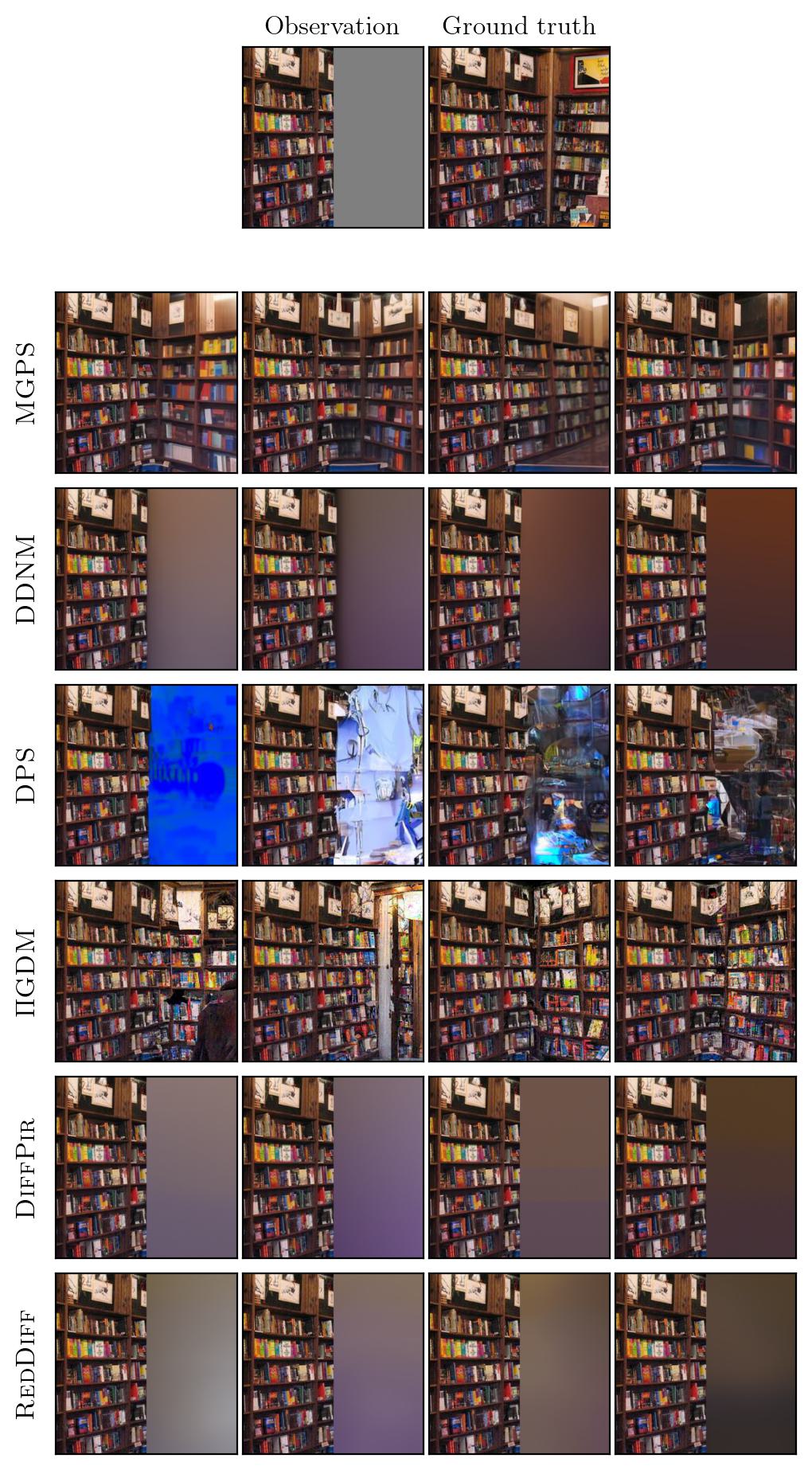}
        \includegraphics[width=0.42\textwidth]{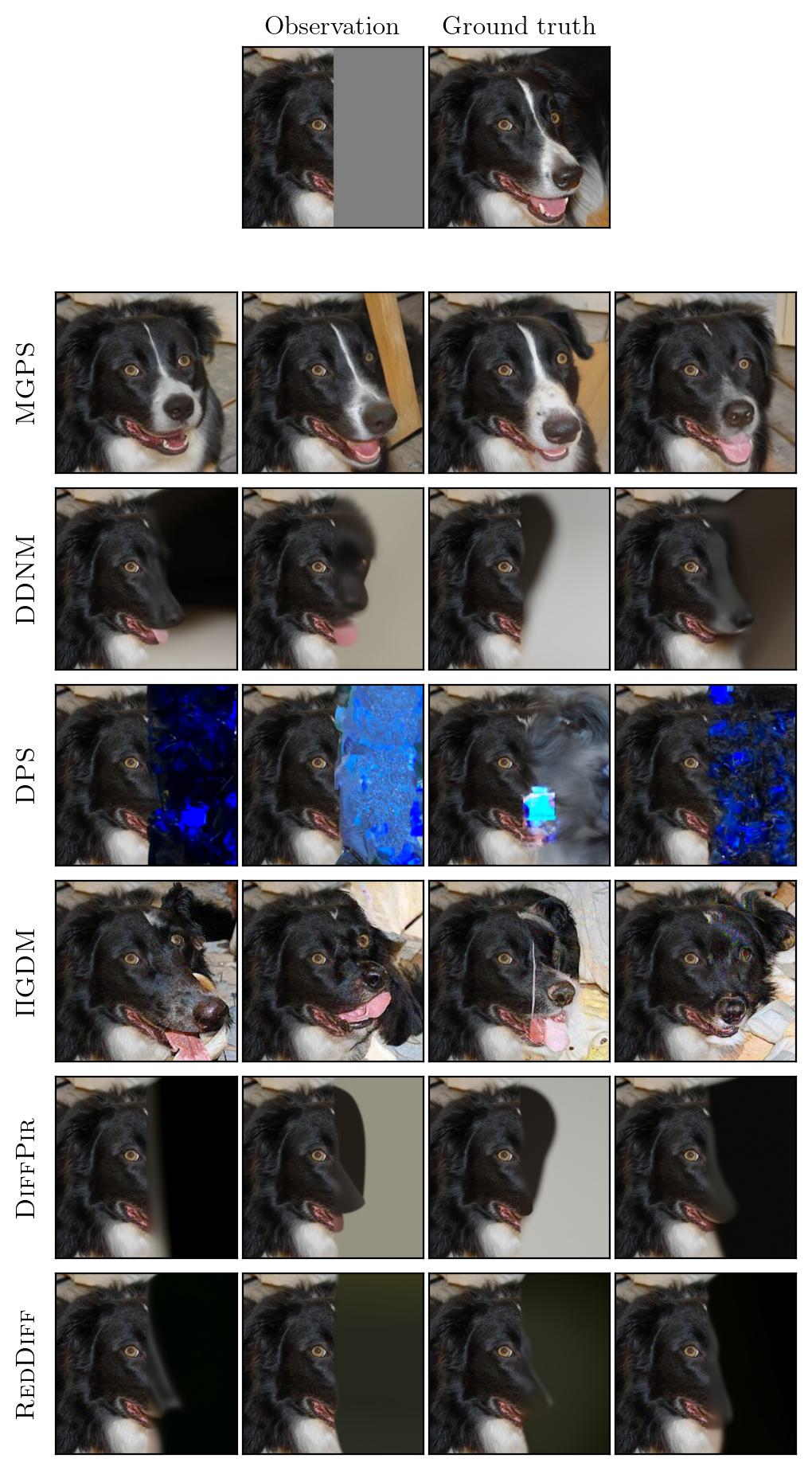}
    }
    \caption{Sample reconstructions with half mask on \imagenet\ dataset.}
    \label{fig:imagnet-inpainting-2}
\end{figure}

\def\prsize{0.49}
\begin{figure}[htb]
    \centering
    \subfigure{
        \includegraphics[width=\prsize\textwidth]{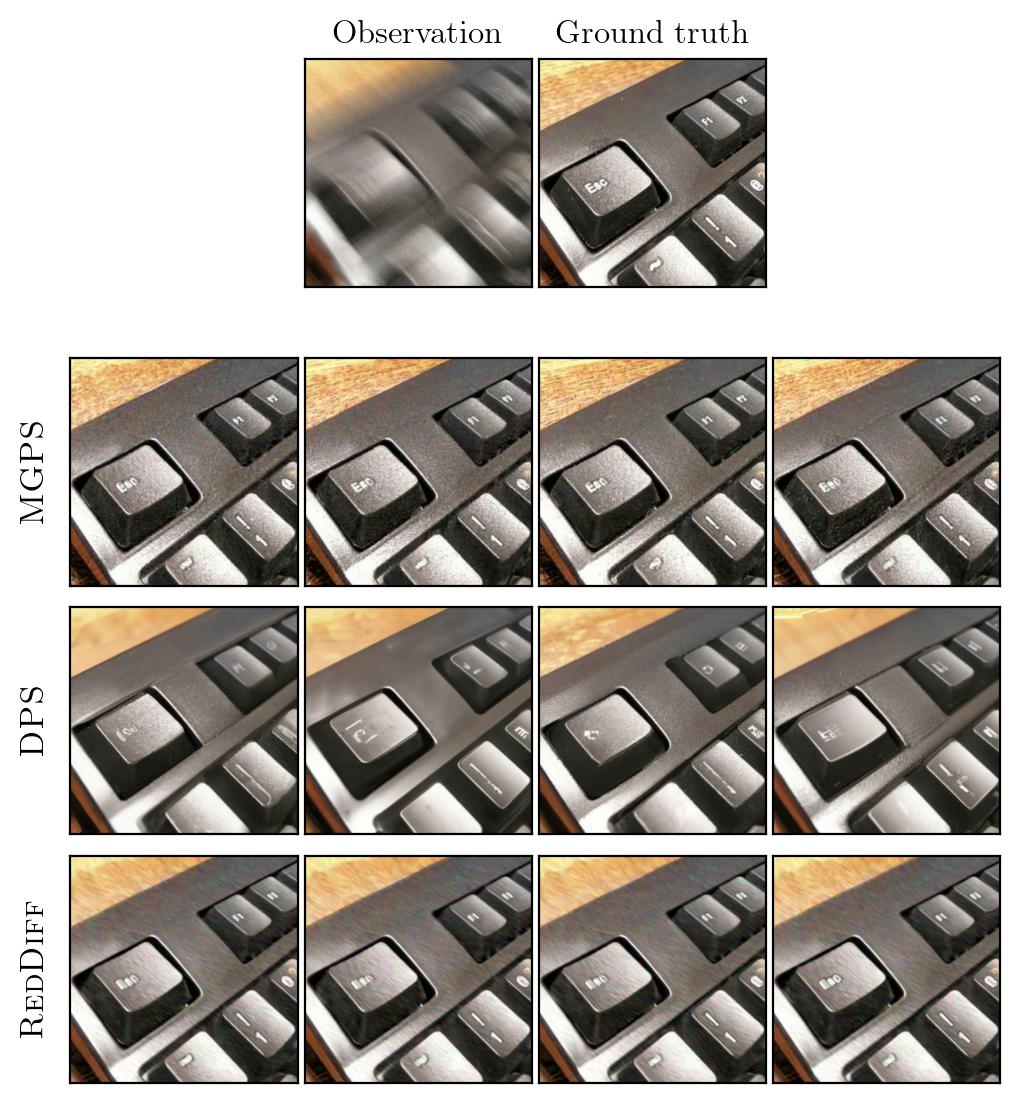}
        \includegraphics[width=\prsize\textwidth]{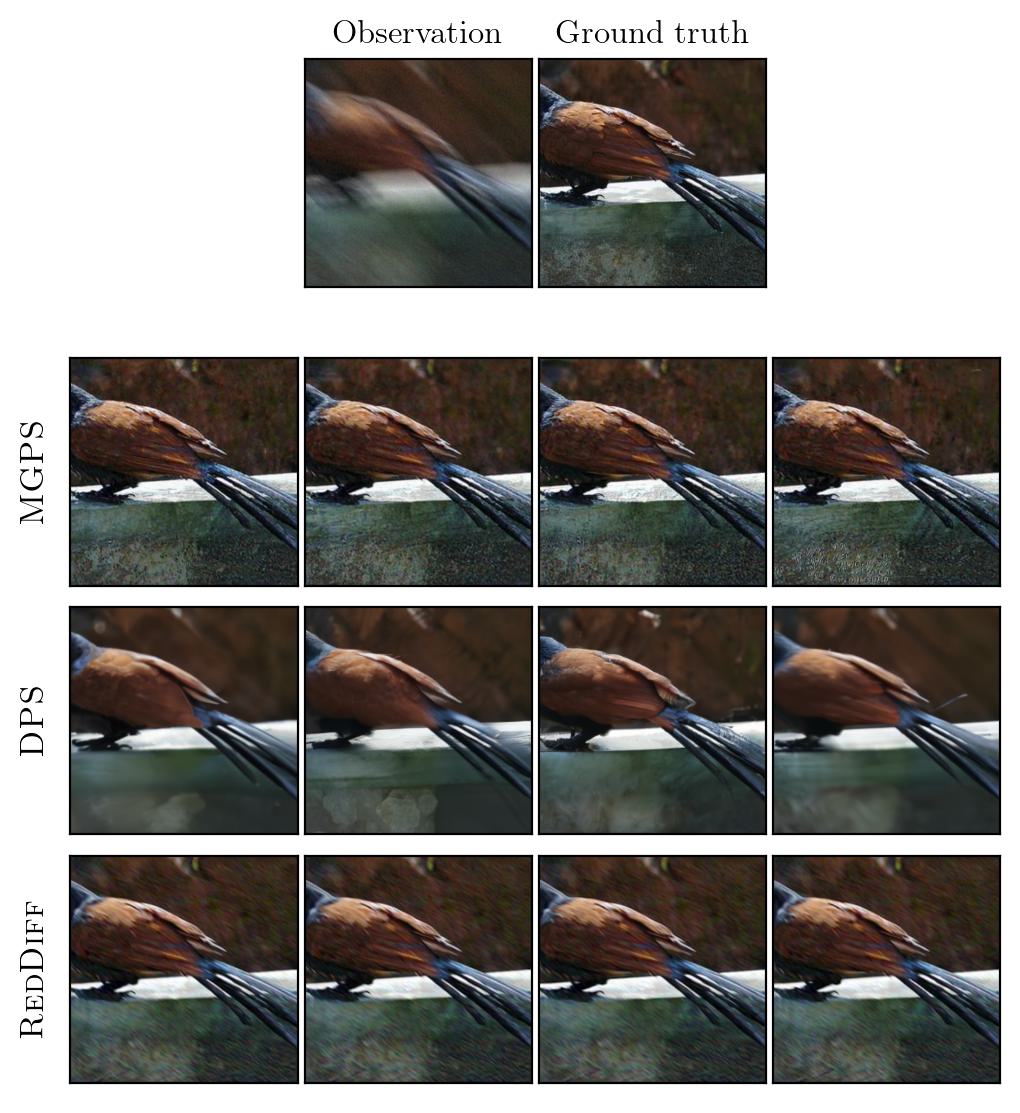}

    }
    \subfigure{
        \includegraphics[width=\prsize\textwidth]{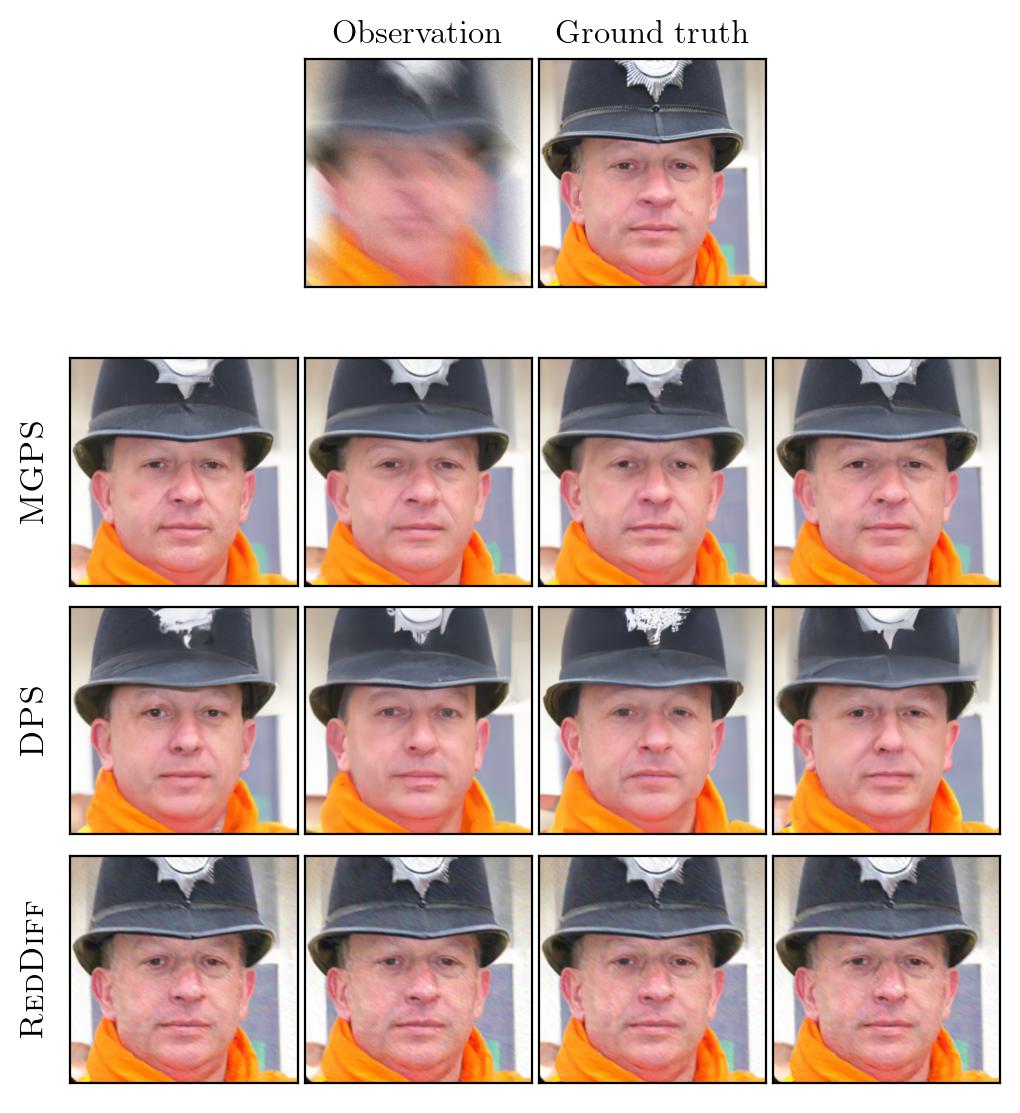}
        \includegraphics[width=\prsize\textwidth]{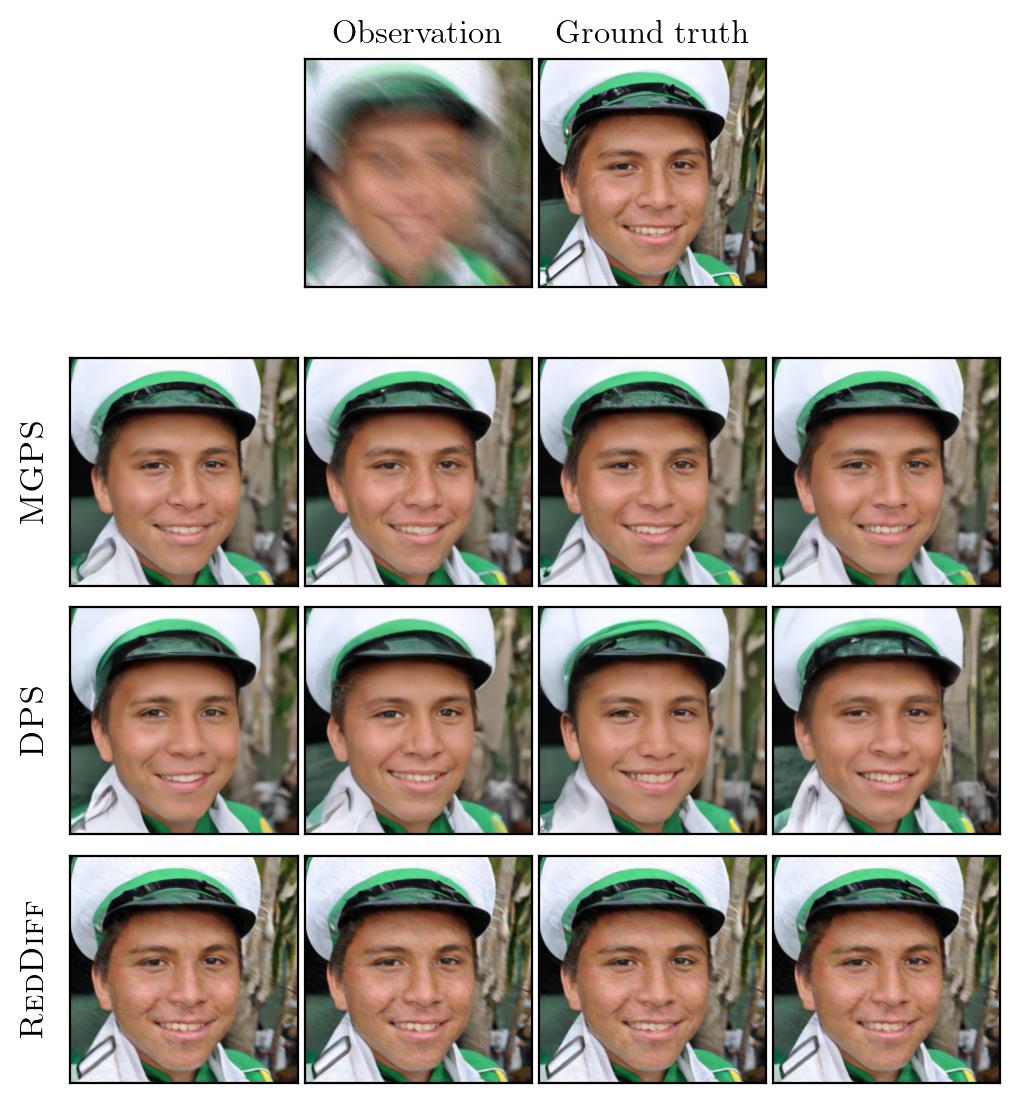}
    }
    \caption{Sample reconstructions on motion deblurring task.}
\end{figure}
\begin{figure}[htb]
    \centering
    \subfigure{
        \includegraphics[width=0.42\textwidth]{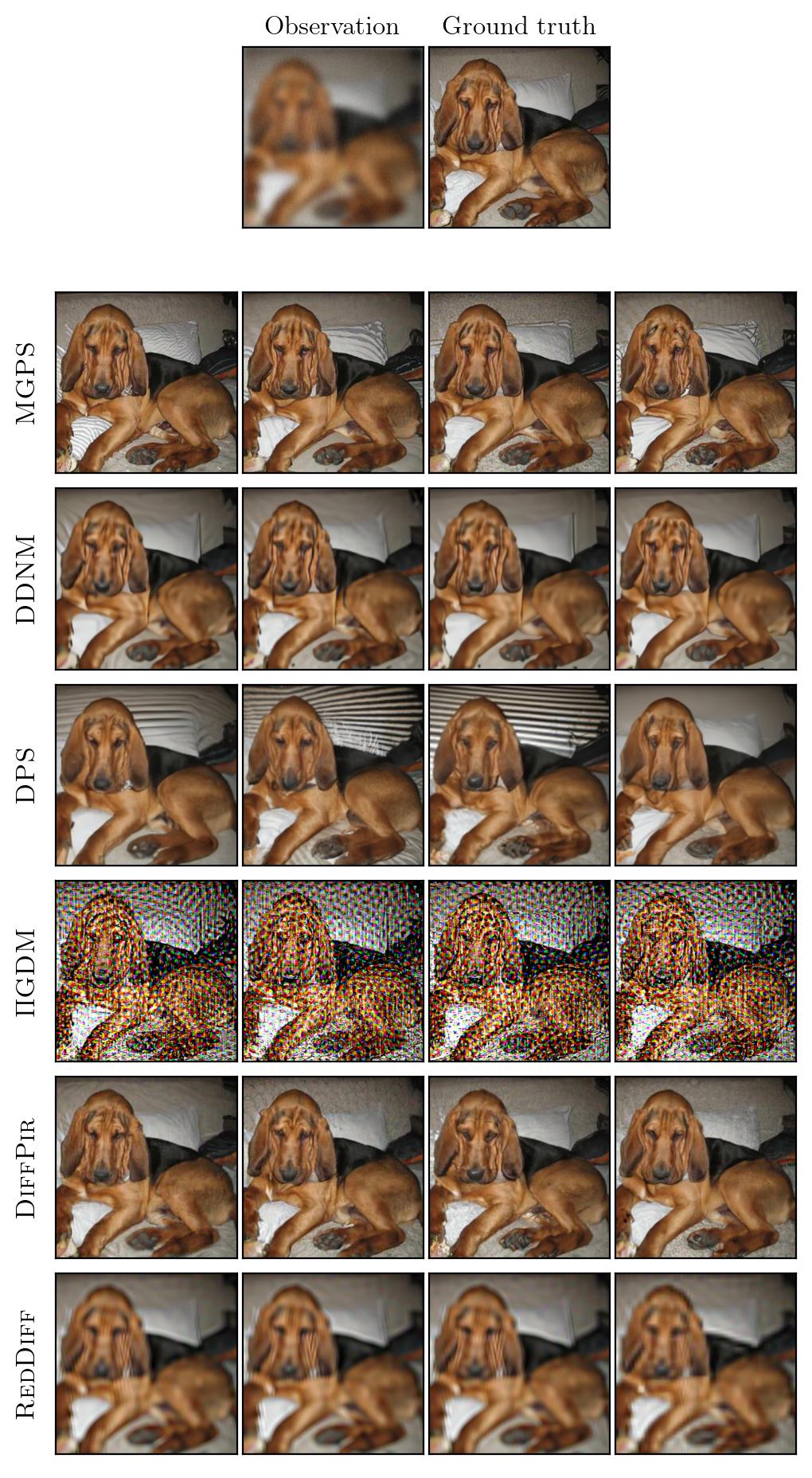}
        \includegraphics[width=0.42\textwidth]{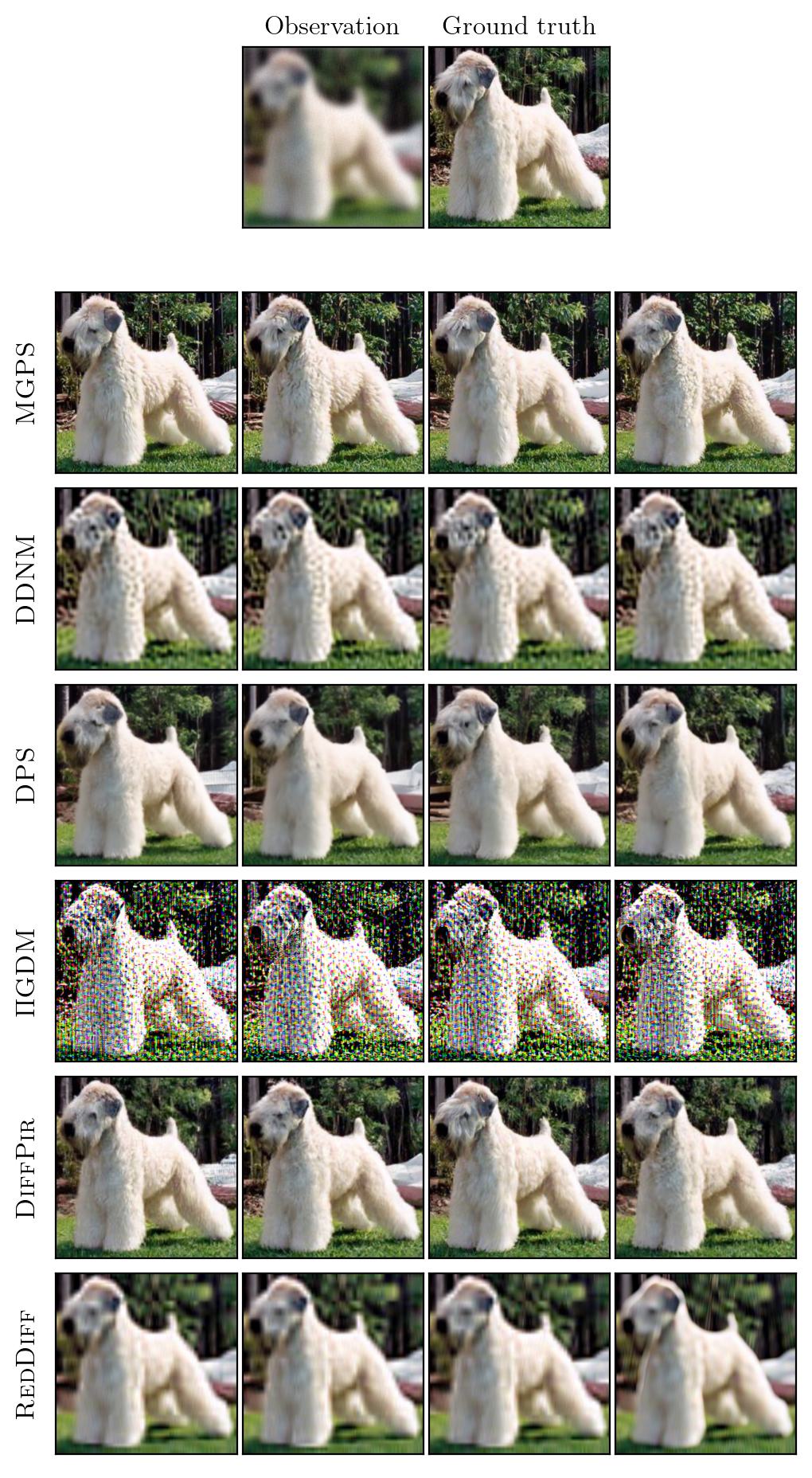}

    }
    \subfigure{
        \includegraphics[width=0.42\textwidth]{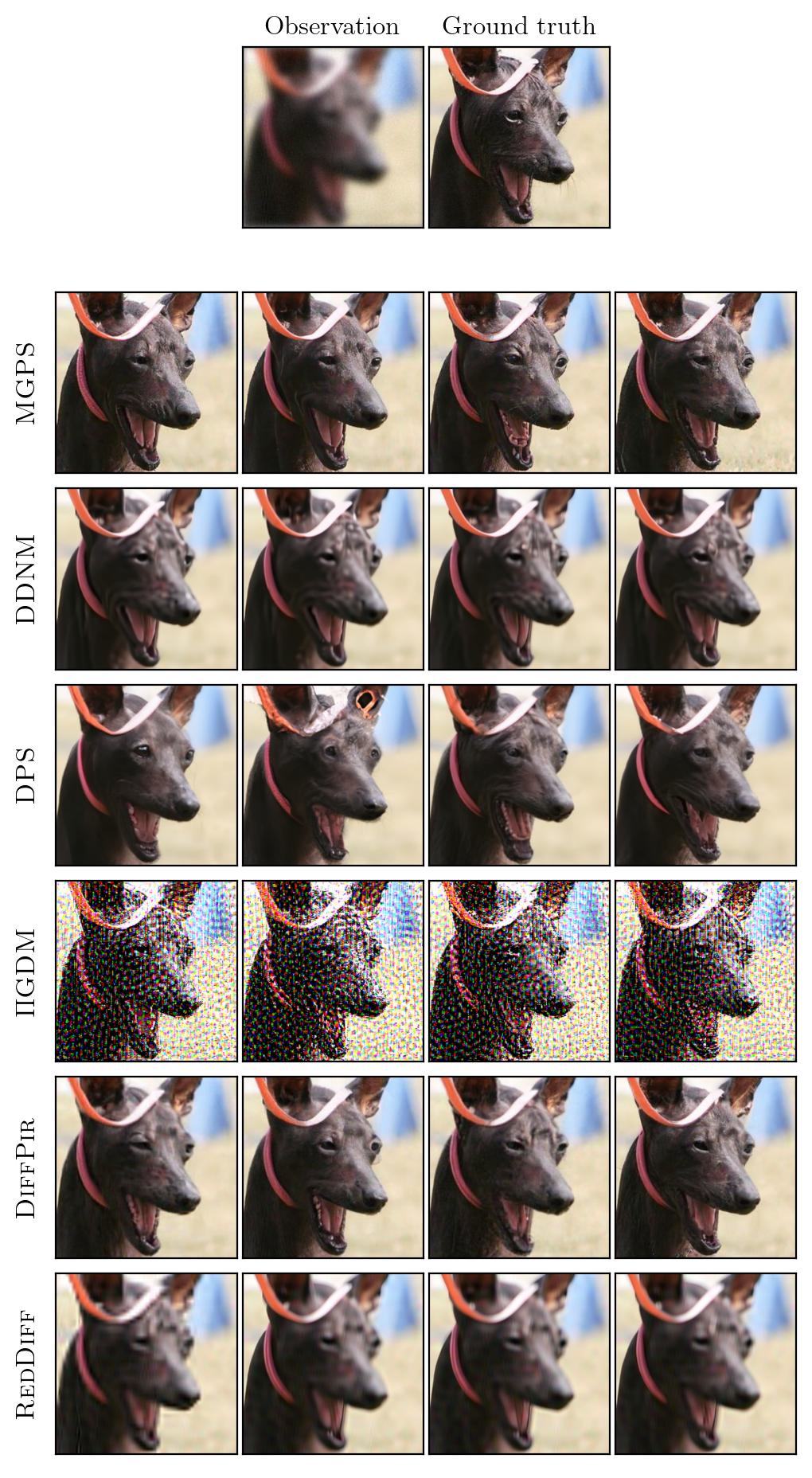}
        \includegraphics[width=0.42\textwidth]{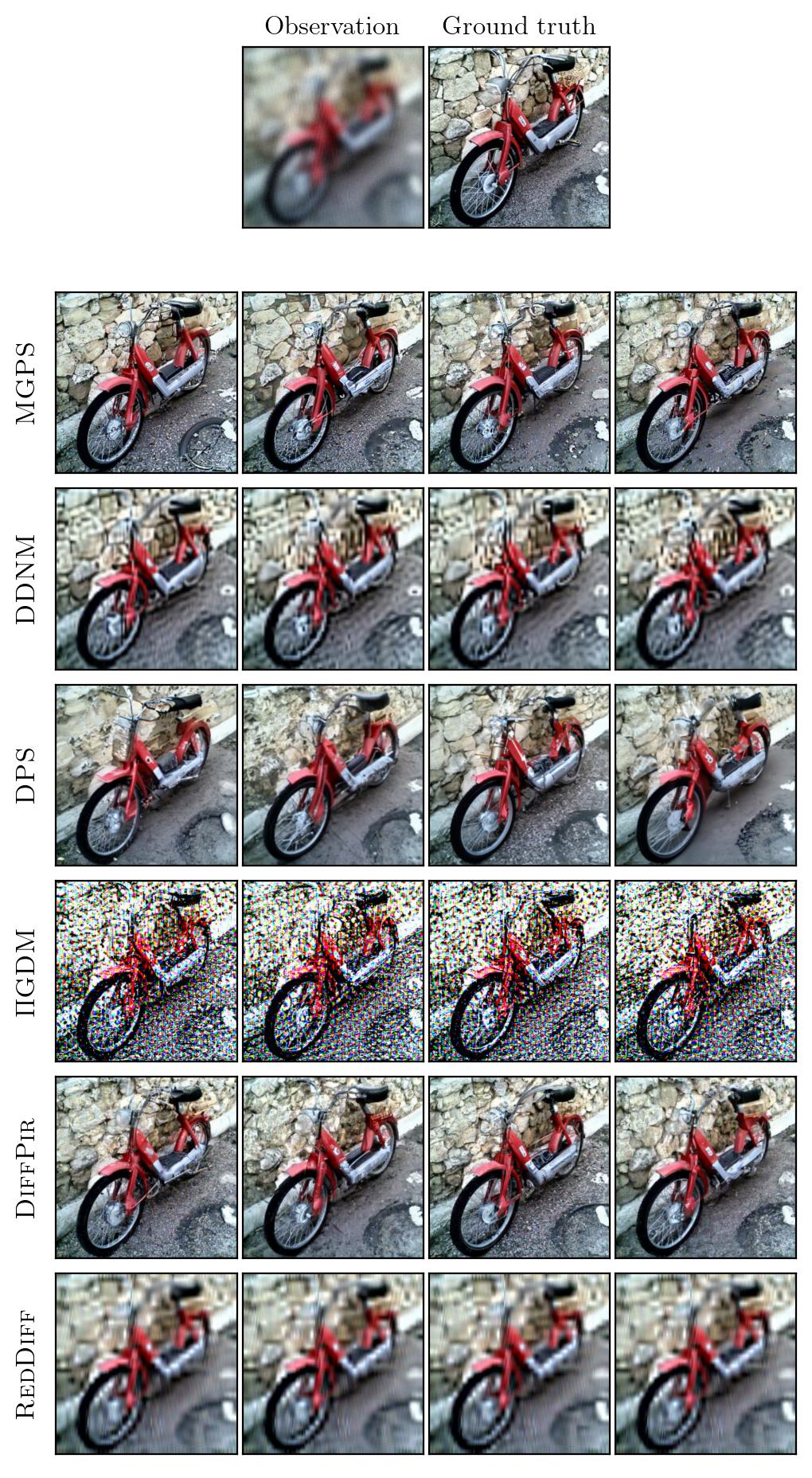}
    }
    \caption{Sample reconstructions on Gaussian deblurring task.}
    \label{fig:imagnet-inpainting-3}
\end{figure}

\begin{figure}[htb]
    \centering
    \subfigure{
        \includegraphics[width=\prsize\textwidth]{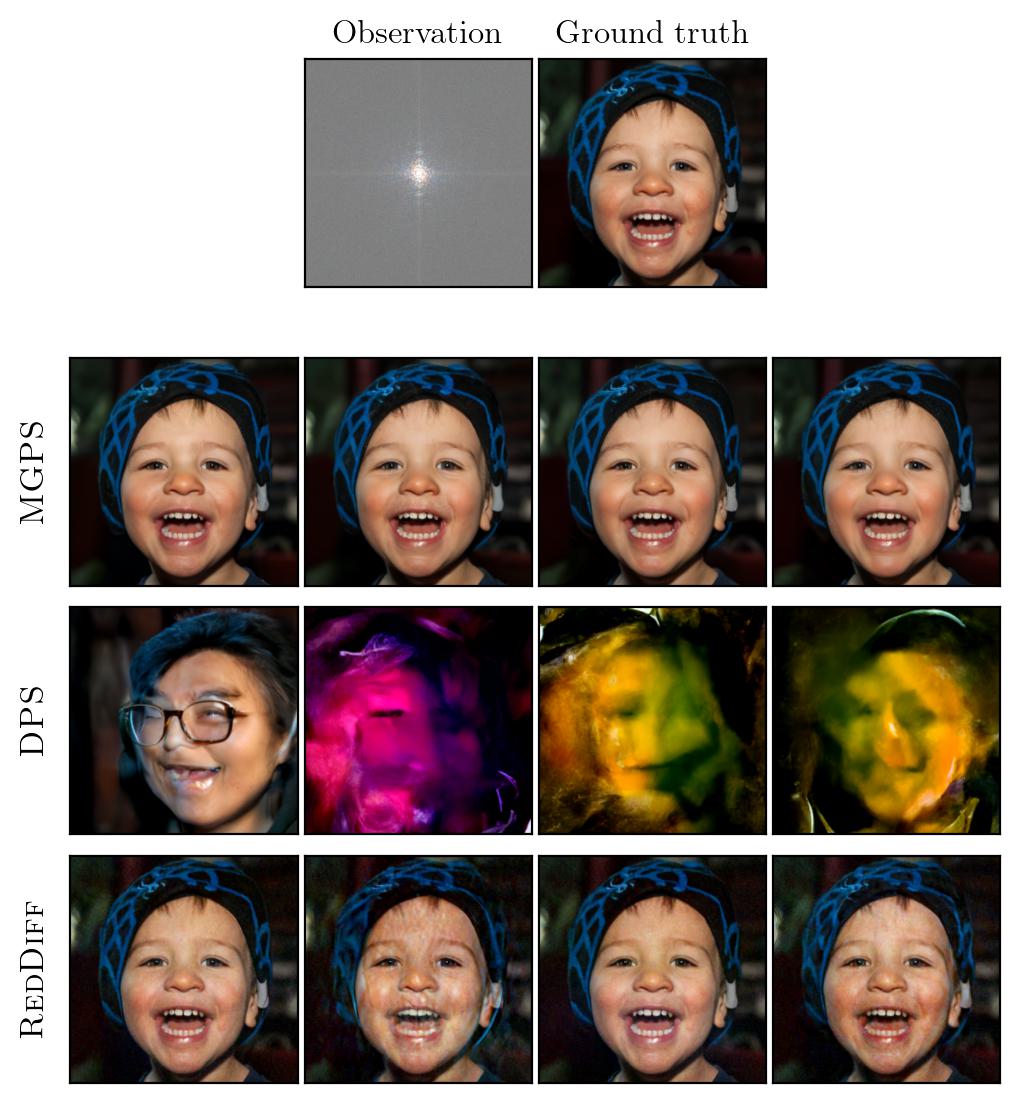}
        \includegraphics[width=\prsize\textwidth]{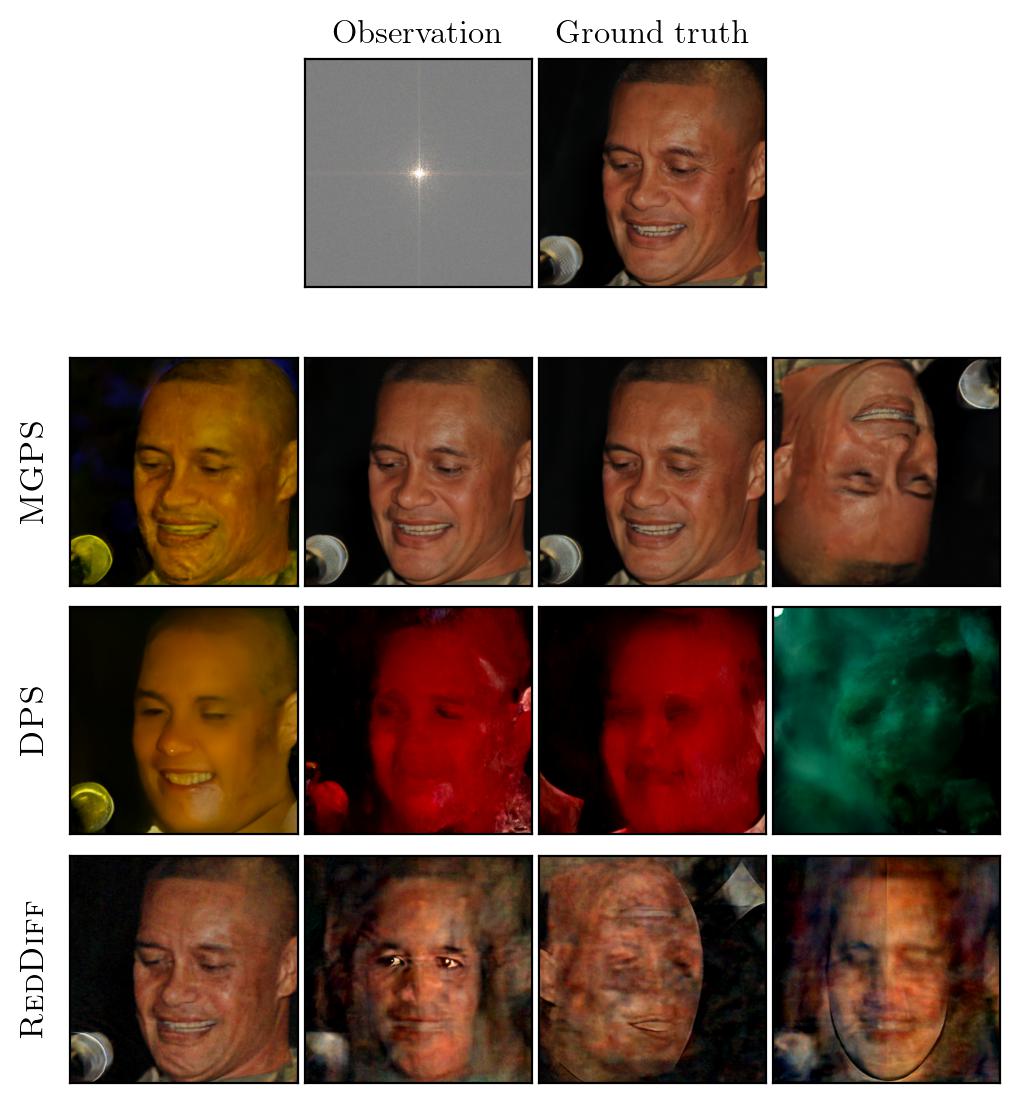}

    }
    \subfigure{
        \includegraphics[width=\prsize\textwidth]{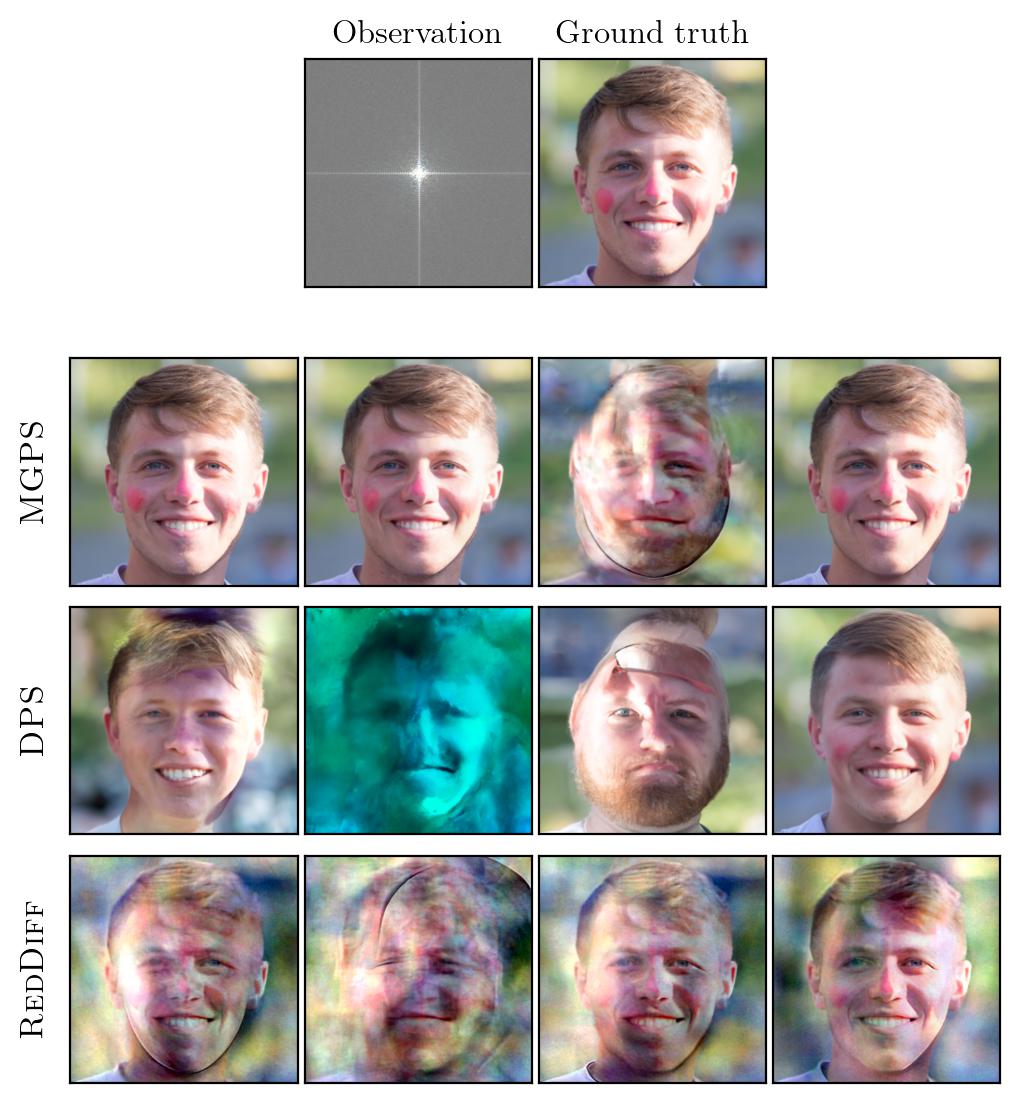}
        \includegraphics[width=\prsize\textwidth]{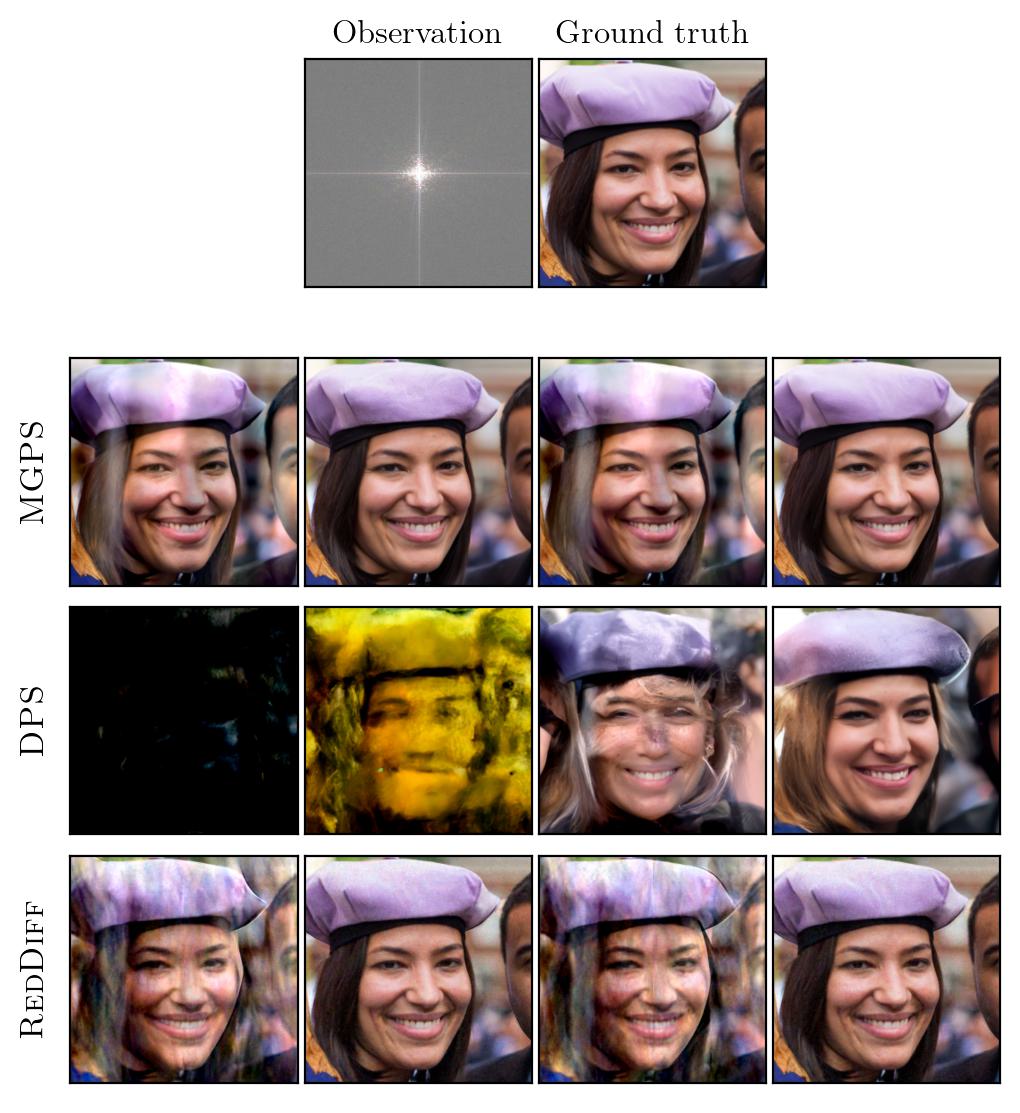}
    }
    \caption{Sample reconstructions on phase retrieval task.}
\end{figure}
\begin{figure}[htb]
    \centering
    \subfigure{
        \includegraphics[width=0.45\textwidth]{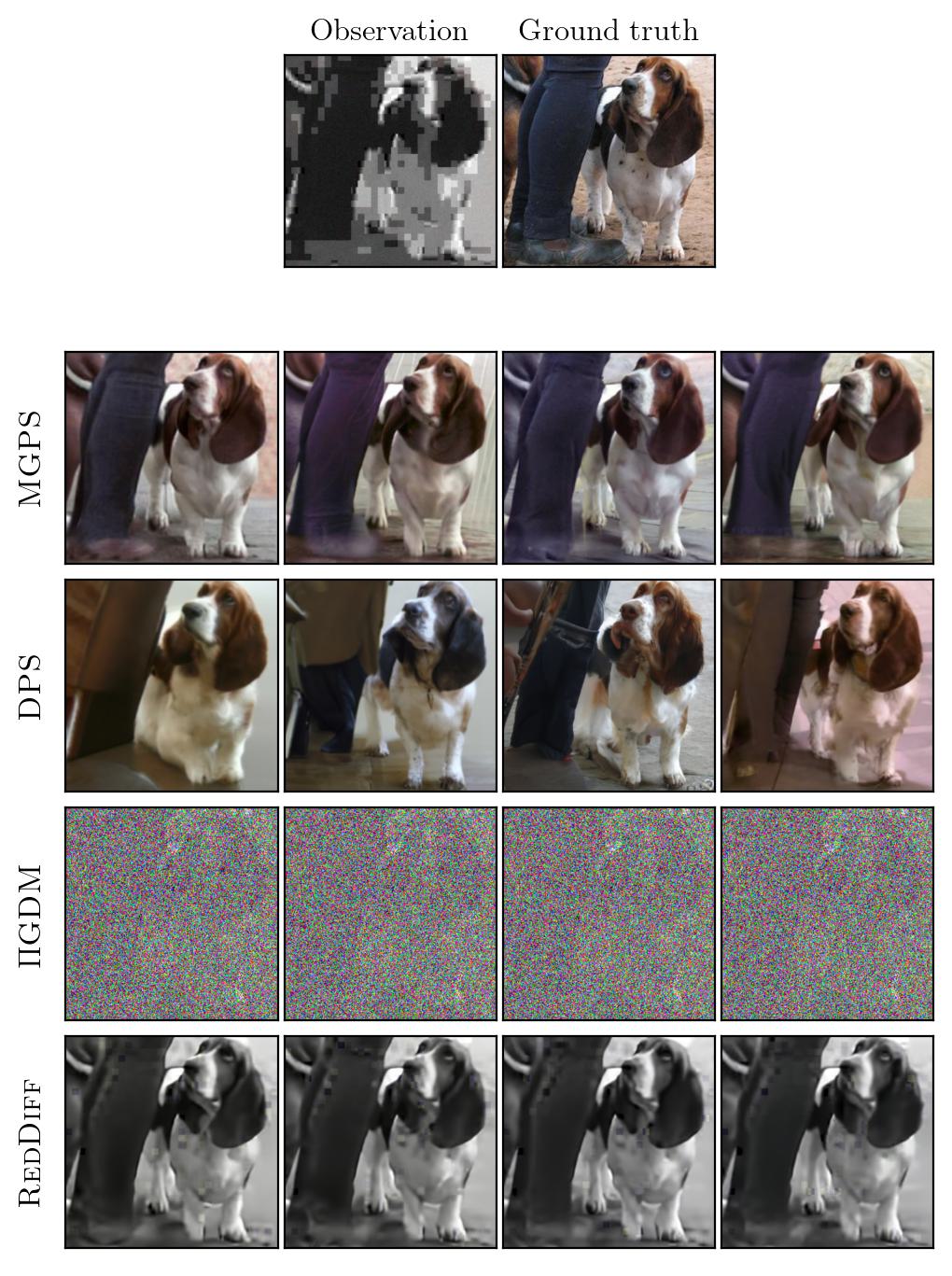}
        \includegraphics[width=0.45\textwidth]{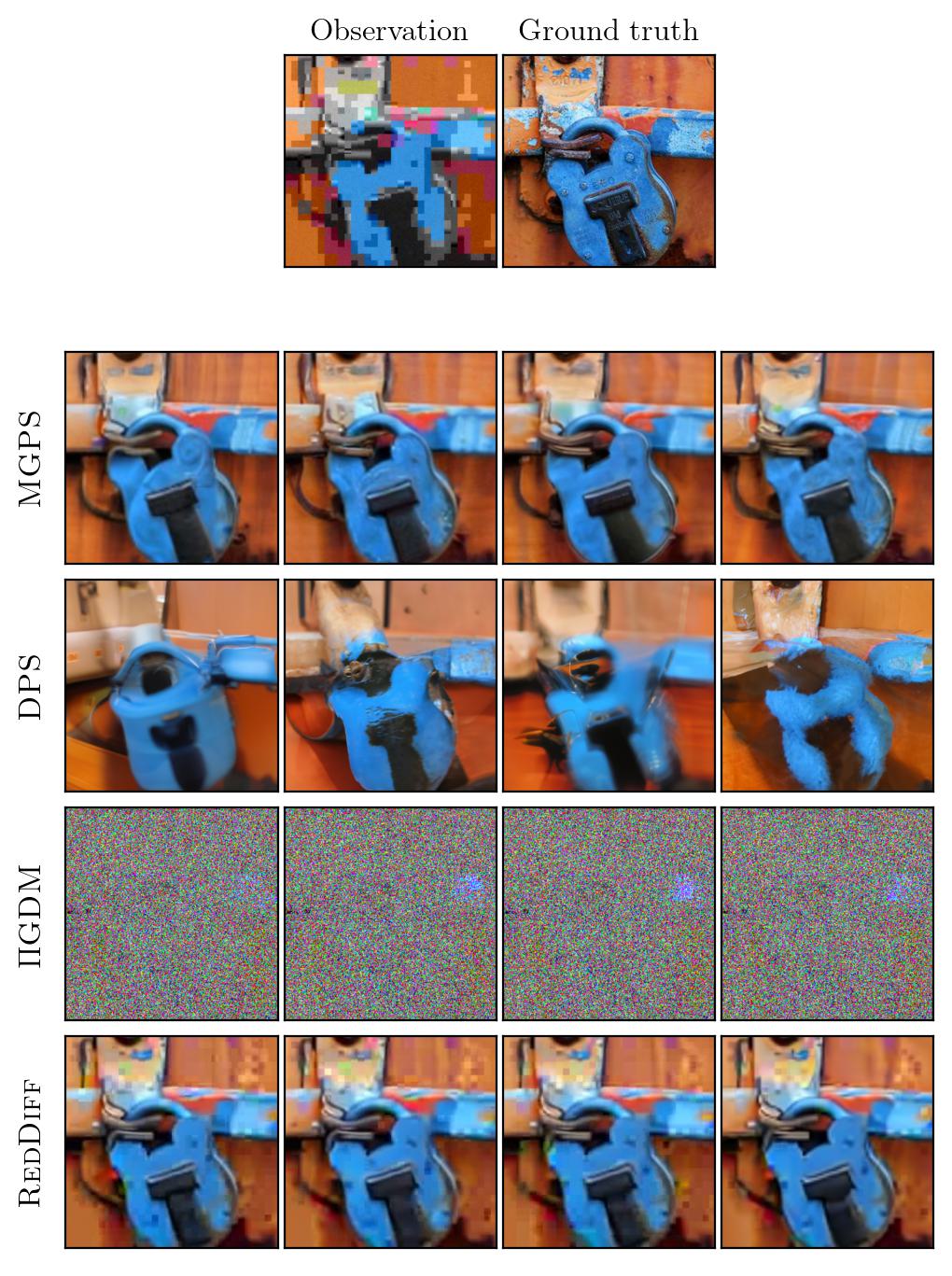}

    }
    \subfigure{
        \includegraphics[width=0.45\textwidth]{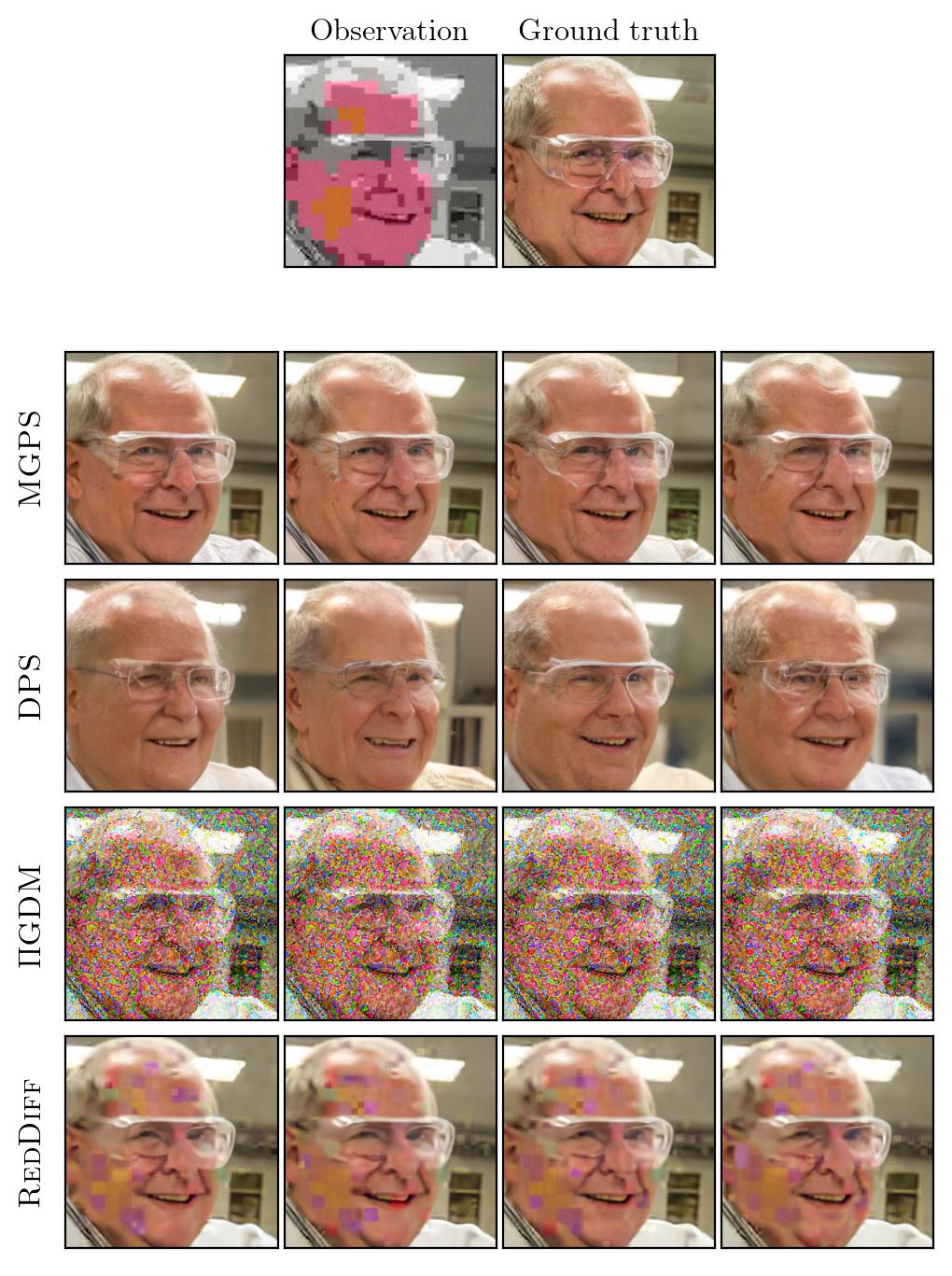}
        \includegraphics[width=0.45\textwidth]{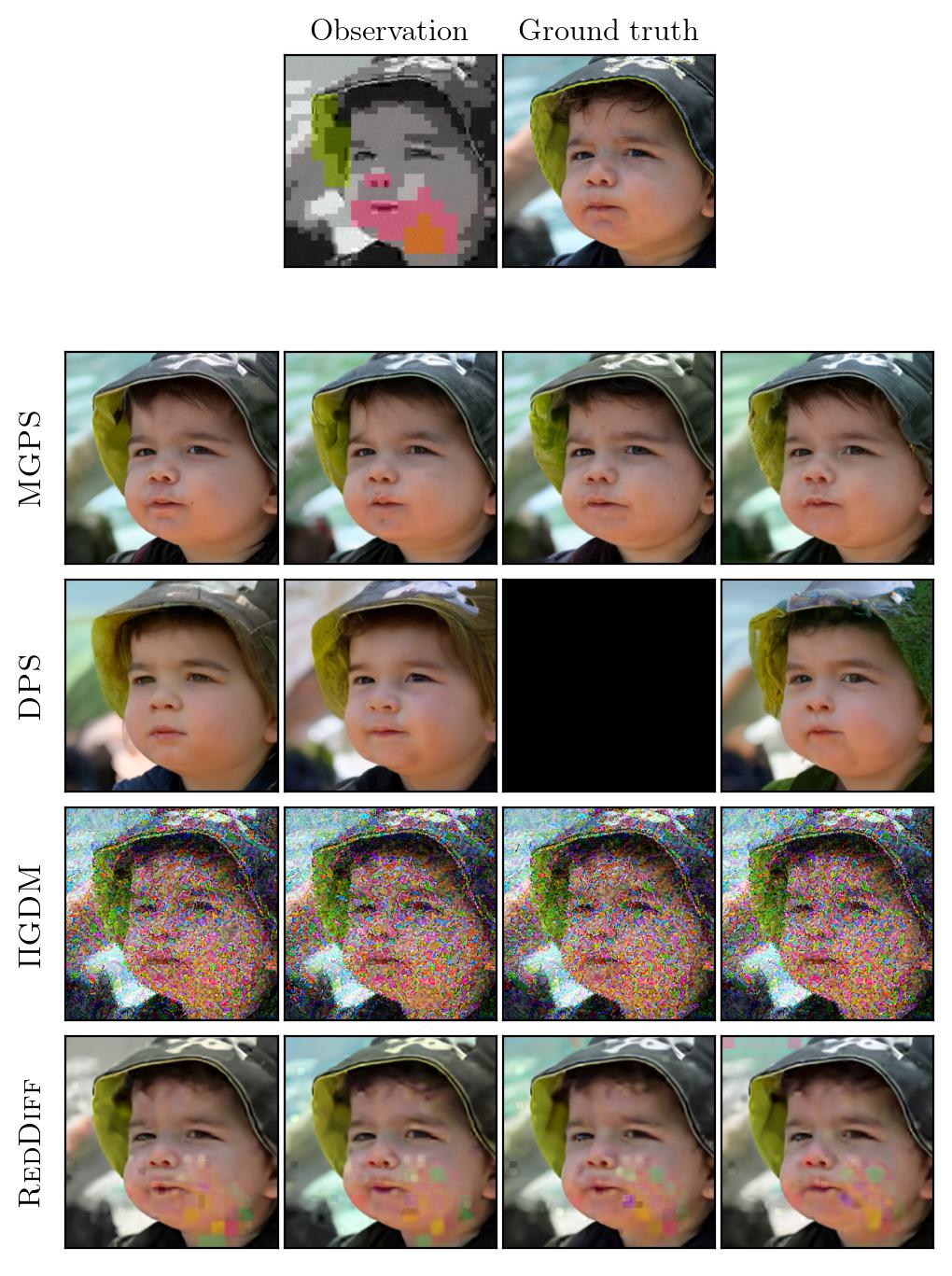}
    }
    \caption{Sample reconstructions on JPEG 2 task.}
\end{figure}
\begin{figure}[htb]
    \centering
    \subfigure{
        \includegraphics[width=0.45\textwidth]{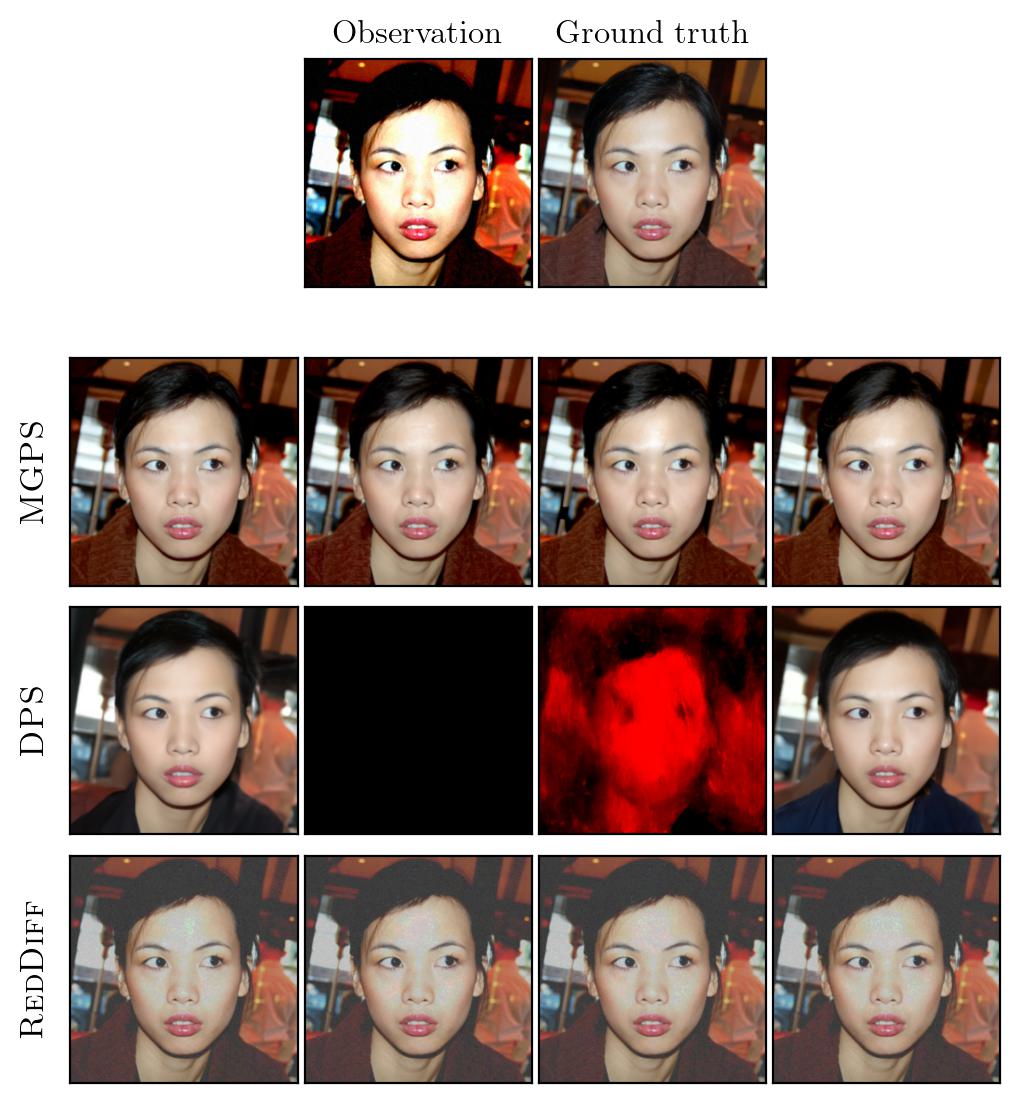}
        \includegraphics[width=0.45\textwidth]{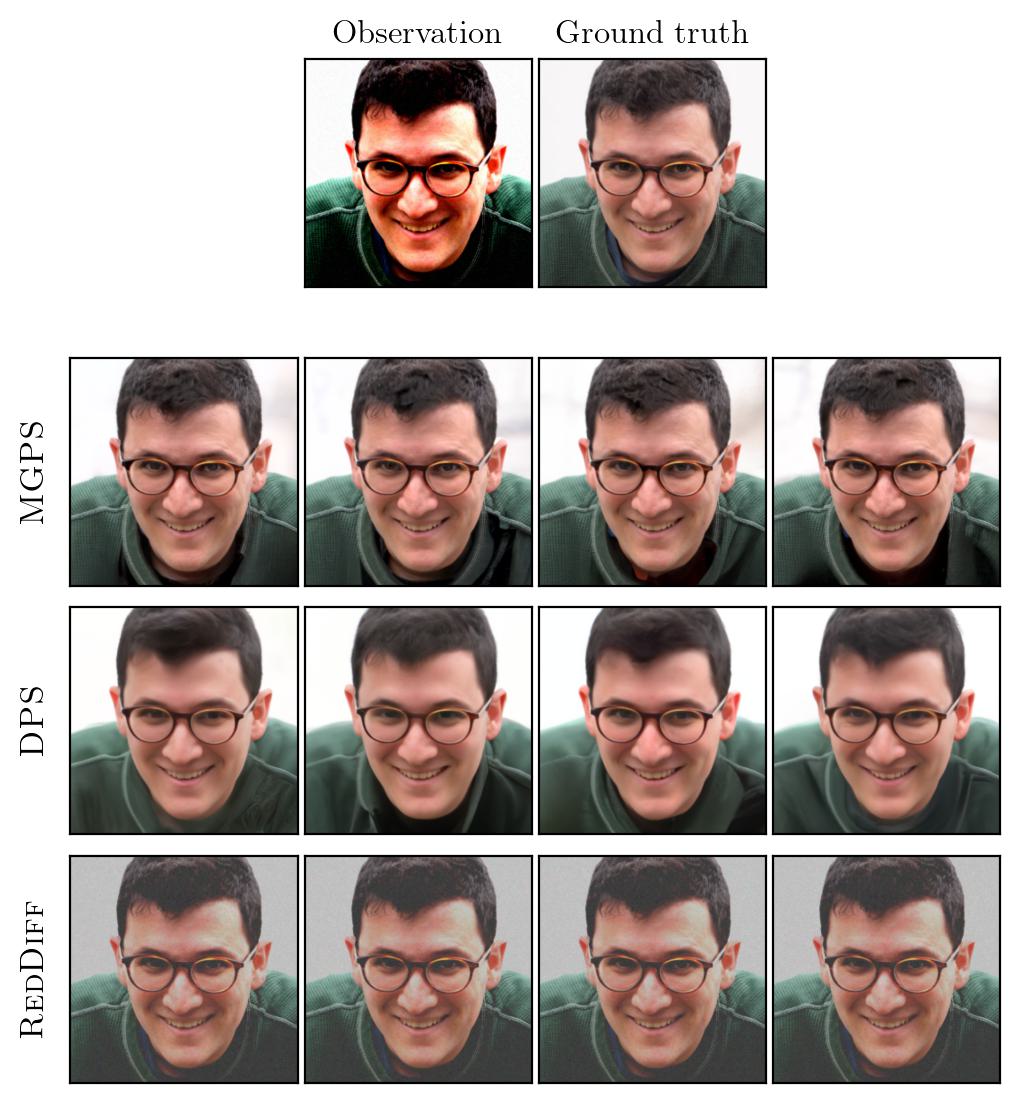}

    }
    \subfigure{
        \includegraphics[width=0.45\textwidth]{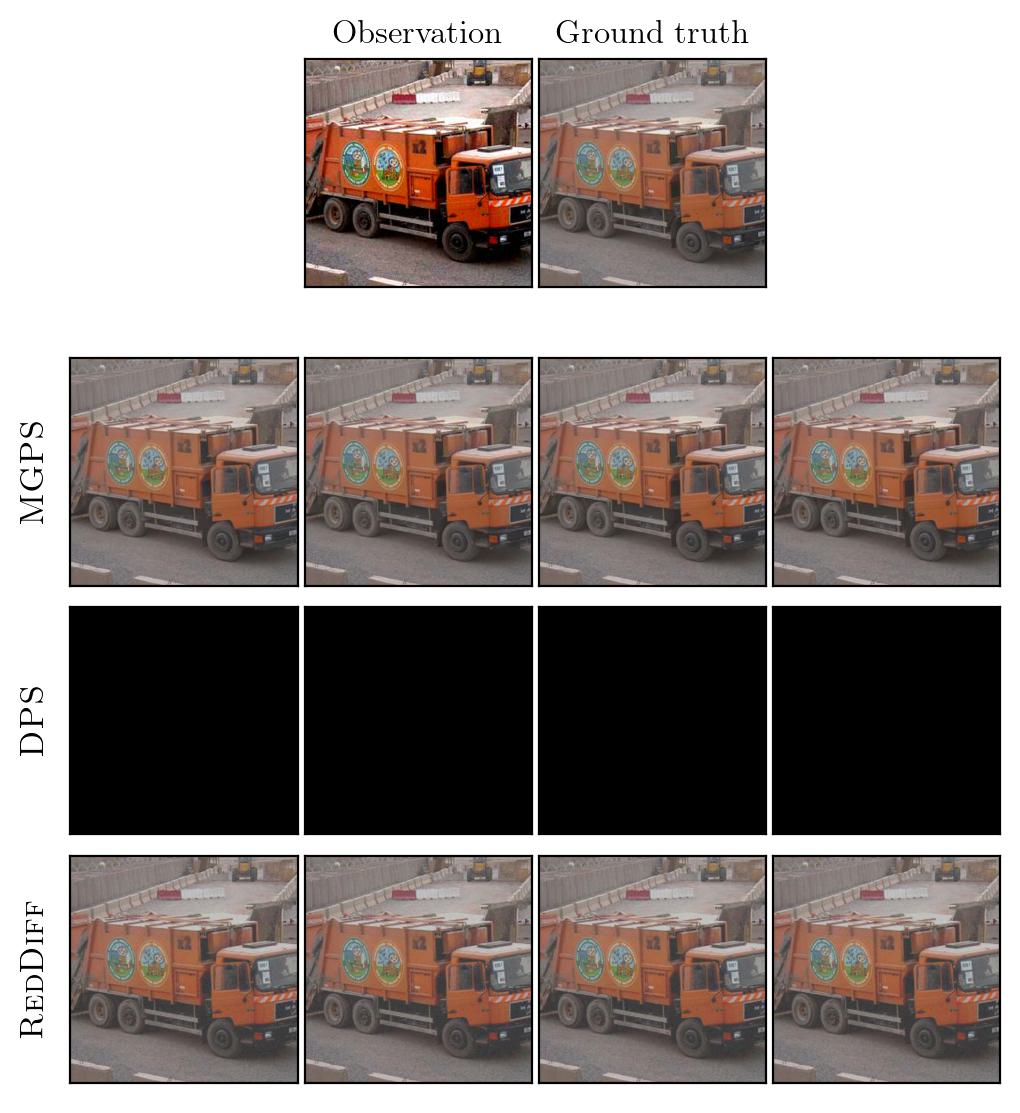}
        \includegraphics[width=0.45\textwidth]{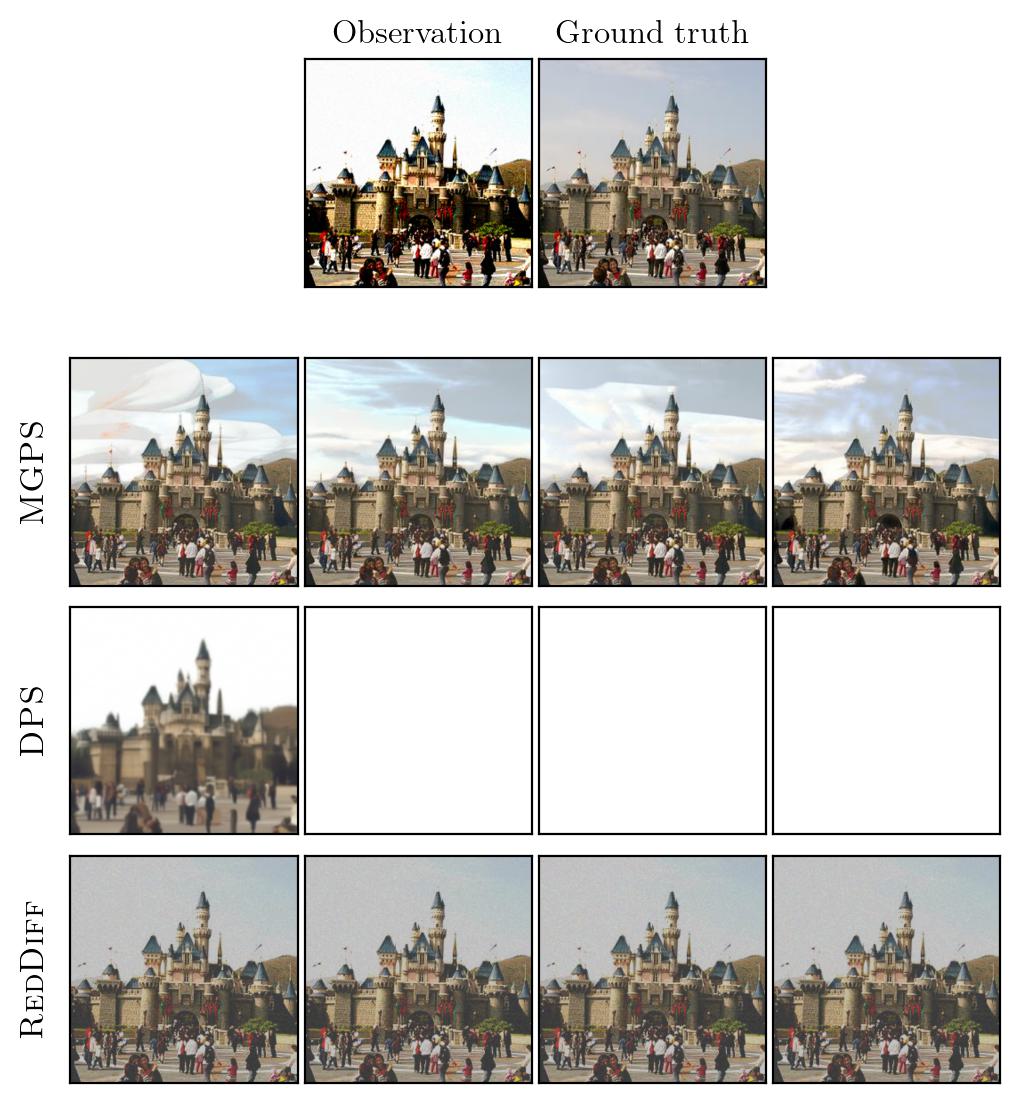}
    }
    \caption{Sample reconstructions on high dynamic range task.}
\end{figure}
\begin{figure}[htb]
    \centering
    \subfigure{
        \includegraphics[width=0.42\textwidth]{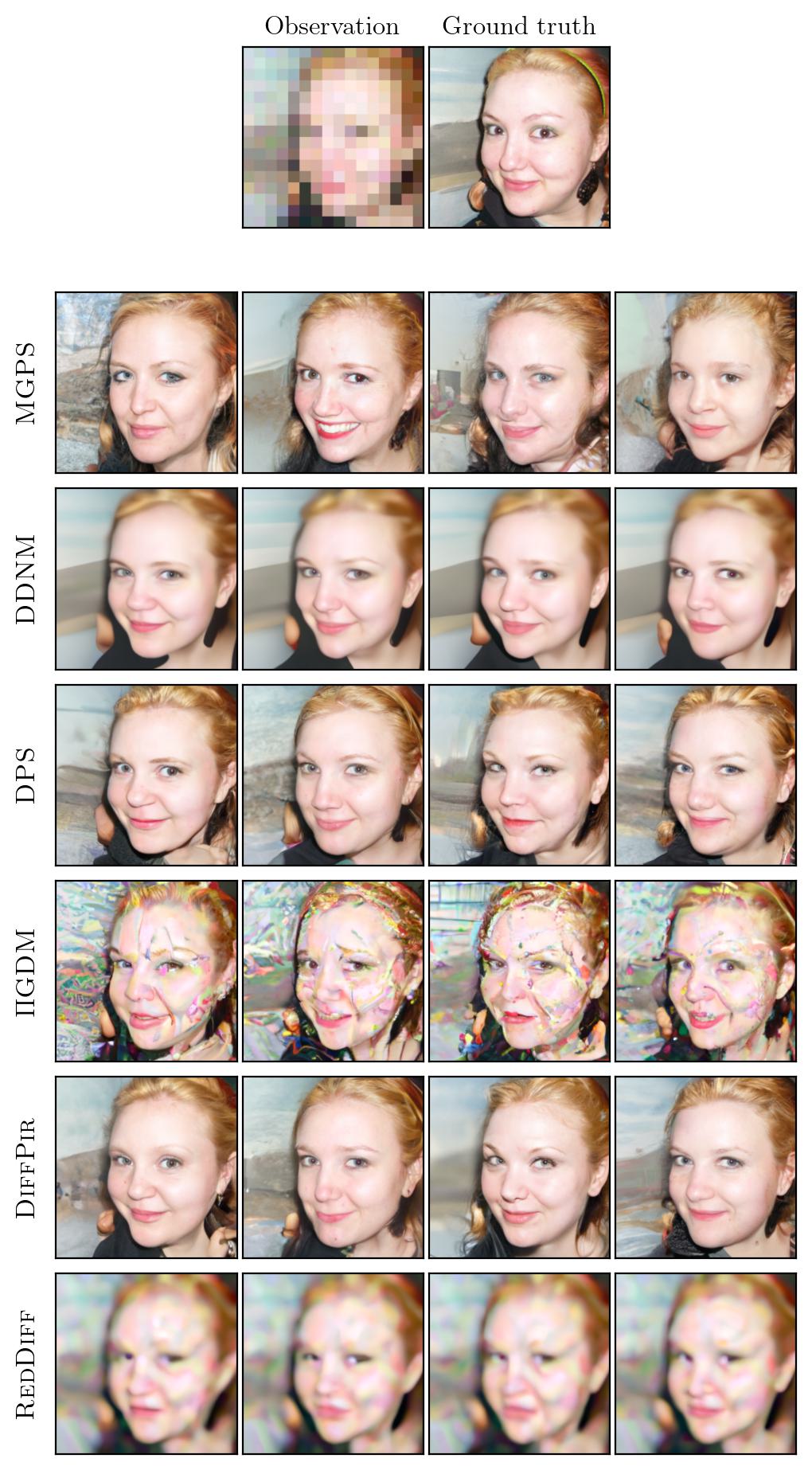}
        \includegraphics[width=0.42\textwidth]{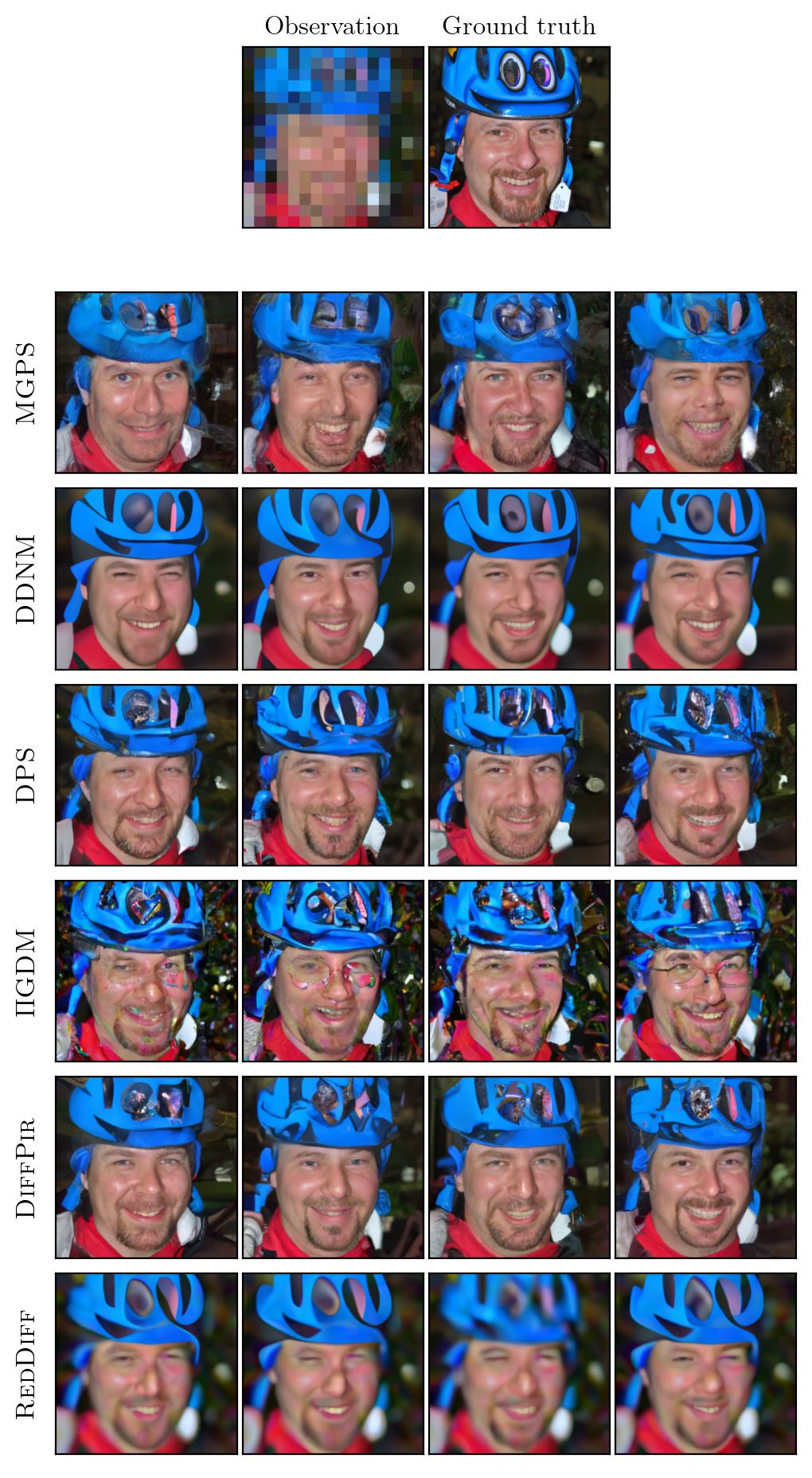}

    }
    \subfigure{
        \includegraphics[width=0.42\textwidth]{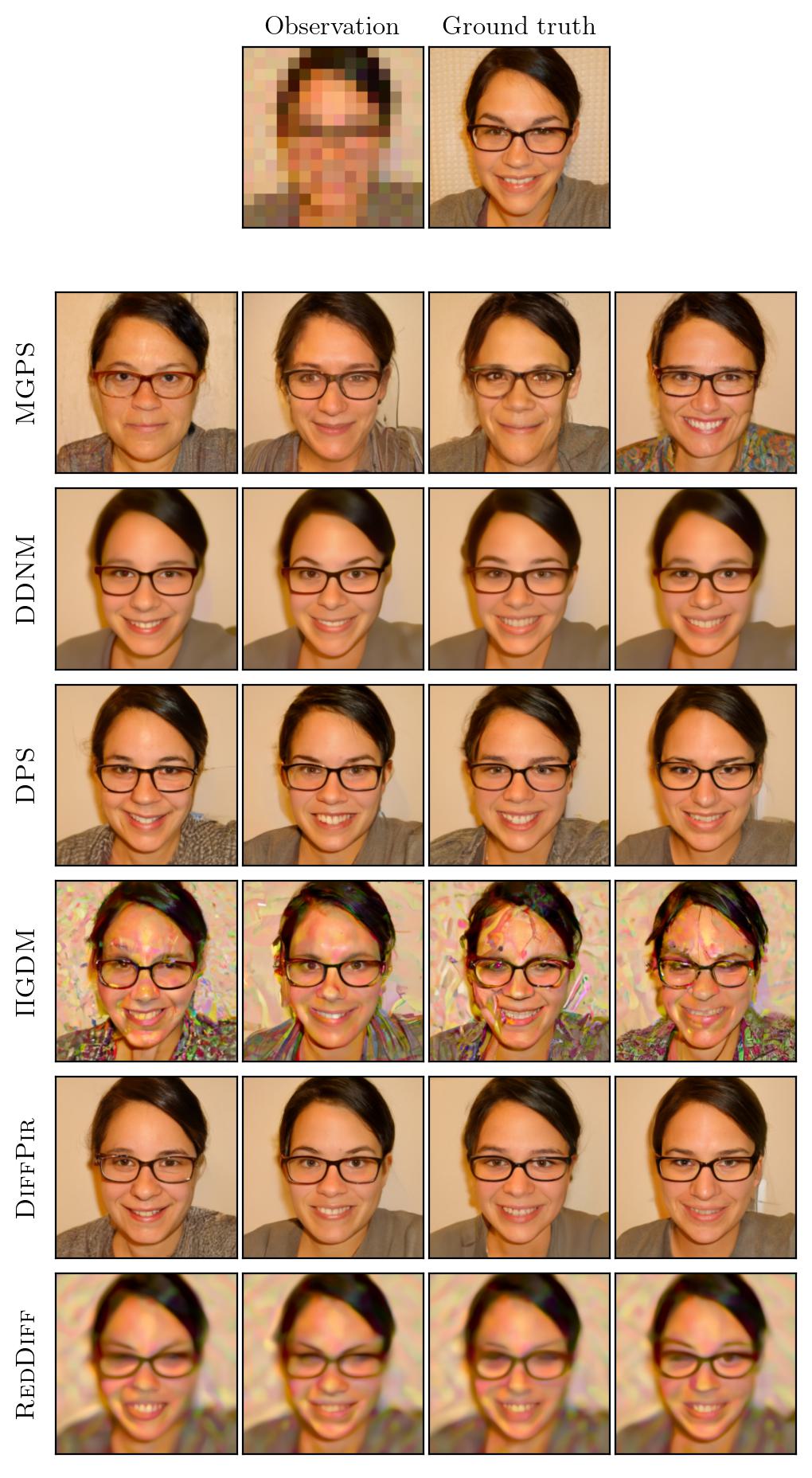}
        \includegraphics[width=0.42\textwidth]{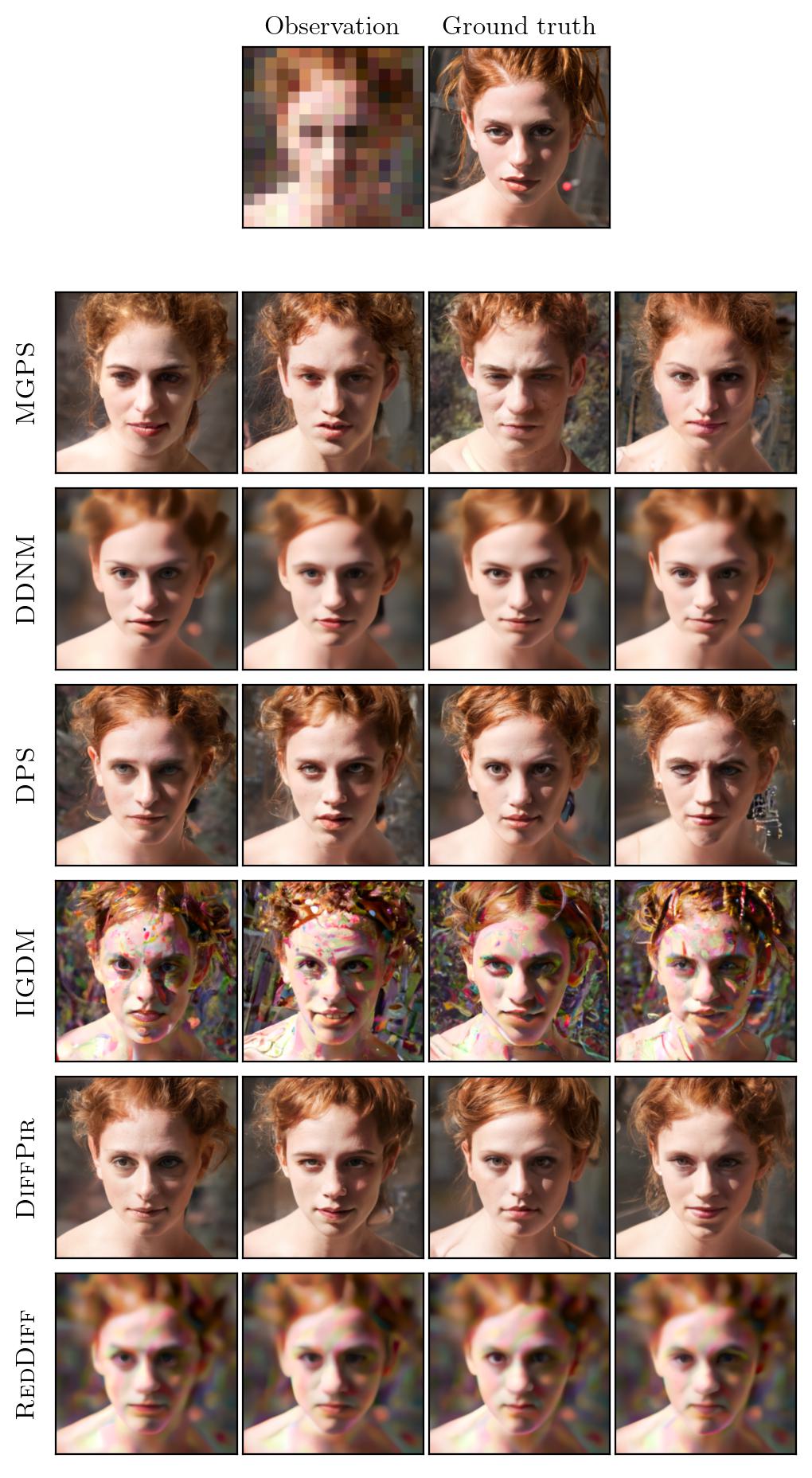}
    }
    \caption{Sample reconstructions on SR (16$\times$) task.}
\end{figure}
\begin{figure}[htb]
    \centering
    \subfigure{
        \includegraphics[width=0.42\textwidth]{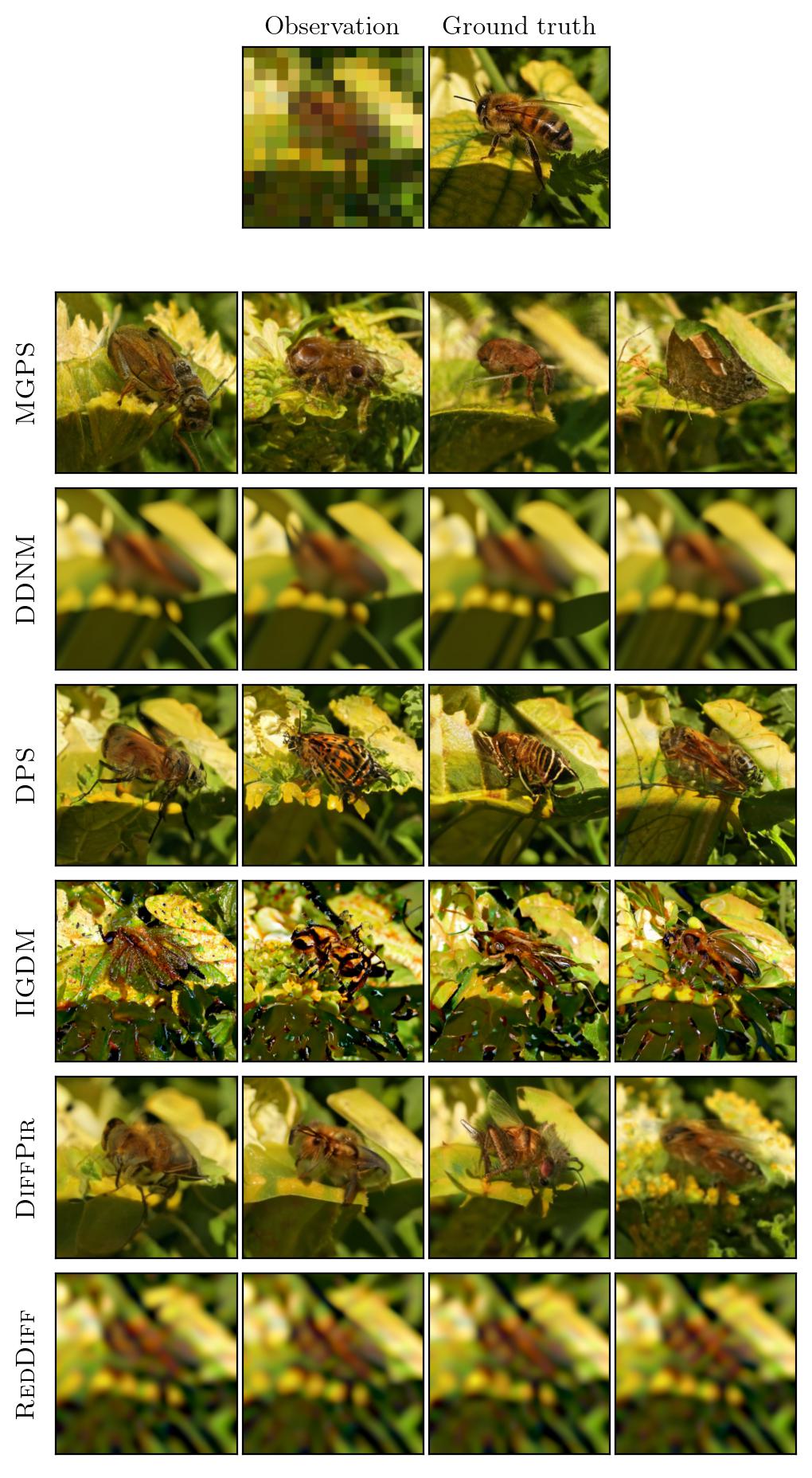}
        \includegraphics[width=0.42\textwidth]{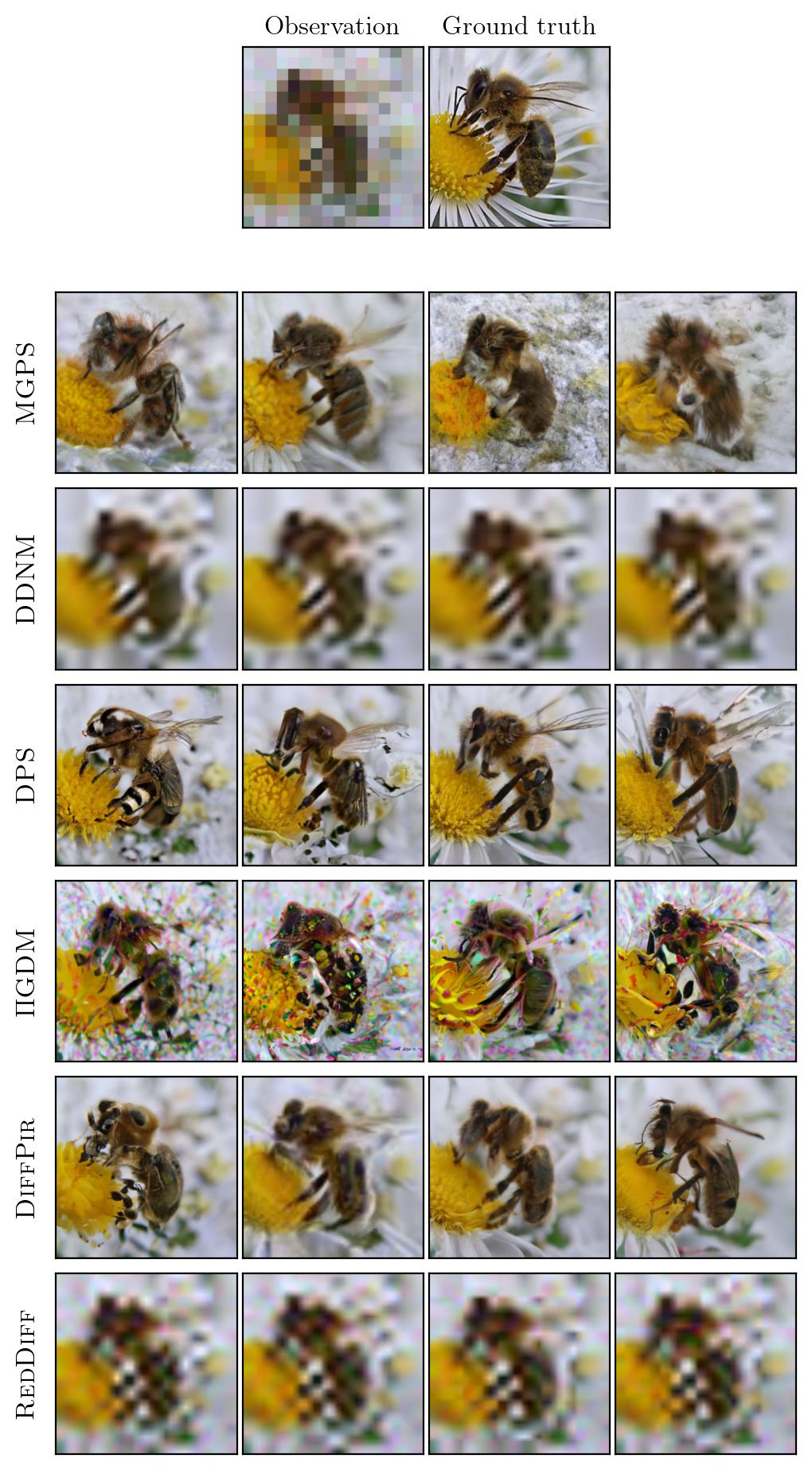}

    }
    \subfigure{
        \includegraphics[width=0.42\textwidth]{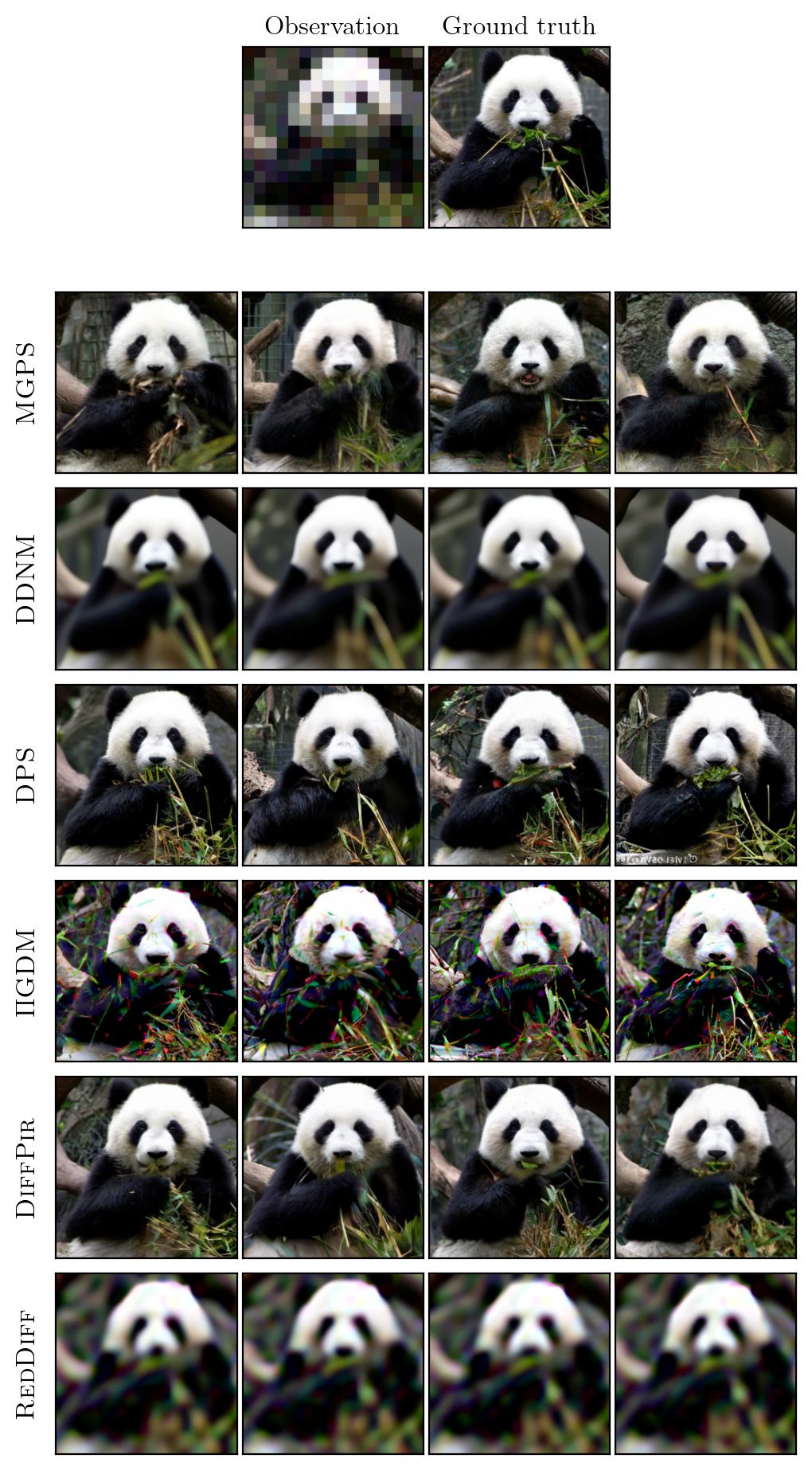}
    }
    \caption{More sample reconstructions on SR (16$\times$) task.}
    \label{fig:imagnet-inpainting-4}
\end{figure}

\begin{figure}[htb]
    \centering
    \subfigure[JPEG 2]{
        \includegraphics[width=0.45\textwidth]{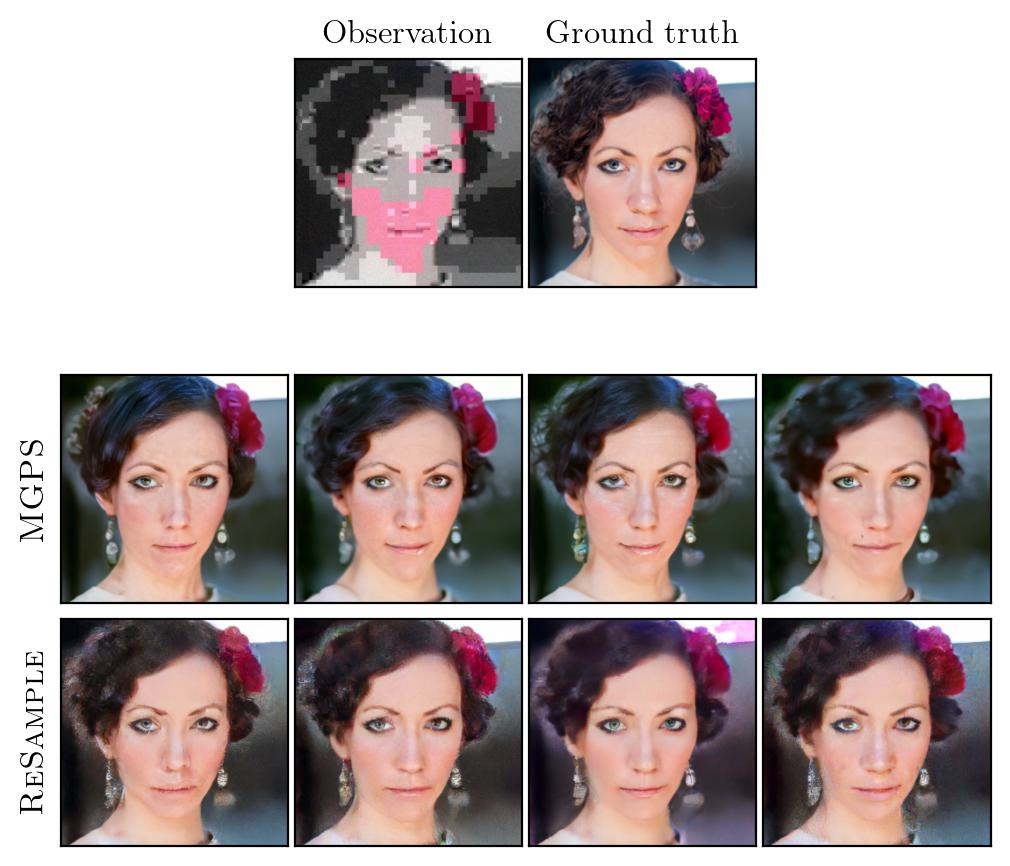}
        \includegraphics[width=0.45\textwidth]{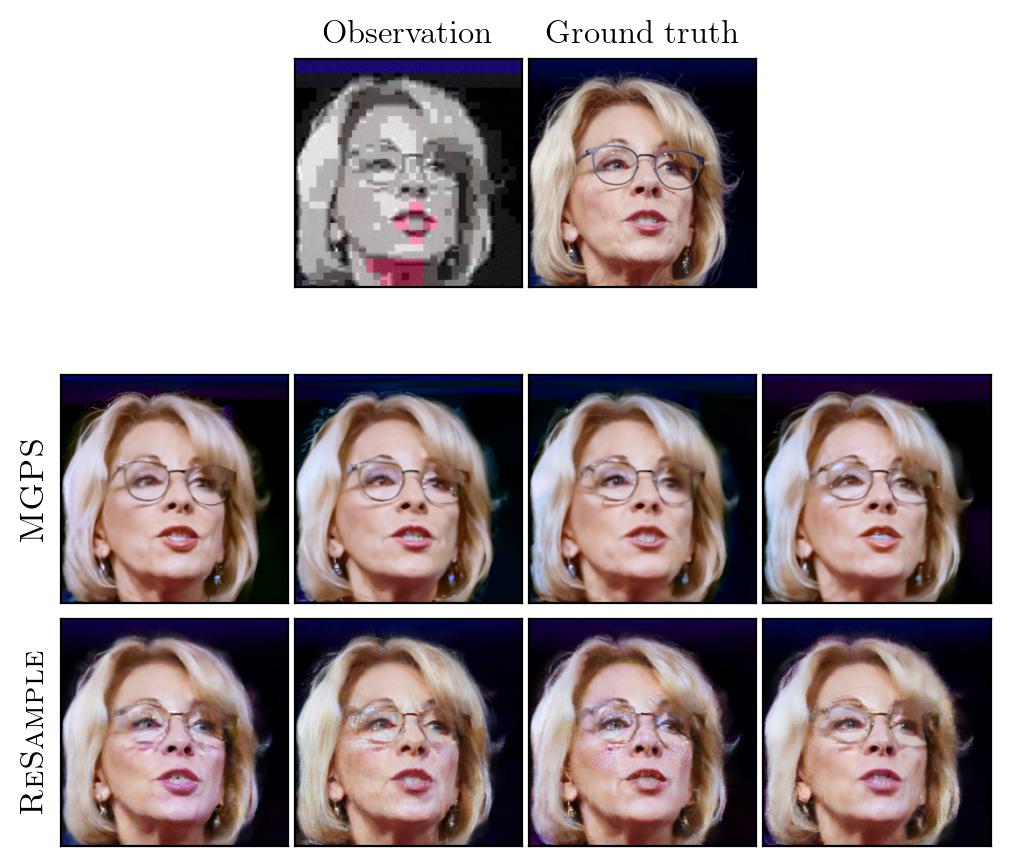}
    }
    \hspace{0.05\textwidth} % Horizontal space between figures
    \subfigure[Outpainting with half mask]{
        \includegraphics[width=0.45\textwidth]{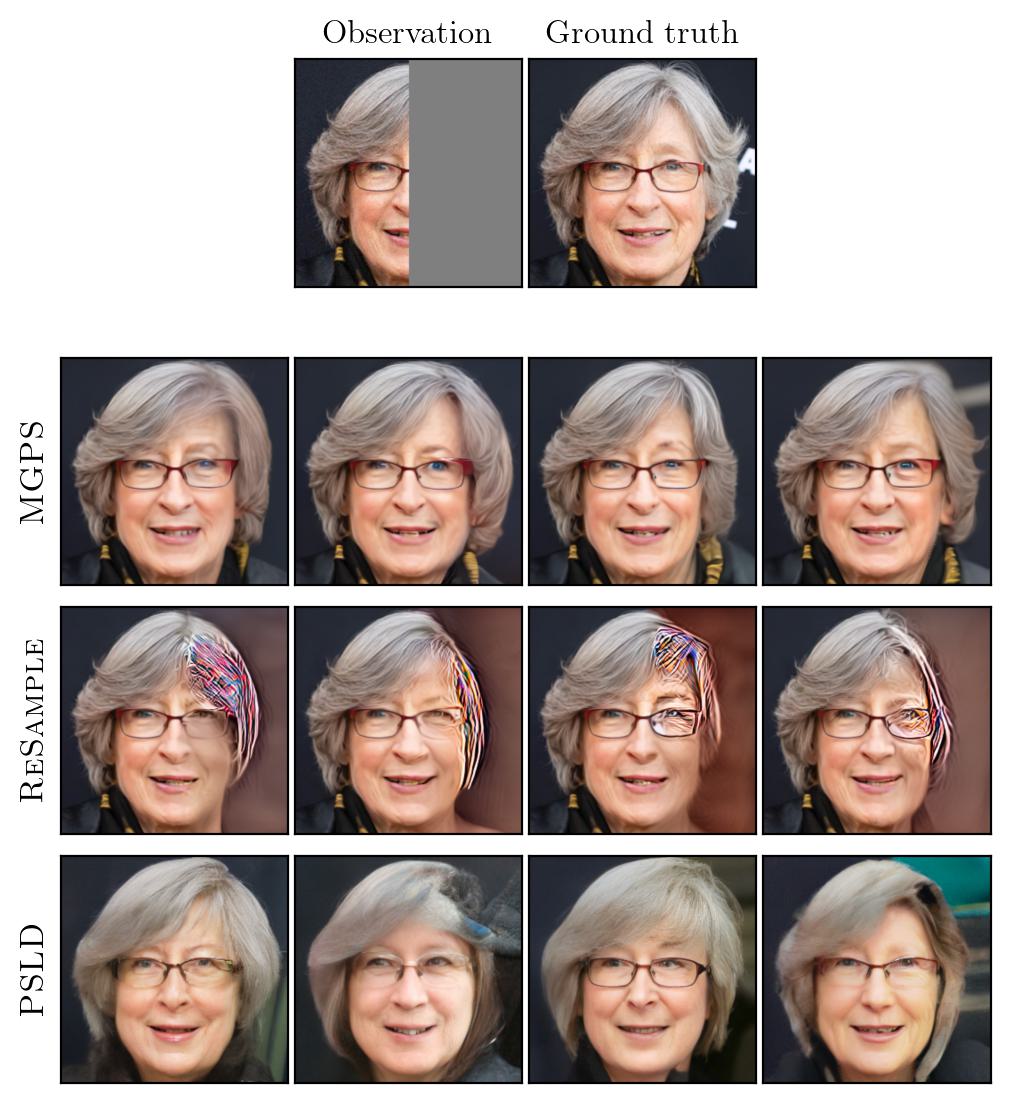}
        \includegraphics[width=0.45\textwidth]{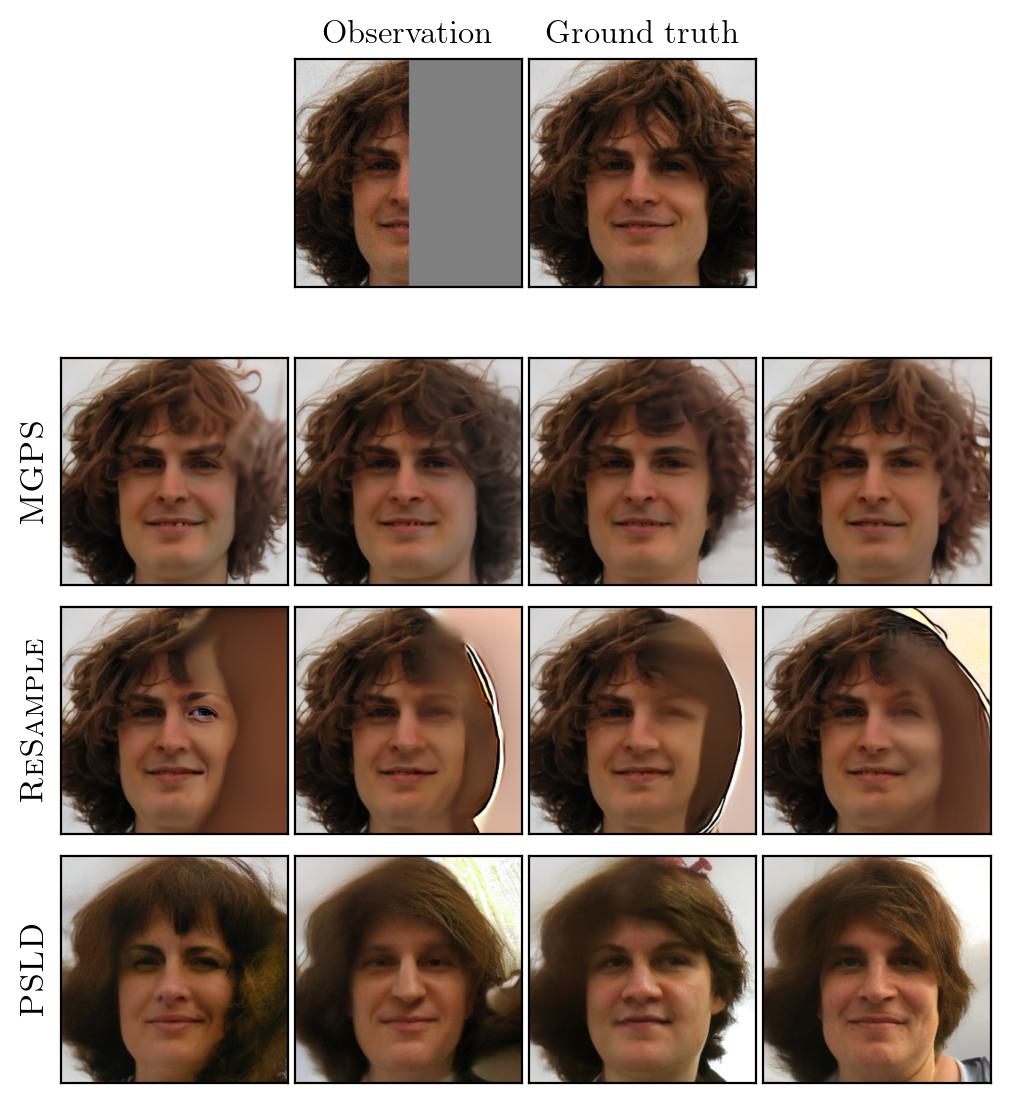}
    }
    \subfigure[SR $4 \times$]{
        \includegraphics[width=0.45\textwidth]{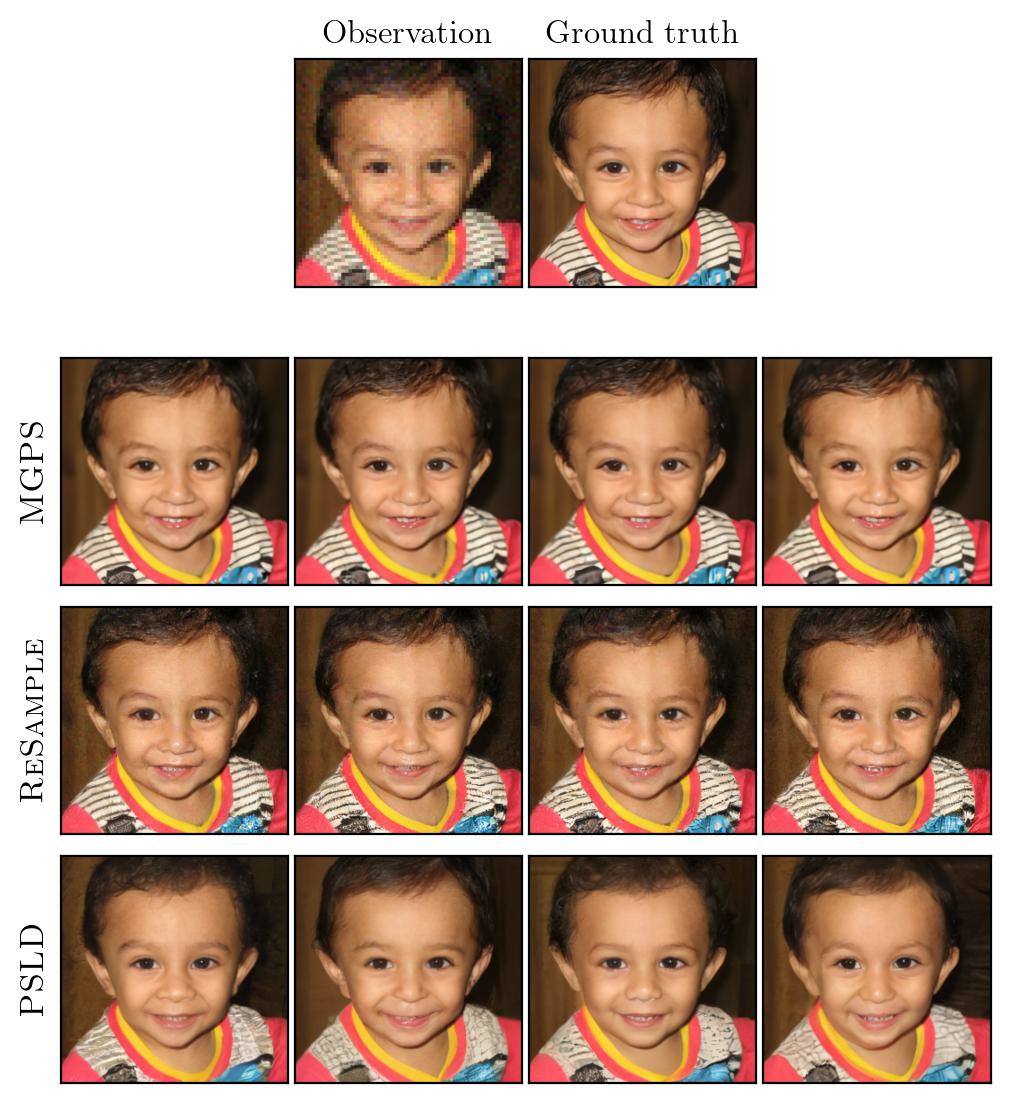}
        \includegraphics[width=0.45\textwidth]{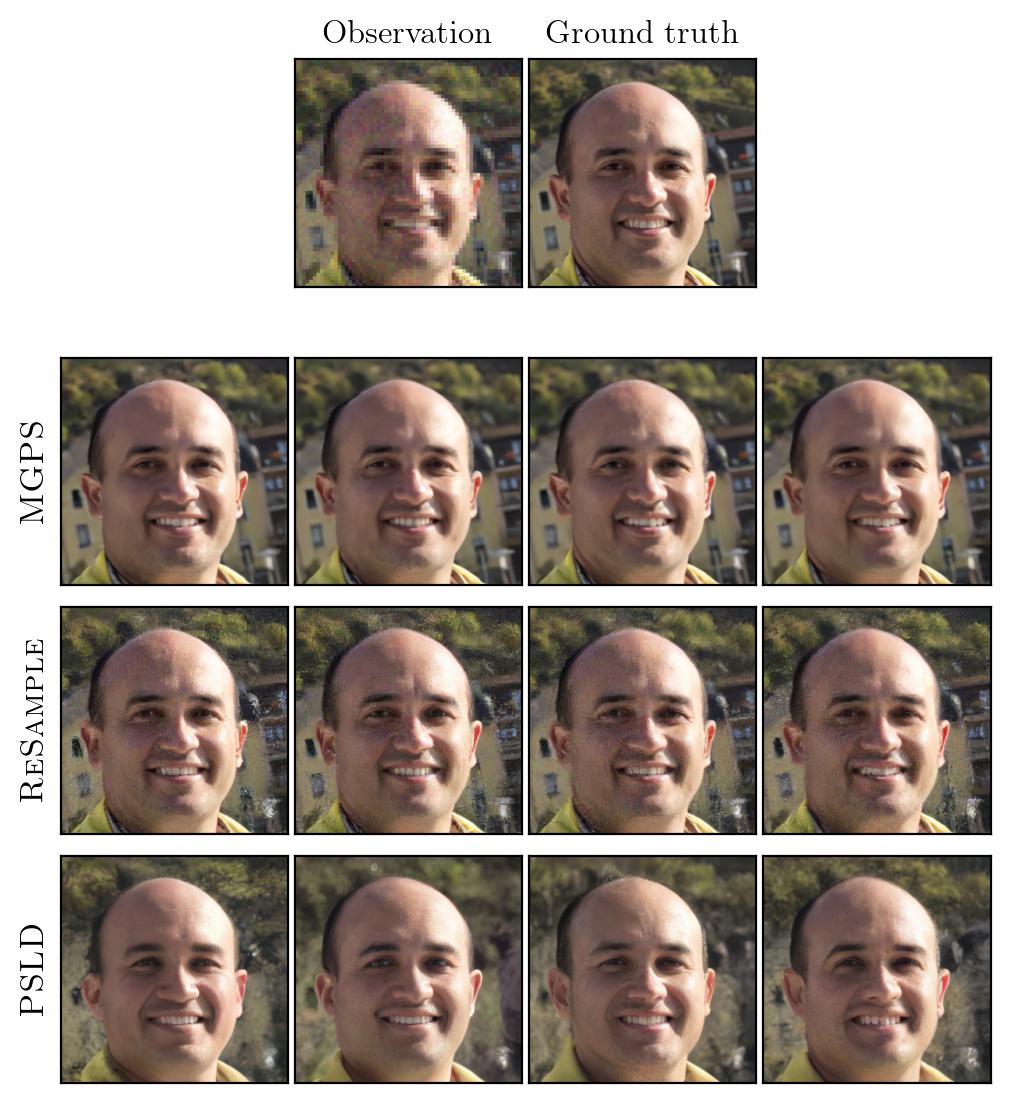}
    }
    \caption{Sample reconstructions with latent diffusion models on \ffhq\ dataset.}
\end{figure}

\def\imsize{0.9}
\begin{figure}[htb]
    \centering
        \includegraphics[width=\imsize\textwidth]{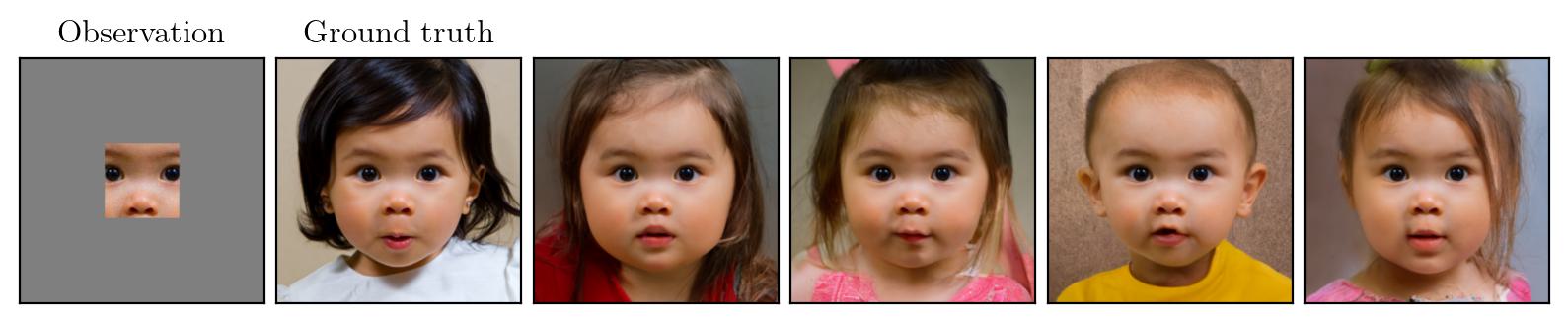}\\
        \includegraphics[width=\imsize\textwidth]{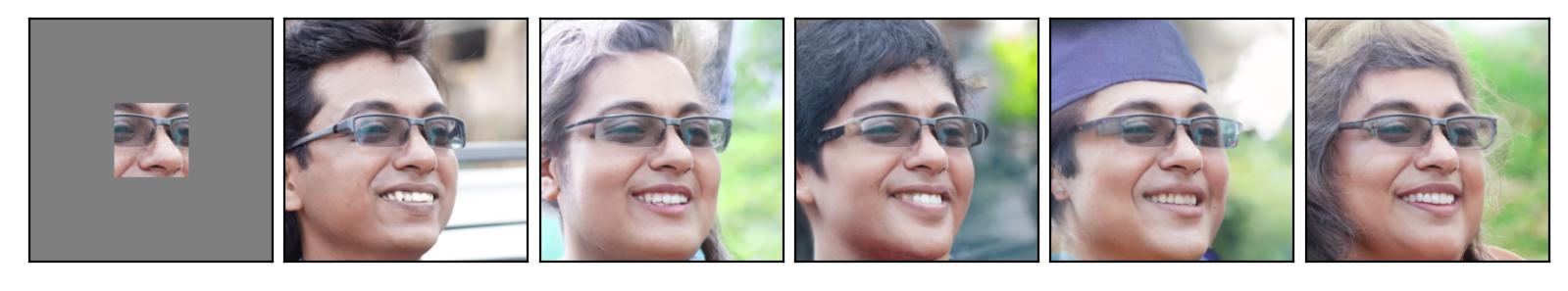}\\
        \includegraphics[width=\imsize\textwidth]{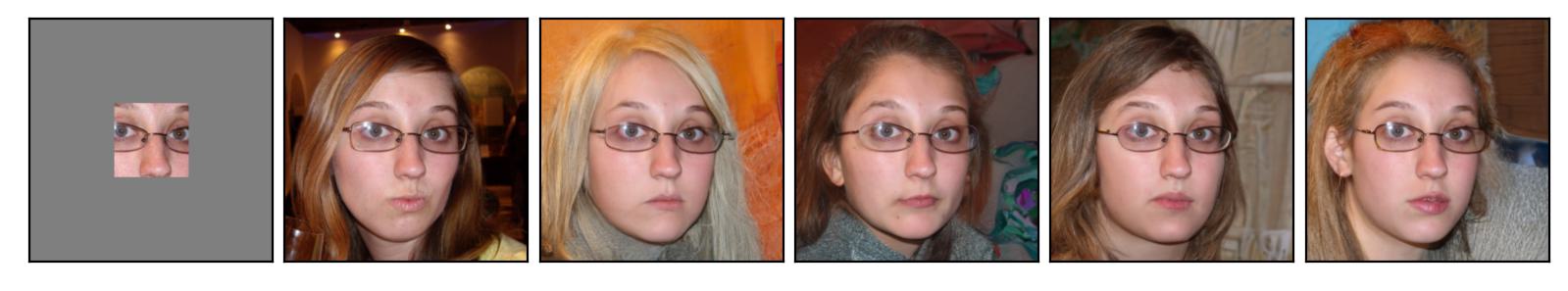}\\
        \includegraphics[width=\imsize\textwidth]{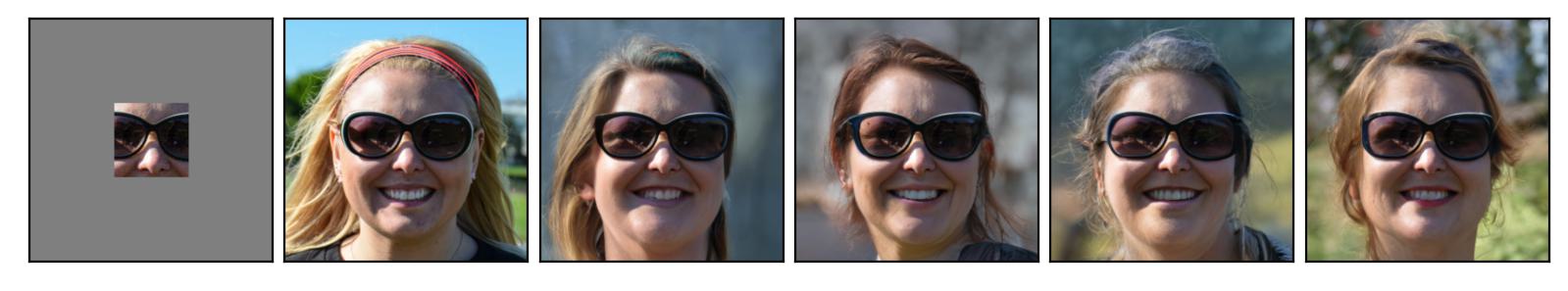}\\
        \includegraphics[width=\imsize\textwidth]{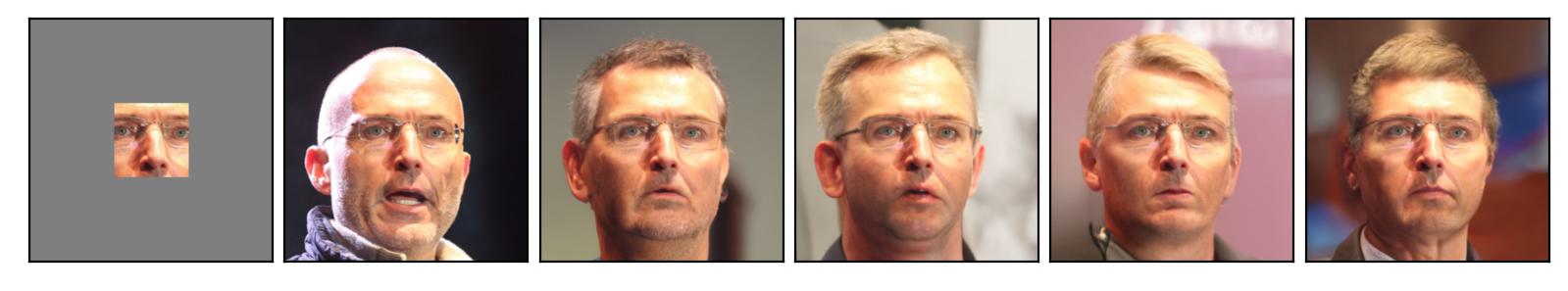}\\
        \includegraphics[width=\imsize\textwidth]{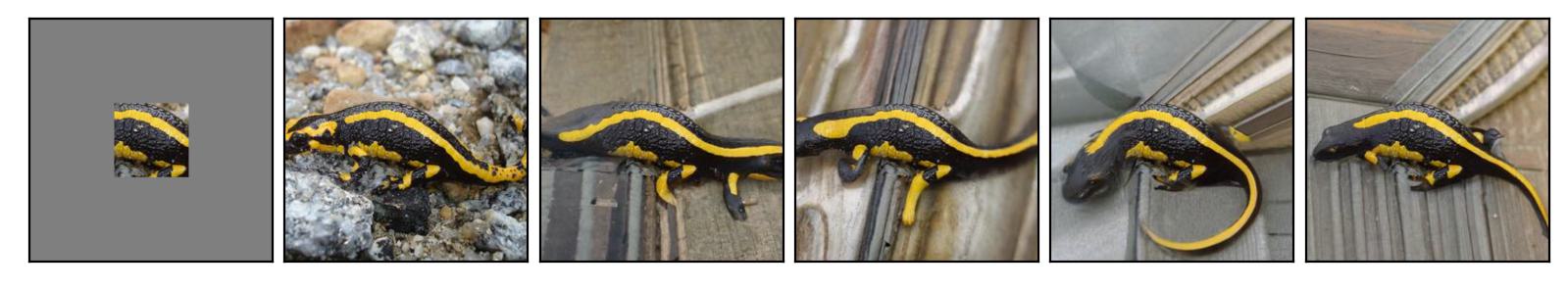}\\
        \includegraphics[width=\imsize\textwidth]{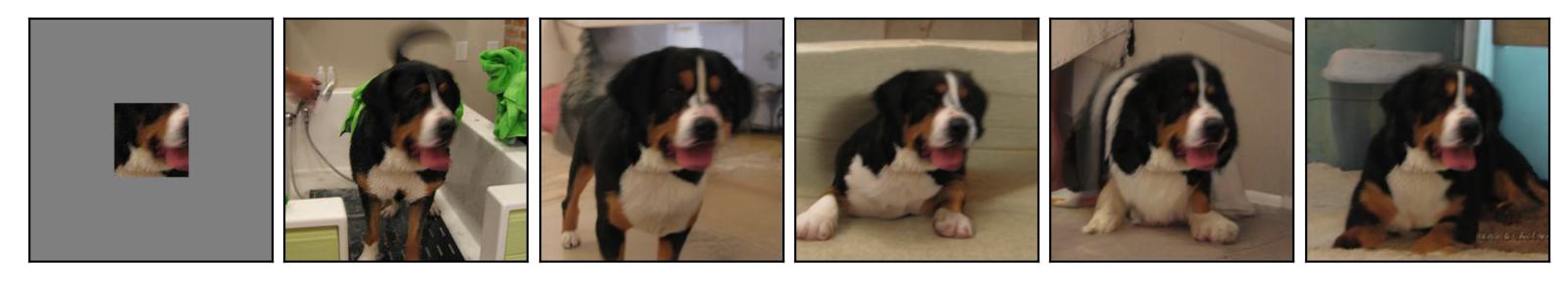}\\
        \includegraphics[width=\imsize\textwidth]{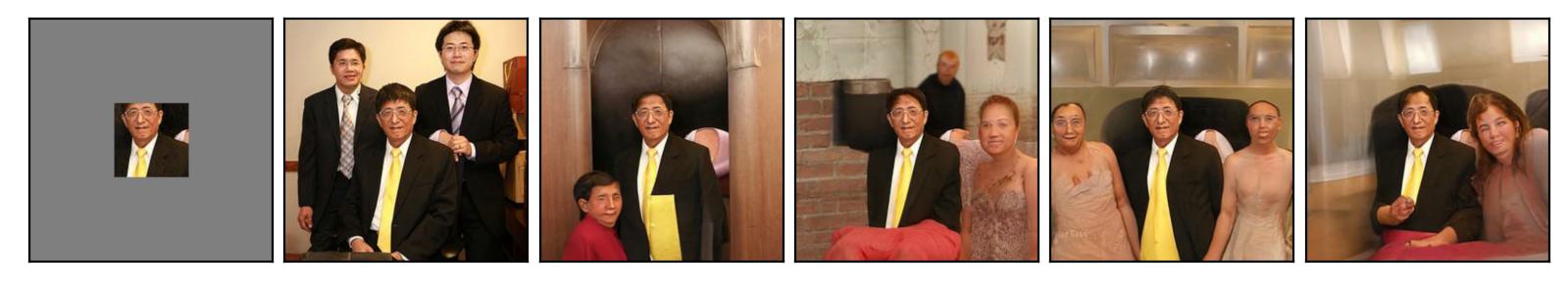}\\

    \caption{More sample reconstructions with \algo\ on the expand task.}
\end{figure}

\def\halfsize{0.8}
\begin{figure}[htb]
    \centering
    \includegraphics[width=\halfsize\textwidth]{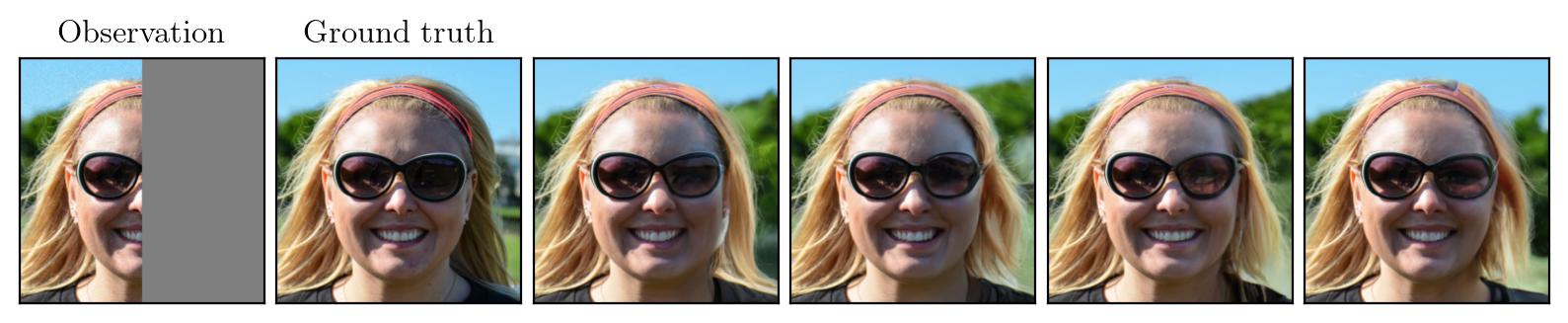}\\
    \includegraphics[width=\halfsize\textwidth]{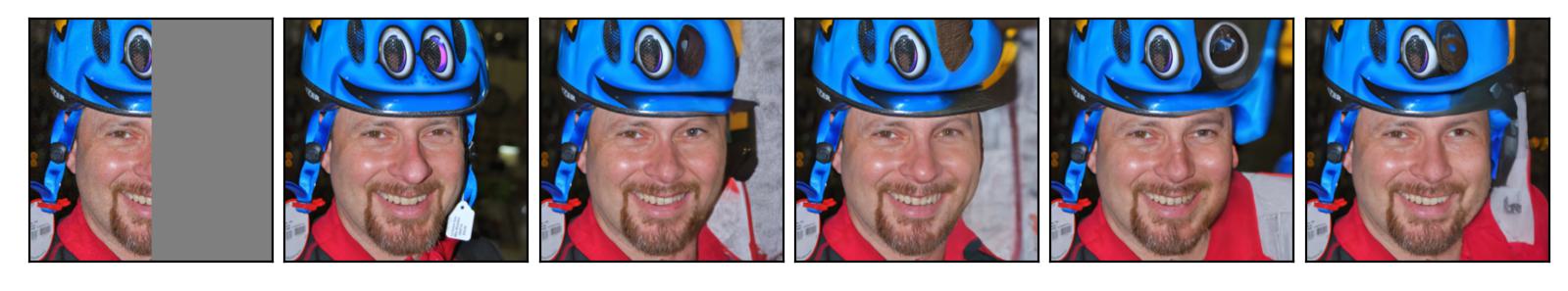}\\
    \includegraphics[width=\halfsize\textwidth]{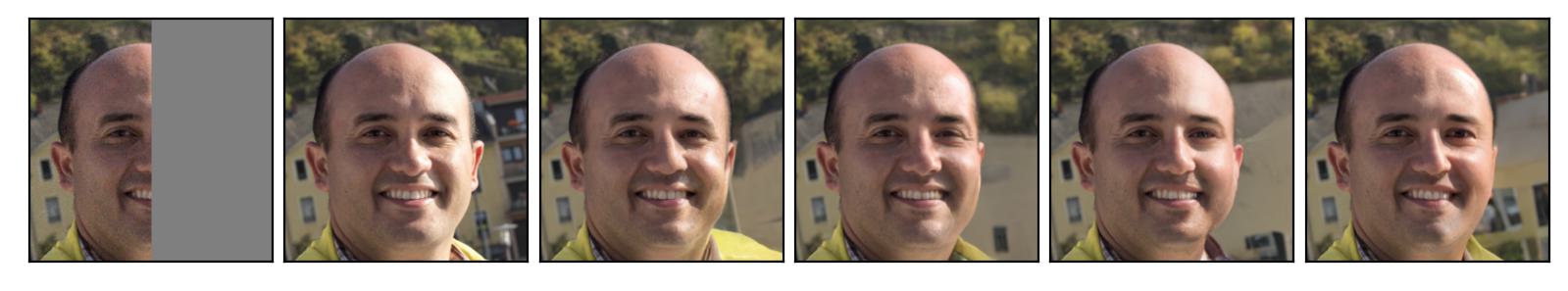}\\
    \includegraphics[width=\halfsize\textwidth]{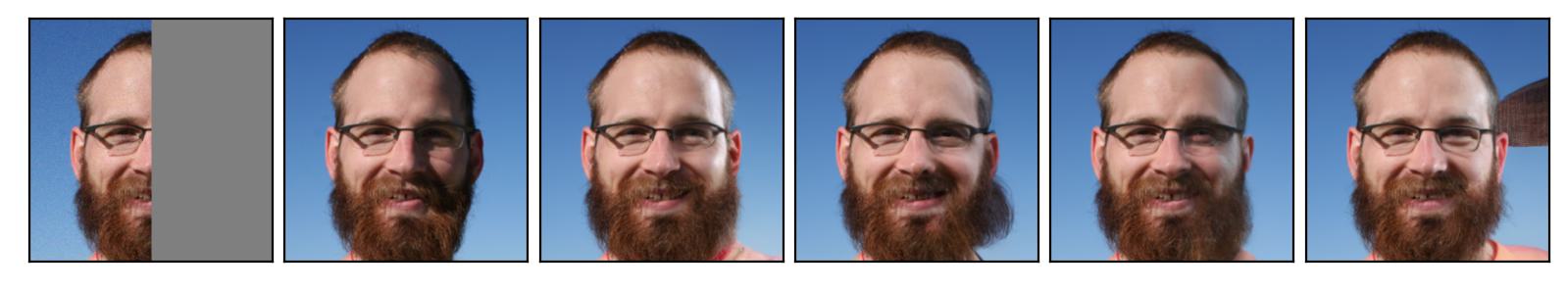}
        \includegraphics[width=\halfsize\textwidth]{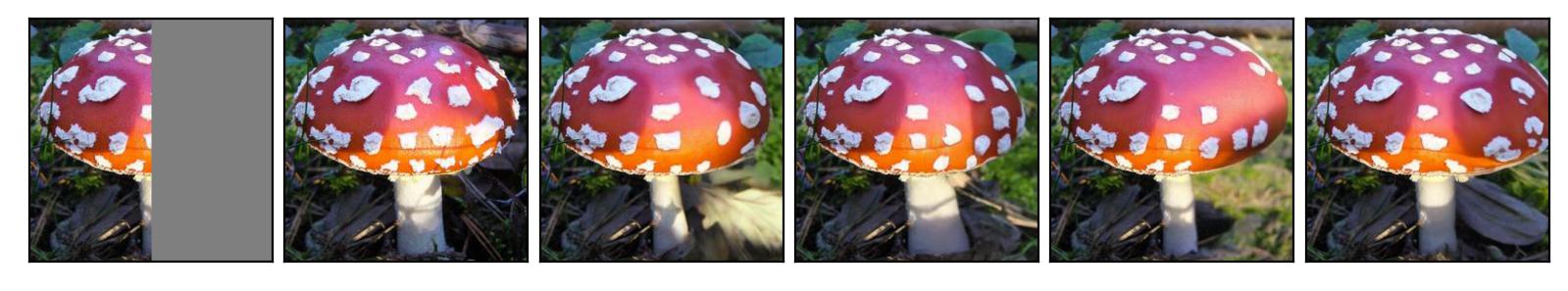}\\
        \includegraphics[width=\halfsize\textwidth]{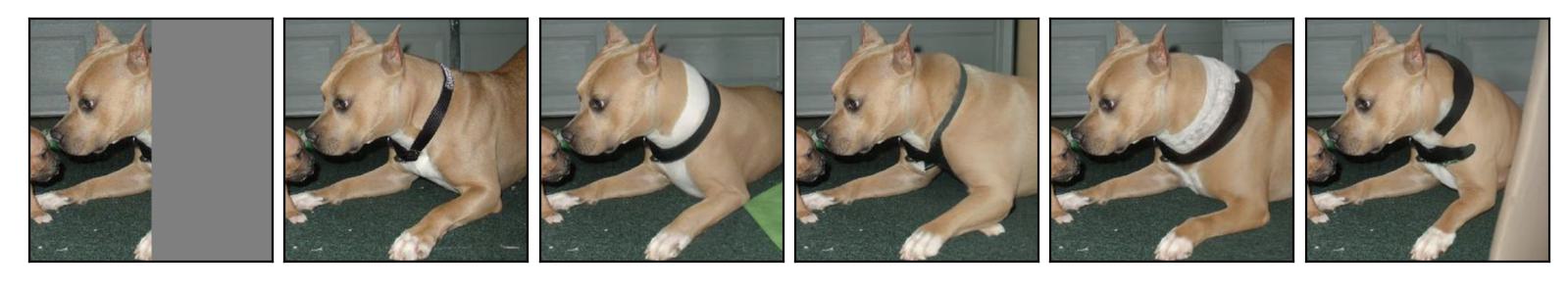}\\
        \includegraphics[width=\halfsize\textwidth]{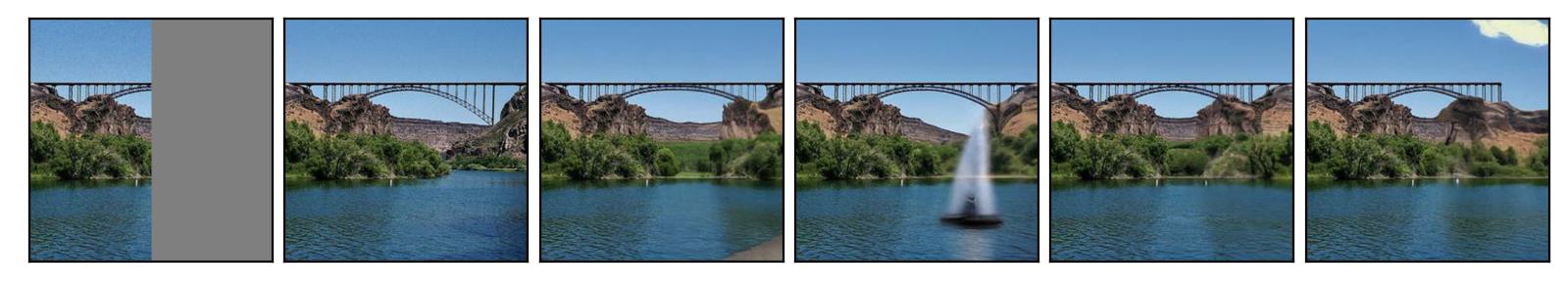}\\
        \includegraphics[width=\halfsize\textwidth]{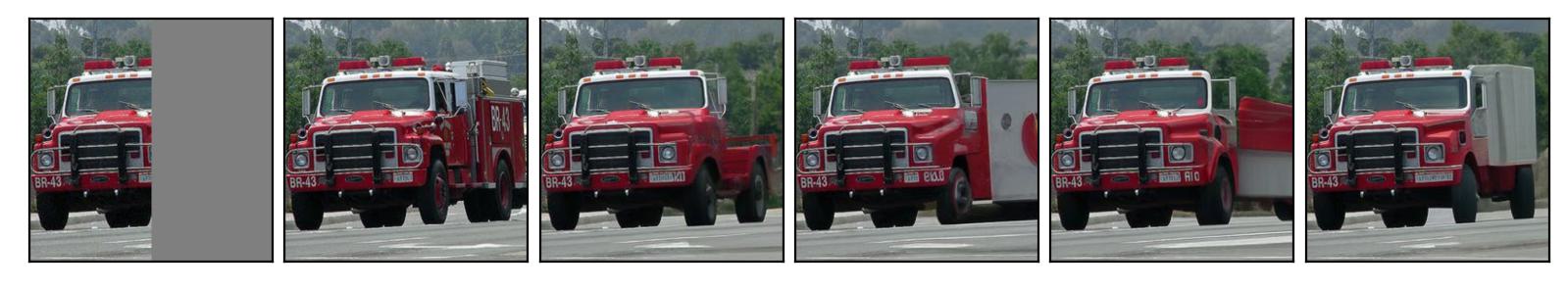}\\
        \includegraphics[width=\halfsize\textwidth]{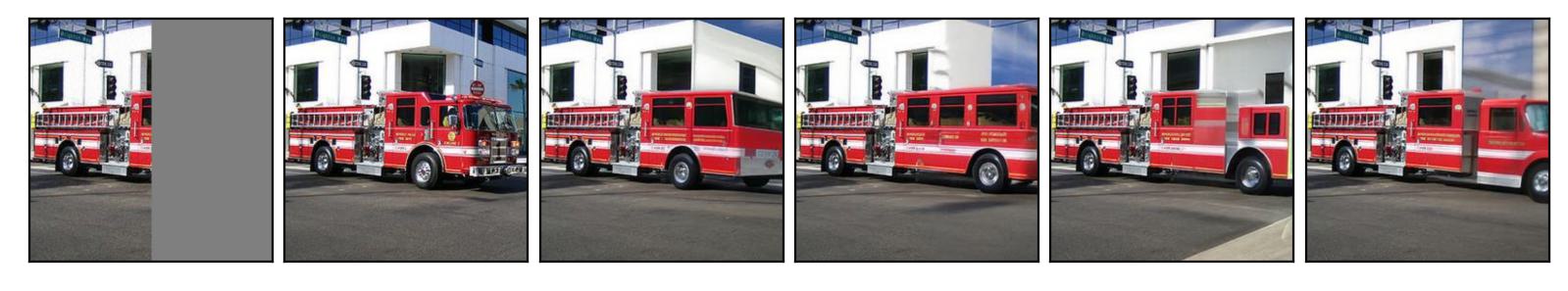}\\
        \includegraphics[width=\halfsize\textwidth]{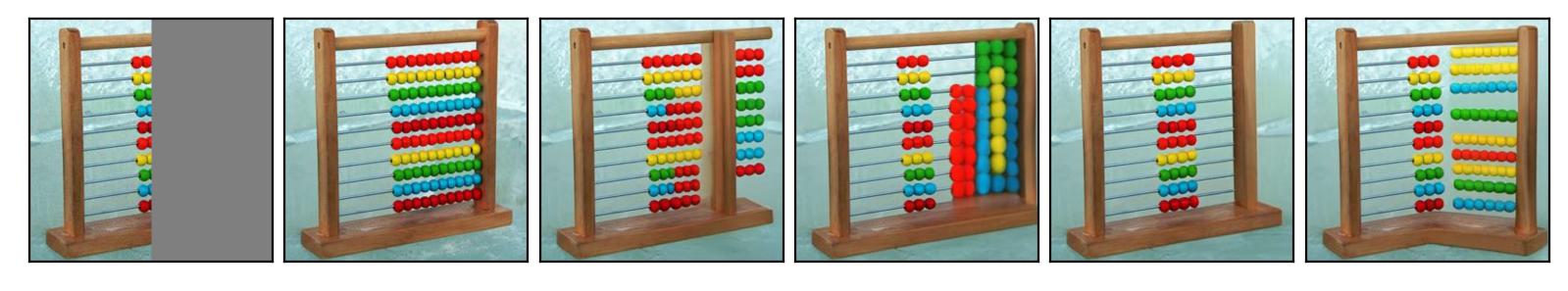}\\
    \caption{More sample reconstructions with \algo\ and half mask.}
\end{figure}

\begin{figure}[t]
    \centering
    \includegraphics[width=1.\textwidth]{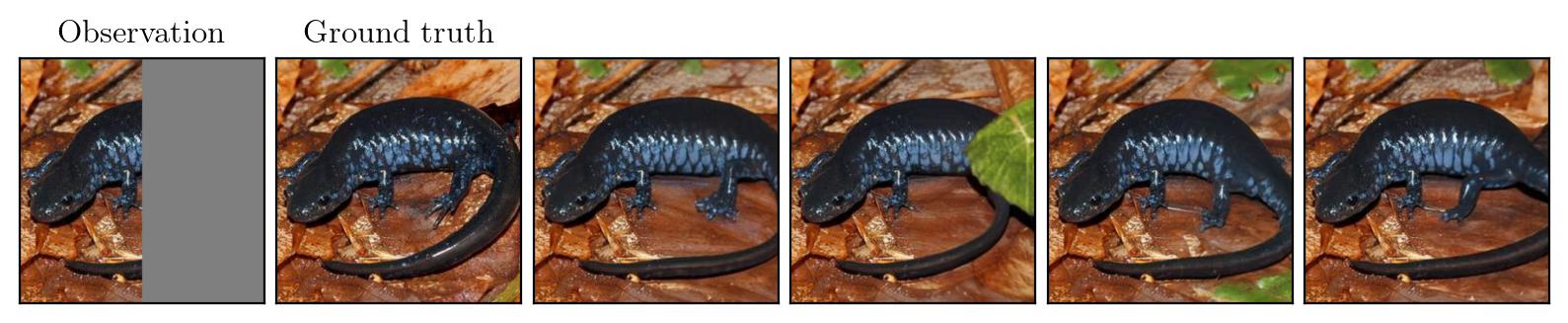}
        \includegraphics[width=1.\textwidth]{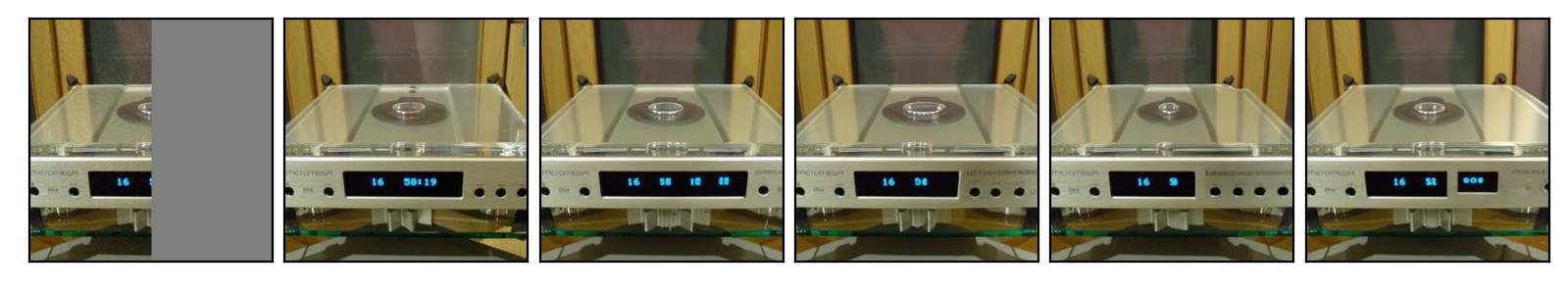}\\
        \includegraphics[width=1.\textwidth]{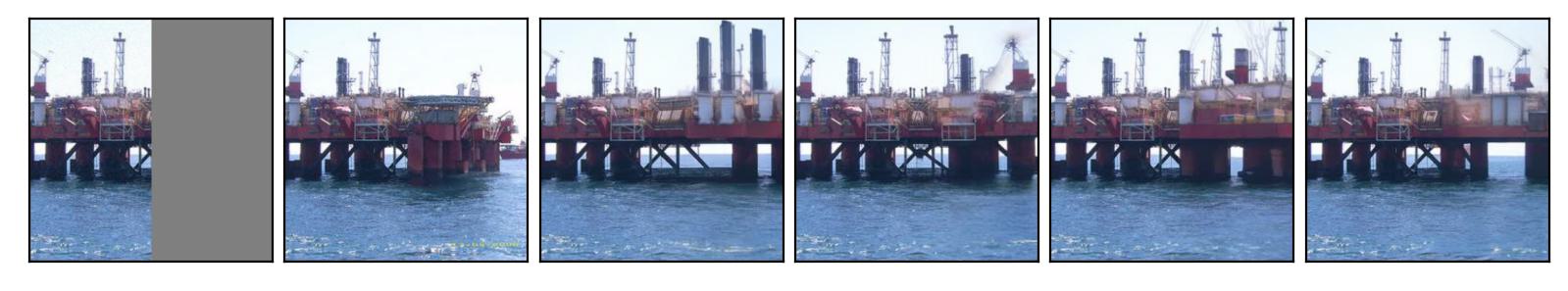}\\
        \includegraphics[width=1.\textwidth]{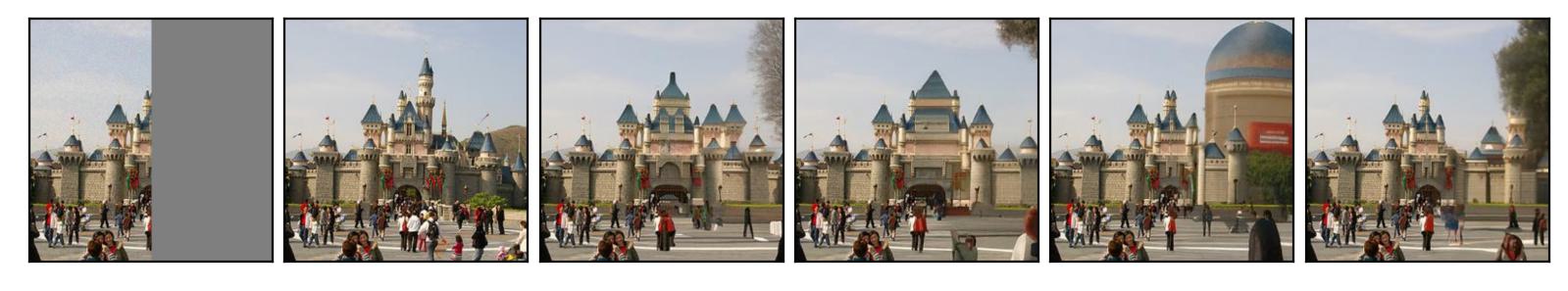}
    \caption{More sample reconstructions with \algo\ and half mask.}
\end{figure}
\revision{
    While it may appear that some of the methods underperform on some tasks/images compared to the original publications, for instance \Cref{fig:imagnet-inpainting-1} and \Cref{fig:imagnet-inpainting-2}, they still produce competitive reconstructions on others; see for example \Cref{fig:ffhq-inpainting-1}, \Cref{fig:imagnet-inpainting-3}, and \Cref{fig:imagnet-inpainting-4}.
    Similar patterns are also displayed in \citet[Figure 10]{zhang2023towards} and \citet[Figure 9]{liu2023img-inpainting}.
    With that being said, we highlight that these discrepancies appear more frequently on the ImageNet dataset than the FFHQ one.
    This can be explained by the fact that \imagenet\ is notoriously challenging due to its diversity, encompassing 1000 classes.
    This also seems to happen on one of the most difficult tasks, namely the half mask one. 
}

\end{document}